\documentclass[10pt,twocolumn,letterpaper]{article}

\usepackage{cvpr}              %

\usepackage{times}
\usepackage{epsfig}
\usepackage{graphicx}
\usepackage{amsmath}
\usepackage{amssymb}
\usepackage{pifont}
\usepackage{array}
\usepackage{comment} 
\usepackage{capt-of}
\usepackage{booktabs}
\usepackage{tabularx}
\usepackage{soul}
\usepackage{siunitx}
\usepackage{multirow}
\usepackage{multicol}
\usepackage[dvipsnames]{xcolor}

\usepackage{appendix}

\usepackage[pagebackref,breaklinks,colorlinks]{hyperref}

\newcommand{\w}{$\mathcal{W}$\xspace}
\newcommand{\wplus}{$\mathcal{W}+$\xspace}

\usepackage[capitalize]{cleveref}
\crefname{section}{Sec.}{Secs.}
\Crefname{section}{Section}{Sections}
\Crefname{table}{Table}{Tables}
\crefname{table}{Tab.}{Tabs.}

\makeatletter
\DeclareRobustCommand\onedot{\futurelet\@let@token\@onedot}
\def\@onedot{\ifx\@let@token.\else.\null\fi\xspace}

\def\etal{\emph{et al}\onedot}
\def\naive{na\"{\i}ve\xspace}
\def\Naive{Na\"{\i}ve\xspace}
\makeatother

\begin{document}

\title{HyperStyle: StyleGAN Inversion with HyperNetworks for Real Image Editing\vspace*{-0.4cm}}

\author{\vspace{0.2cm}Yuval Alaluf$^*$ \hspace{0.65cm} Omer Tov$^*$ \hspace{0.65cm} Ron Mokady \hspace{0.65cm} Rinon Gal \hspace{0.65cm} Amit Bermano \\
Blavatnik School of Computer Science, Tel Aviv University \vspace{-0.1cm}
}

\maketitle

\def\thefootnote{*}\footnotetext{Denotes equal contribution}

\vspace*{-0.8cm}
\begin{abstract}
\vspace{-0.2cm}
The inversion of real images into StyleGAN's latent space is a well-studied problem. Nevertheless, applying existing approaches to real-world scenarios remains an open challenge, due to an inherent trade-off between reconstruction and editability: latent space regions which can accurately represent real images typically suffer from degraded semantic control. Recent work proposes to mitigate this trade-off by fine-tuning the generator to add the target image to well-behaved, editable regions of the latent space. While promising, this fine-tuning scheme is impractical for prevalent use as it requires a lengthy training phase for each new image. In this work, we introduce this approach into the realm of encoder-based inversion. We propose HyperStyle, a hypernetwork that learns to modulate StyleGAN's weights to faithfully express a given image in editable regions of the latent space. A naive modulation approach would require training a hypernetwork with over three billion parameters. Through careful network design, we reduce this to be in line with existing encoders. HyperStyle yields reconstructions comparable to those of optimization techniques with the near real-time inference capabilities of encoders. Lastly, we demonstrate HyperStyle's effectiveness on several applications beyond the inversion task, including the editing of out-of-domain images which were never seen during training. Code is available on our project page: \small{\url{https://yuval-alaluf.github.io/hyperstyle/}}.
\end{abstract}
\vspace{-10pt}

\begin{figure}
    \centering
    \vspace{-0.3cm}
    \includegraphics[width=\linewidth]{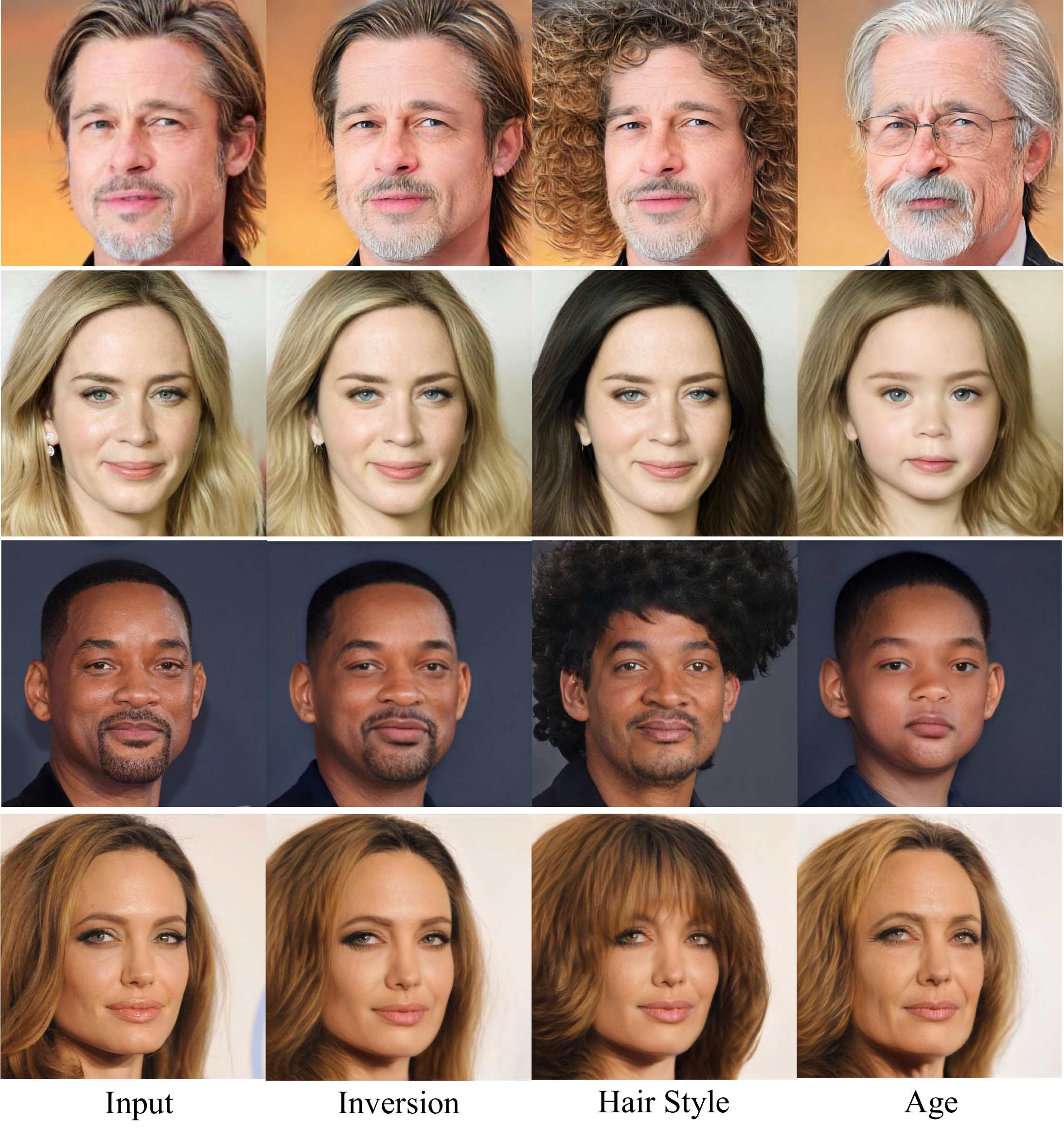}
    \vspace{-0.75cm}
    \caption{
    Given a desired input image, our hypernetworks \textit{learn} to modulate a pre-trained StyleGAN network to achieve accurate image reconstructions in editable regions of the latent space. 
    Doing so enables one to effectively apply techniques such as StyleCLIP~\cite{patashnik2021styleclip} and InterFaceGAN~\cite{shen2020interpreting} for editing real images.
    }
    \vspace{-0.2cm}
    \label{fig:teaser}
\end{figure}

\vspace{-0.55cm}
\section{Introduction}
\vspace{-0.03cm}
Generative Adversarial Networks (GANs)~\cite{Goodfellow2014GenerativeAN}, and in particular StyleGAN~\cite{karras2019style,karras2020analyzing,karras2020training,aliasfreeKarras2021} have become the gold standard for image synthesis.
Thanks to their semantically rich latent representations, many works have utilized these models to facilitate diverse and expressive editing through latent space manipulations~\cite{harkonen2020ganspace,shen2020interpreting,menon2020pulse,Chen2021DeepFaceEditingDF,alaluf2021matter,Lewis2021TryOnGANBT,Collins2020EditingIS,bau2021paint,patashnik2021styleclip}.
Yet, a significant challenge in adopting these approaches for real-world applications is the ability to edit \textit{real} images. For editing a real photo, one must first find its corresponding latent representation via a process commonly referred to as GAN inversion~\cite{zhu2016generative}. While the inversion process is a well-studied problem, it remains an open challenge.

Recent works~\cite{abdal2020image2stylegan++,zhu2020improved,zhu2020domain,tov2021designing} have demonstrated the existence of a \textit{distortion-editability} trade-off:
one may invert an image into well-behaved~\cite{zhu2020improved} regions of StyleGAN's latent space and attain good editability. However, these regions are typically less expressive, resulting in reconstructions that are less faithful to the original image.
Recently, Roich~\etal~\cite{roich2021pivotal} showed that one may side-step this trade-off by considering a different approach to inversion.
Rather than searching for a latent code that most accurately reconstructs the input image, they fine-tune the generator in order to insert a target identity into well-behaved regions of the latent space. In doing so, they demonstrate state-of-the-art reconstructions while retaining a high level of editability. Yet, this approach relies on a costly per-image optimization of the generator, requiring up to a minute per image.

A similar \textit{time-accuracy} trade-off can be observed in classical inversion approaches. On one end of the spectrum, latent vector optimization approaches~\cite{lipton2017precise,creswell2018inverting,abdal2019image2stylegan,abdal2020image2stylegan++,karras2020analyzing} achieve impressive reconstructions, but are impractical at scale, requiring several minutes per image. On the other end, encoder-based approaches leverage rich datasets to learn a mapping from images to their latent representations. These approaches operate in a fraction of a second but are typically less faithful in their reconstructions.

In this work, we aim to bring the generator-tuning technique of Roich~\etal~\cite{roich2021pivotal} to the realm of interactive applications by adapting it to an encoder-based approach. We do so by introducing a hypernetwork~\cite{ha2016hypernetworks} that \textit{learns} to refine the generator weights with respect to a given input image.
The hypernetwork is composed of a lightweight feature extractor (e.g., ResNet~\cite{he2015deep}) and a set of refinement blocks, one for each of StyleGAN's convolutional layers. Each refinement block is tasked with predicting offsets for the weights of the convolutional filters of its corresponding layer.
A major challenge in designing such a network is the number of parameters comprising each convolutional block that must be refined. {\Naive}ly predicting an offset for each parameter would require a hypernetwork with over three billion parameters.
We explore several avenues for reducing this complexity: sharing offsets between parameters, sharing network weights between different hypernetwork layers, and an approach inspired by depthwise-convolutions~\cite{howard2017mobilenets}.
Lastly, we observe that reconstructions can be further improved through an iterative refinement scheme~\cite{alaluf2021restyle} which gradually predicts the desired offsets over a small number of forward passes through the hypernetwork. 
By doing so, our approach, HyperStyle, essentially learns to ``optimize" the generator in an efficient manner.

The relation between HyperStyle and existing generator-tuning approaches can be viewed as similar to the relation between encoders and optimization inversion schemes.
Just as encoders find a \textit{desired latent code} via a learned network, our hypernetwork efficiently finds a \textit{desired generator} with no image-specific optimization. 

We demonstrate that HyperStyle achieves a significant improvement over current encoders. Our reconstructions even rival those of optimization schemes, while being several orders of magnitude faster.
We additionally show that HyperStyle preserves the appealing structure and semantics of the original latent space, allowing one to leverage off-the-shelf editing techniques on the resulting inversions,  see \cref{fig:teaser}.
Finally, we show that HyperStyle generalizes well to out-of-domain images, such as paintings and animations, even when unobserved during the training of the hypernetwork itself.
This hints that the hypernetwork does not only learn to correct specific flawed attributes, but rather learns to refine the generator in a more general sense.

\section{Background and Related Work}~\label{sec:rw}

\vspace{-1.1cm}
\paragraph{Hypernetworks}
Introduced by Ha~\etal~\cite{ha2016hypernetworks}, hypernetworks are neural networks tasked with predicting the weights of a primary network. By training a hypernetwork over a large data collection, the primary network's weights are adjusted with respect to specific inputs, yielding a more expressive model. Hypernetworks have been applied to a wide range of applications including semantic segmentation~\cite{nirkin2021hyperseg}, 3D modeling~\cite{littwin2019deep,spurek2021hyperpocket}, neural architecture search~\cite{zhang2018graph}, and continual learning~\cite{von2019continual}, among others. 

\vspace{-0.45cm}
\paragraph{Latent Space Manipulation}
A widely explored application for generative models is their use for the editing of real images.
Considerable effort has gone into leveraging StyleGAN~\cite{karras2019style,karras2020analyzing} for such tasks, owing to its highly-disentangled latent spaces.
Many methods have been proposed for finding semantic latent directions using varying levels of supervision. These range from full-supervision in the form of semantic labels~\cite{shen2020interpreting,abdal2020styleflow,denton2019detecting, goetschalckx2019ganalyze} and facial priors~\cite{tewari2020pie, tewari2020stylerig} to unsupervised approaches~\cite{harkonen2020ganspace,shen2020closedform,voynov2020unsupervised,wang2021a}. Others have explored self-supervised approaches~\cite{jahanian2019steerability,spingarn2020gan,plumerault2020controlling}, the mixing of latent codes to produce local edits~\cite{collins2020editing,kafri2021stylefusion,chong2021retrieve}, and the use of contrastive language-image (CLIP) models ~\cite{radford2021learning} to achieve new editing capabilities~\cite{patashnik2021styleclip,xia2021tedigan,gal2021stylegannada}.
Applying these methods to real images requires one to first perform an accurate inversion of the given image. 

\vspace{-0.45cm}
\paragraph{GAN Inversion}
GAN inversion~\cite{zhu2016generative} is the process of obtaining a latent code that can be passed to the generator to reconstruct a given image.
Generally, inversion methods either directly optimize the latent vector to minimize the error for a given image~\cite{lipton2017precise,creswell2018inverting,abdal2019image2stylegan,abdal2020image2stylegan++,semantic2019bau,zhu2020improved,zhu2016generative,yeh2017semantic,gu2020image}, train an encoder over a large number of samples to learn a mapping from an image to its latent representation~\cite{perarnau2016invertible, luo2017learning, guan2020collaborative,pidhorskyi2020adversarial,guan2020collaborative,richardson2020encoding,tov2021designing,alaluf2021restyle,kang2021gan,kim2021exploiting,wang2021HFGI}, or use a hybrid approach combining both~\cite{zhu2016generative,zhu2020domain}. Among encoder-based methods, Alaluf \etal~\cite{alaluf2021restyle} iteratively refine the predicted latent code through a small number of forward passes through the network. Our work adopts this idea and applies it to the generator weight offsets predicted by the hypernetwork.
Finally, in a concurrent work, Dinh~\etal~\cite{dinh2021hyperinverter} also explore the use of hypernetworks for achieving higher fidelity inversions.

\vspace{-0.1cm}
\begin{figure*}[tb]
    \centering
    \setlength{\belowcaptionskip}{-5pt}
    \includegraphics[width=0.97\linewidth]{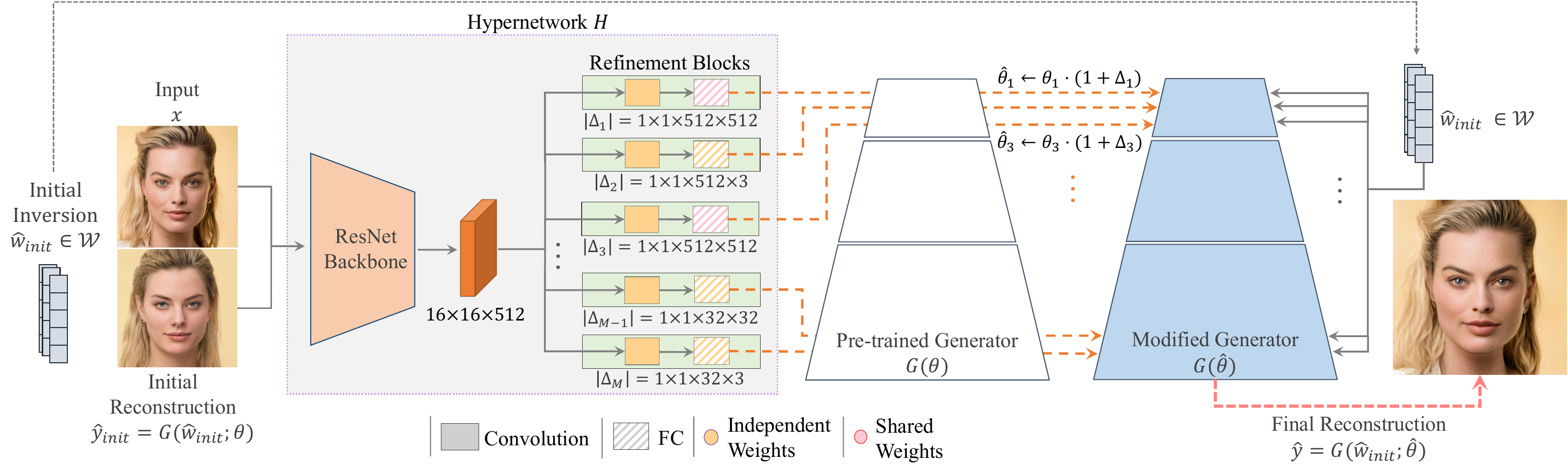}
    \vspace{-0.25cm}
    \caption{
    The HyperStyle scheme. 
    Given an image $x$, we begin with an initial, approximate latent code $\hat{w}_{init} \in \mathcal{W}$ with a corresponding reconstruction $\hat{y}_{init} = G(\hat{w}_{init}; \theta)$ obtained using a pre-trained generator $G$ with weights $\theta$.
    Given inputs $x$ and $\hat{y}_{init}$, our hypernetwork $H$ predicts a set of offsets $\Delta_\ell$ used to modulate $G$'s weights at various input layers $\ell$.
    This results in a modified generator $G$ parameterized by new weights $\hat{\theta}$, shown in blue.
    To predict the desired offsets for the given image, we incorporate multiple Refinement Blocks, one for each generator layer we wish to modify.
    The final reconstruction $\hat{y} = G(\hat{w}_{init}; \hat{\theta})$ is then synthesized using the modified generator. 
    }
    \vspace{-0.2cm}
    \label{fig:overview}
\end{figure*}

\vspace{-0.43cm}
\paragraph{Distortion-Editability}
Typically, latent traversal and inversion methods concern themselves with one of two spaces: \w, obtained via StyleGAN's mapping network and \wplus, where each layer of the generator is assigned a different latent code $w_i \in \mathcal{W}$.
Images inverted into $\mathcal{W}$ show a high degree of editability: they can be modified through latent space traversal with minimal corruption. However, $\mathcal{W}$ offers poor expressiveness, limiting the range of images that can be faithfully reconstructed. Therefore, many prior works invert into the extended \wplus space, achieving reduced distortion at the cost of inferior editability.
Tov~\etal~\cite{tov2021designing} suggest balancing the two by designing an encoder that predicts codes in \wplus residing close to $\mathcal{W}$.
Others have explored similar ideas for optimization~\cite{zhu2020improved}.

\paragraph{Generator Tuning}
To leverage the visual quality of a pre-trained generator, most works avoid altering the generator weights when performing the inversion.
Nonetheless, some works have explored performing a per-image tuning of the generator to obtain more accurate inversions. 
Pan~\etal~\cite{pan2020exploiting} invert BigGAN~\cite{brock2018large} by randomly sampling noise vectors, selecting the one that best matches the real image, and optimizing it simultaneously with the generator weights in a progressive manner.
Roich~\etal~\cite{roich2021pivotal} and Hussien~\etal~\cite{hussein2020image} invert images into a pre-trained GAN by first recovering a latent code which approximately reconstructs the target image and then fine-tuning the generator weights for improve image-specific details. 
Bau~\etal~\cite{semantic2019bau} explored the use of a neural network to predict feature modulations to improve GAN inversion.
However, the aforementioned works require a lengthy optimization for every input, typically requiring minutes per image. As such, these methods are often inapplicable to real-world scenarios at scale. 
In contrast, we train a hypernetwork over a large set of images, resulting in a \textit{single} network used to refine the generator for any given image. Importantly, this is achieved in near real-time and is more suitable for interactive settings.
\setlength{\abovedisplayskip}{3.5pt}
\setlength{\belowdisplayskip}{3.5pt}

\vspace{-0.125cm}
\section{Method}~\label{sec:method}%
\vspace{-0.7cm}
\subsection{Preliminaries}~\label{sec:preliminaries}%
When solving the GAN inversion task, our goal is to identify a latent code that minimizes the reconstruction distortion with respect to a given target image $x$:
\begin{align}~\label{eq:inversion}
\hat{w} = \underset{w}{\arg\min} \; \mathcal{L} \left ( x, G \left ( w; \theta \right ) \right ),
\end{align}
where $G(w; \theta)$ is the image produced by a \textit{pre-trained} generator $G$ parameterized by weights $\theta$, over the latent $w$. $\mathcal{L}$ is the loss objective, usually $L_2$ or LPIPS~\cite{zhang2018perceptual}. 
Solving \cref{eq:inversion} via optimization typically requires several minutes per image. To reduce inference times, an encoder $E$ can be trained over a large set of images $\{x^i\}_{i=1}^N$ to minimize:
\begin{align}
\sum_{i=1}^N \; \mathcal{L} \left ( x^i, G \left ( E \left ( x^i \right ); \theta \right ) \right ).
\end{align}
This results in a fast inference procedure $\hat{w} = E(x)$. 
A latent manipulation $f$ can then be applied over the inverted code $\hat{w}$ to obtain an edited image $G(f(\hat{w}); \theta)$.

Recently, Roich \etal~\cite{roich2021pivotal} propose injecting new identities into the well-behaved regions of StyleGAN's latent space. Given a target image, they use an optimization process to find an initial latent $\hat{w}_{init} \in \mathcal{W}$ leading to an approximate reconstruction. This is followed by a fine-tuning session where the generator weights are adjusted so that the same latent better reconstructs the specific image:
\begin{align} \label{eq:pti_objective}
\hat{\theta} = \underset{\theta}{\arg\min} \; \mathcal{L} ( x,  G  ( \hat{w}_{init}; \theta ) ),
\end{align}
where $\hat{\theta}$ represents the new generator weights.
The final reconstruction is obtained by utilizing the initial inversion and altered weights: $\hat{y} = G(\hat{w}_{init}; \hat{\theta})$.

\subsection{Overview}

Our method HyperStyle aims to perform the identity-injection operation by efficiently providing modified weights for the generator, as illustrated in \cref{fig:overview}.
We begin with an image $x$, a generator $G$ parameterized by weights $\theta$, and an initial inverted latent code $\hat{w}_{init} \in \mathcal{W}$. 
Using these weights and $\hat{w}_{init}$, we generate the initial reconstructed image $\hat{y}_{init} = G(\hat{w}_{init}; \theta)$. 
To obtain such a latent code we employ an off-the-shelf encoder~\cite{tov2021designing}.

Our goal is to predict a new set of weights $\hat{\theta}$ that minimizes the objective defined in \cref{eq:pti_objective}.
To this end, we present our hypernetwork $H$, tasked with predicting these weights. To assist the hypernetwork in inferring the desired modifications, we pass as input both the target image $x$ and the initial, approximate image reconstruction $\hat{y}_{init}$.
The predicted weights are thus given by: $\hat{\theta} = H\left(\hat{y}_{init}, x\right)$.
We train $H$ over a large collection of images with the goal of minimizing the distortion of the reconstructions:
\begin{equation}~\label{eq:objective}
\sum_{i=1}^N \; \mathcal{L} \left ( x^i,  G \left ( \hat{w}^i_{init} ; H \left (\hat{y}^i_{init}, x^i \right ) \right ) \right ).
\end{equation}
Given the hypernetwork predictions, the final reconstruction can be obtained as $\hat{y} = G(\hat{w}_{init}; \hat{\theta})$.

Owing to the reconstruction-editability trade-off outlined in \cref{sec:rw}, the initial latent code should reside within the well-behaved (i.e., editable) regions of StyleGAN's latent space. To this end, we employ a pre-trained e4e encoder~\cite{tov2021designing} into \w that is kept fixed throughout the training of the hypernetwork.
As shall be shown, by tuning around such a code, one can apply the same editing techniques as used with the original generator.

In practice, rather than directly predicting the new generator weights, our hypernetwork predicts a set of offsets with respect to the original weights. In addition, we follow ReStyle~\cite{alaluf2021restyle} and perform a small number of passes (e.g., $5$) through the hypernetwork to gradually refine the predicted weight offsets, resulting in higher-fidelity inversions.

In a sense, one may view HyperStyle as \emph{learning to optimize} the generator, but doing so in an efficient manner. 
Moreover, by learning to modify the generator, HyperStyle is given more freedom to determine how to best project an image into the generator, even when out of domain. This is in contrast to standard encoders which are restricted to encoding into existing latent spaces.

\subsection{Designing the HyperNetwork}~\label{sec:design}
The StyleGAN generator contains approximately $30$M parameters. On one hand, we wish our hypernetworks to be expressive, allowing us to control these parameters for enhancing the reconstruction. 
On the other hand, control over too many parameters would result in an inapplicable network requiring significant resources for training. Therefore, the design of the hypernetwork is challenging, requiring a delicate balance between expressive power and the number of trainable parameters involved.

We denote the weights of the $\ell$-th convolutional layer of StyleGAN by $\theta_\ell = \{\theta^{i,j}_\ell\}_{i,j=0}^{C^{out}_\ell,C^{in}_\ell}$ where $\theta^{i,j}_\ell$ denotes the weights of the $j$-th channel in the $i$-th filter. 
Here, $C^{out}_\ell$ represents the total number of filters, each with $C^{in}_\ell$ channels.
Let $M$ be the total number of layers. The generator weights are then denoted as $\{\theta_\ell\}_{\ell=1}^{M}$. 
Our hypernetwork produces offsets $\Delta_\ell$ for each modified layer $\ell$. These offsets are then multiplied by the corresponding layer weights $\theta_{\ell}$ and added to the original weights in a channel-wise fashion:
\setlength{\abovedisplayskip}{5pt}
\setlength{\belowdisplayskip}{5pt}
\begin{equation}~\label{eq:weight_update}
    \hat{\theta}^{i,j}_{\ell} := \theta^{i,j}_{\ell} \cdot(1 + \Delta^{i,j}_\ell),
\end{equation}
where $\Delta^{i,j}_\ell$ is the scalar applied to the $j$-th channel of the $i$-th filter.
Learning an offset per channel reduces the number of hypernetwork parameters by $88\%$ compared to predicting an offset for each generator parameter (see \cref{tb:parameter_count}). Later experiments verify that this does not harm expressiveness. 

\begin{table}
    \small
    \setlength{\tabcolsep}{2.5pt}
    \centering
    { \small
    \textbf{HyperStyle Trainable Parameters} \\
    
        \begin{tabular}{c | c | c }
        \toprule
        Delta-Per Channel & Shared Refinement & Number of Parameters \\
        \midrule
        & & $3.07$B \\
        & \ding{51} & $1.40$B \\
        \ding{51} & & $367$M \\
        \ding{51} & \ding{51} & $332$M \\
        \bottomrule
        \end{tabular}

        \begin{tabularx}{0.47\textwidth}{@{}c | *2{>{\centering\arraybackslash}X}@{}}
        \toprule
        \quad{}\; pSp~\cite{richardson2020encoding} / e4e~\cite{tov2021designing}: $267$M \quad{}\; & ReStyle~\cite{alaluf2021matter}: $205$M \\
        \bottomrule
        \end{tabularx}

    \vspace{-0.25cm}
    }
    \caption{Our final hypernetwork configuration, consisting of an offset predicted per channel and Shared Refinement blocks reduces the number of parameters by $89\%$ compared to a \naive network design. We compare this to the size of existing encoders.}
    \vspace{-0.35cm}
    \label{tb:parameter_count}
\end{table}

\begin{figure}[tb]
    \centering
    \setlength{\belowcaptionskip}{-10pt}
    \hspace*{-0.4cm}
    \includegraphics[width=1.05\linewidth]{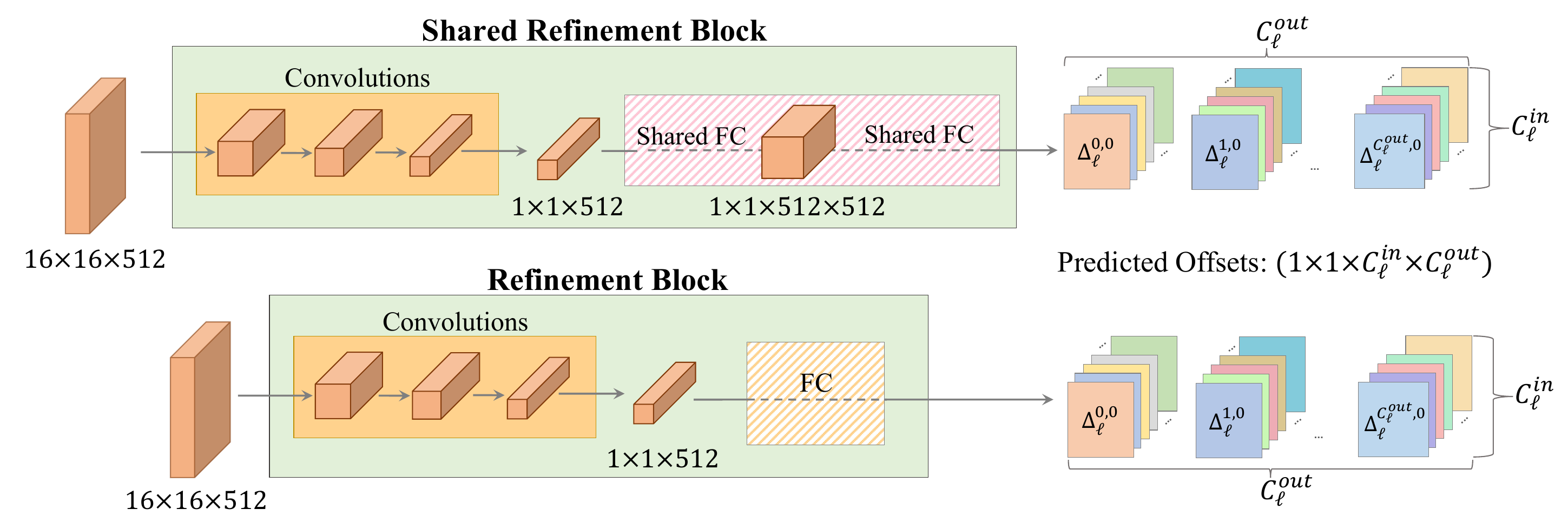}
    \vspace{-0.65cm}
    \caption{The Refinement Block (shown on bottom) consists of a series of down-sampling convolutions and a fully-connected layer resulting in an output of size $1\times1\times C^{in}_\ell \times C^{out}_\ell$. 
    The Shared Refinement Block (top) is introduced to further reduce the network parameters and encourage information sharing between layers.}
    \vspace{-0.1cm}
    \label{fig:refinement_blocks}
\end{figure}

To process the input images, we incorporate a ResNet34 ~\cite{he2015deep} backbone that receives a $6$-channel input $\left( x^i, ~y^i_{init} \right)$ and outputs a $16\times16\times512$ feature map. 
This shared backbone is then followed by a set of \textit{Refinement Blocks}, each producing the modulation of a single generator layer.
Consider layer $\ell$ with parameters $\theta_\ell$ of size $k_\ell \times
k_\ell \times C^{in}_{\ell} \times C^{out}_{\ell}$ where $k_\ell$ is the kernel size. The corresponding Refinement Block receives the feature map extracted by the backbone and outputs an offset of size $1\times1\times C^{in}_{\ell} \times C^{out}_\ell$. The offset is then replicated to match the $k_\ell \times k_\ell$ kernel dimension of $\theta_\ell$. Finally, the new weights of layer $\ell$ are updated using \cref{eq:weight_update}. The Refinement Block is illustrated in \cref{fig:refinement_blocks}. 

To further reduce the number of trainable parameters, we introduce a \textit{Shared Refinement Block}, inspired by the original hypernetwork~\cite{ha2016hypernetworks}. 
These output heads consist of independent convolutional layers used to down-sample the input feature map. They are then followed by two fully-connected layers shared across multiple generator layers, as illustrated in \cref{fig:refinement_blocks}.
Here, the fully-connected weights are shared across the non-toRGB layers with dimension $3\times3\times512\times512$, i.e., the largest generator convolutional blocks.
As demonstrated in Ha~\etal~\cite{ha2016hypernetworks} this allows for information sharing between the output heads, yielding improved reconstruction quality. Detailed layouts of the Refinement Blocks are given in Appendix~\ref{supp:hyperstyle_architecture}.

\newcommand{\textapprox}{\raisebox{0.5ex}{\texttildelow}}

Combining the Shared Refinement Blocks and per-channel predictions, our final configuration contains $2.7$B fewer parameters (\textapprox$89\%$) than a \naive hypernetwork. We summarize the total number of parameters of different hypernetwork variants in \cref{tb:parameter_count}.
We refer the reader to \cref{sec:ablation} where we validate our design choices and explore additional avenues for reducing the number of parameters.

\vspace{-0.4cm}
\paragraph{Which layers are refined?}
The choice of which layers to refine is of great importance.
It allows us to reduce the output dimension while focusing the hypernetwork on the more meaningful generator weights.
Since we invert one identity at a time, any changes to the affine transformation layers can be reproduced by a respective re-scaling of the convolution weights. 
Moreover, we find that altering the toRGB layers harms the editing capabilities of the GAN. We hypothesize that modifying these layers mainly alters the pixel-wise texture and color~\cite{wu2021stylespace},
changes that do not translate well under global edits such as pose (see Appendix~\ref{supp:ablation_study} for examples).
Therefore, we restrict ourselves to modifying only the non-toRGB convolutions. 

Lastly, we follow Karras~\etal~\cite{karras2020analyzing} and split the generator layers into three levels of detail --- coarse, medium, fine --- each controlling different aspects of the generated image.
As the initial inversions tend to capture coarse details,
we further restrict our hypernetwork to output offsets for the medium and fine generator layers.

\vspace{-0.05cm}
\subsection{Iterative Refinement}~\label{sec:iterative_refinement}
To further improve the inversion quality, we adopt the iterative refinement scheme suggested by Alaluf~\etal~\cite{alaluf2021matter}. This enables us to perform several passes through our hypernetwork for a single image inversion. Each added step allows the hypernetwork to gradually refine its predicted weight offsets, resulting in stronger expressive power and a more accurate inversion.

We perform $T$ passes. For the first pass, we use the initial reconstruction $\hat{y}_{0} = G(\hat{w}_{init}; \theta)$.
For each refinement step $t\ge1$, we predict a set of offsets $\Delta_{t} = H(\hat{y}_{t-1},x)$
used to obtain the modified weights $\hat{\theta}_{t}$ and updated reconstruction $\hat{y}_{t} = G(\hat{w}_{init}; \hat{\theta}_{t})$.
The weights at step $t$ are defined as the accumulated modulation across all previous steps:
\setlength{\abovedisplayskip}{3.5pt}
\setlength{\belowdisplayskip}{4pt}
\begin{equation}
    \hat{\theta}_{\ell,t} := \theta \cdot(1 + \sum_{i=1}^{t}\Delta_{\ell,i}).
\end{equation}
The number of refinement steps is set to $T=5$ during training.
Following Alaluf~\etal~\cite{alaluf2021restyle} we compute the losses at each refinement step. 
Note, $\hat{w}_{init}$ remains fixed during the iterative process.
The final inversion $\hat{y}$ is the reconstruction obtained at the last step.

\subsection{Training Losses}
Similar to encoder-based methods, our training is guided by an image-space reconstruction objective. 
We apply a weighted combination of the pixel-wise $L_2$ loss and LPIPS perceptual loss \cite{zhang2018perceptual}. For the facial domain, we further apply an identity-based similarity loss~\cite{richardson2020encoding} by employing a pre-trained facial recognition network~\cite{deng2019arcface} to preserve the facial identity. As suggested by Tov \etal~\cite{tov2021designing}, we apply a MoCo-based similarity loss for non-facial domains. The final loss objective is given by: 
\begin{equation}~\label{eq:rec_loss}
      \mathcal{L}_{2}(x, \hat{y}) + \lambda_{\text{LPIPS}}\mathcal{L}_{\text{LPIPS}}(x, \hat{y}) + \lambda_{sim}\mathcal{L}_{sim}(x, \hat{y}).
\end{equation}

\section{Experiments}

\begin{figure*}
    \centering
    \setlength{\belowcaptionskip}{-6pt}
    \setlength{\tabcolsep}{1.5pt}
    {\small
    \begin{tabular}{c | c c | c c c c c}

        \vspace{-0.05cm}

        \includegraphics[width=0.115\textwidth]{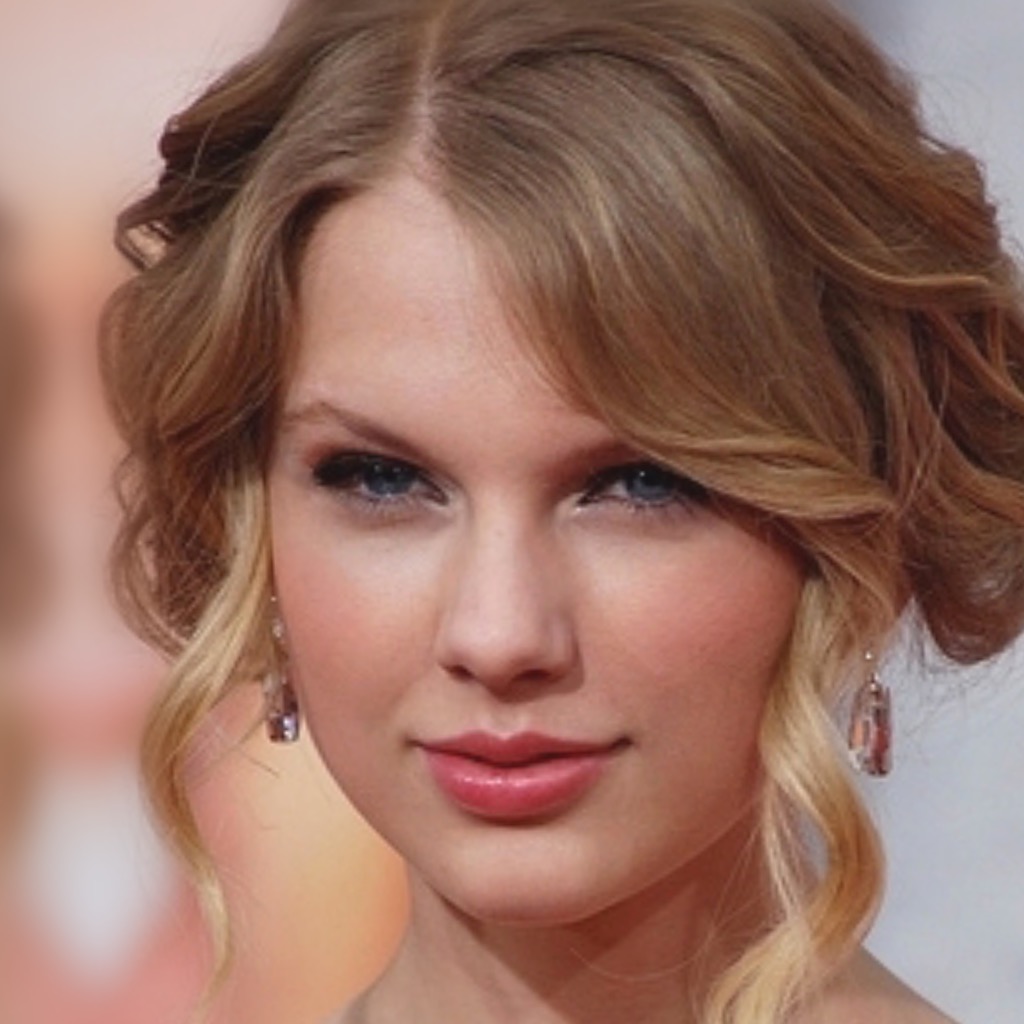} &
        \includegraphics[width=0.115\textwidth]{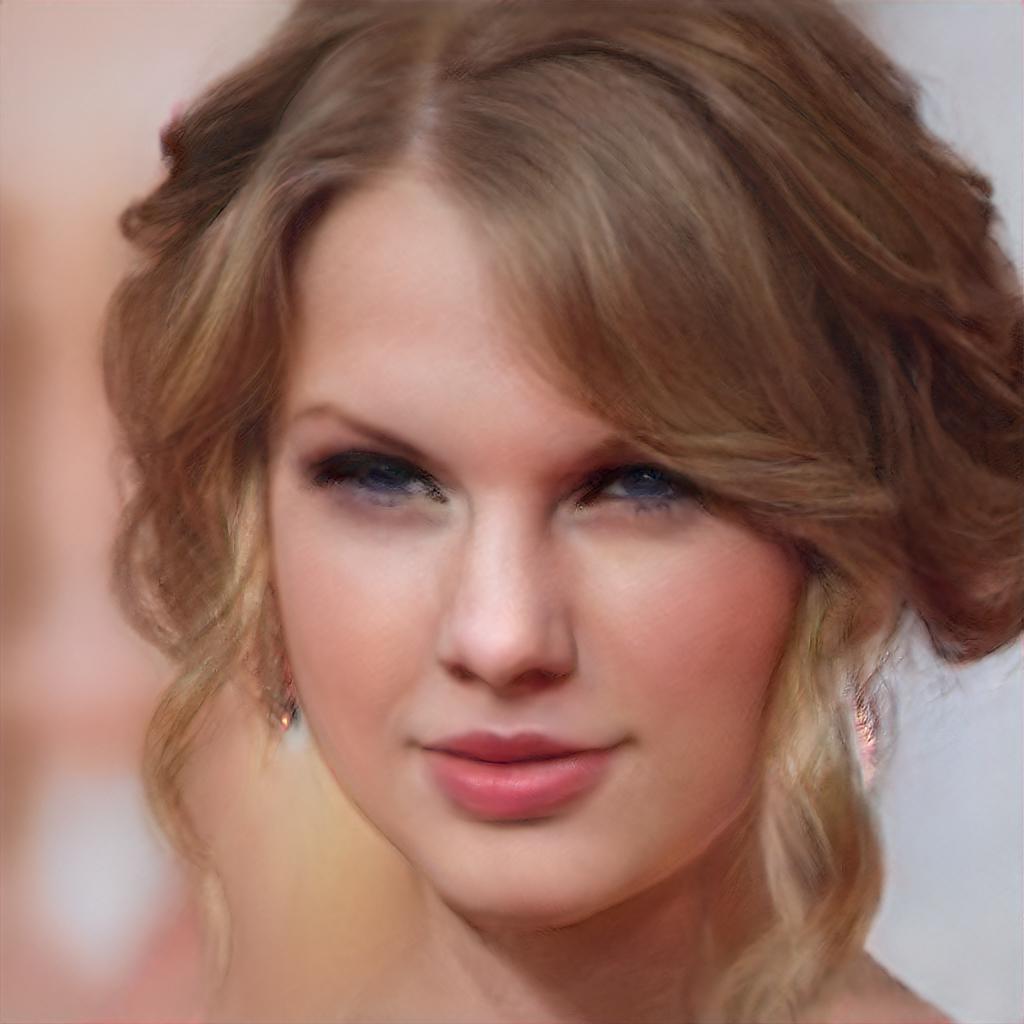} &
        \includegraphics[width=0.115\textwidth]{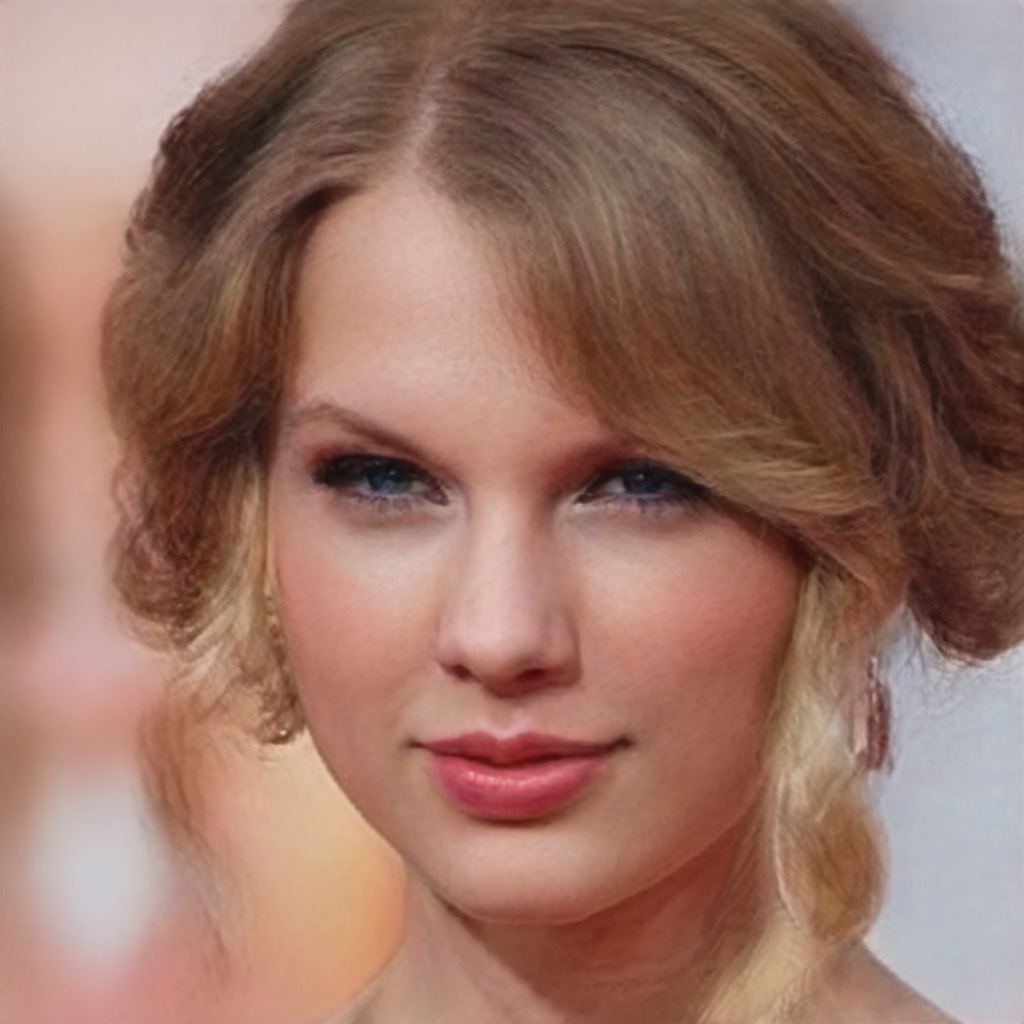} &
        \includegraphics[width=0.115\textwidth]{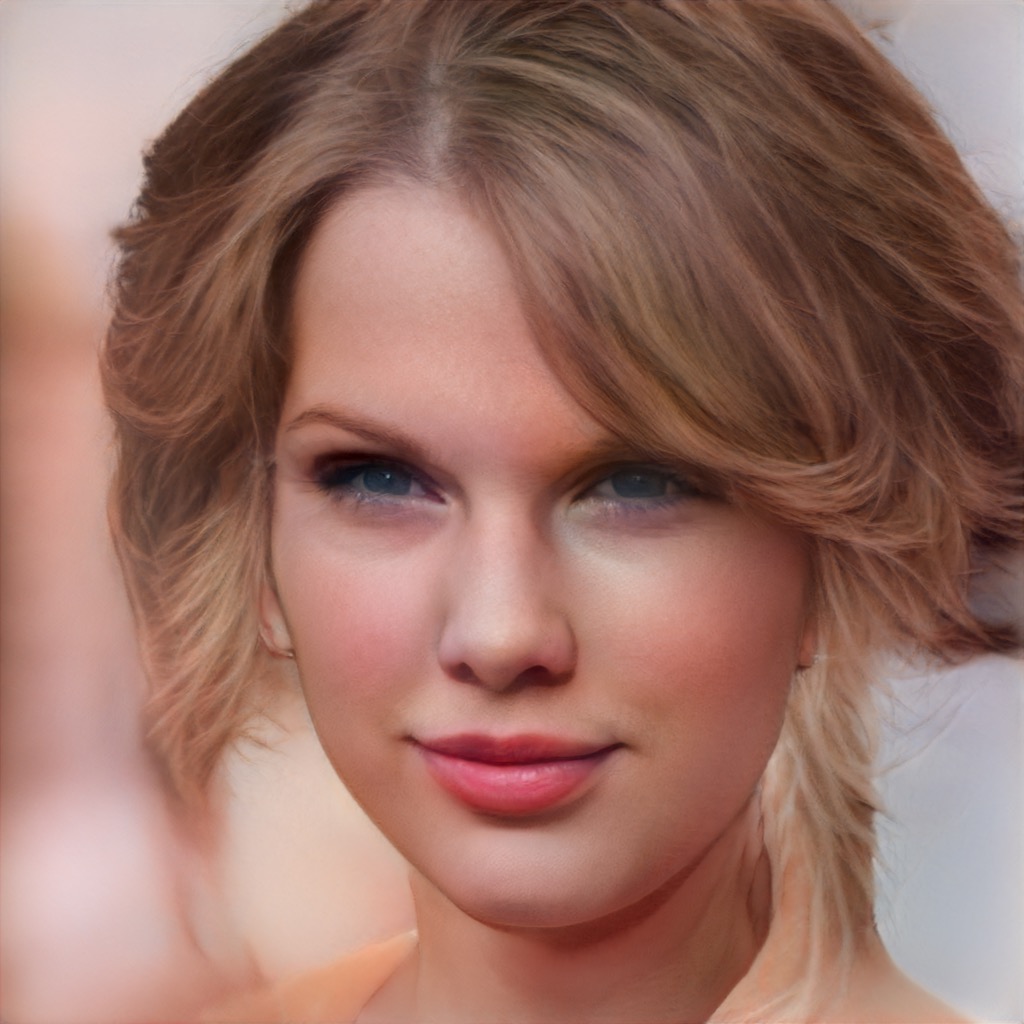} &
        \includegraphics[width=0.115\textwidth]{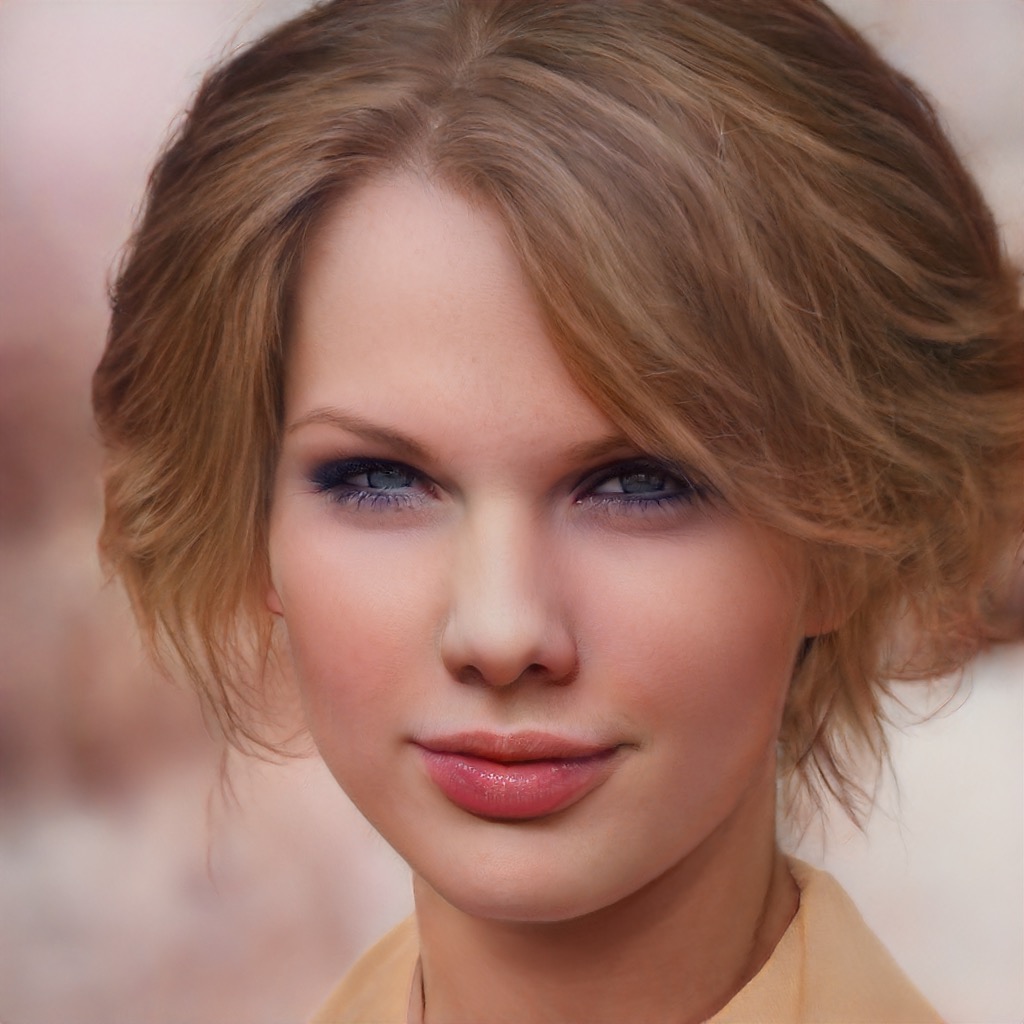} &
        \includegraphics[width=0.115\textwidth]{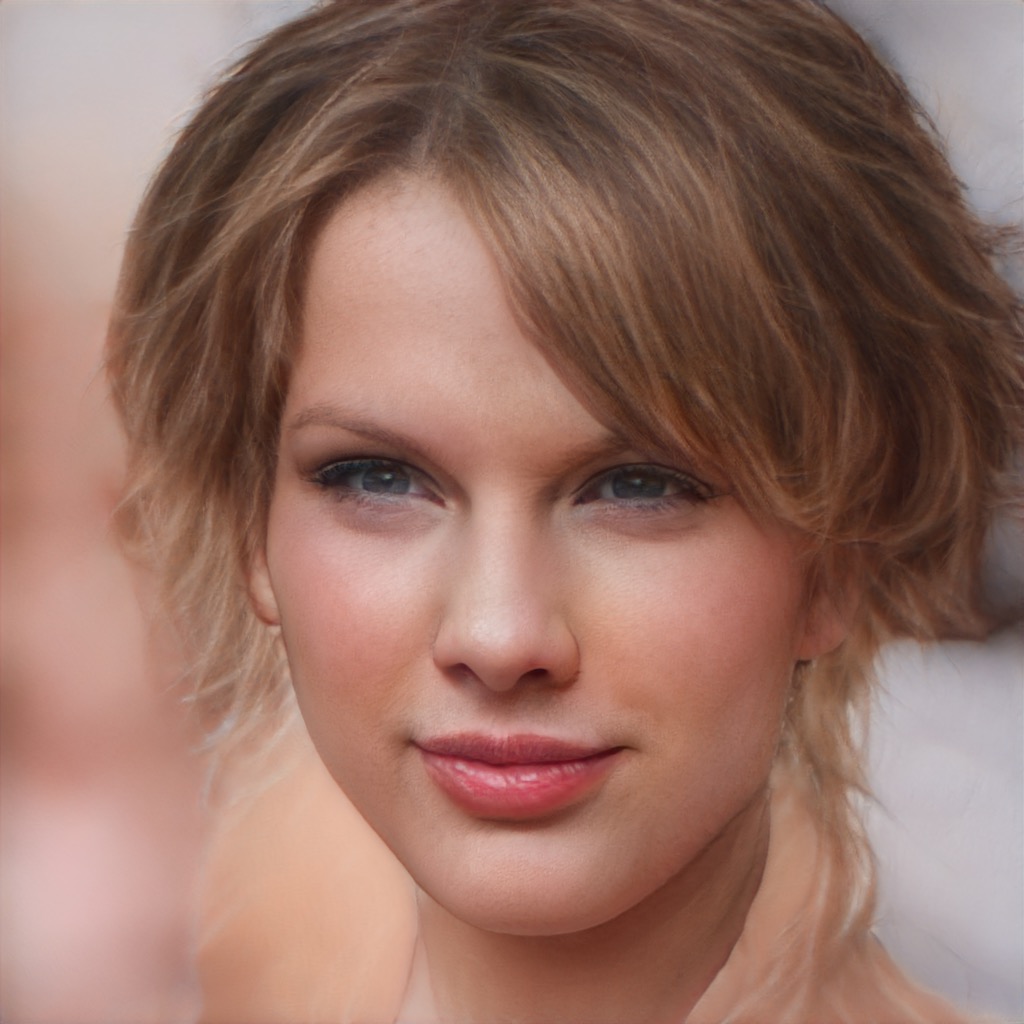} &
        \includegraphics[width=0.115\textwidth]{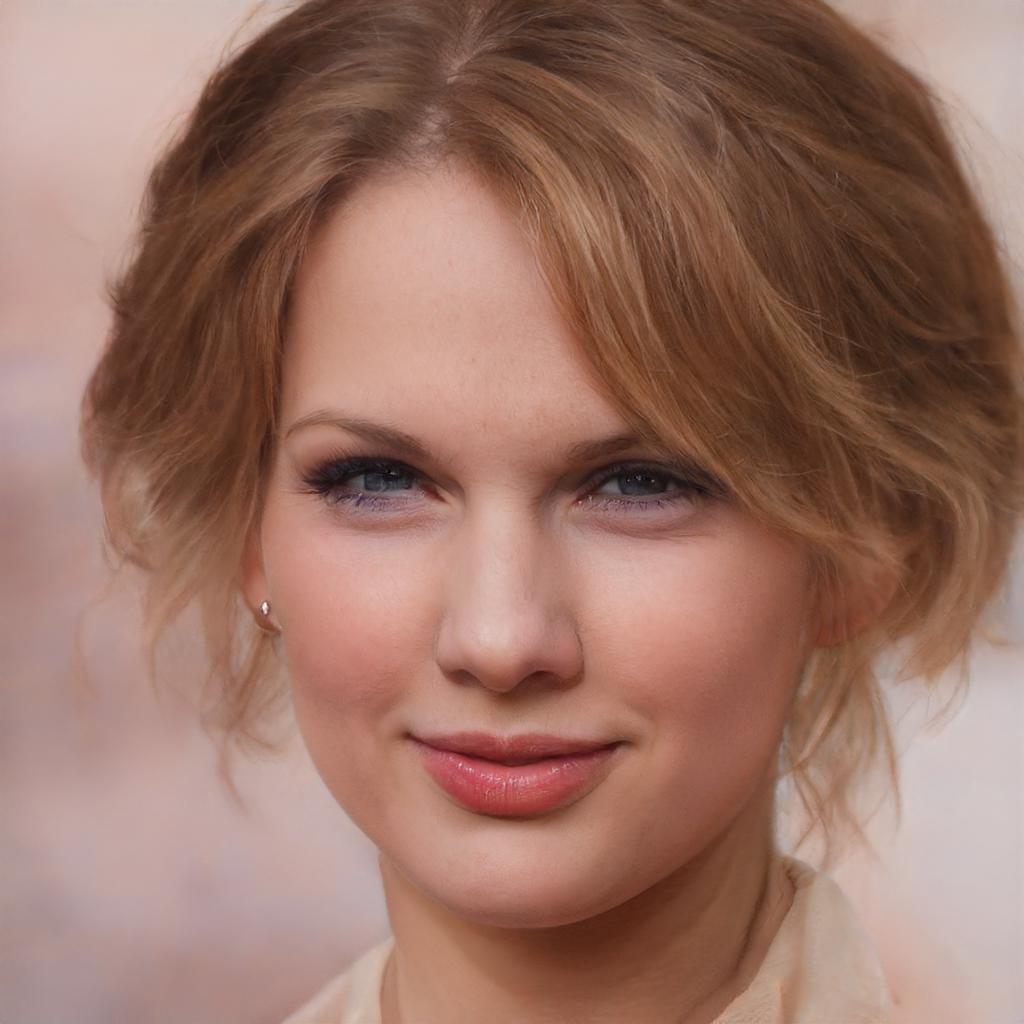} &
        \includegraphics[width=0.115\textwidth]{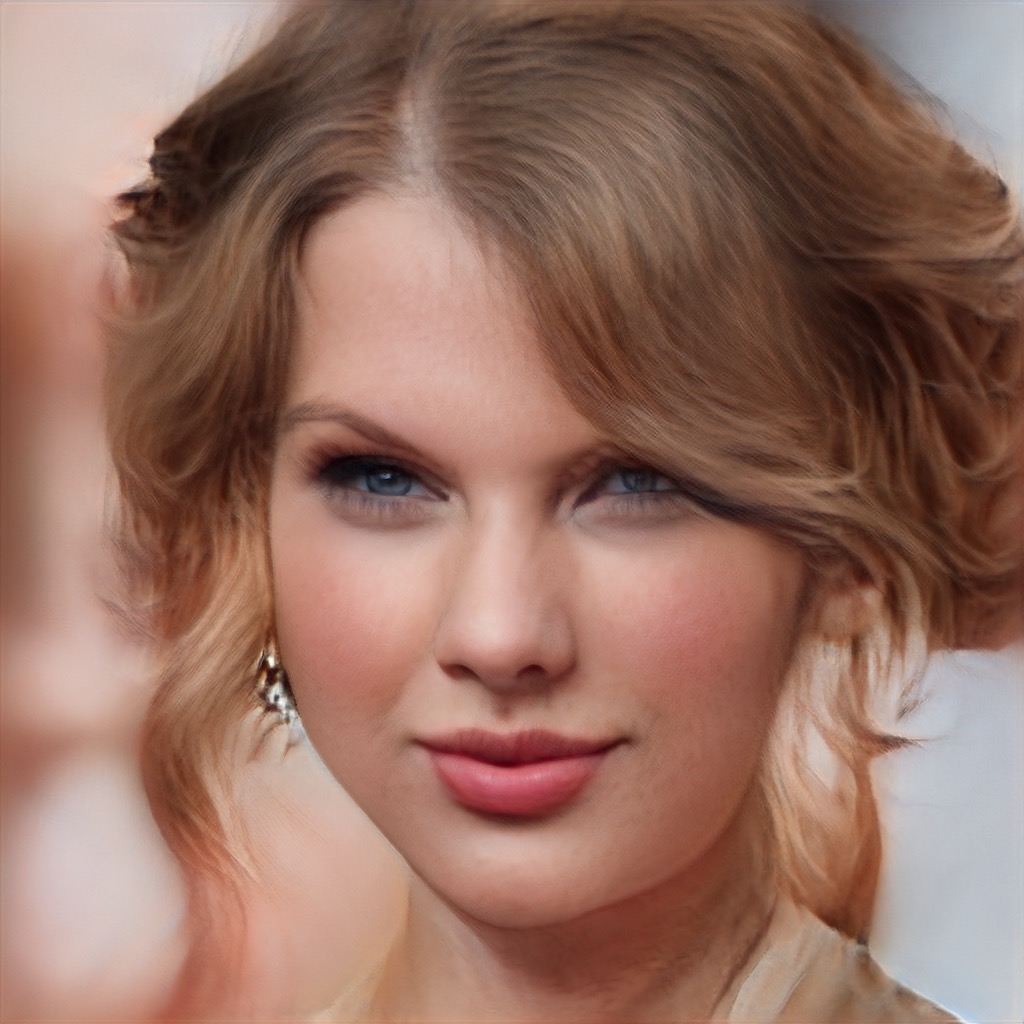} \\

        \vspace{-0.05cm}

        \includegraphics[width=0.115\textwidth]{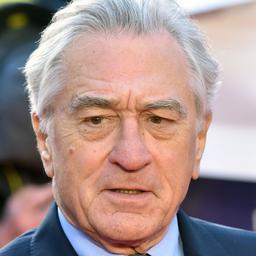} &
        \includegraphics[width=0.115\textwidth]{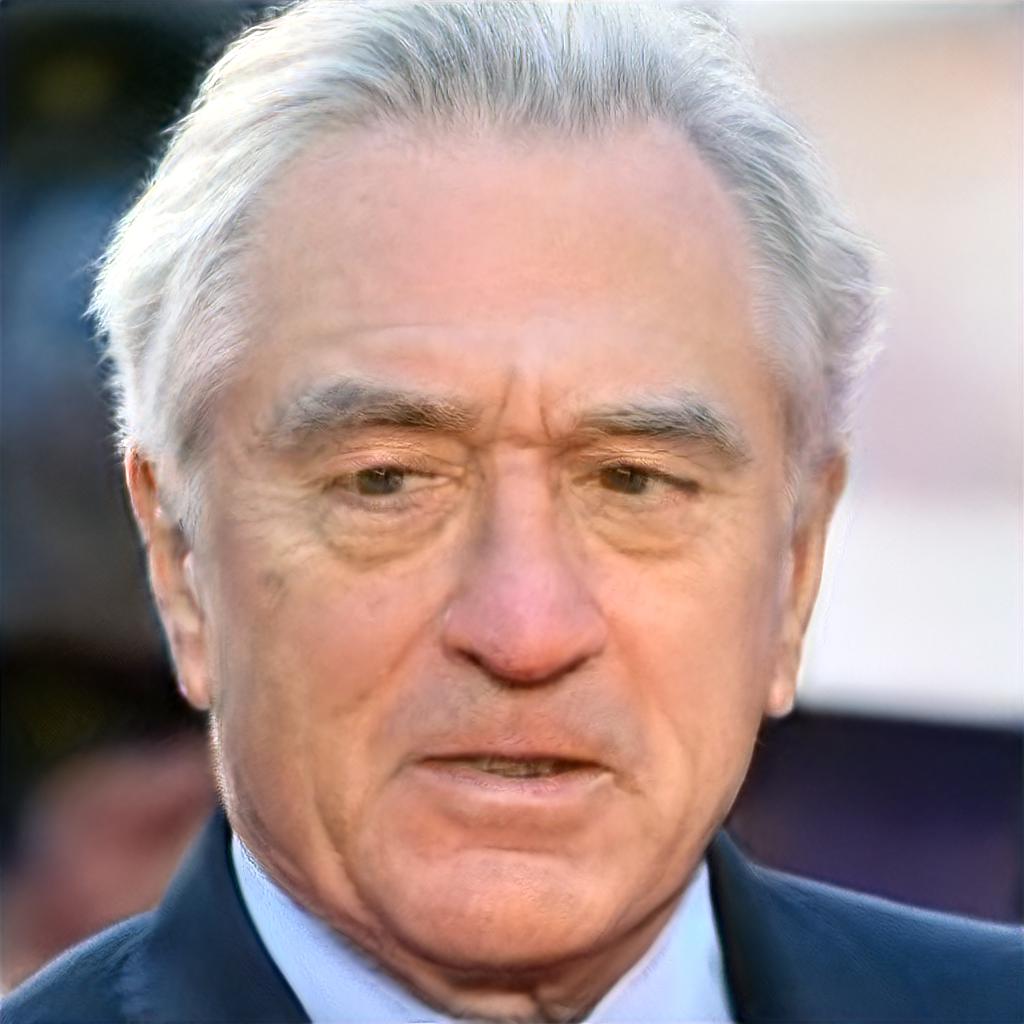} &
        \includegraphics[width=0.115\textwidth]{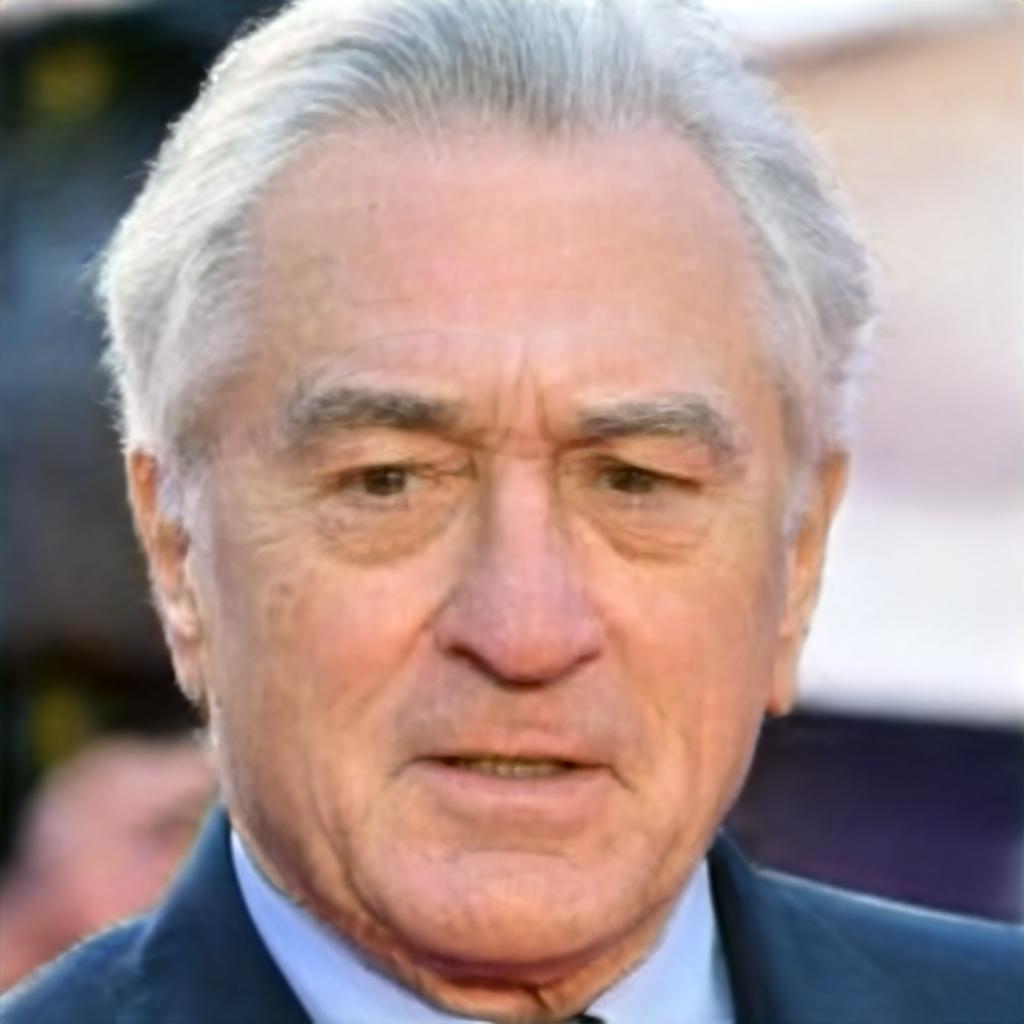} &
        \includegraphics[width=0.115\textwidth]{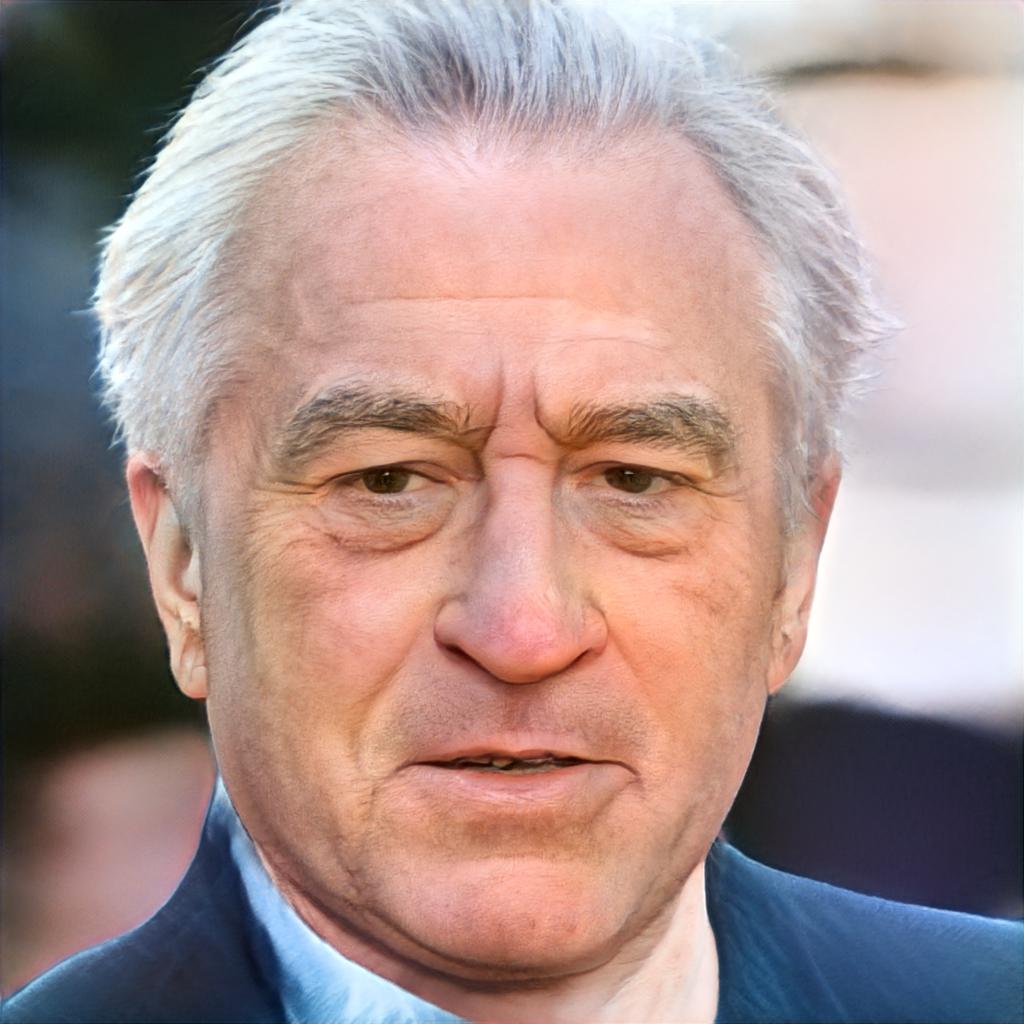} &
        \includegraphics[width=0.115\textwidth]{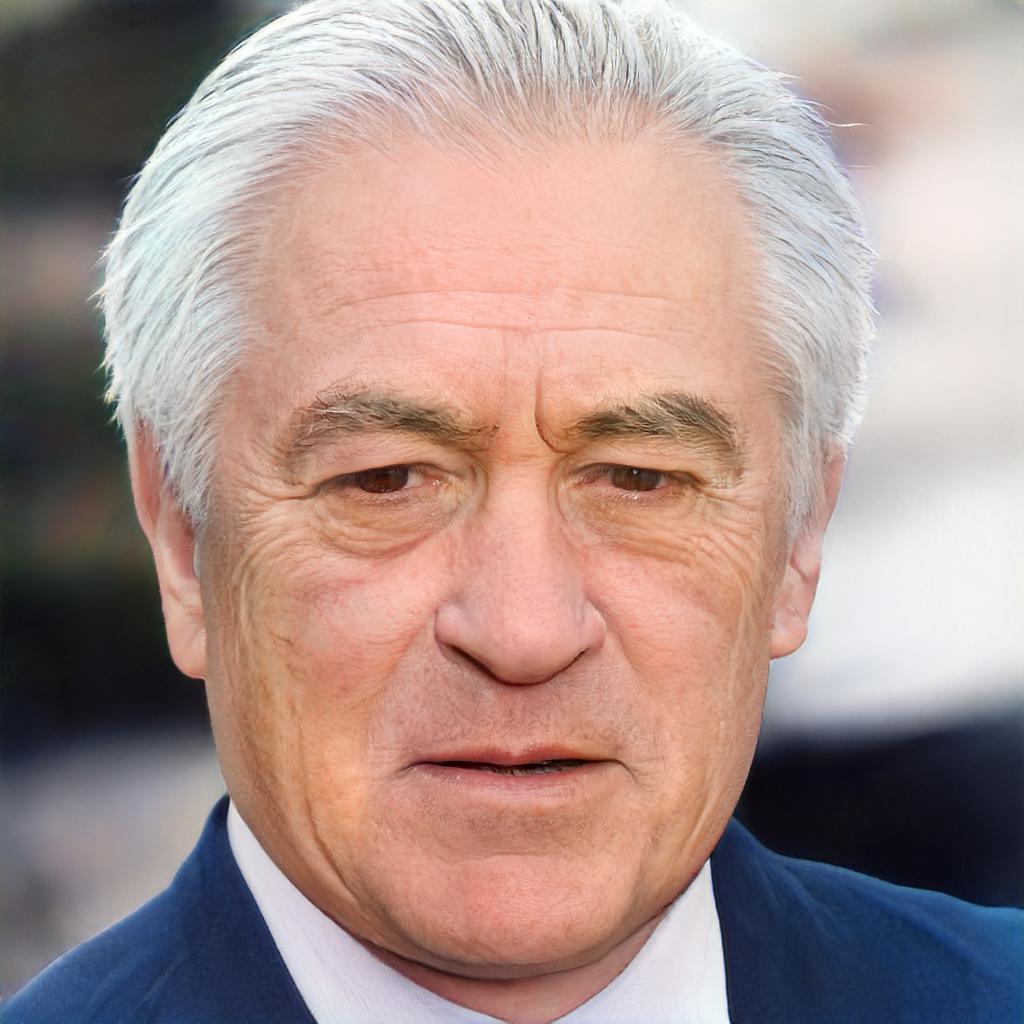} &
        \includegraphics[width=0.115\textwidth]{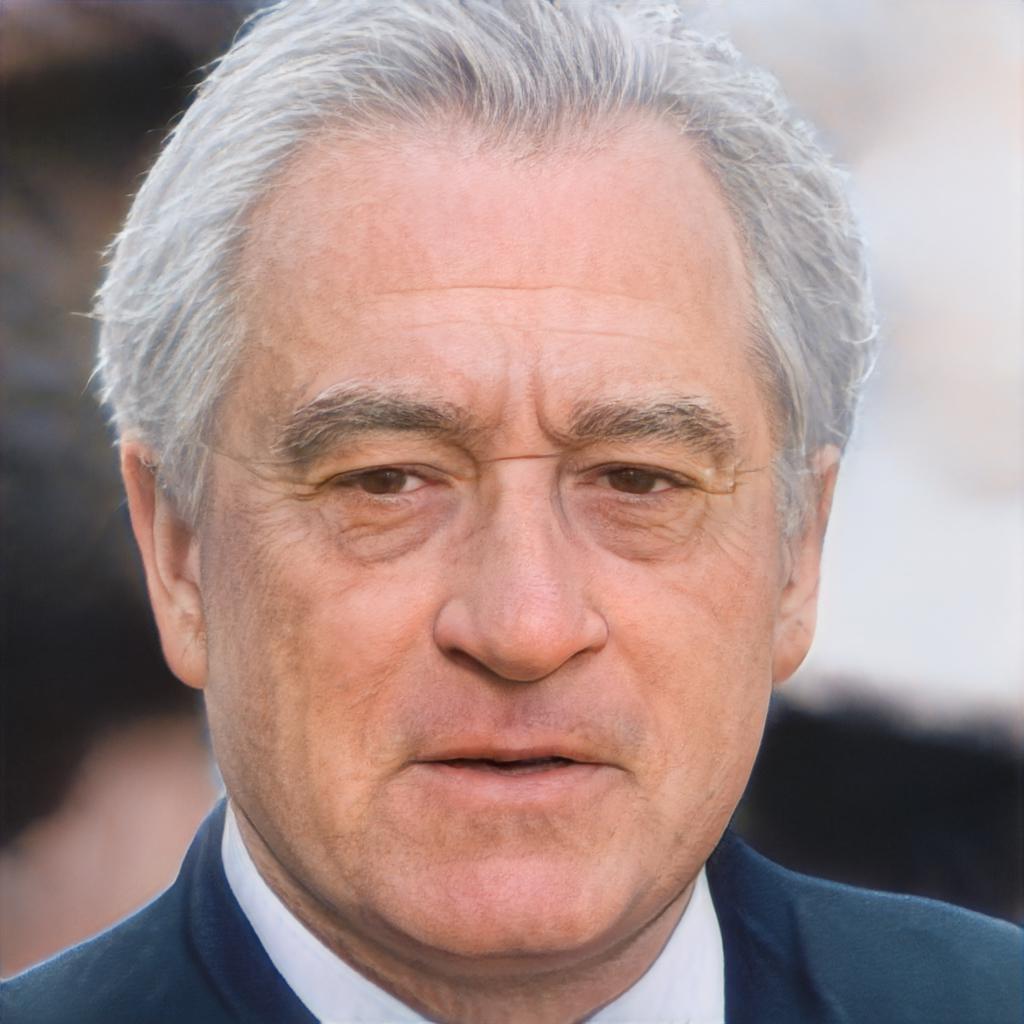} &
        \includegraphics[width=0.115\textwidth]{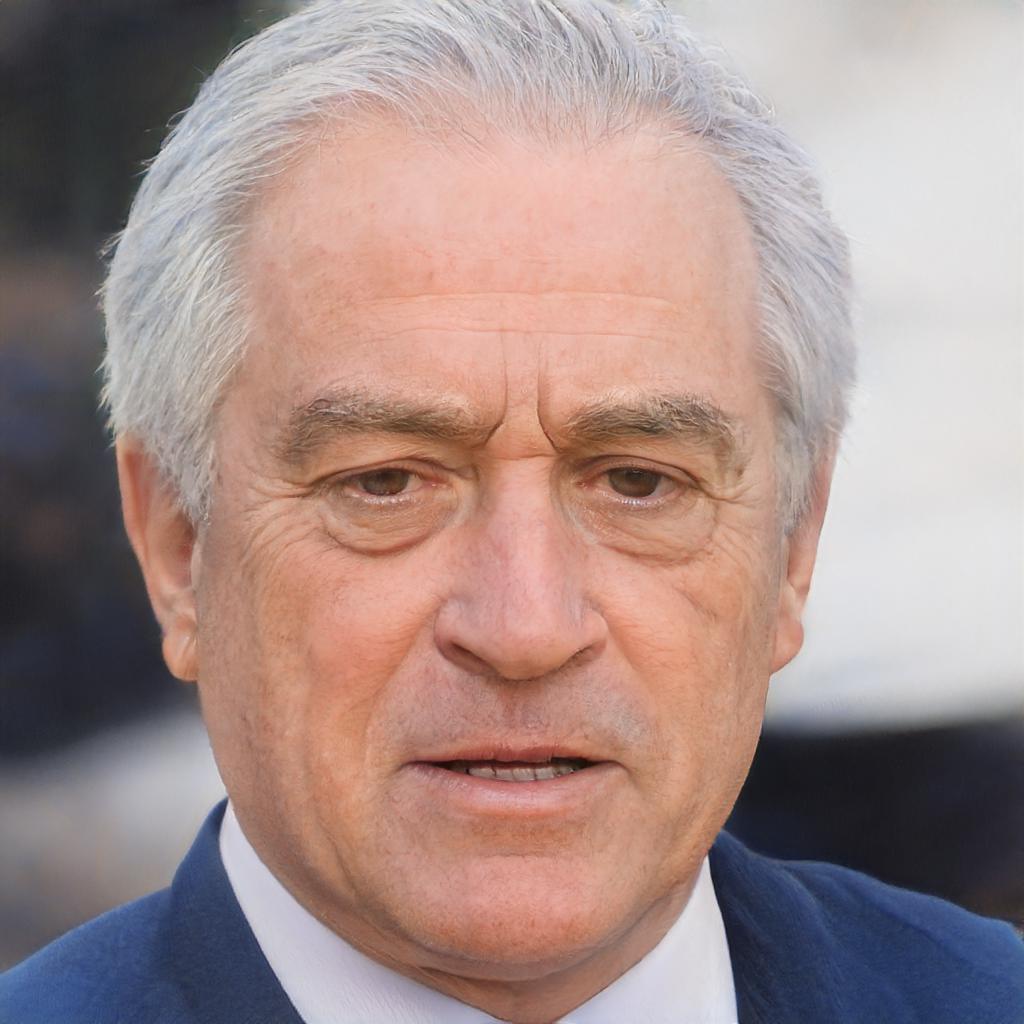} &
        \includegraphics[width=0.115\textwidth]{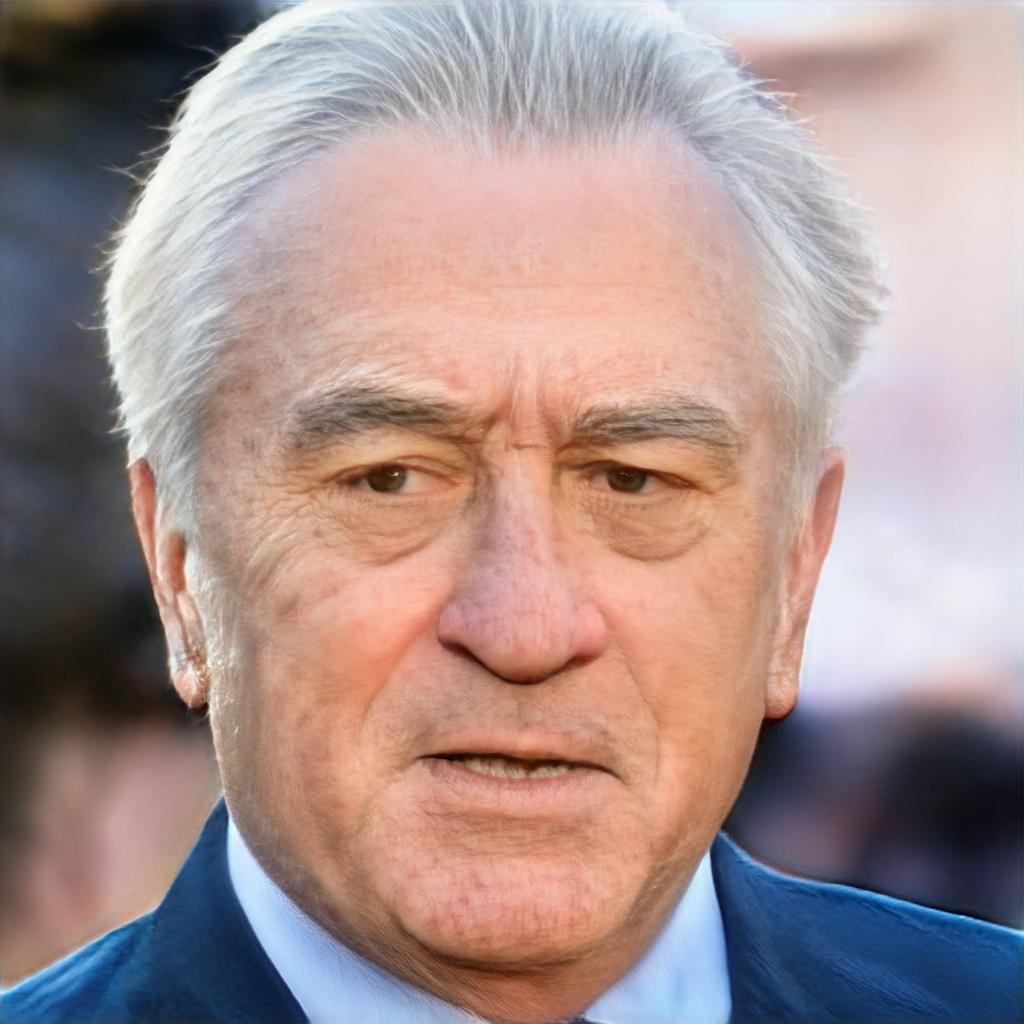} \\
        
        \vspace{-0.05cm}
        
        \includegraphics[width=0.115\textwidth]{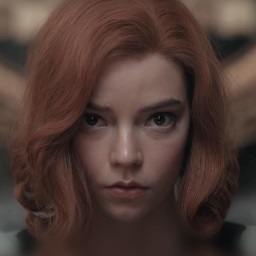} &
        \includegraphics[width=0.115\textwidth]{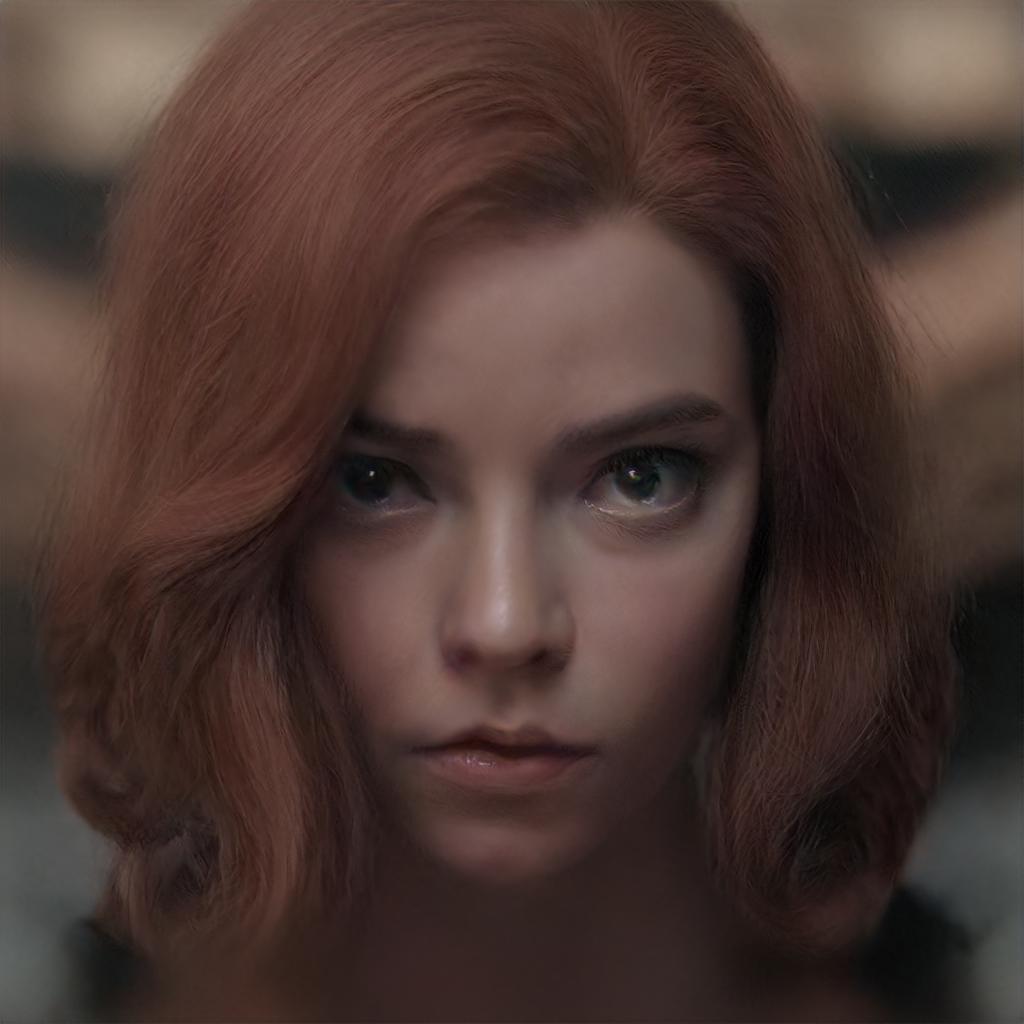} &
        \includegraphics[width=0.115\textwidth]{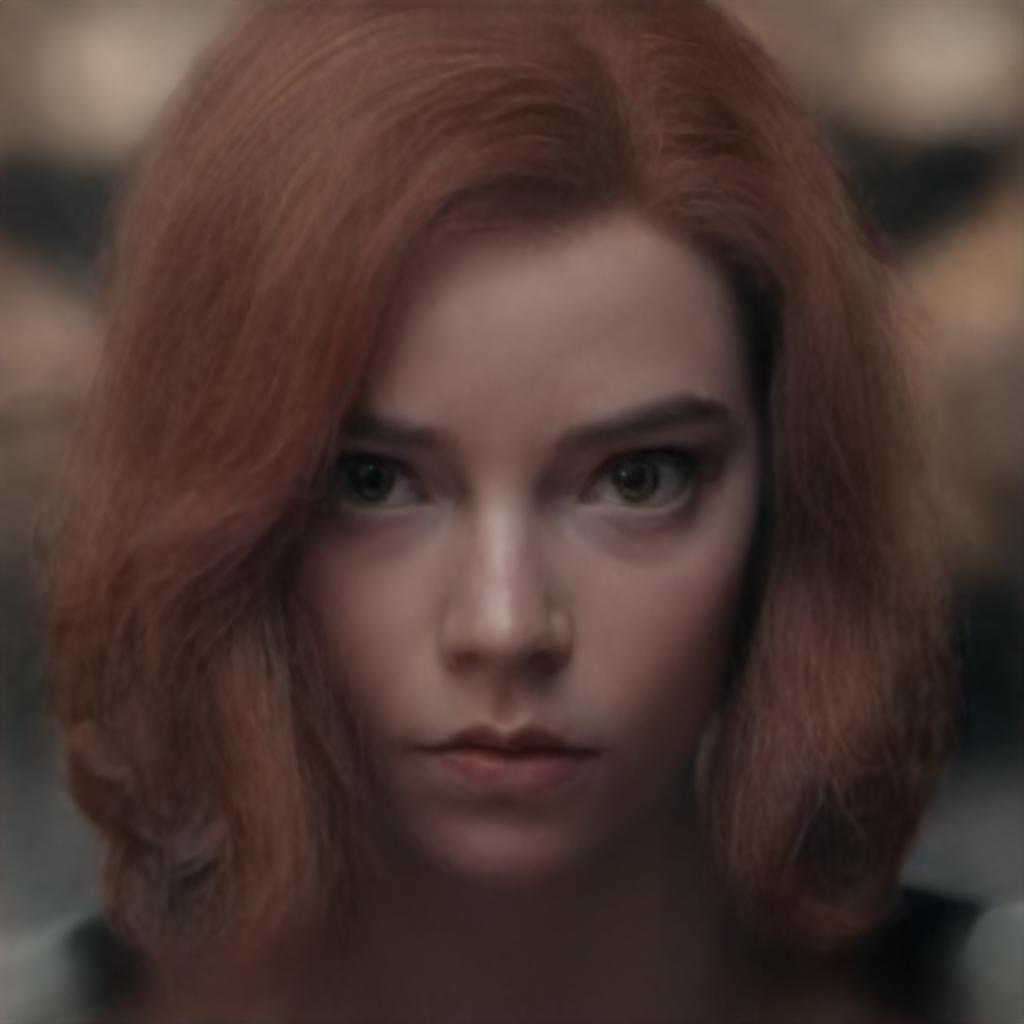} &
        \includegraphics[width=0.115\textwidth]{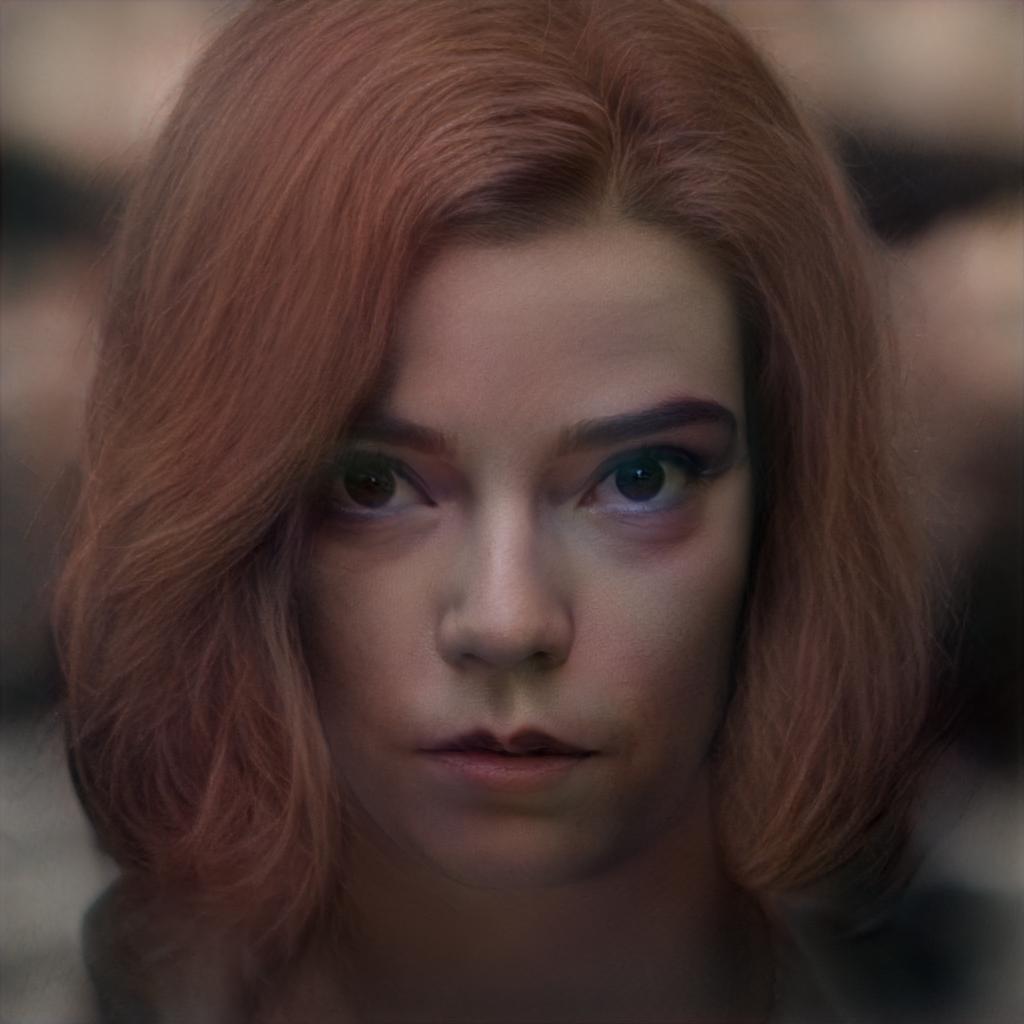} &
        \includegraphics[width=0.115\textwidth]{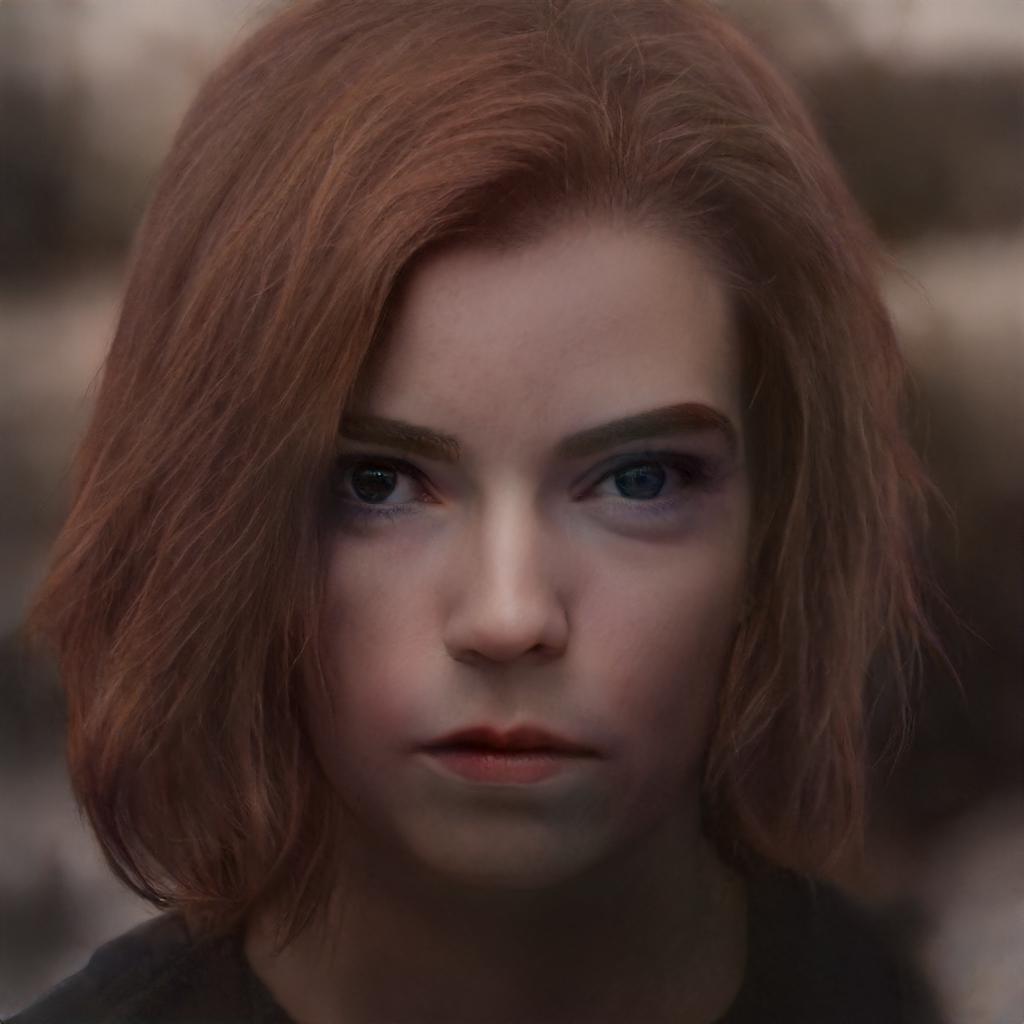} &
        \includegraphics[width=0.115\textwidth]{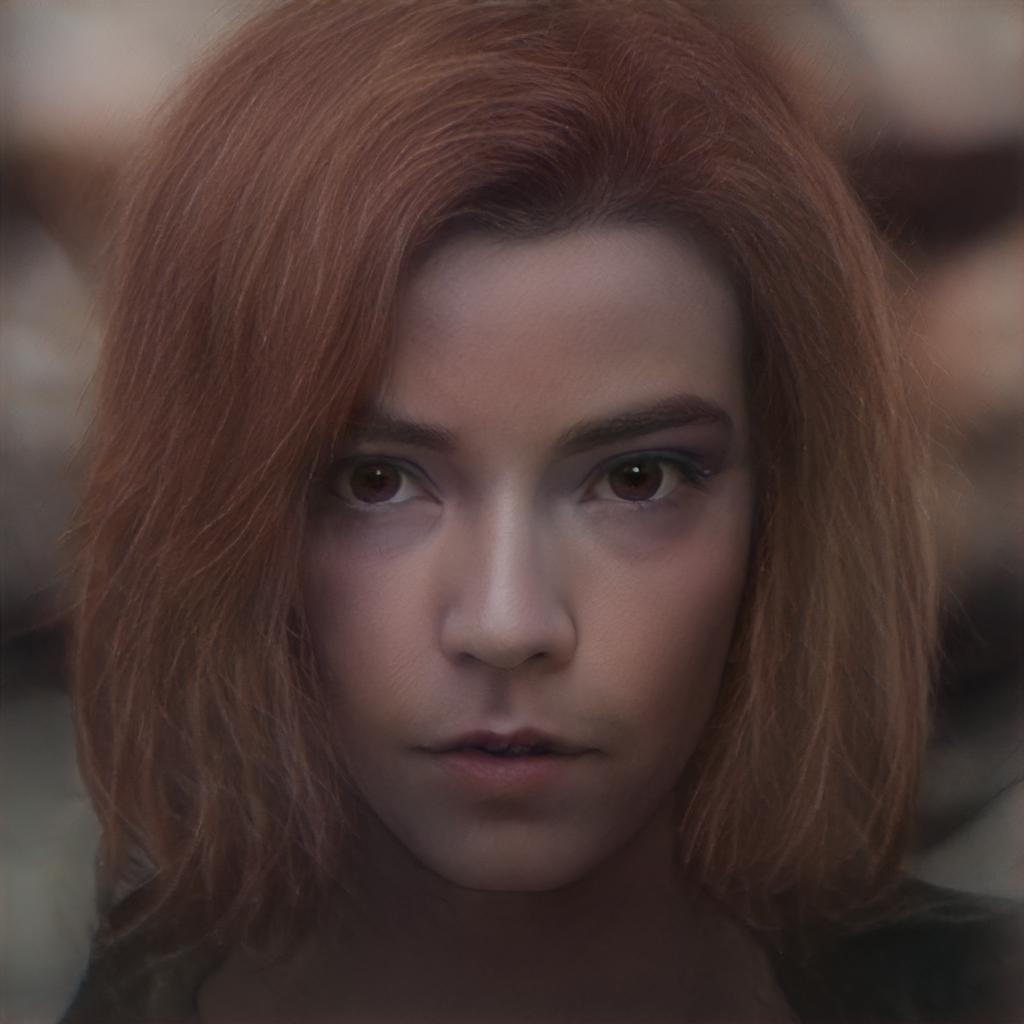} &
        \includegraphics[width=0.115\textwidth]{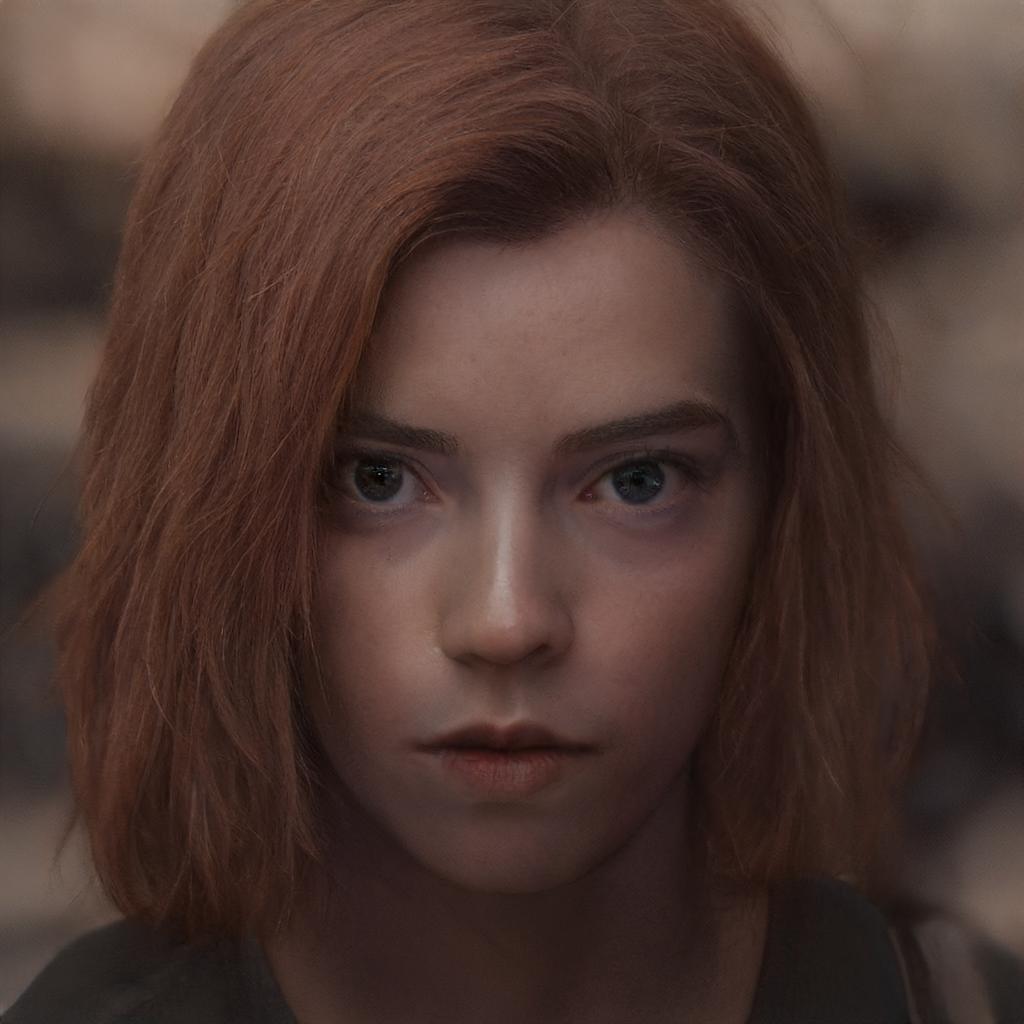} &
        \includegraphics[width=0.115\textwidth]{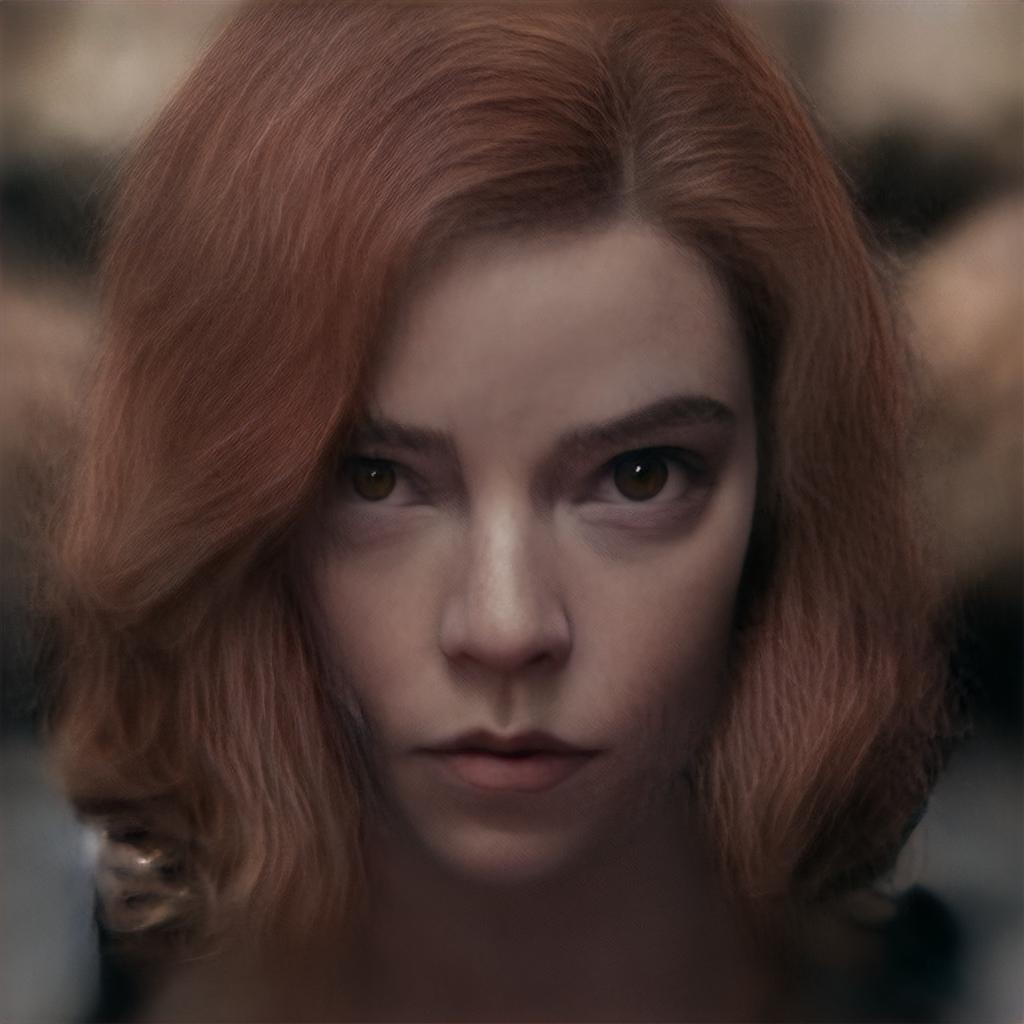} \\

        Input & Optimization & PTI & $\text{ReStyle}_{pSp}$ &  $\text{ReStyle}_{e4e}$ & pSp & e4e & HyperStyle
        
    \end{tabular}
    
    }
    \vspace{-0.2cm}
    \caption{Reconstruction quality comparison. 
    HyperStyle achieves comparable results to optimization techniques while requiring a fraction of the inference time. Compared to encoders, our method attains superior results in terms of the preservation of identity and finer details such as hairstyles and clothing. Additional results are presented in Appendix~\ref{supp:comparisons}.
    Best viewed zoomed-in.}
    \vspace{-0.25cm}
    \label{fig:inversion_comparison}
\end{figure*}
\begin{table}
    \small
    \setlength{\tabcolsep}{2.65pt}
    \centering
    { \small
    \begin{tabular}{l@{\,\, }| c c c c c}
        \toprule
        Method & $\uparrow$ ID & $\uparrow$ MS-SSIM & $\downarrow$ LPIPS & $\downarrow$ $L_2$ & $\downarrow$ Time (s) \\
        \midrule
        StyleGAN2~\cite{karras2020analyzing} &
        \multicolumn{1}{c}{$0.78$} &
        \multicolumn{1}{c}{$0.90$} &
        \multicolumn{1}{c}{$0.09$} &
        \multicolumn{1}{c}{$0.020$} &
        \multicolumn{1}{c}{$227.55$} \\ 
        PTI~\cite{roich2021pivotal} &
        \multicolumn{1}{c}{$0.85$} &
        \multicolumn{1}{c}{$0.92$} &
        \multicolumn{1}{c}{$0.09$} &
        \multicolumn{1}{c}{$0.015$} &
        \multicolumn{1}{c}{$55.715$} \\ 
        \midrule
        IDInvert~\cite{zhu2020domain} &
        \multicolumn{1}{c}{$0.18$} &
        \multicolumn{1}{c}{$0.68$} &
        \multicolumn{1}{c}{$0.22$} &
        \multicolumn{1}{c}{$0.061$} &
        \multicolumn{1}{c}{$0.04$} \\
        pSp~\cite{richardson2020encoding} &
        \multicolumn{1}{c}{$0.56$} &
        \multicolumn{1}{c}{$0.76$} &
        \multicolumn{1}{c}{$0.17$} &
        \multicolumn{1}{c}{$0.034$} &
        \multicolumn{1}{c}{$0.106$} \\
        e4e~\cite{tov2021designing} & 
        \multicolumn{1}{c}{$0.50$} &
        \multicolumn{1}{c}{$0.72$} &
        \multicolumn{1}{c}{$0.20$} &
        \multicolumn{1}{c}{$0.052$} &
        \multicolumn{1}{c}{$0.106$} \\
        $\text{ReStyle}_{pSp}$~\cite{alaluf2021restyle} & 
        \multicolumn{1}{c}{$0.66$} & 
        \multicolumn{1}{c}{$0.79$} &
        \multicolumn{1}{c}{$0.13$} & 
        \multicolumn{1}{c}{$0.030$} &
        \multicolumn{1}{c}{$0.366$} \\
        $\text{ReStyle}_{e4e}$~\cite{alaluf2021restyle} & 
        \multicolumn{1}{c}{$0.52$} & 
        \multicolumn{1}{c}{$0.74$} &
        \multicolumn{1}{c}{$0.19$} & 
        \multicolumn{1}{c}{$0.041$} &
        \multicolumn{1}{c}{$0.366$} \\
        \midrule
        HyperStyle & 
        \multicolumn{1}{c}{$0.76$} &
        \multicolumn{1}{c}{$0.84$} &
        \multicolumn{1}{c}{$0.09$} & 
        \multicolumn{1}{c}{$0.019$} &
        \multicolumn{1}{c}{$1.234$} \\
        \bottomrule
    \end{tabular}
    }
    \vspace{-0.3cm}
    \caption{Quantitative reconstruction results on the human facial domain measured over the CelebA-HQ~\cite{karras2017progressive,liu2015deep} test set.} 
    \label{tb:quantitative_inversion}
    \vspace{-0.45cm}
\end{table}

\vspace{-0.15cm}
\paragraph{Datasets and Baselines}
For the human facial domain we use FFHQ~\cite{karras2019style} for training and the CelebA-HQ test set~\cite{karras2017progressive,liu2015deep} for quantitative evaluations. On the cars domain, we use the Stanford Cars dataset~\cite{KrauseStarkDengFei-Fei_3DRR2013}. Additional results on AFHQ Wild~\cite{choi2020stargan} are provided in Appendix~\ref{supp:comparisons}.
We compare our results to the state-of-the-art encoders pSp~\cite{richardson2020encoding}, e4e~\cite{tov2021designing}, and ReStyle~\cite{alaluf2021restyle} applied over both pSp and e4e. A visual comparison with IDInvert~\cite{zhu2020domain} is provided in Appendix~\ref{supp:comparisons}. For a comparison with optimization techniques, we compare to PTI~\cite{roich2021pivotal} and the latent vector optimization into $\mathcal{W}+$ from Karras~\etal~\cite{karras2020analyzing}.

\subsection{Reconstruction Quality}

\paragraph{Qualitative Evaluation}
We begin with a qualitative comparison, provided in \cref{fig:inversion_comparison}. 
While optimization techniques are typically able to achieve accurate reconstructions, they come with a high computational cost. HyperStyle offers visually comparable results with an inference time several orders of magnitude faster. 
Furthermore, PTI may struggle when inverting a low-resolution input (2nd row), yielding a blurred reconstruction due to its inherent design of over-fitting to the target image. 
Our hypernetwork, meanwhile, is trained on a large image collection and is therefore less likely to re-create such resolution-based artifacts.
In addition, compared to single-shot encoders (pSp and e4e), HyperStyle better captures the input identity (3rd row). When compared to the more recent ReStyle~\cite{alaluf2021restyle} encoders, HyperStyle is still able to better reconstruct finer details such as complex hairstyles (1st row) and clothing (2nd row).

\vspace{-0.3cm}
\paragraph{Quantitative Evaluation}
In \cref{tb:quantitative_inversion}, we present a quantitative evaluation focusing on the time-accuracy trade-off.
Along with the inference time of each method, we report the pixel-wise $L_2$ distance, the LPIPS~\cite{zhang2018perceptual} distance, and the MS-SSIM~\cite{wang2003multiscale} score between each reconstruction and source. We additionally measure identity similarity using a pre-trained facial recognition network~\cite{huang2020curricularface}.
For HyperStyle and ReStyle~\cite{alaluf2021restyle}, we performed multiple iterative steps until the metric scores stopped improving or until $10$ iterations were reached. For optimization, we use at most $1,500$ steps, while for PTI we perform at most $350$ pivotal tuning steps.

As presented in \cref{tb:quantitative_inversion}, HyperStyle's performance consistently surpasses that of the encoder-based methods. Surprisingly, it even achieves results on par with the StyleGAN2 optimization~\cite{karras2020analyzing}, while being nearly $200$ times faster. Overall, HyperStyle demonstrates optimization-level reconstructions achieved with encoder-like inference times.

\begin{figure*}
    \centering
    \setlength{\belowcaptionskip}{-6pt}
    \setlength{\tabcolsep}{1.5pt}
    {\small
    \begin{tabular}{c c | c c | c c c c c}
        
        \vspace{-0.0615cm}
        
        \raisebox{0.225in}{\rotatebox{90}{$-$Age}} &
        \includegraphics[width=0.1125\textwidth]{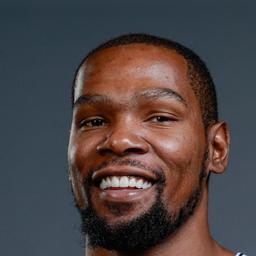} &
        \includegraphics[width=0.1125\textwidth]{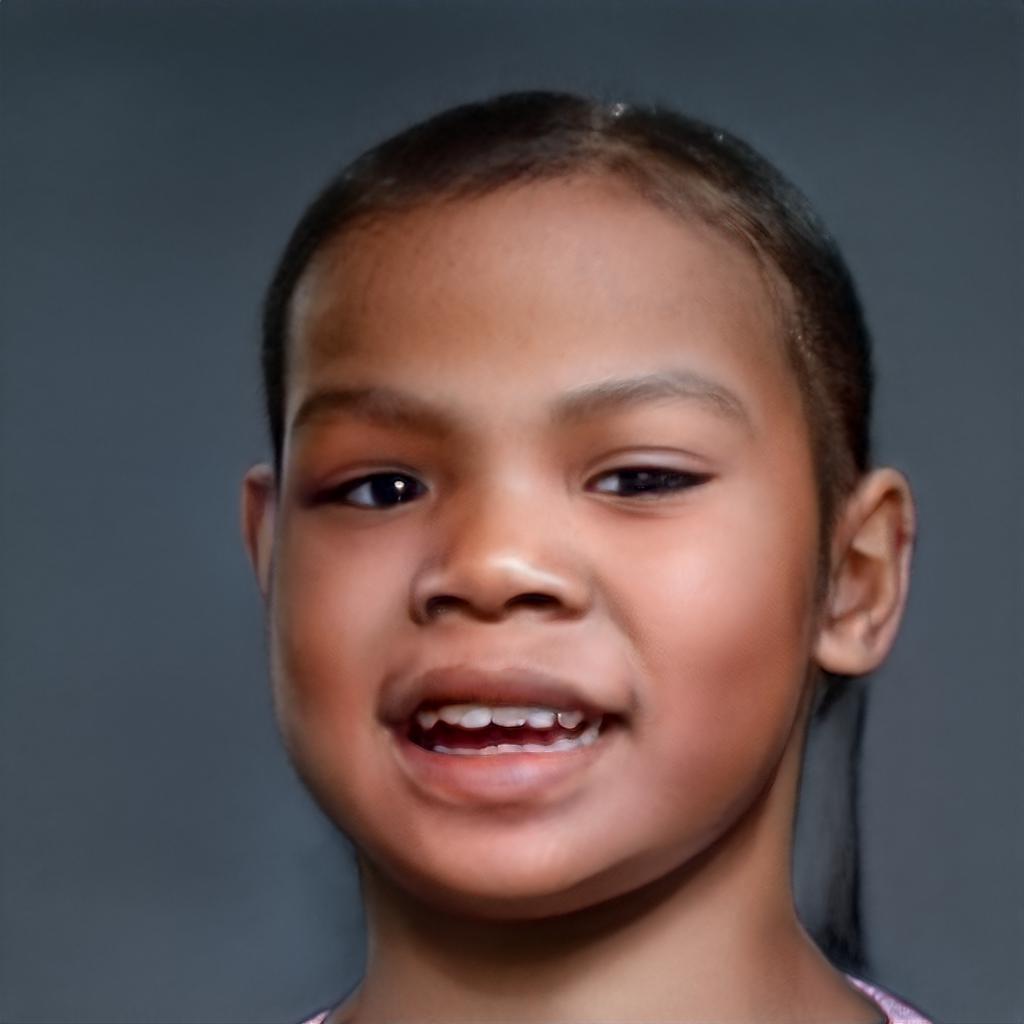} &
        \includegraphics[width=0.1125\textwidth]{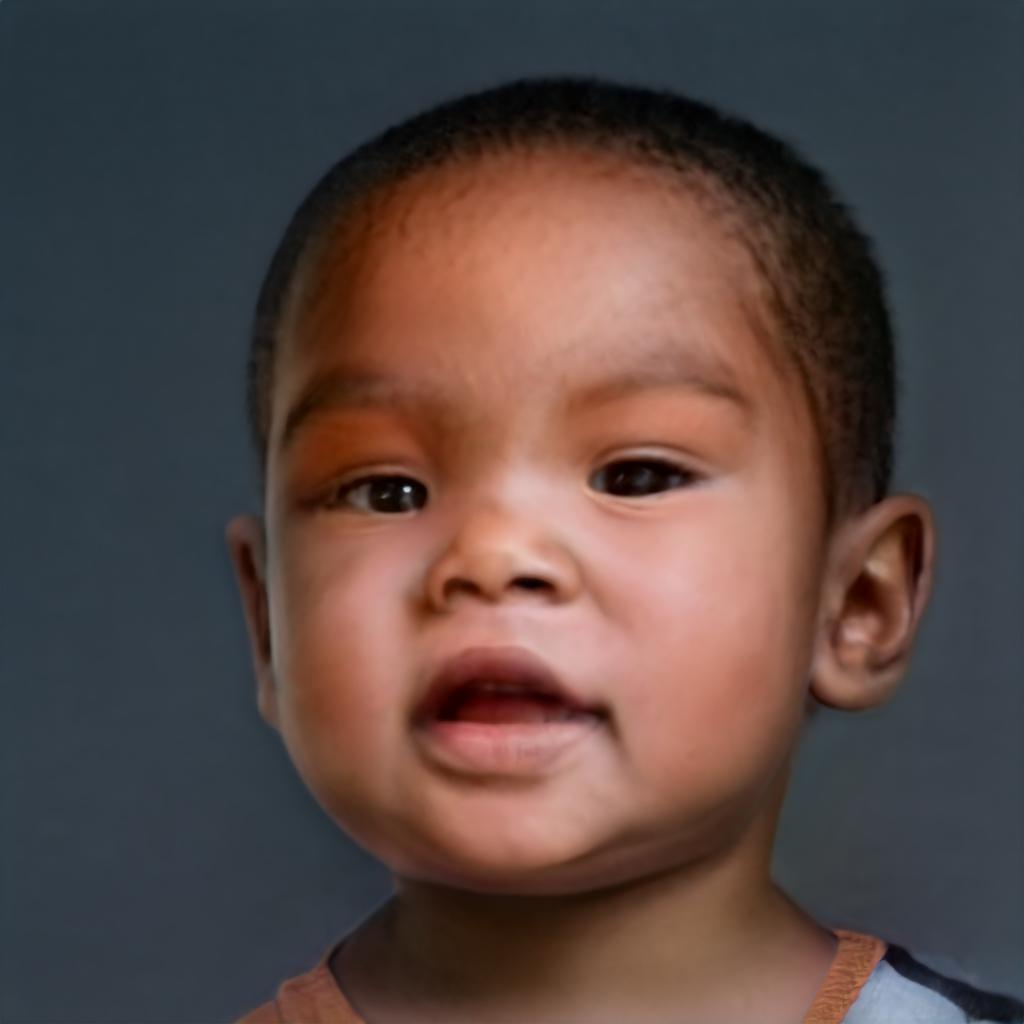} &
        \includegraphics[width=0.1125\textwidth]{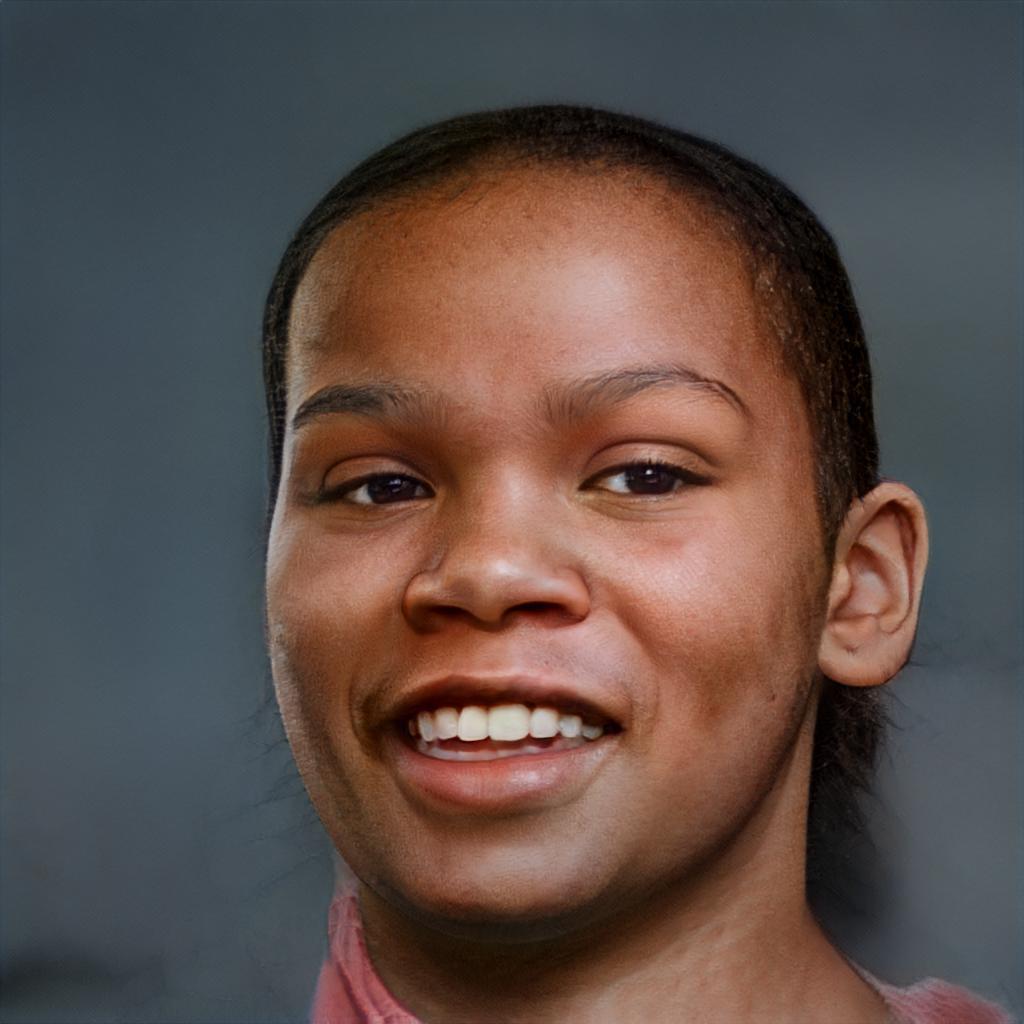} &
        \includegraphics[width=0.1125\textwidth]{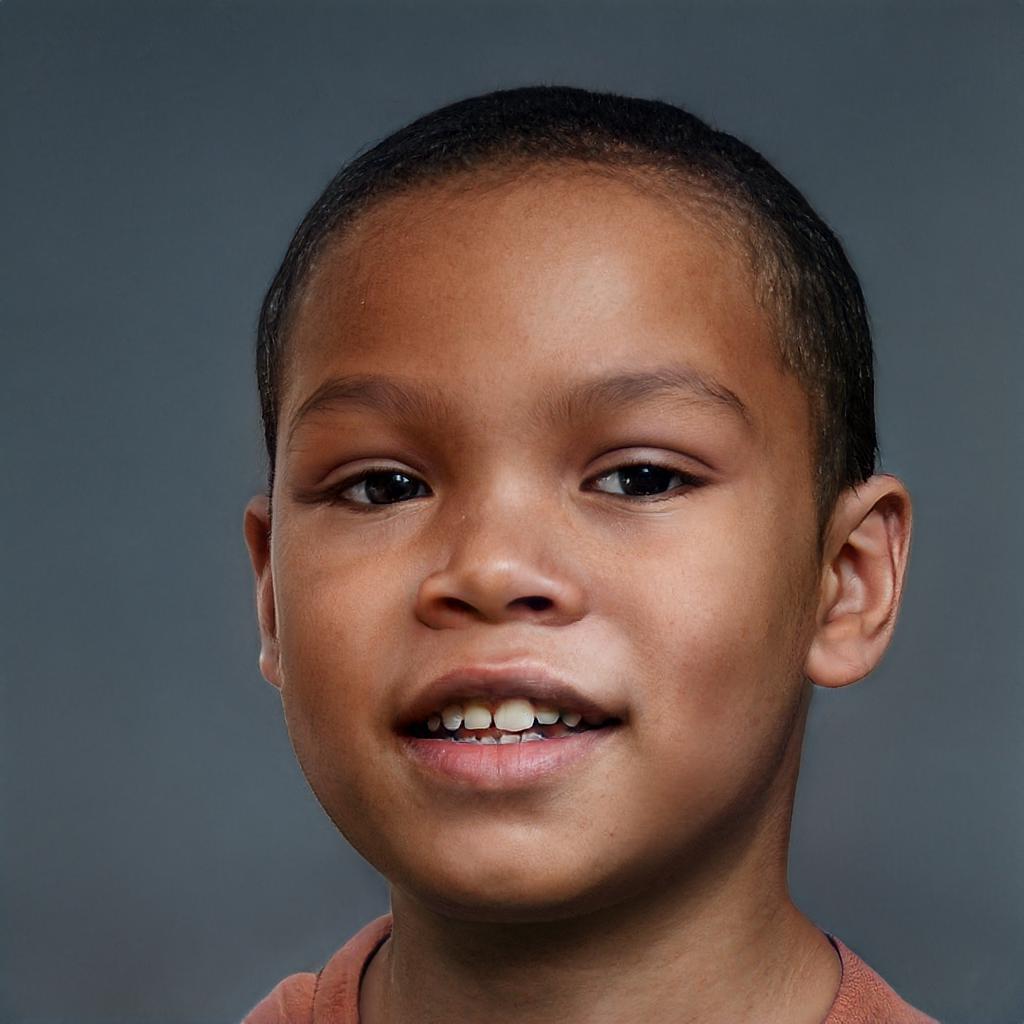} &
        \includegraphics[width=0.1125\textwidth]{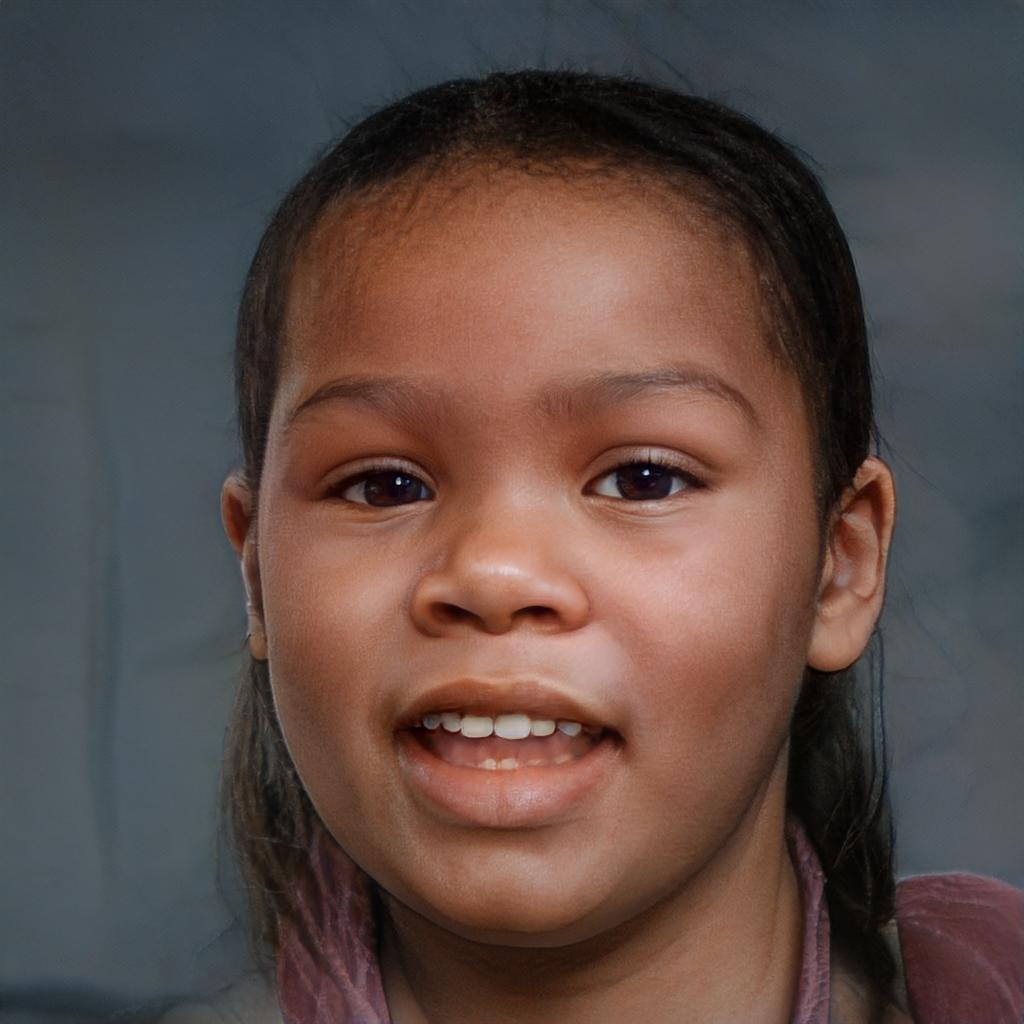} &
        \includegraphics[width=0.1125\textwidth]{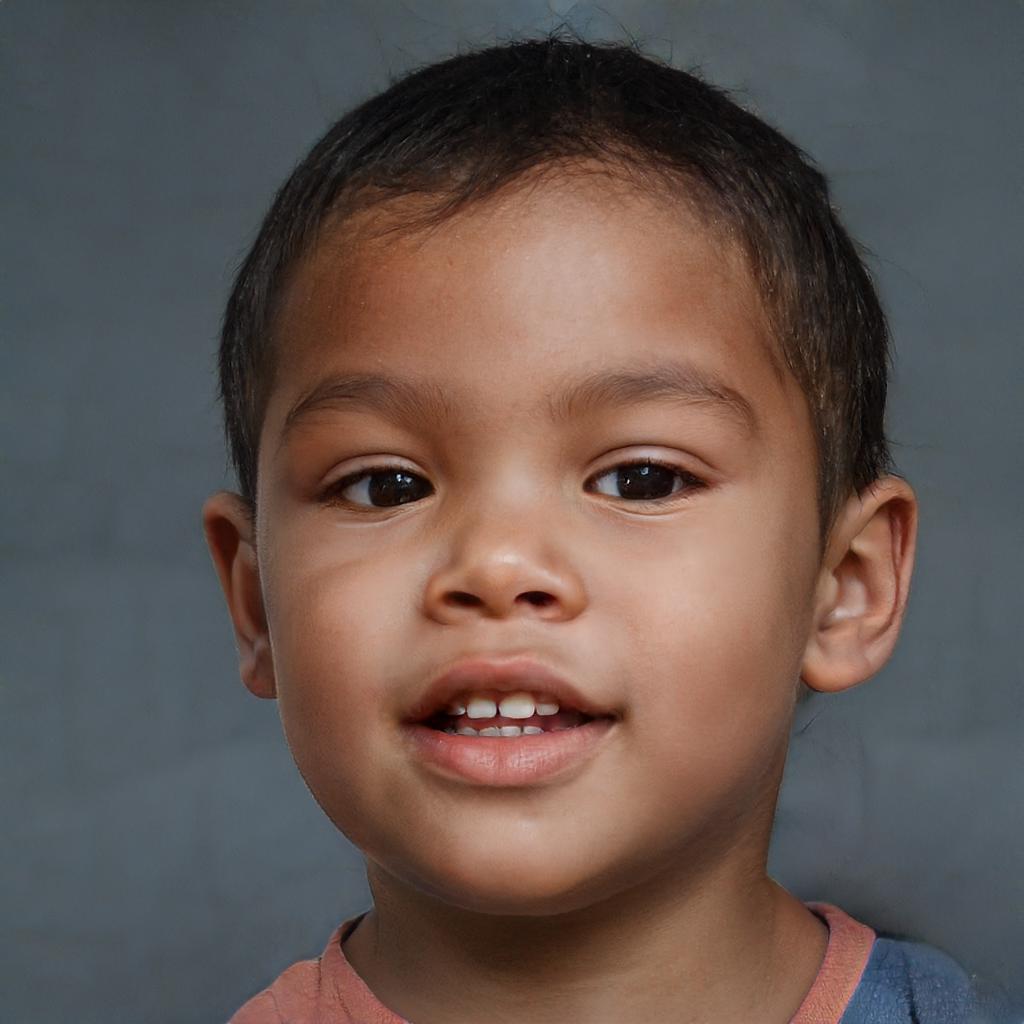} &
        \includegraphics[width=0.1125\textwidth]{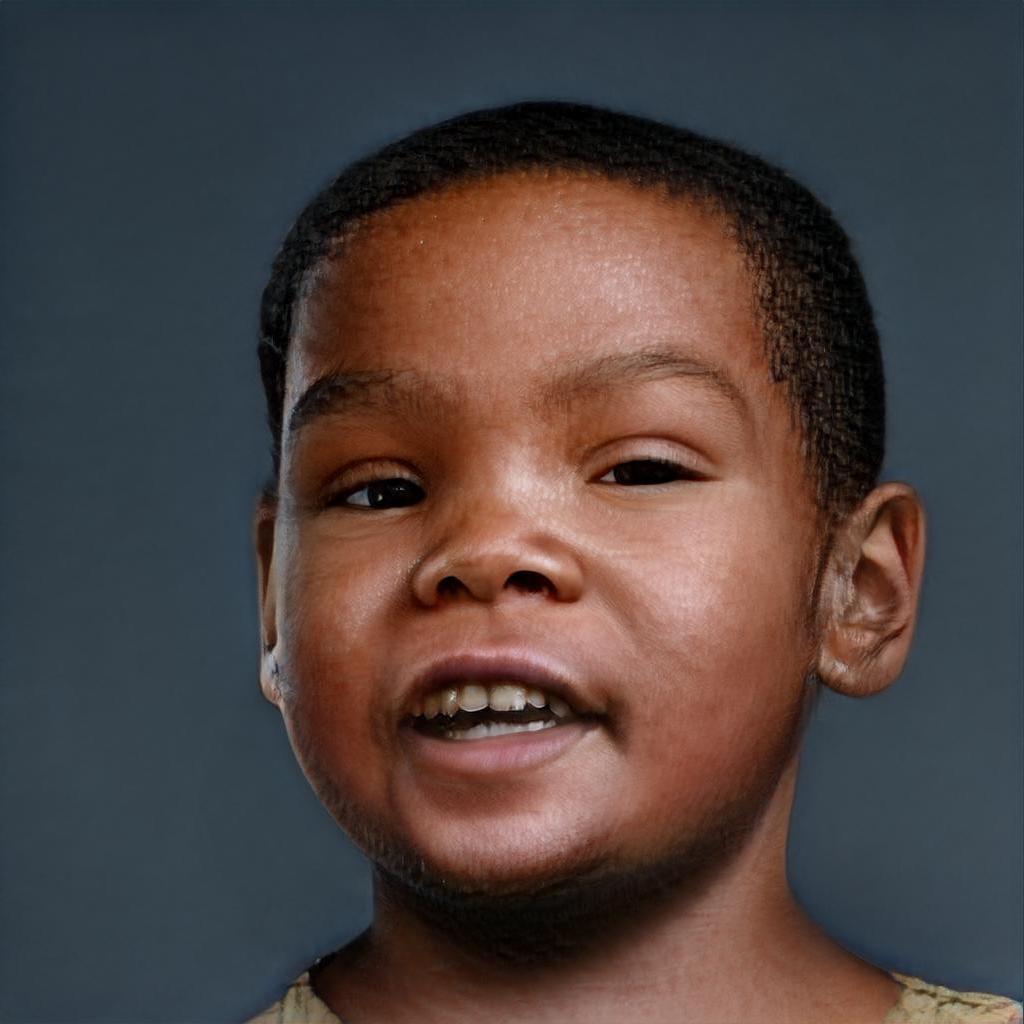} \\

        \vspace{-0.0615cm}

        \raisebox{0.075in}{\rotatebox{90}{$+$Dark Hair}} &
        \includegraphics[width=0.1125\textwidth]{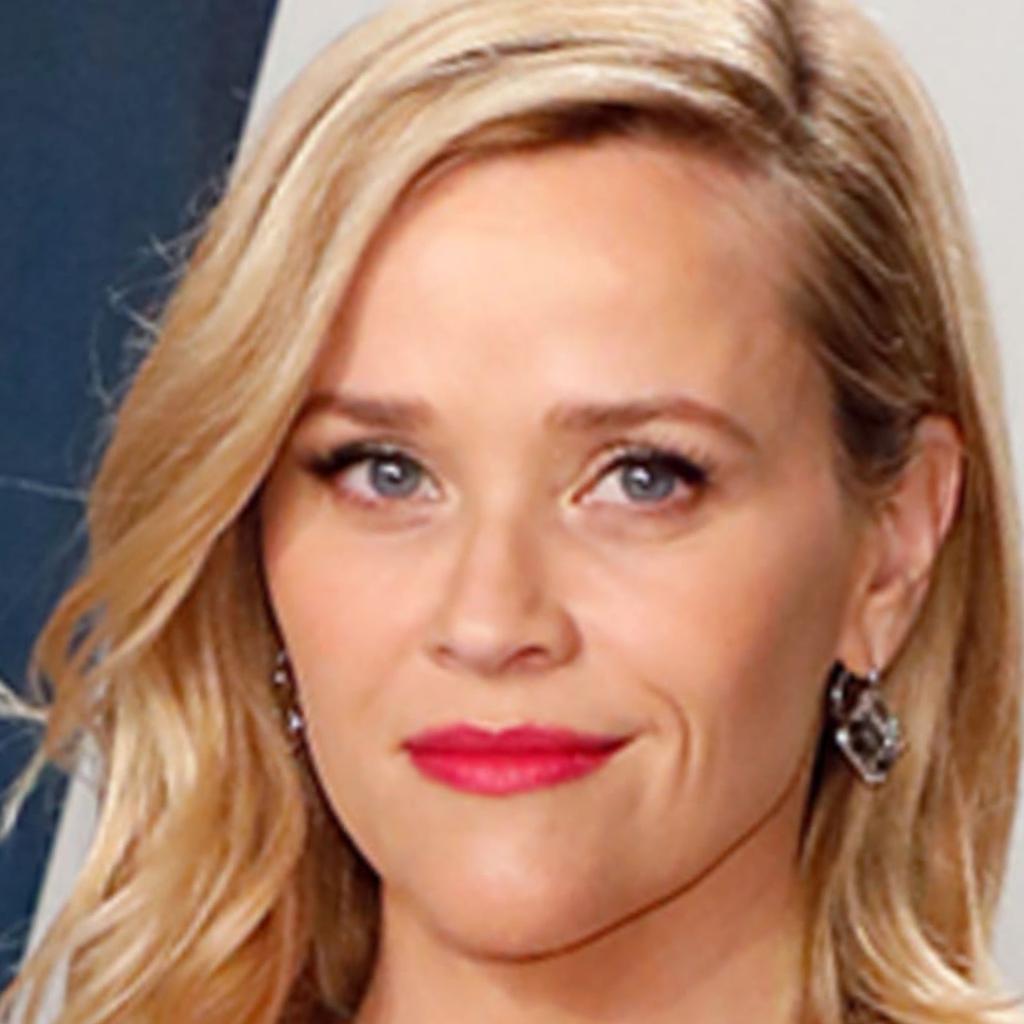} &
        \includegraphics[width=0.1125\textwidth]{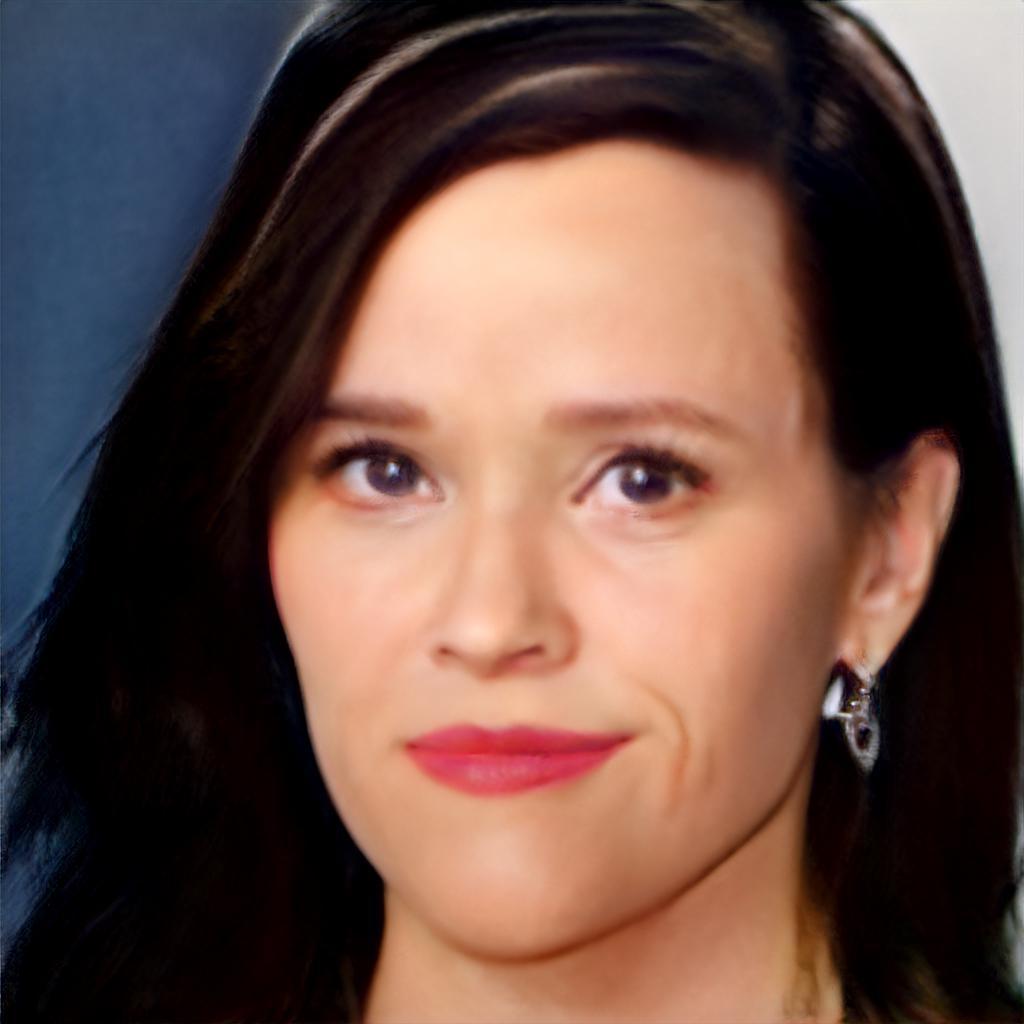} &
        \includegraphics[width=0.1125\textwidth]{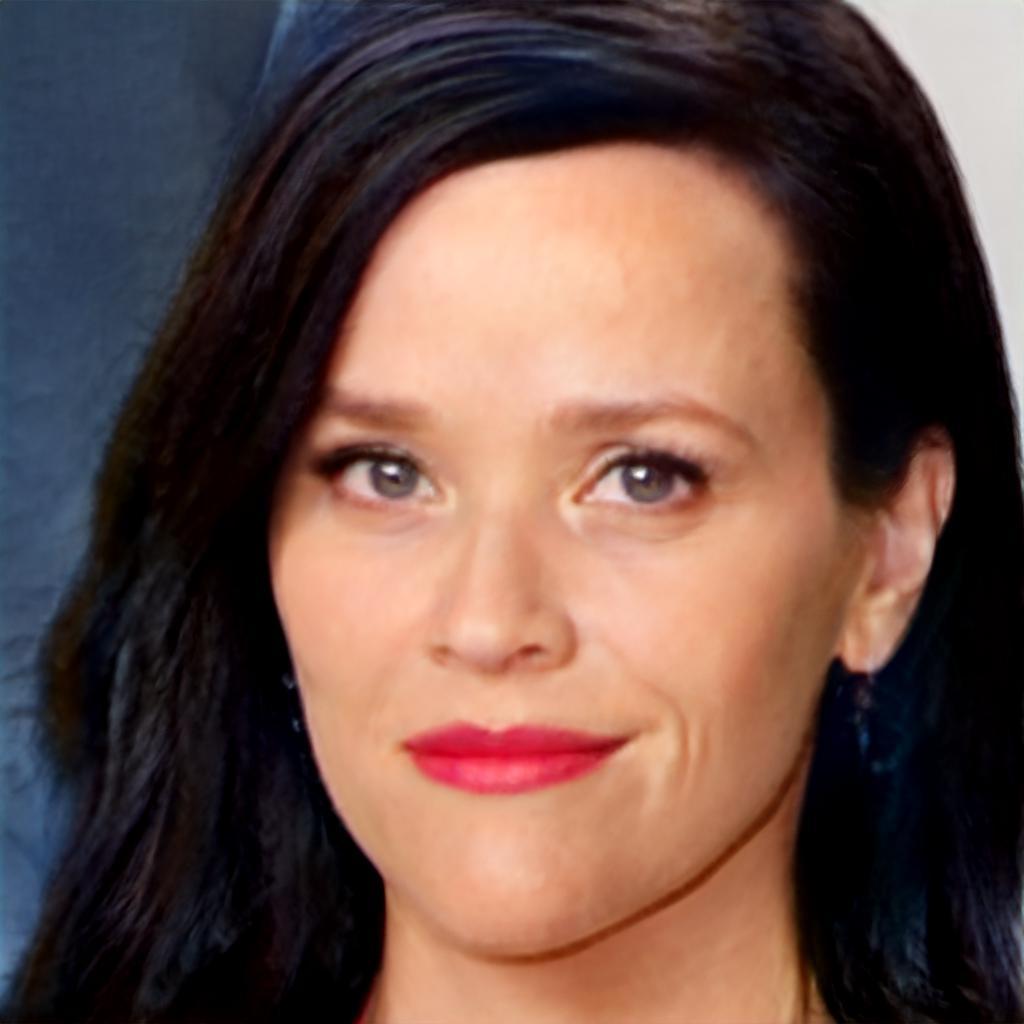} &
        \includegraphics[width=0.1125\textwidth]{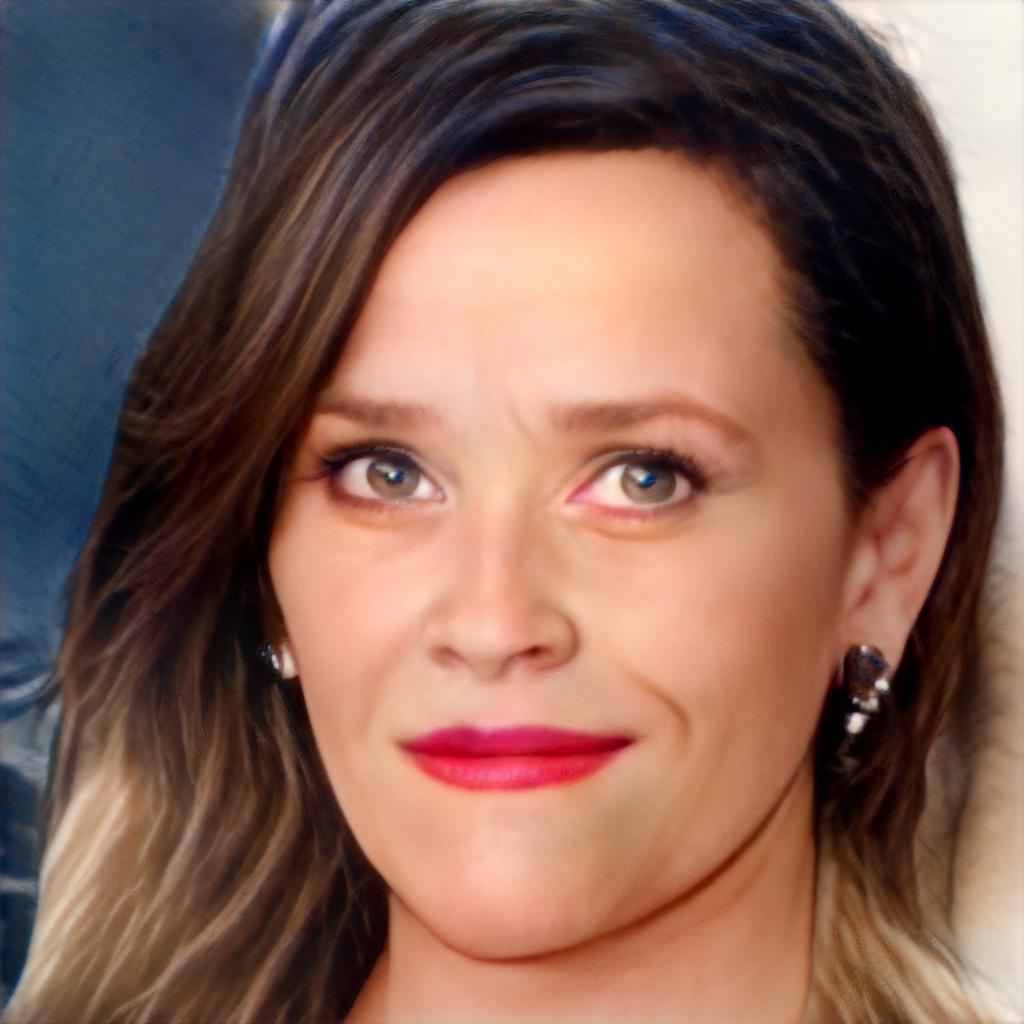} &
        \includegraphics[width=0.1125\textwidth]{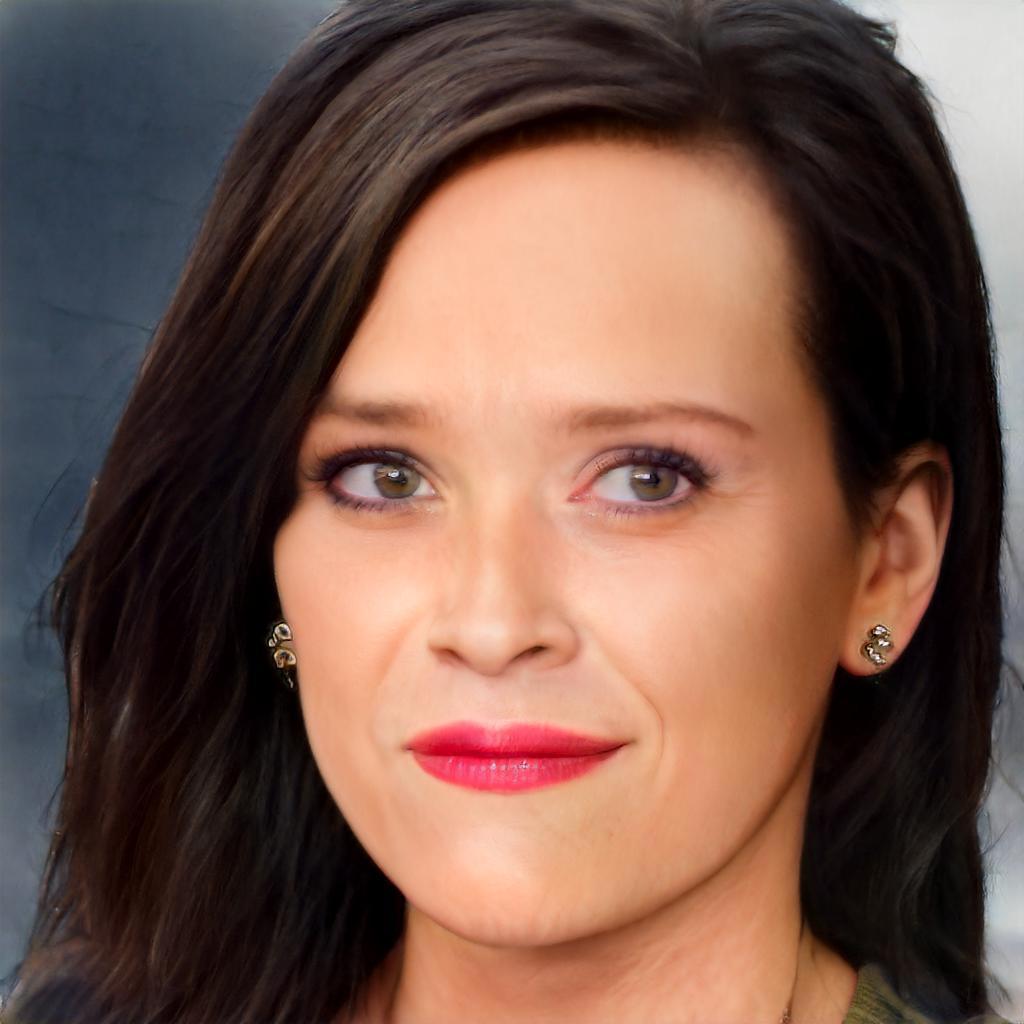} &
        \includegraphics[width=0.1125\textwidth]{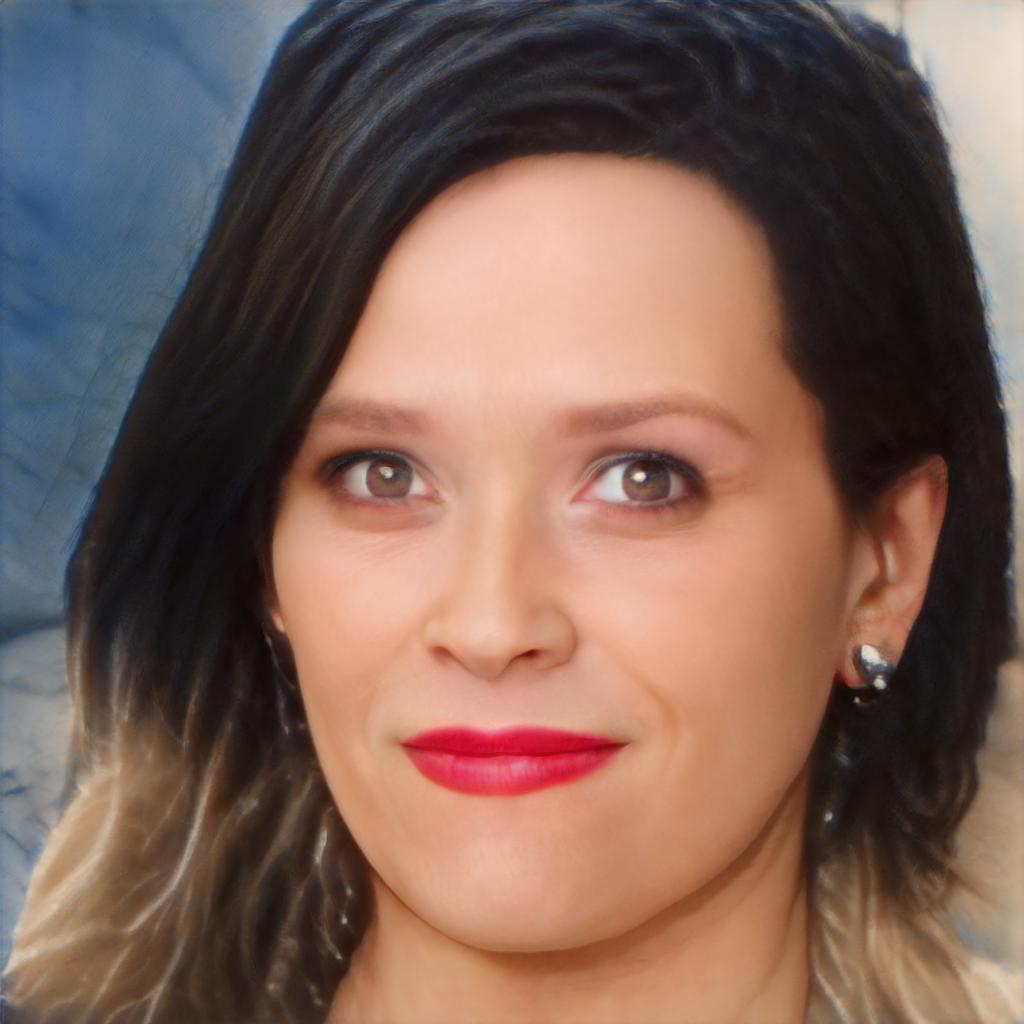} &
        \includegraphics[width=0.1125\textwidth]{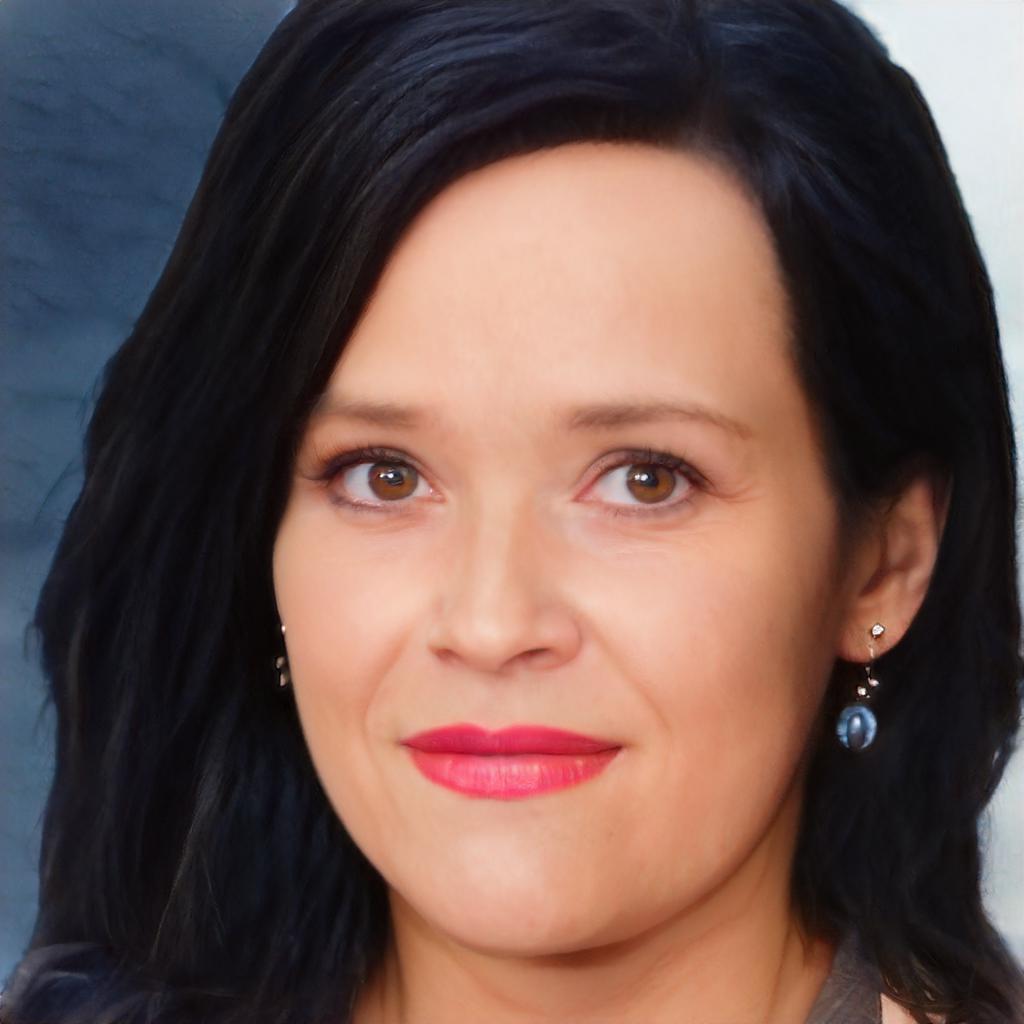} &
        \includegraphics[width=0.1125\textwidth]{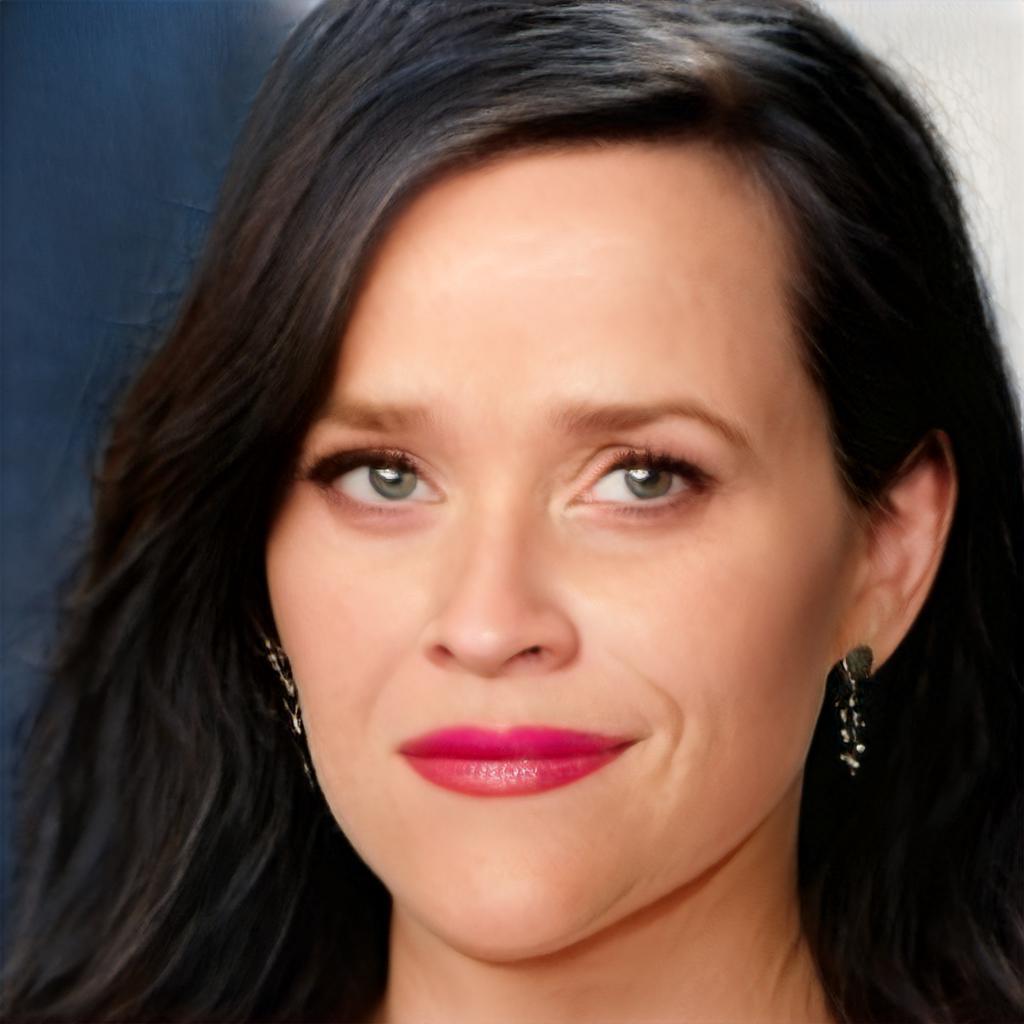} \\
        
        \vspace{-0.0615cm}
        
        \raisebox{0.175in}{\rotatebox{90}{$+$Smile}} &
        \includegraphics[width=0.1125\textwidth]{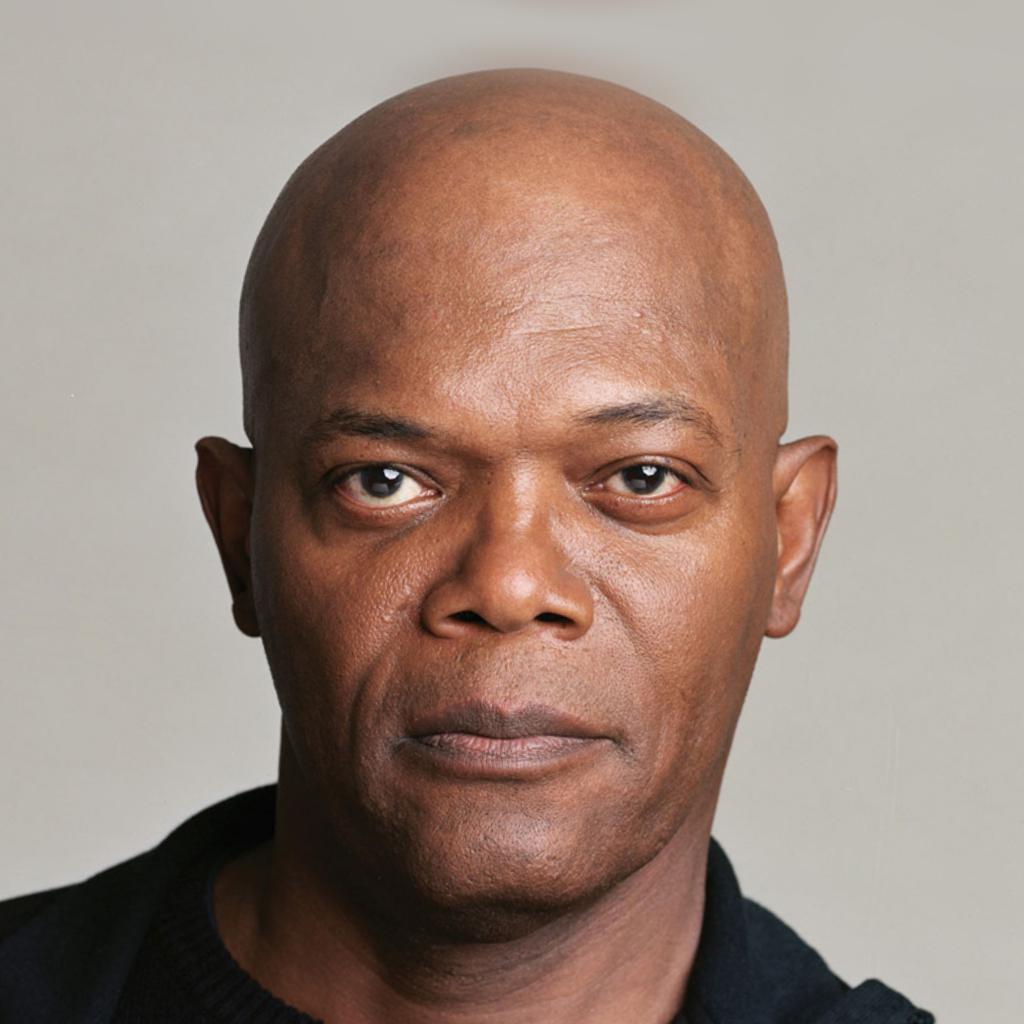} &
        \includegraphics[width=0.1125\textwidth]{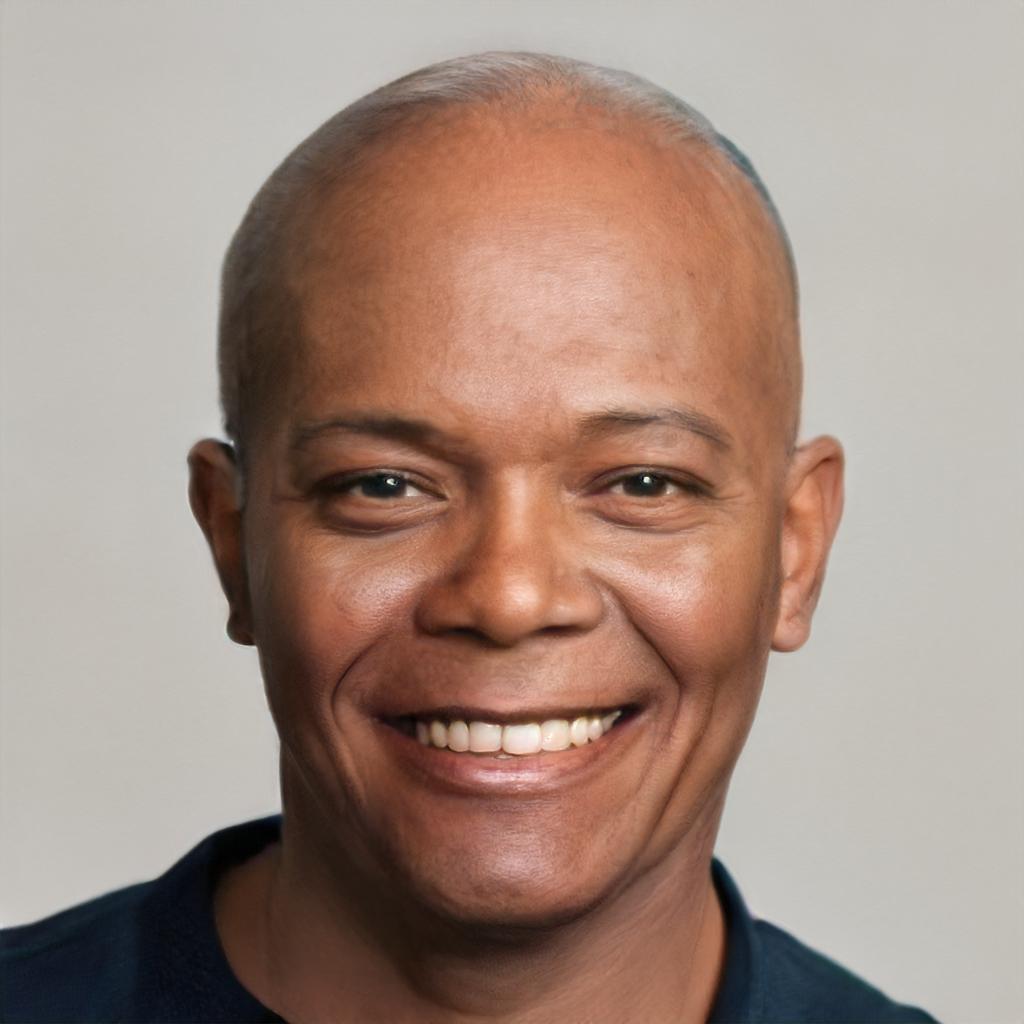} &
        \includegraphics[width=0.1125\textwidth]{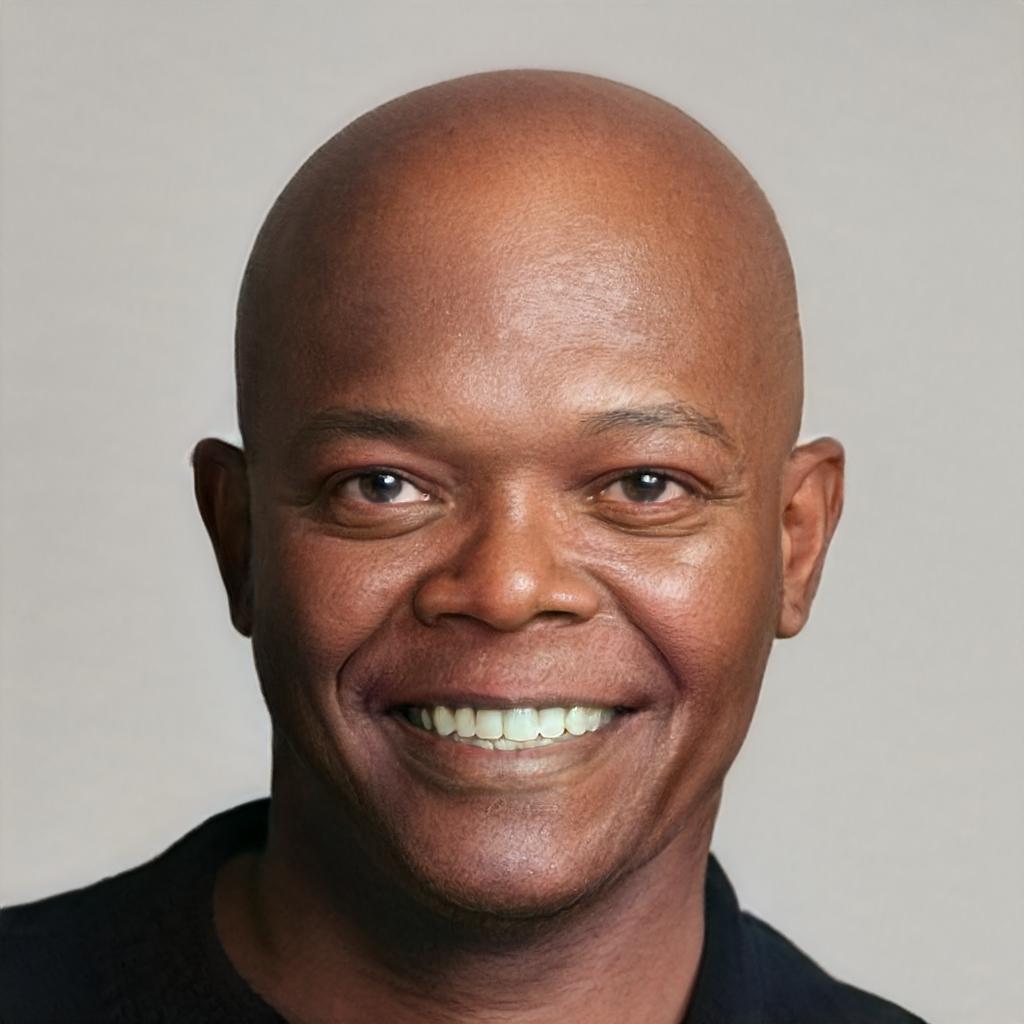} &
        \includegraphics[width=0.1125\textwidth]{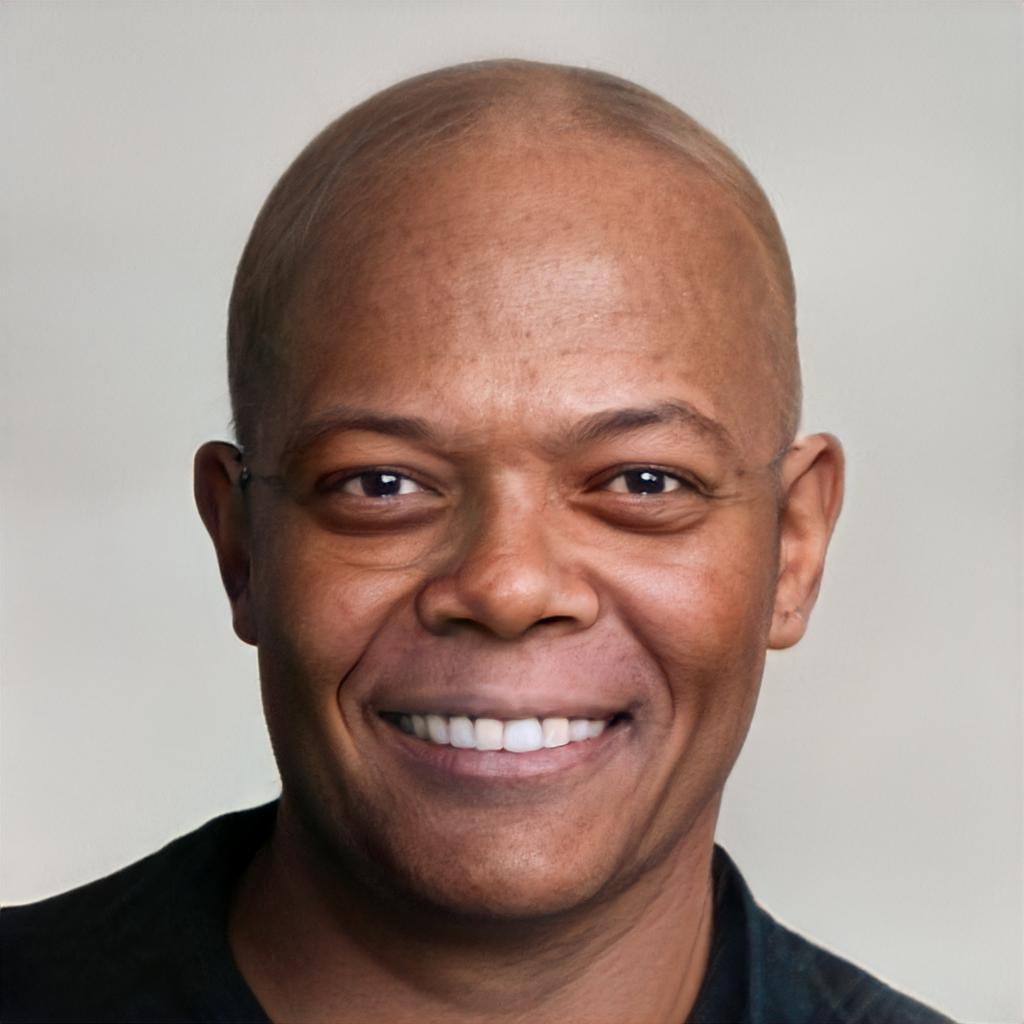} &
        \includegraphics[width=0.1125\textwidth]{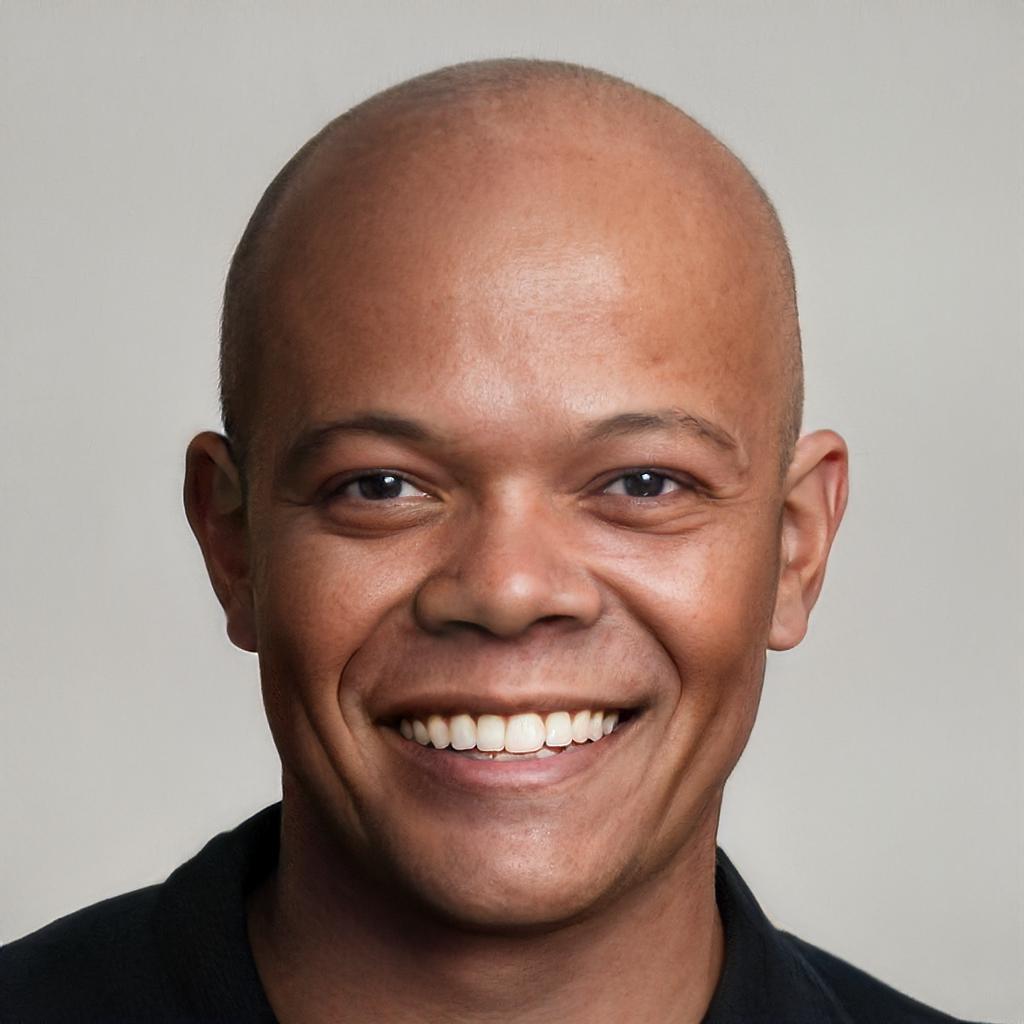} &
        \includegraphics[width=0.1125\textwidth]{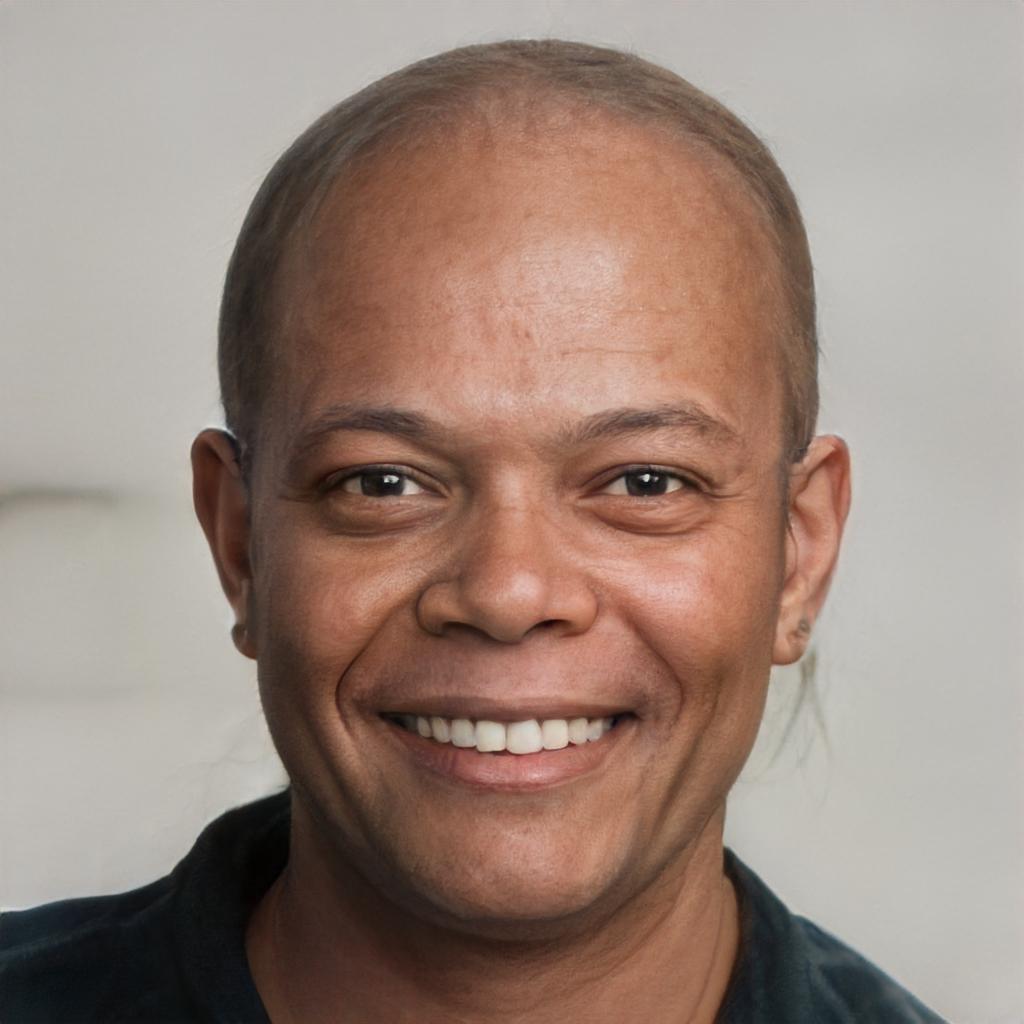} &
        \includegraphics[width=0.1125\textwidth]{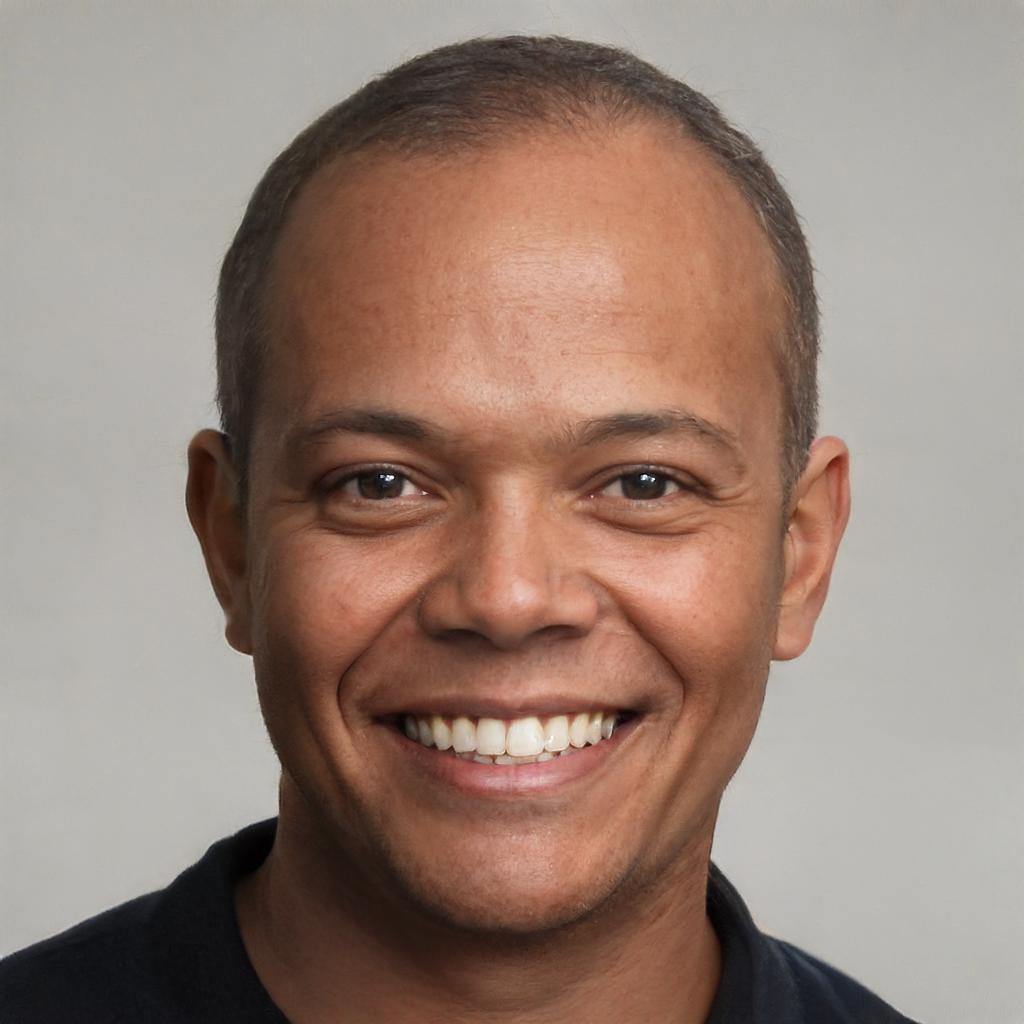} &
        \includegraphics[width=0.1125\textwidth]{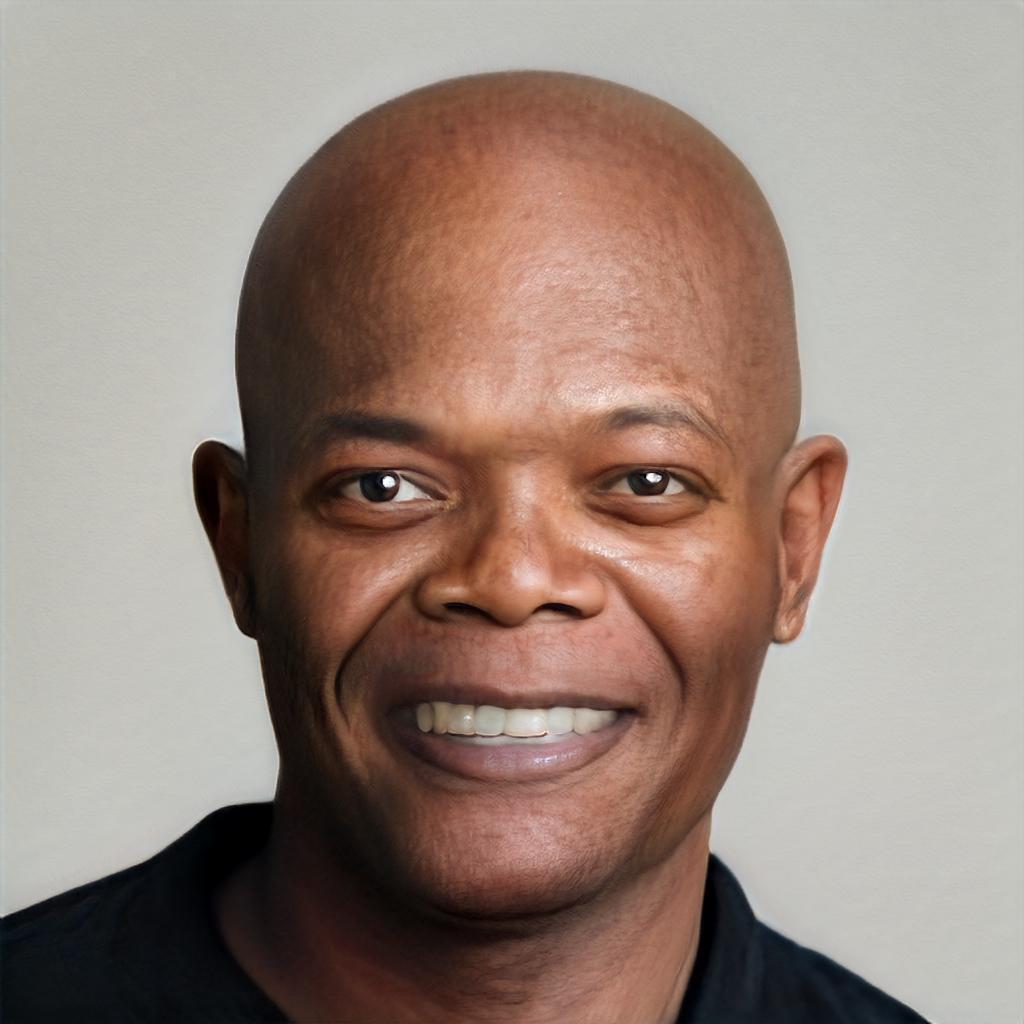} \\
        
        \vspace{-0.0615cm}
        
        \raisebox{0.15in}{\rotatebox{90}{$+$Frontal}} &
        \includegraphics[width=0.1125\textwidth]{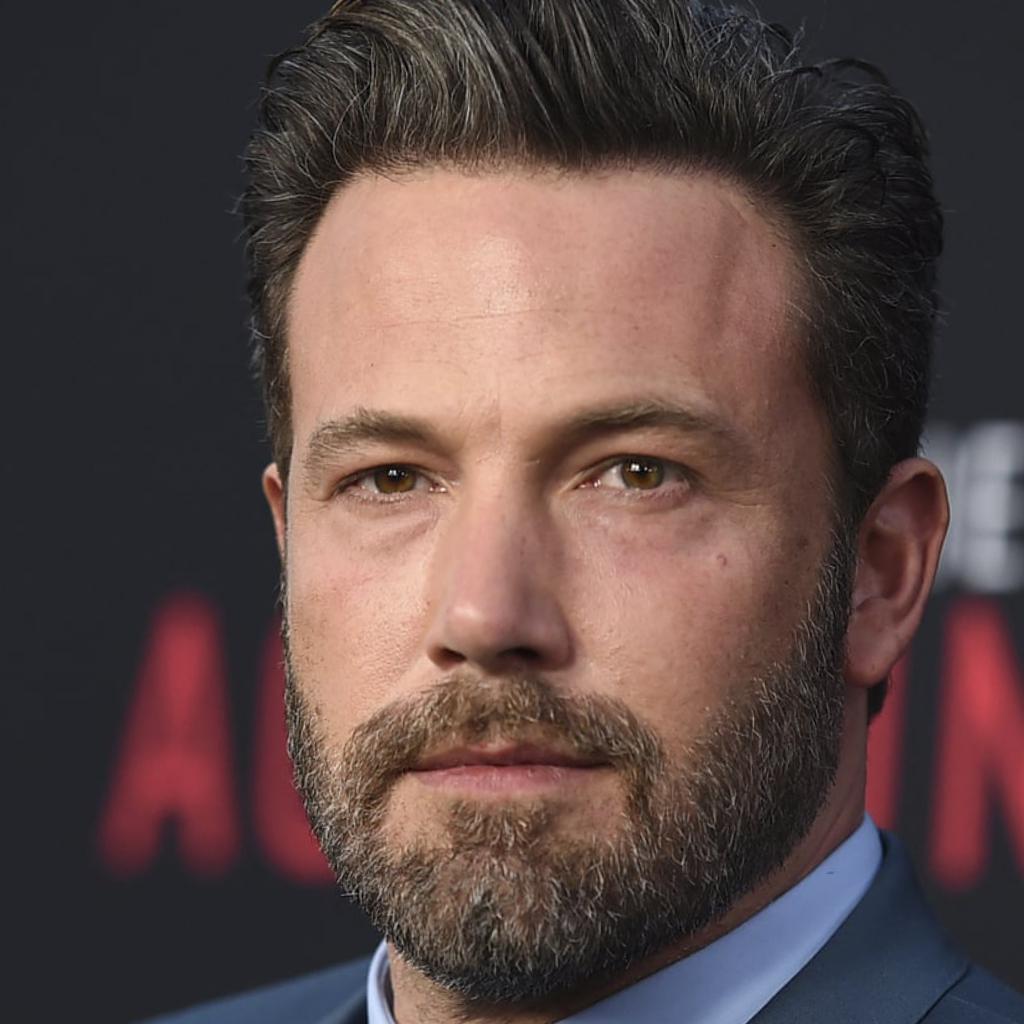} &
        \includegraphics[width=0.1125\textwidth]{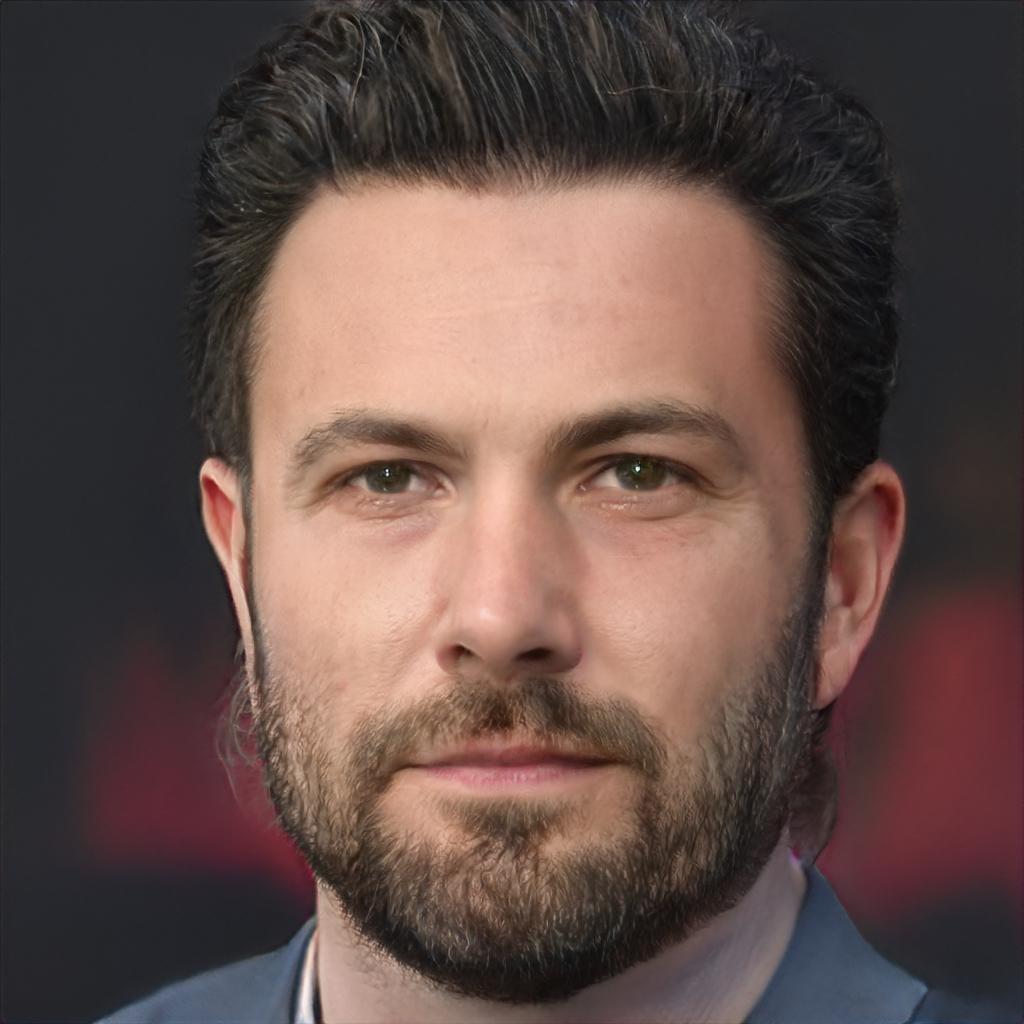} &
        \includegraphics[width=0.1125\textwidth]{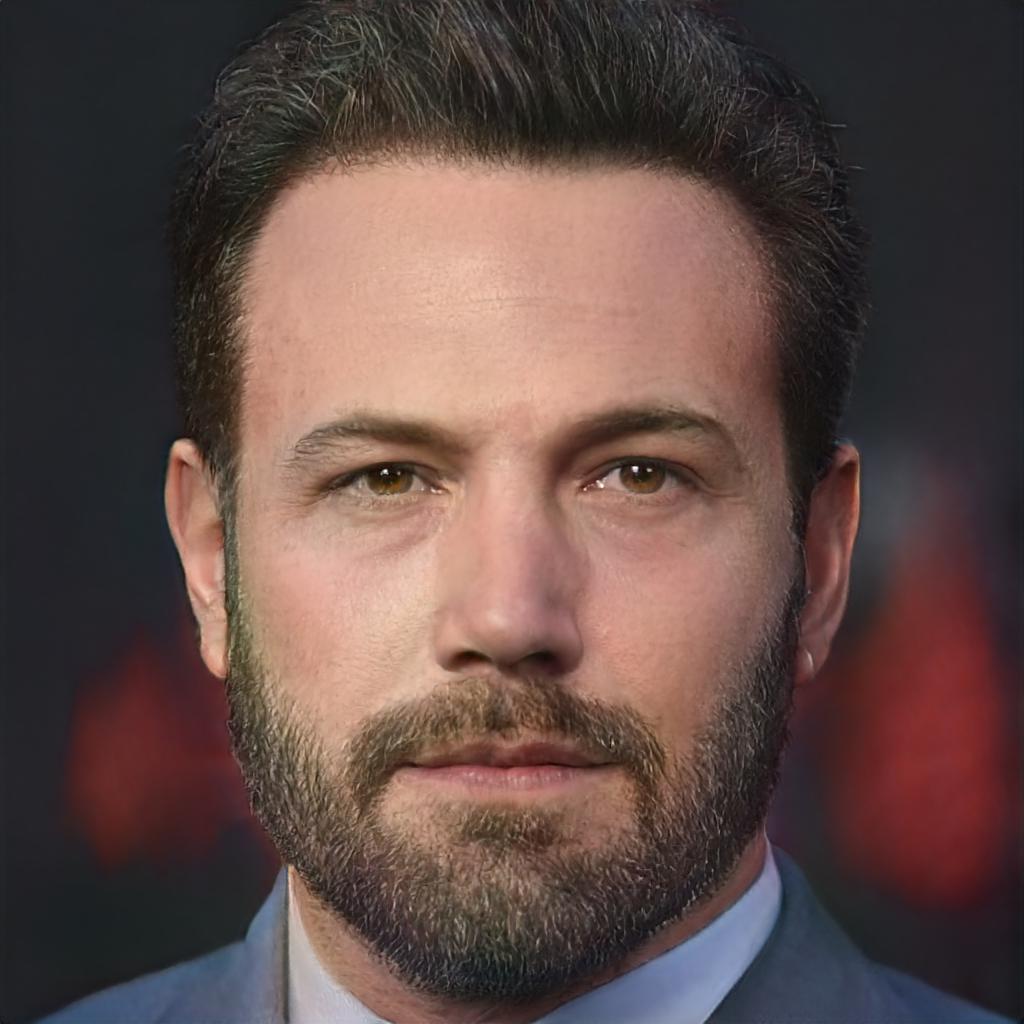} &
        \includegraphics[width=0.1125\textwidth]{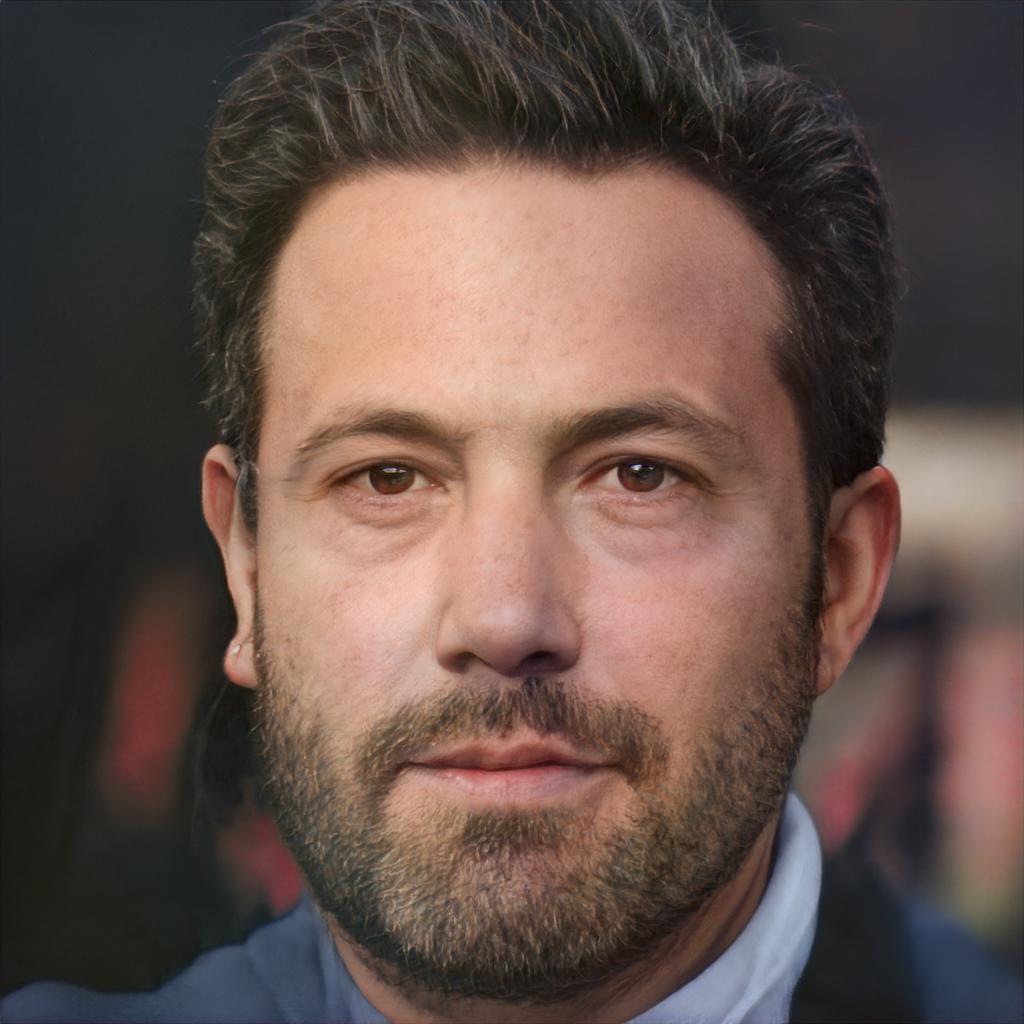} &
        \includegraphics[width=0.1125\textwidth]{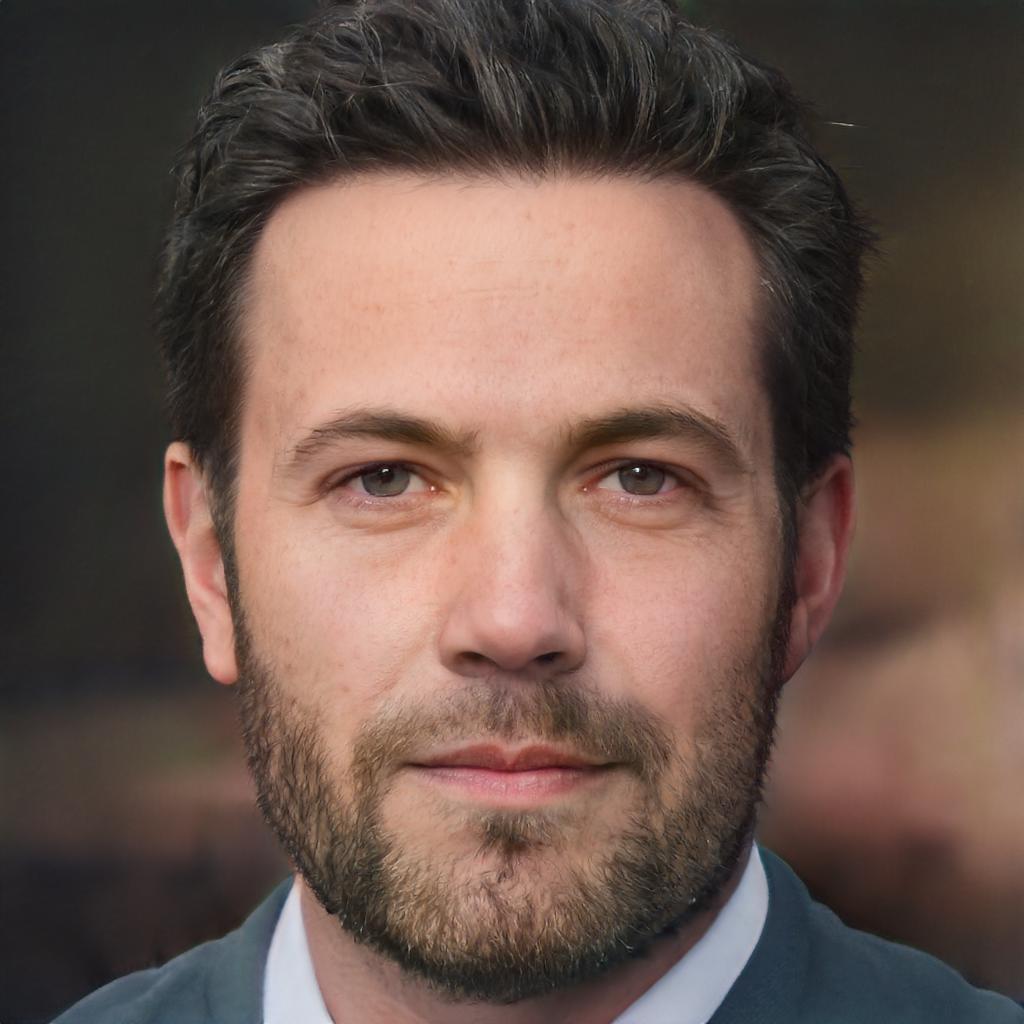} &
        \includegraphics[width=0.1125\textwidth]{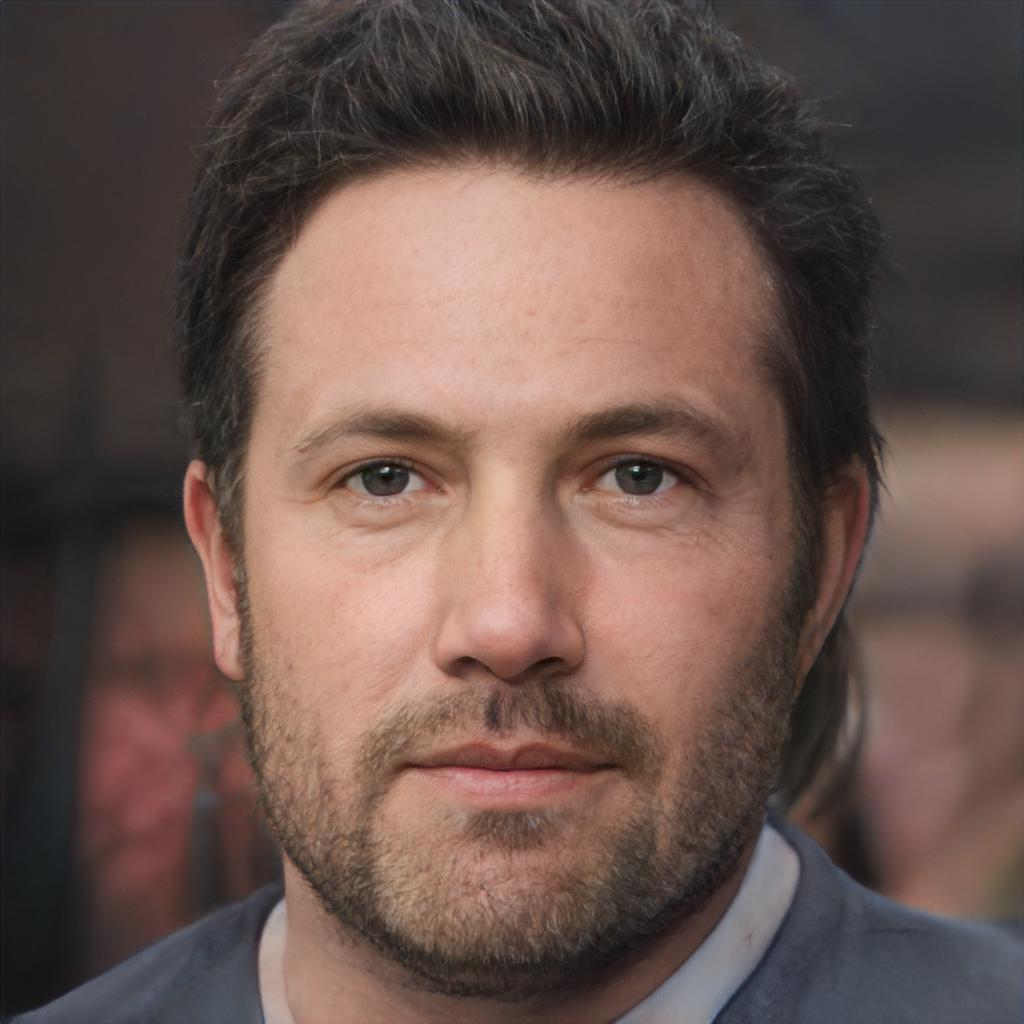} &
        \includegraphics[width=0.1125\textwidth]{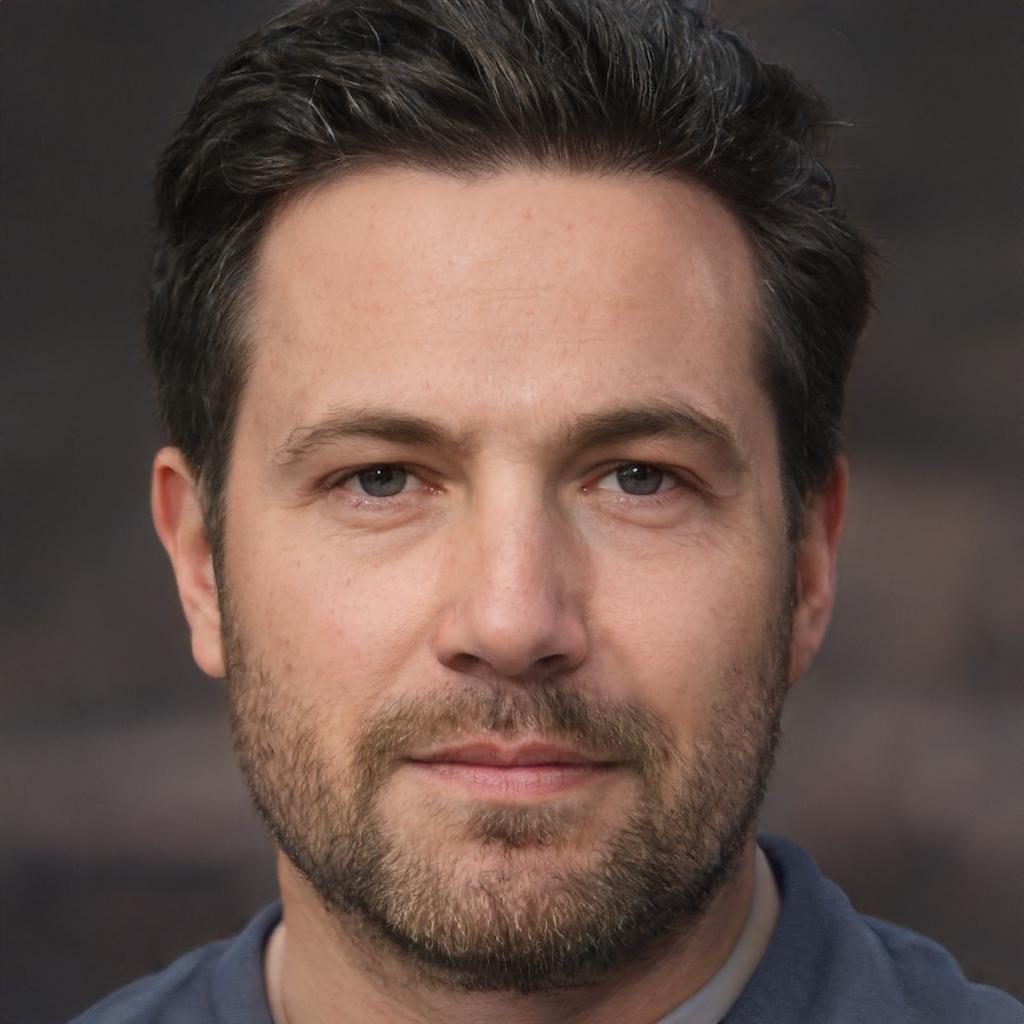} &
        \includegraphics[width=0.1125\textwidth]{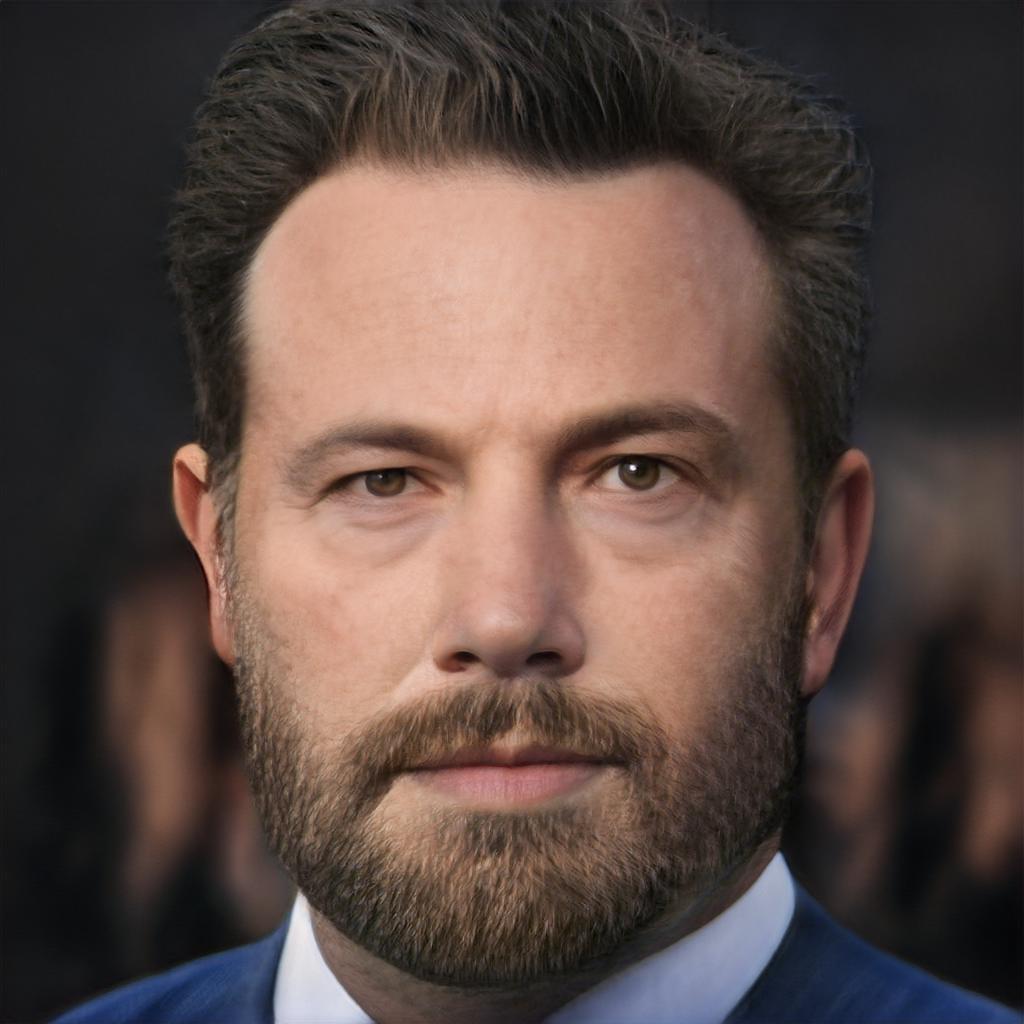} \\
        
        \vspace{-0.0615cm}
        
        \raisebox{0.1in}{\rotatebox{90}{$+$Color}} &
        \includegraphics[width=0.1125\textwidth]{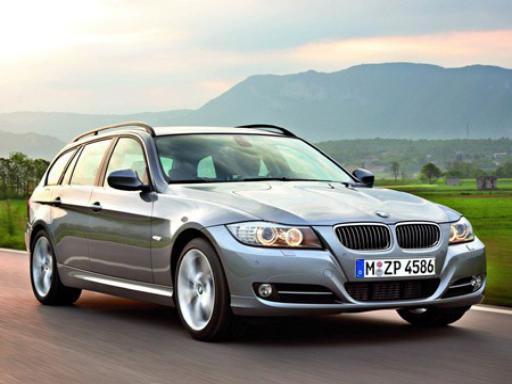} &
        \includegraphics[width=0.1125\textwidth]{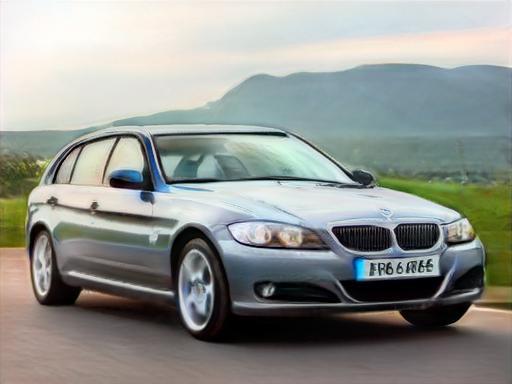} &
        \includegraphics[width=0.1125\textwidth]{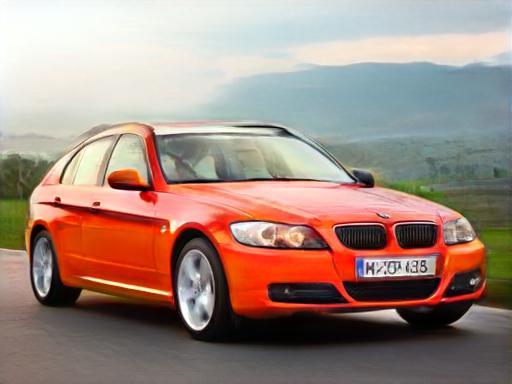} &
        \includegraphics[width=0.1125\textwidth]{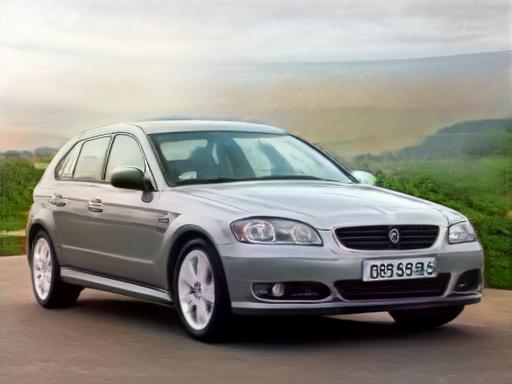} &
        \includegraphics[width=0.1125\textwidth]{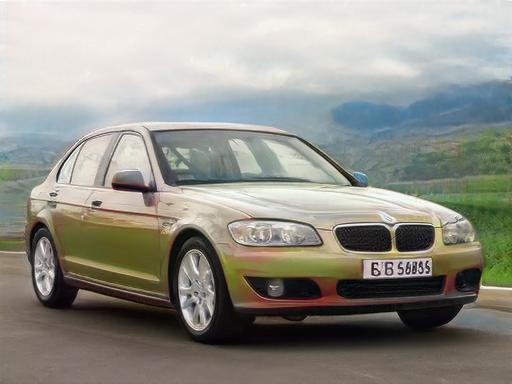} &
        \includegraphics[width=0.1125\textwidth]{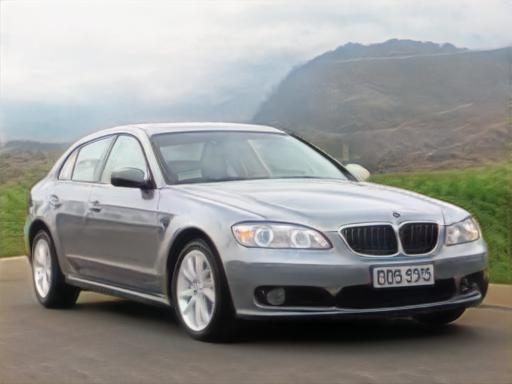} &
        \includegraphics[width=0.1125\textwidth]{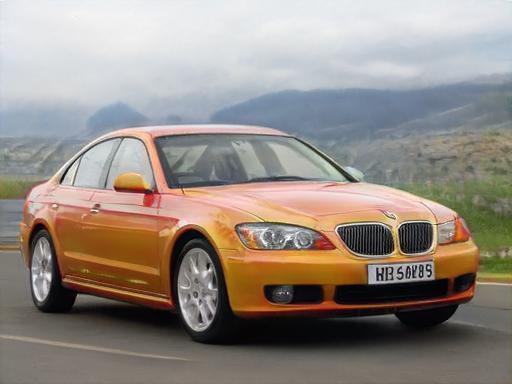} &
        \includegraphics[width=0.1125\textwidth]{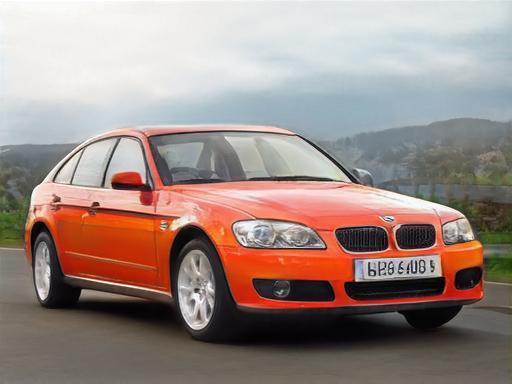} \\
        
        \vspace{-0.0615cm}
        
        \raisebox{0.1in}{\rotatebox{90}{$+$Cube}} &
        \includegraphics[width=0.1125\textwidth]{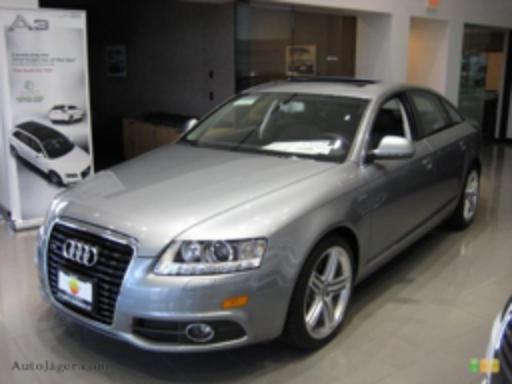} &
        \includegraphics[width=0.1125\textwidth]{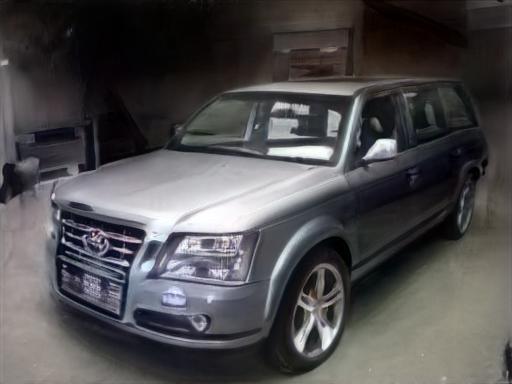} &
        \includegraphics[width=0.1125\textwidth]{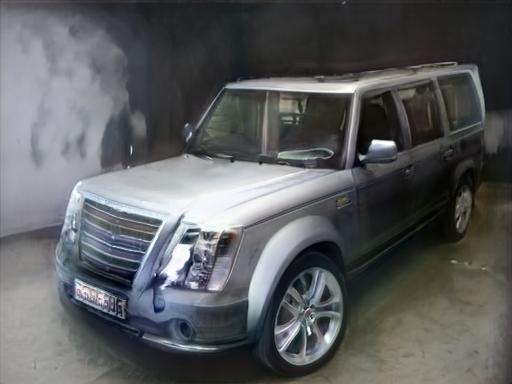} &
        \includegraphics[width=0.1125\textwidth]{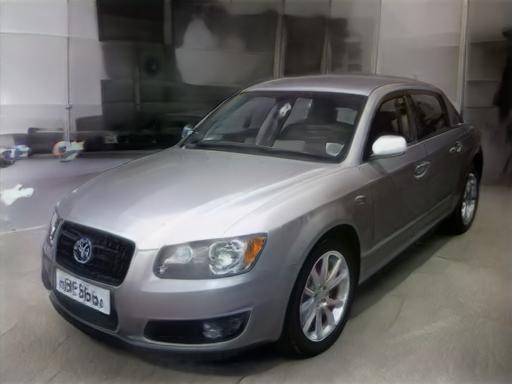} &
        \includegraphics[width=0.1125\textwidth]{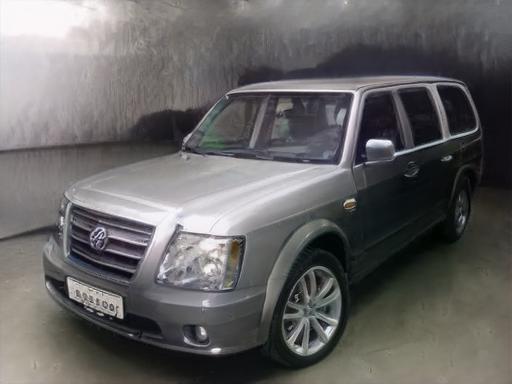} &
        \includegraphics[width=0.1125\textwidth]{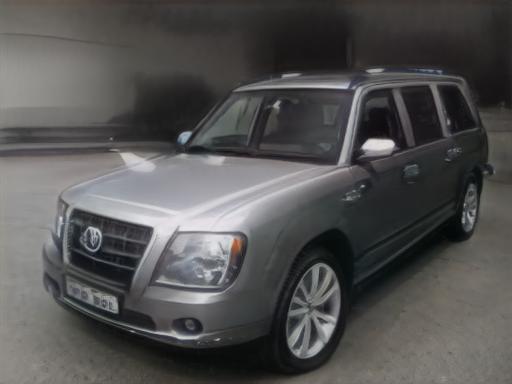} &
        \includegraphics[width=0.1125\textwidth]{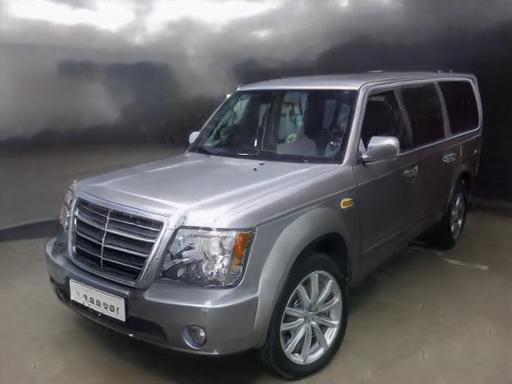} &
        \includegraphics[width=0.1125\textwidth]{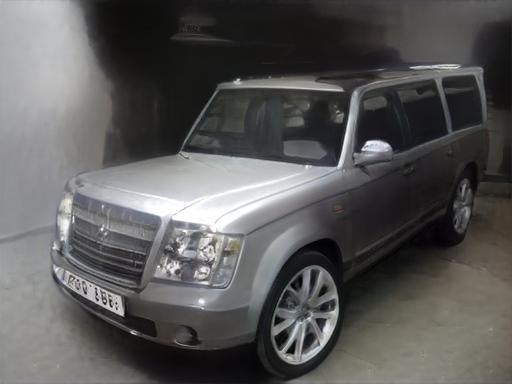} \\

        & Input & Optimization & PTI & $\text{ReStyle}_{pSp}$ &  $\text{ReStyle}_{e4e}$ & pSp & e4e & HyperStyle
        
    \end{tabular}
    
    }
    \vspace{-0.225cm}
    \caption{Editing quality comparison. We perform various edits~\cite{patashnik2021styleclip,shen2020interpreting,harkonen2020ganspace} over latent codes obtained by each inversion method. HyperStyle achieves both realistic, faithful edits and a high level of identity preservation across the different edits. Best viewed zoomed-in.}
    \vspace{-0.225cm}
    \label{fig:editing}
\end{figure*}

\subsection{Editability via Latent Space Manipulations}
Good inversion methods should provide not only meticulous reconstructions but also \textit{highly-editable} latent codes. We thus evaluate the editability of our produced inversions. We do so by analyzing two key aspects. 
One aspect of interest is the range of modifications that an inverted latent can support (e.g., how much the pose can be changed). The other is how well the identity is preserved along this range.

\vspace{-0.2cm}
\paragraph{Qualitative Evaluation} 
As shown in ~\cref{fig:editing}, our method successfully achieves realistic and meaningful edits, while being faithful to the input identity.
The inversions of optimization, pSp, and $\textit{ReStyle}_{pSp}$ reside in poorly-behaved latent regions of $\mathcal{W}+$. Therefore, their editing is less meaningful and introduces significant artifacts. For instance, in the cars domain, they struggle in making notable changes to the car color and shape. In the 4th row, these methods fail to either preserve the original identity or perform a full frontalization. On the other end of the reconstruction-editability trade-off, e4e and $\textit{ReStyle}_{e4e}$ are more editable but cannot faithfully preserve the original identity, as demonstrated in the 3rd and 4th rows. In contrast, HyperStyle and PTI, which invert into the well-behaved \w space, are more robust in their editing capabilities while successfully retaining the original identity. Yet, HyperStyle requires a significantly lower inference overhead to achieve these results.

\vspace{-0.385cm}
\paragraph{Quantitative Evaluation}
Comparing the editability of inversion methods is challenging since applying the same editing step size to latent codes obtained with different methods results in different editing strengths. This would introduce unwanted bias to the identity similarity measure, as the less-edited images may tend to be more similar to the source. 
To address this, we edit using a \textit{range} of various step sizes and plot the measured identity similarity along this range, resulting in a continuous similarity curve for each inversion method. 
This allows us to validate the identity preservation with respect to a fixed editing magnitude, as well as examine the range of edits supported. Ideally, an inversion method should achieve high identity similarity across a wide range of editing strengths. 
We measure the editing magnitude using trait-specific classifiers (HopeNet~\cite{Ruiz_2018_CVPR_Workshops} for pose and the classifier from Lin~\etal~\cite{lin2021anycost} for smile extent). 
As before, identity similarity is measured using the CurricularFace method~\cite{huang2020curricularface}.

\begin{figure}
    \setlength{\tabcolsep}{1pt}
    \centering
    \hspace*{-0.25cm}
    \includegraphics[width=1\linewidth]{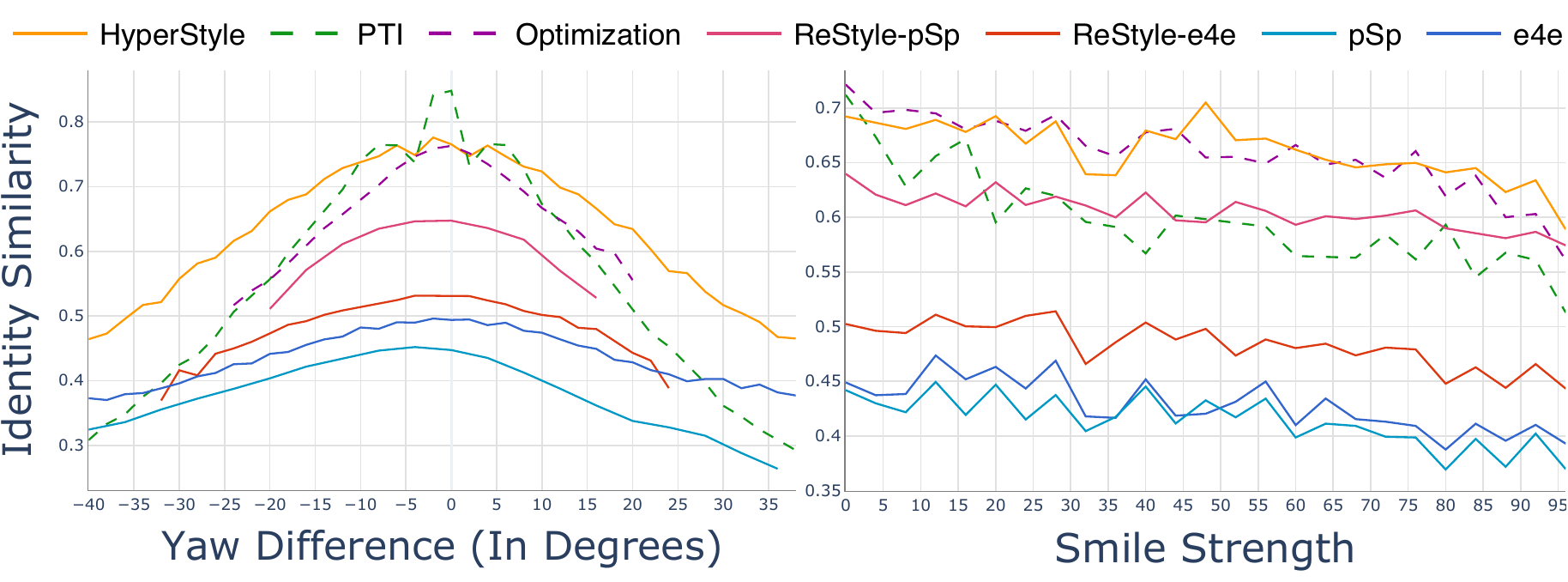}
    \vspace{-0.35cm}
    \caption{Quantitative editing metrics. For each method, we compute the identity similarity between the original and edited images as a function of editing magnitude. 
    }
    \vspace{-0.5cm}
    \label{fig:editing_quantitative_comparison}
\end{figure}

As can be seen in \cref{fig:editing_quantitative_comparison}, HyperStyle consistently outperforms other encoder-based methods in terms of identity preservation while supporting an equal or greater editing range. 
Compared to optimization-based techniques, HyperStyle achieves similar identity preservation and editing range yet does so substantially faster.

These results highlight the appealing nature of HyperStyle. With respect to other encoders, HyperStyle achieves superior reconstruction quality while providing strong editability and fast inference. Additionally, compared to optimization techniques, HyperStyle achieves comparable reconstruction and editability at a fraction of the time, making it more suitable for real-world use at scale. 
This places HyperStyle favorably on both the reconstruction-editability and the time-accuracy trade-off curves.

\vspace{-0.1cm}
\subsection{Ablation Study}~\label{sec:ablation}
We now validate the design choices described in \cref{sec:method}. Results are summarized in \cref{tb:ablation_study_res}. 
First, we investigate the choice of layers refined by the hypernetwork. We observe that training only the medium and fine non-toRGB layers achieves comparable performance, a slimmer network, and faster inference. Notably, we also find that altering toRGB layers may harm editability. 
Second, we find the iterative scheme to be more accurate with fewer artifacts. 
Finally, we validate the effectiveness of the Shared Refinement Block and the information sharing it provides. 
Visual comparisons of all ablations can be found in Appendix~\ref{supp:ablation_study}.

\newcolumntype{P}[1]{>{\centering\arraybackslash}p{#1}}

\begin{table}
    \small
    \setlength{\tabcolsep}{2pt}
    \centering
    
    {\small

            \begin{tabular}{p{0.15\textwidth} | c | P{0.0315\textwidth} | c c c c c}
            \toprule
            Method & Layers & Iters & $\uparrow$ ID & $\downarrow$ LPIPS & $\downarrow$ $L_2$ & $\downarrow$ Time \\
            \midrule
            
            \multirow{3}{*}{\shortstack[l]{No Iterative \\ Refinement}} &
            C,M,F,R &
            $1$ &
            $0.68$ &
            $0.10$ &
            $0.02$ &
            $0.17$ \\ 
            
            &
            C,M,F &
            $1$ &
            $0.67$ &
            $0.10$ &
            $0.02$ &
            $0.16$ \\ 
            
            &
            M,F &
            $1$ &
            $0.66$ &
            $0.11$ &
            $0.021$ &
            $0.15$ \\ 
            
            \midrule

            \textbf{HyperStyle} & 
            \textbf{M,F} &
            $\textbf{10}$ &
            $0.76$ &
            $0.09$ &
            $0.019$ &
            $1.23$ \\ 
            
            \midrule
            
            HyperStyle + Coarse &
            C,M,F &
            $10$ &
            $0.74$ &
            $0.10$ &
            $0.02$ &
            $1.54$ \\

            \midrule 
            
            HyperStyle w/o \newline Shared Refinement &
            \multirow{2}{*}{M,F} &
            \multirow{2}{*}{$10$} &
            \multirow{2}{*}{$0.68$} &
            \multirow{2}{*}{$0.12$} &
            \multirow{2}{*}{$0.022$} &
            \multirow{2}{*}{$1.36$} \\ 
            
            \midrule 
            \midrule 
            
            Separable Convs. &
            M,F &
            $10$ &
            $0.71$ &
            $0.10$ &
            $0.019$ &
            $1.28$ \\

            \bottomrule
            \end{tabular}
    }
    \vspace{-0.3cm}
    \caption{
    Ablation study. We validate the hypernetwork components and design choices: the importance of different layers --- coarse (C), medium (M), fine (F), and toRGB (R) --- as well as the iterative refinement scheme and Shared Refinement.
    We also explore separable convolutions as an alternative refinement head.
    } 
    \vspace{-0.4cm}
    \label{tb:ablation_study_res}
\end{table}

\vspace{-0.4cm}
\paragraph{Separable Convolutions} Our final configuration uses shared offsets for each convolutional kernel. An important question is whether this constrains the network too strongly. To answer this, we design an alternative refinement head, inspired by separable convolutions~\cite{howard2017mobilenets}. Rather than predicting offsets for an entire 
$k\times k \times C^{in} \times C^{out}$ filter in one step, we decompose it into two slimmer predictions: $k\times k \times C^{in} \times 1$ and $k\times k \times 1 \times C^{out}$. The final offset block is then given by their product. This allows us to predict an offset for every parameter of the kernel, potentially increasing the network's expressiveness. 
We observe (\cref{tb:ablation_study_res}) that the increased flexibility of predicting an offset per parameter does not improve reconstruction, indicating that simpler, per-channel predictions are sufficient.

\vspace{-0.1cm}
\subsection{Additional Applications}~\label{sec:additional_applications}

\vspace{-1.125cm}
\paragraph{Domain Adaptation}
Many works \cite{gal2021stylegannada,pinkney2020resolution,wu2021stylealign,ojha2021fewshot} have explored fine-tuning a pre-trained StyleGAN towards semantically similar domains.
This process maintains a correspondence between semantic attributes in the two latent spaces, allowing translation between domains~\cite{wu2021stylealign}.
Yet, some features, such as facial hair or hair color, may be lost during this translation. 
To address this, we use HyperStyle trained on the source generator to modify the fine-tuned target generator.  
Namely, given an input image, we can take the weight offsets predicted with respect to the source generator and apply them to the target generator. 
The image in the new domain is then obtained by passing the image's original latent code to the modified target generator.

\begin{figure}
\setlength{\tabcolsep}{1pt}
    \centering
    { \small 
    \begin{tabular}{c c c c c c}

     \vspace{-0.08cm}

    \raisebox{0.1in}{\rotatebox{90}{Toonify}} &
    \includegraphics[width=0.184\linewidth]{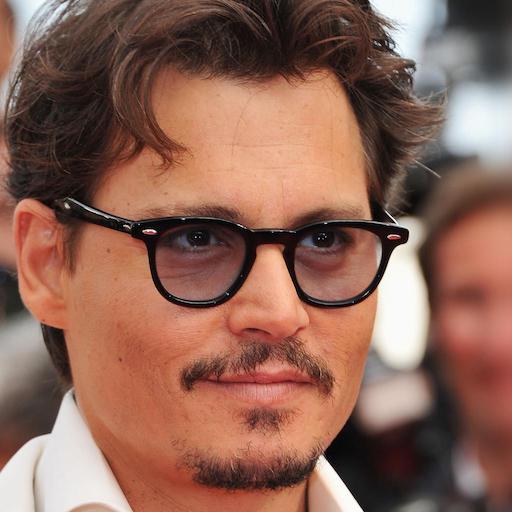} &
    \includegraphics[width=0.184\linewidth]{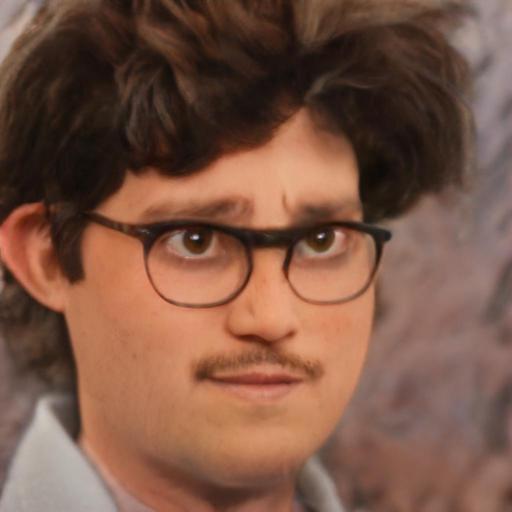} &
    \includegraphics[width=0.184\linewidth]{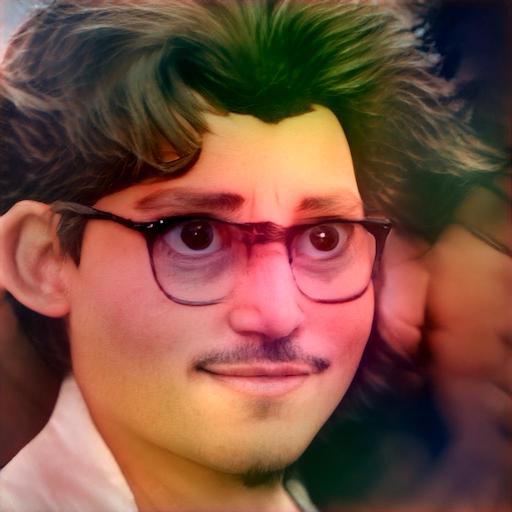} &
    \includegraphics[width=0.184\linewidth]{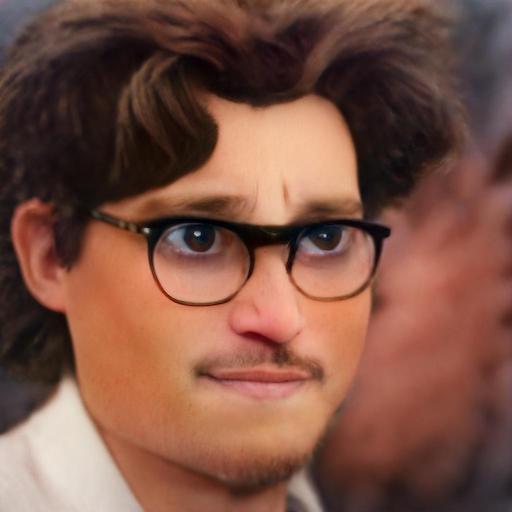} &
    \includegraphics[width=0.184\linewidth]{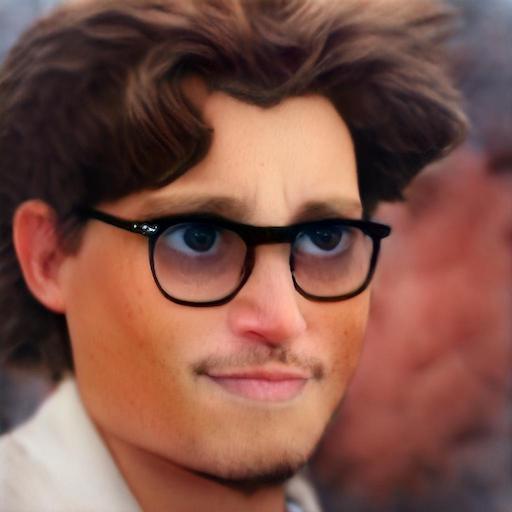}  \\

     \vspace{-0.08cm}

    \raisebox{0.125in}{\rotatebox{90}{Disney}} &
    \includegraphics[width=0.184\linewidth]{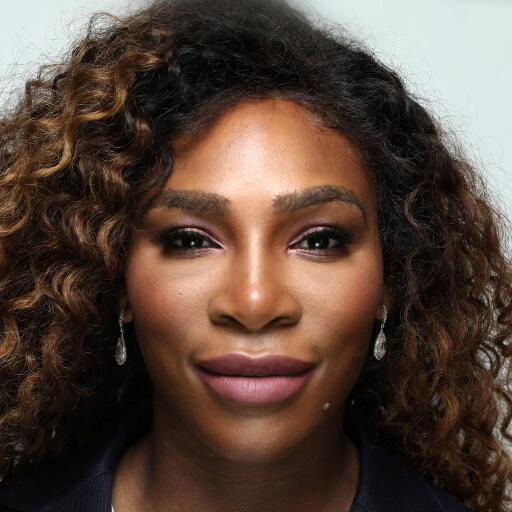} &
    \includegraphics[width=0.184\linewidth]{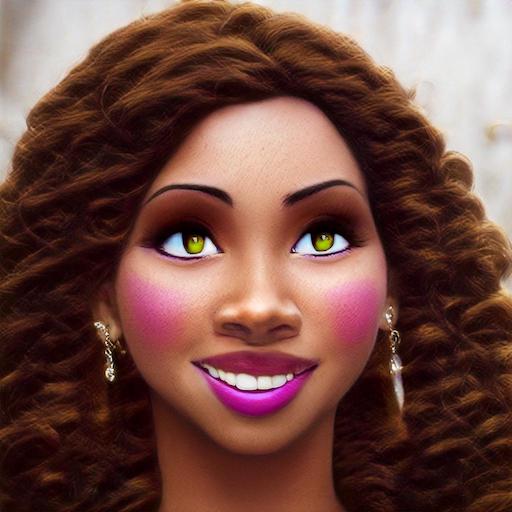} &
    \includegraphics[width=0.184\linewidth]{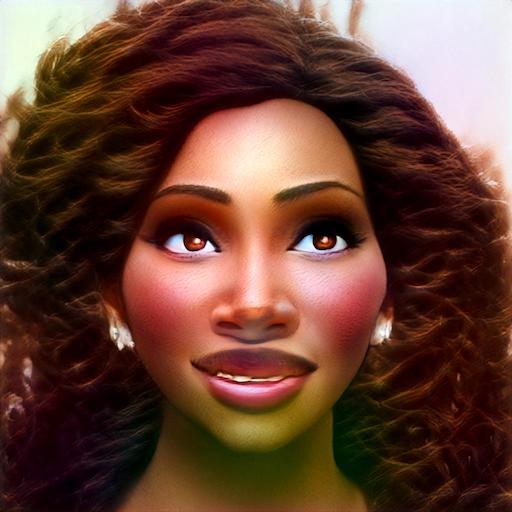} &
    \includegraphics[width=0.184\linewidth]{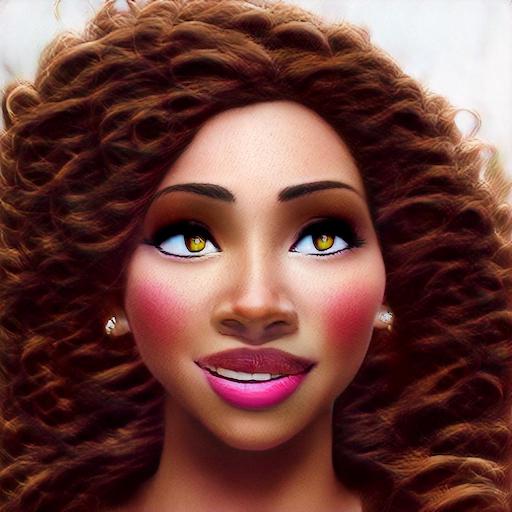} &
    \includegraphics[width=0.184\linewidth]{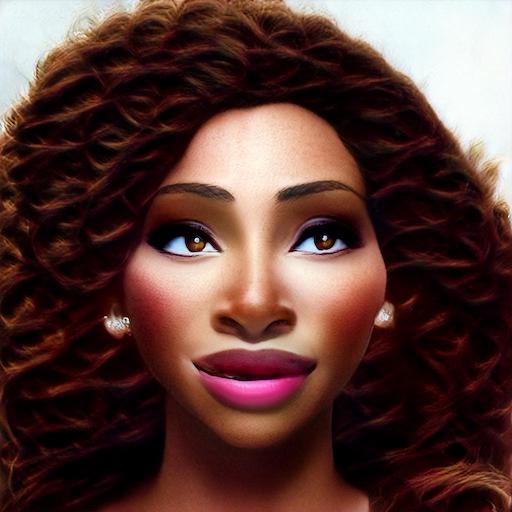}  \\
    
     \vspace{-0.08cm}
    
    \raisebox{0.175in}{\rotatebox{90}{Pixar}} &
    \includegraphics[width=0.184\linewidth]{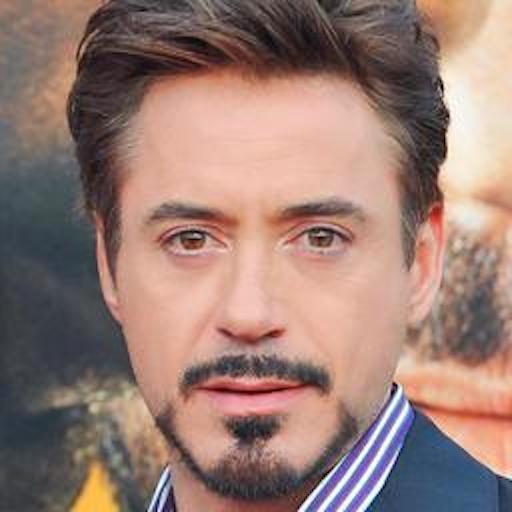} &
    \includegraphics[width=0.184\linewidth]{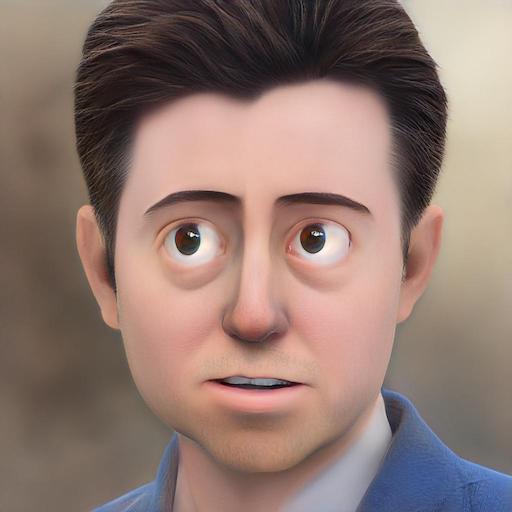} &
    \includegraphics[width=0.184\linewidth]{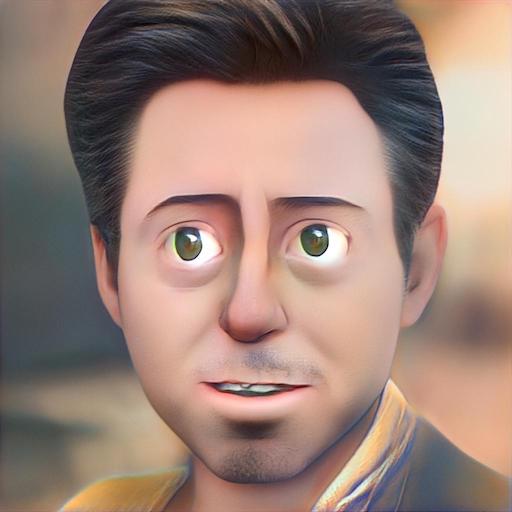} &
    \includegraphics[width=0.184\linewidth]{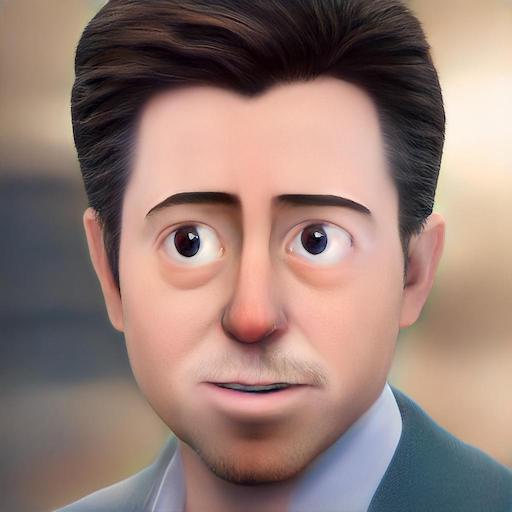} &
    \includegraphics[width=0.184\linewidth]{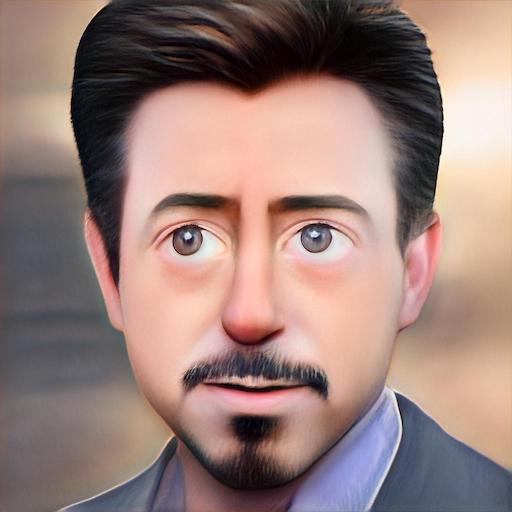}  \\
    
     \vspace{-0.08cm}

    & Input & e4e & $\text{ReStyle}_{pSp}$ & $\text{ReStyle}_{e4e}$ & HyperStyle
    
    \end{tabular}
    }
    \vspace{-0.225cm}
    \caption{Weight offsets predicted by HyperStyle trained on FFHQ are also applicable for modifying fine-tuned generators (e.g., Toonify \cite{pinkney2020resolution} and StyleGAN-NADA \cite{gal2021stylegannada}). Our refinement leads to improved identity preservation while retaining target style.}
    \vspace{-0.35cm}
    \label{fig:domain_adaption}
\end{figure}

\cref{fig:domain_adaption} shows examples of applying weight offsets over various fine-tuned generators. 
As shown, when no offsets are applied, important details are lost. However, HyperStyle leads to more faithful translations preserving identity without harming the target style.
Importantly, the translations are attained with no domain-specific hypernetwork training.

\begin{figure}
\setlength{\tabcolsep}{1pt}
    \centering
    { \small 
    \begin{tabular}{c c c c c c}

    \vspace{-0.08cm}

    \includegraphics[width=0.1535\linewidth]{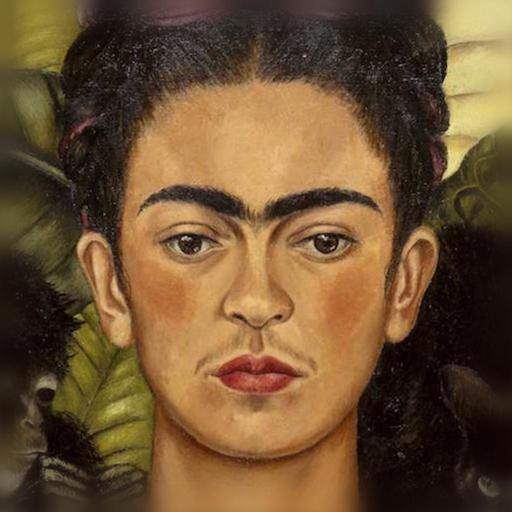} &
    \includegraphics[width=0.1535\linewidth]{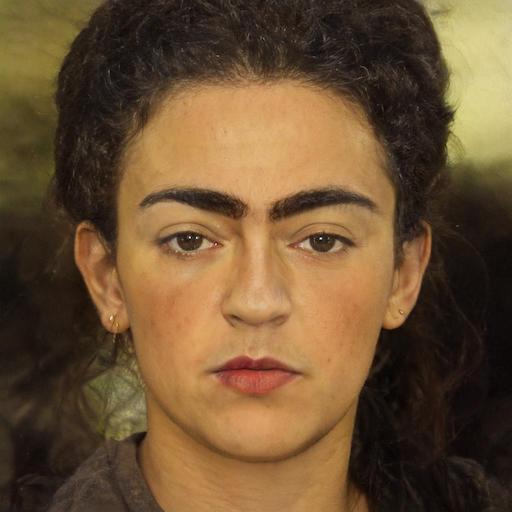} &
    \includegraphics[width=0.1535\linewidth]{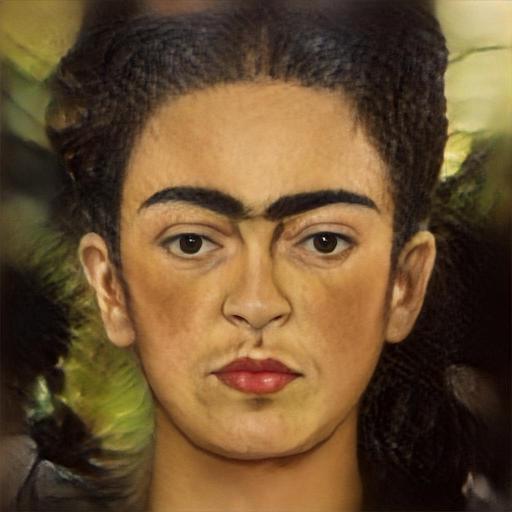} &
    \includegraphics[width=0.1535\linewidth]{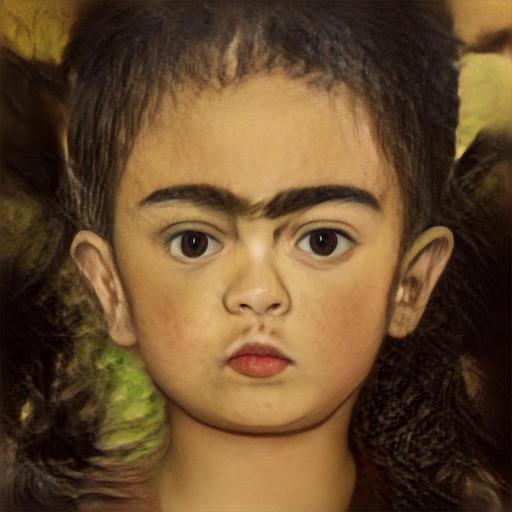} &
    \includegraphics[width=0.1535\linewidth]{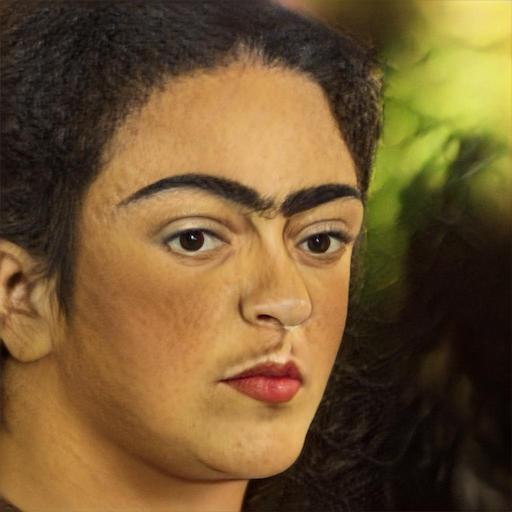} &
    \includegraphics[width=0.1535\linewidth]{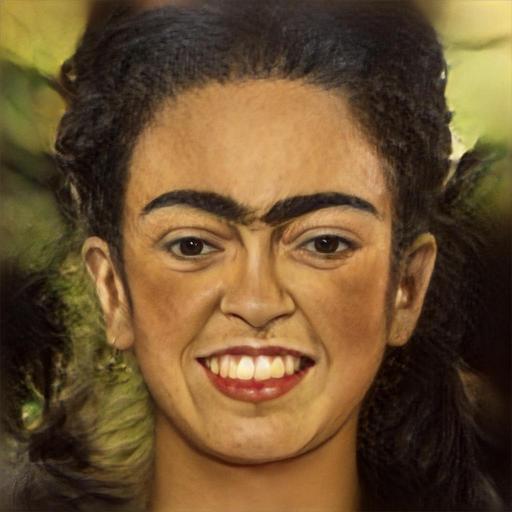} \\

    \vspace{-0.08cm}

    \includegraphics[width=0.1535\linewidth]{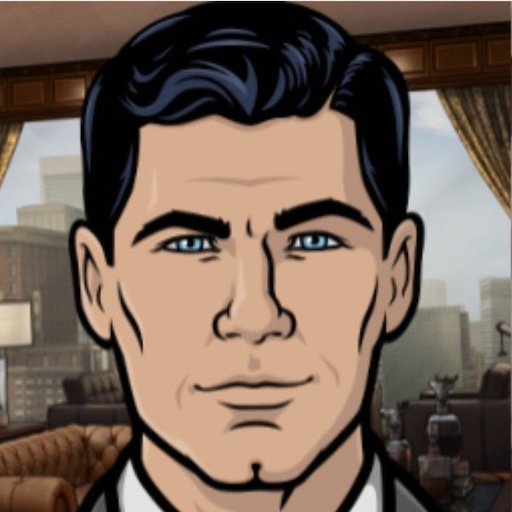} &
    \includegraphics[width=0.1535\linewidth]{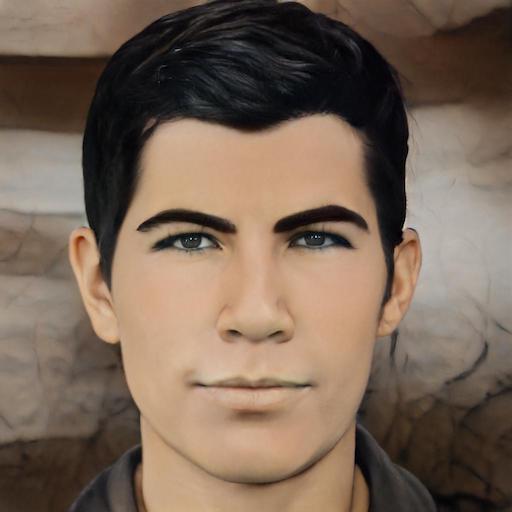} &
    \includegraphics[width=0.1535\linewidth]{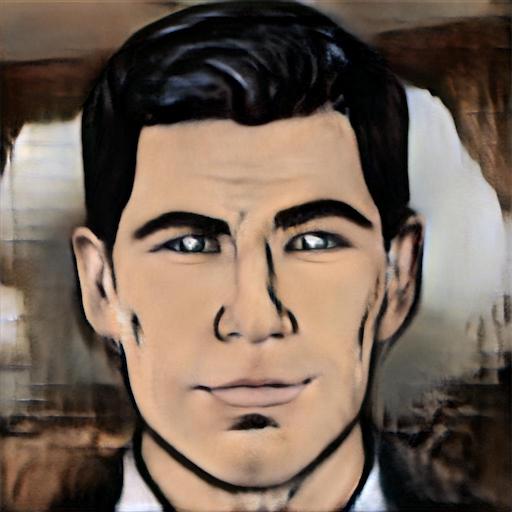} &
    \includegraphics[width=0.1535\linewidth]{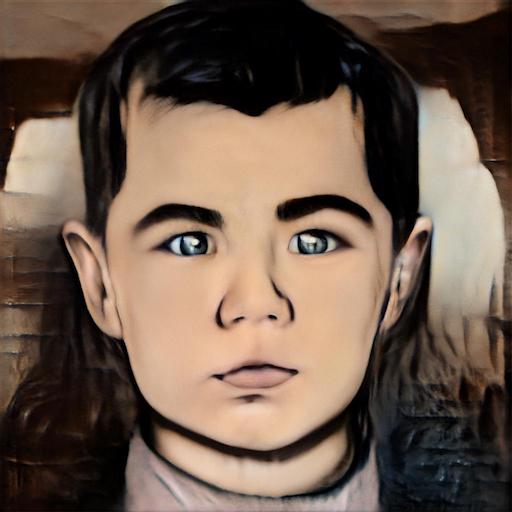} &
    \includegraphics[width=0.1535\linewidth]{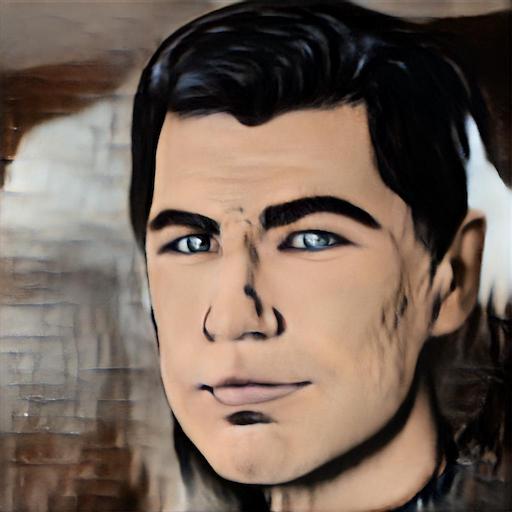} &
    \includegraphics[width=0.1535\linewidth]{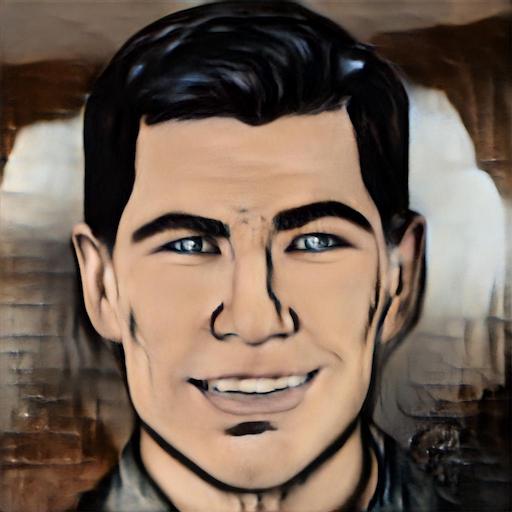} \\
    
    \vspace{-0.08cm}
    
    \includegraphics[width=0.1535\linewidth]{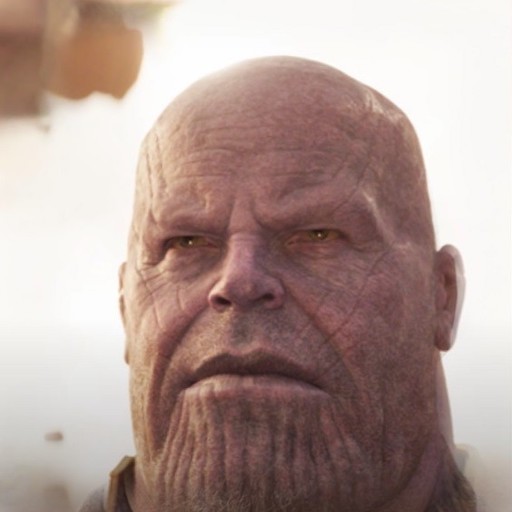} &
    \includegraphics[width=0.1535\linewidth]{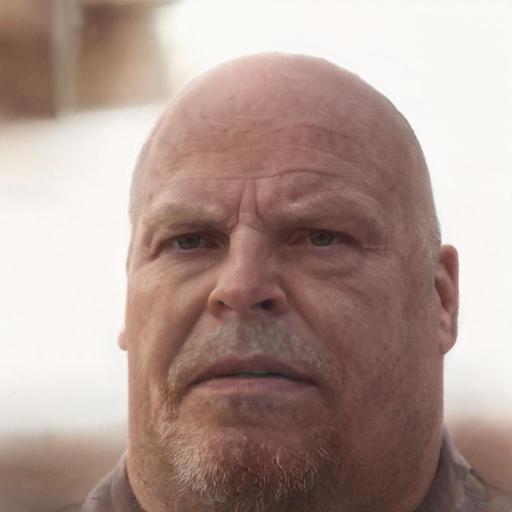} &
    \includegraphics[width=0.1535\linewidth]{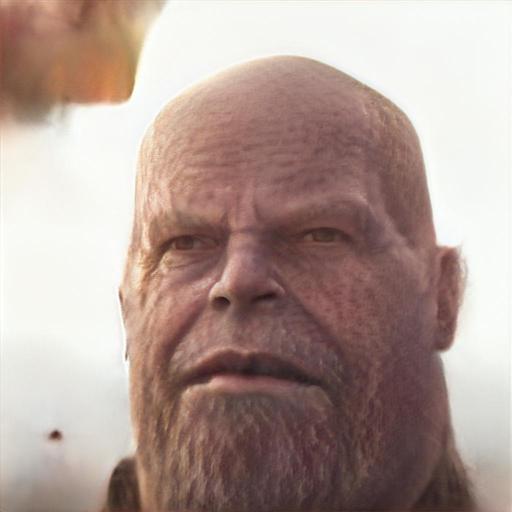} &
    \includegraphics[width=0.1535\linewidth]{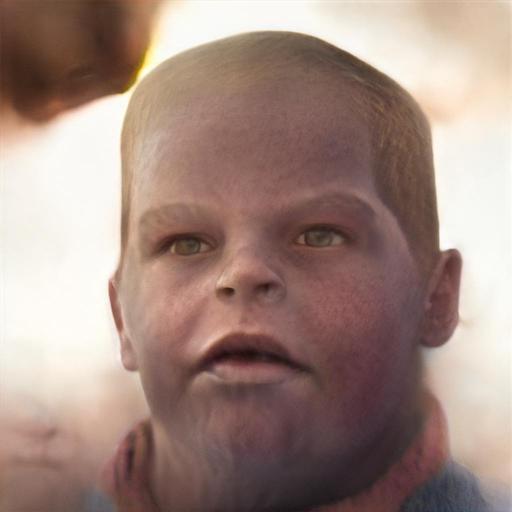} &
    \includegraphics[width=0.1535\linewidth]{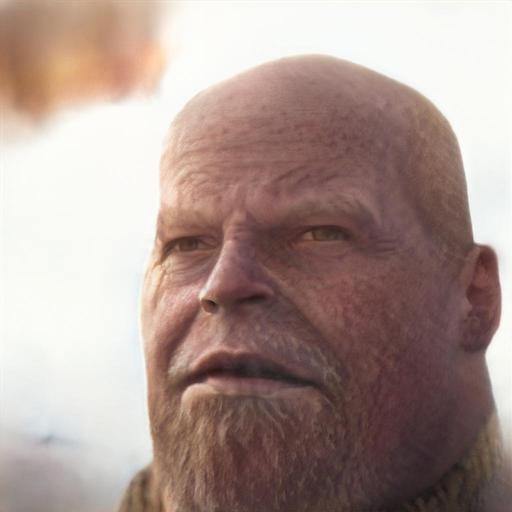} &
    \includegraphics[width=0.1535\linewidth]{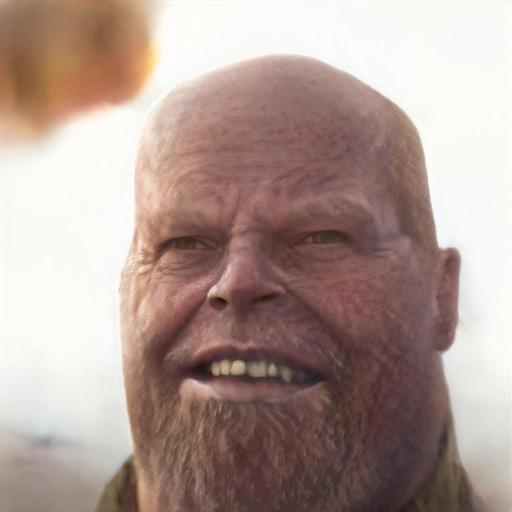} \\

    \vspace{-0.08cm}

    \includegraphics[width=0.1535\linewidth]{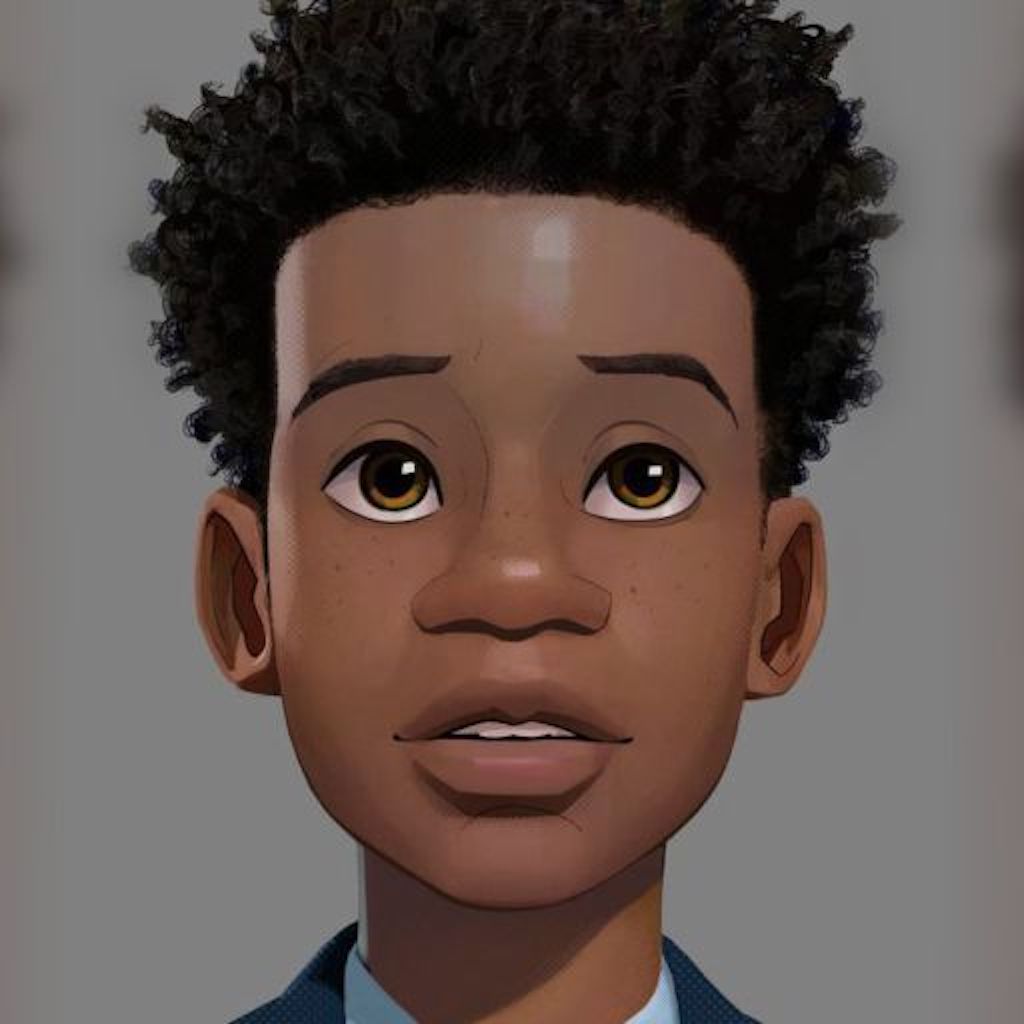} &
    \includegraphics[width=0.1535\linewidth]{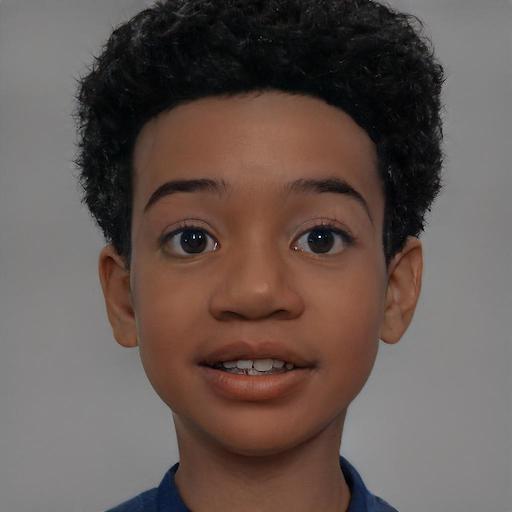} &
    \includegraphics[width=0.1535\linewidth]{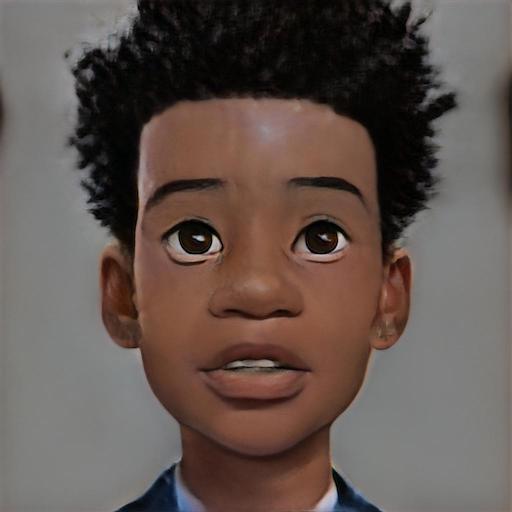} &
    \includegraphics[width=0.1535\linewidth]{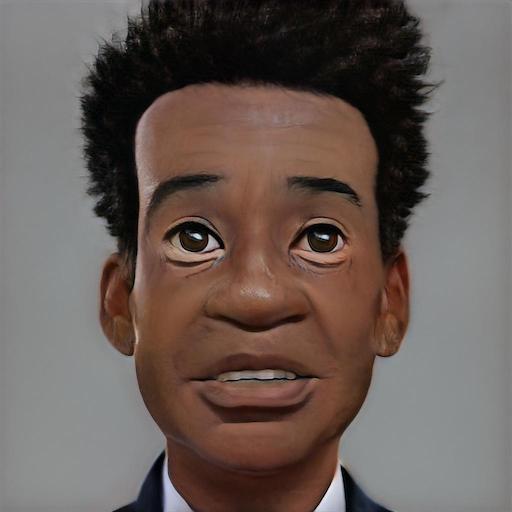} &
    \includegraphics[width=0.1535\linewidth]{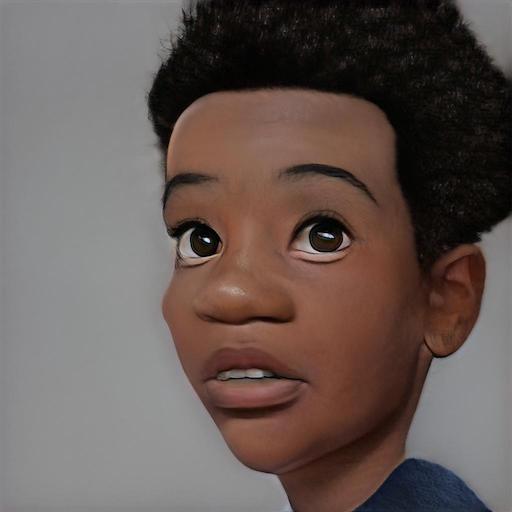} &
    \includegraphics[width=0.1535\linewidth]{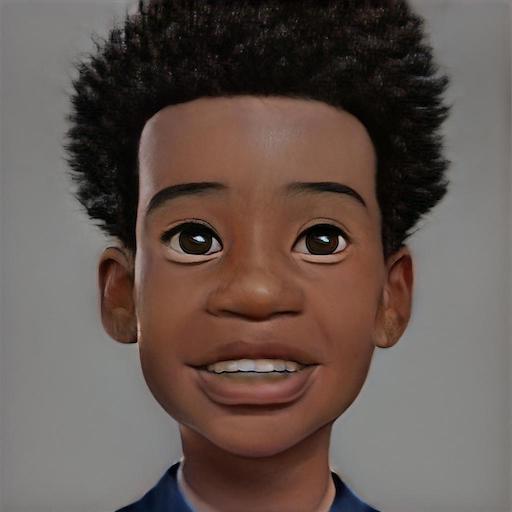} \\
    
    Input & e4e & HyperStyle & \multicolumn{3}{c}{ ------ HyperStyle Edits ------ }
    
    \end{tabular}
    }
    \vspace{-0.325cm}
    \caption{Trained only on real images, our method successfully generalizes to challenging styles not observed during training, 
    even without generator fine-tuning.
    } \vspace{-0.5cm}
    \label{fig:out_of_domain}

\end{figure}

\vspace{-0.45cm}
\paragraph{Editing Out-of-Domain Images}
To this point, we have discussed handling images from the same domain as used for training. 
If our hypernetwork has indeed learned to generalize, it should not be sensitive to the domain of the input.
As may be expected, standard encoders cannot handle out-of-domain images well (see \cref{fig:out_of_domain} for e4e and Appendix~\ref{supp:comparisons} for others). 
By adjusting the pre-trained generator towards a given out-of-domain input, HyperStyle enables editing diverse images, without explicitly training a new generator on their domain.
This points to improved expressiveness and generalization. It seems the hypernetwork does not just fix poorly reconstructed attributes but learns to adapt the generator in a more general sense.
We find these results to be a promising direction for manipulating out-of-domain images without having to train new generators or perform lengthy per-image tuning. 

\section{Conclusions}
We introduced HyperStyle, a novel approach for StyleGAN inversion. We leverage recent advancements in hypernetworks to achieve optimization-level reconstructions at encoder-like inference times. 
In a sense, HyperStyle \textit{learns} to efficiently optimize the generator for a given target image. Doing so mitigates the reconstruction-editability trade-off and enables the effective use of existing editing techniques on a wide range of inputs.
In addition, HyperStyle generalizes surprisingly well, even to out-of-domain images neither the hypernetwork nor the generator have seen during training.
Looking forward, further broadening generalization away from the training domain is highly desirable. This includes robustness to unaligned images and unstructured domains. The former may potentially be addressed through StyleGAN3~\cite{aliasfreeKarras2021} while the latter would probably warrant training on a richer set of images.
In summary, we believe this approach to be an essential step towards interactive and semantic in-the-wild image editing and may open the door for many intriguing real-world scenarios.

\section*{Acknowledgements}
We would like to thank Or Patashnik, Elad Richardson, Daniel Roich, Rotem Tzaban, and Oran Lang for their early feedback and discussions. This work was supported in part by Len Blavatnik and the Blavatnik Family Foundation.

{\small
\bibliographystyle{ieee_fullname}
\bibliography{main}
}

\clearpage
\appendix
\appendixpage
\section{Broader Impact}
HyperStyle enables accurate and highly editable inversions of real images. While our tool aims to empower content creators, it can also be used to generate more convincing deep-fakes~\cite{suwajanakorn2017synthesizing} and aid in the spread of disinformation~\cite{vaccari2020disinformation}. However, powerful tools already exist for the detection of GAN-synthesized imagery~\cite{wang2019cnngenerated,mandelli2020training}, for example through frequency analysis~\cite{dzanic2019fourier,swagan2021gal}. These tools continually evolve, which gives us hope that any potential misuse of our method can be mitigated.

Another cause for concern is the bias that generative networks inherit from their training data~\cite{merler2019diversity}. Our model was similarly trained on such a biased set, and as a result, may display degraded performance when dealing with images from minority classes~\cite{menon2020pulse}. However, we have demonstrated that our model successfully generalizes beyond its training set, and allows us to similarly shift the GAN beyond its original domain. These properties allow us to better preserve minority traits when compared to prior works, and we hope that this benefit would similarly enable fairer treatment of minorities in downstream tasks.

\section{Ablation Study: Qualitative Comparisons}~\label{supp:ablation_study}
\begin{figure}
\setlength{\tabcolsep}{1pt}
    \centering
    { \small 
    \begin{tabular}{c c c c c c}

    \raisebox{0.125in}{\rotatebox{90}{Input}} &
    \includegraphics[width=0.185\linewidth]{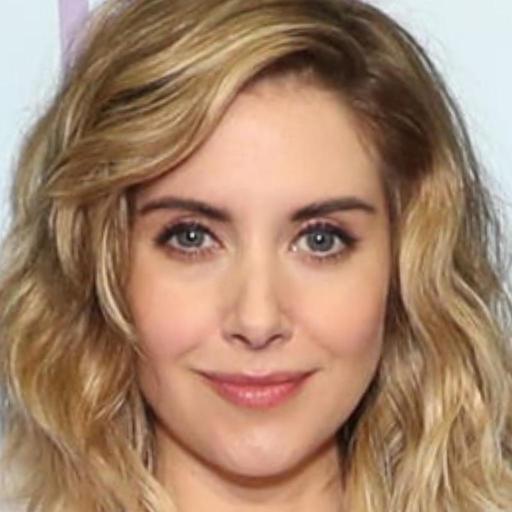} &
    \includegraphics[width=0.185\linewidth]{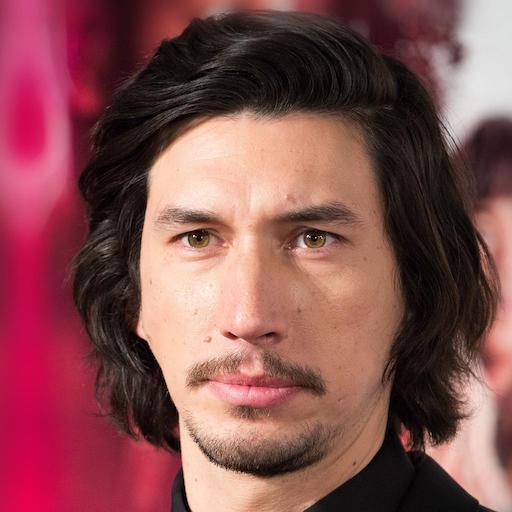} &
    \includegraphics[width=0.185\linewidth]{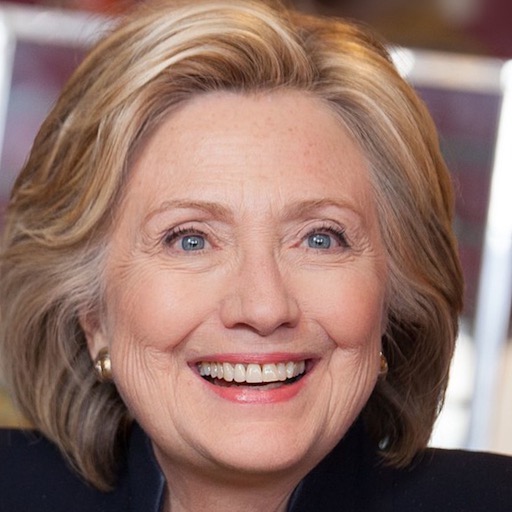} &
    \includegraphics[width=0.185\linewidth]{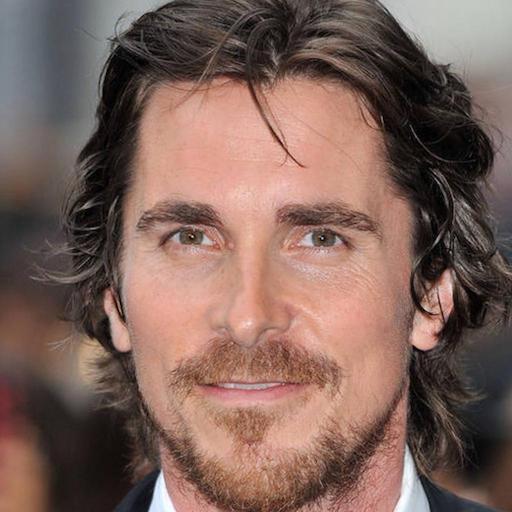} &
    \includegraphics[width=0.185\linewidth]{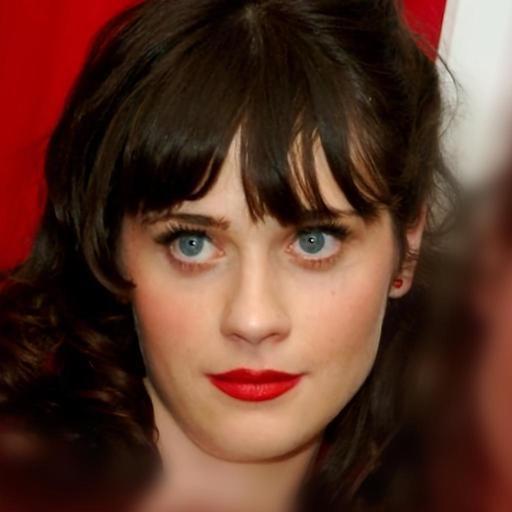} \\

    \raisebox{0.025in}{\rotatebox{90}{HyperStyle}} &
    \includegraphics[width=0.185\linewidth]{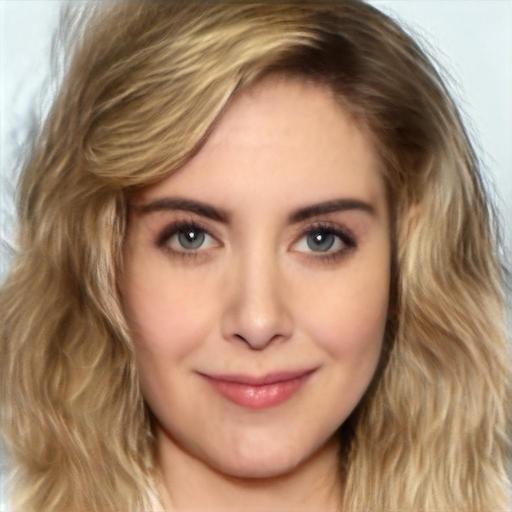} &
    \includegraphics[width=0.185\linewidth]{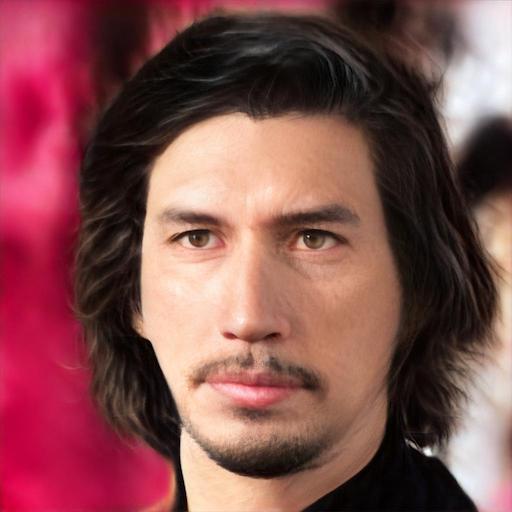} &
    \includegraphics[width=0.185\linewidth]{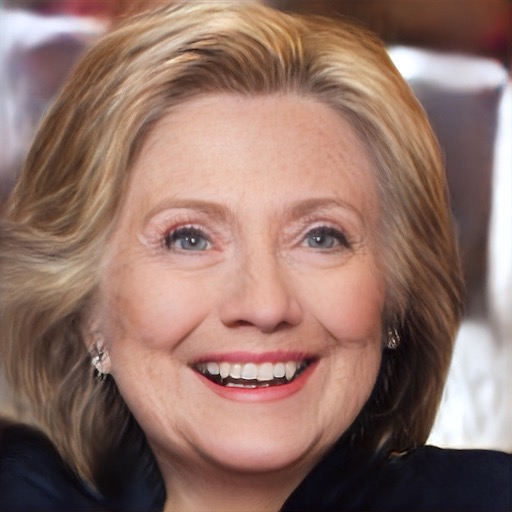} &
    \includegraphics[width=0.185\linewidth]{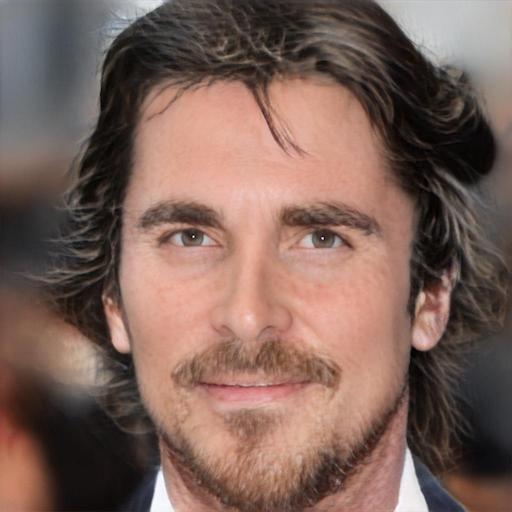} &
    \includegraphics[width=0.185\linewidth]{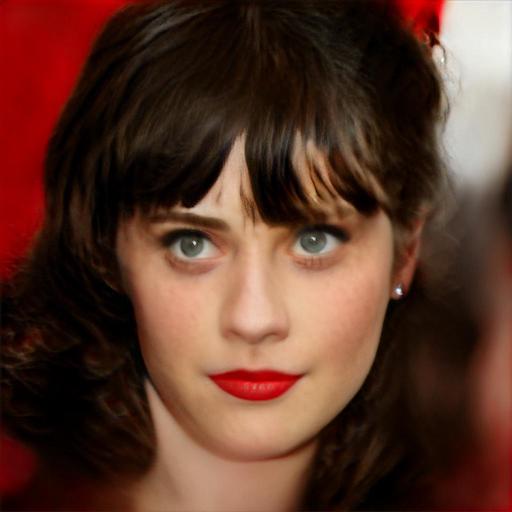} \\
    
    \raisebox{0.125in}{\rotatebox{90}{No IR}} &
    \includegraphics[width=0.185\linewidth]{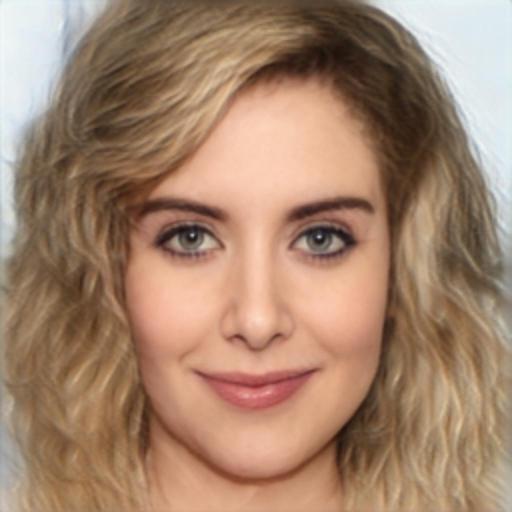} &
    \includegraphics[width=0.185\linewidth]{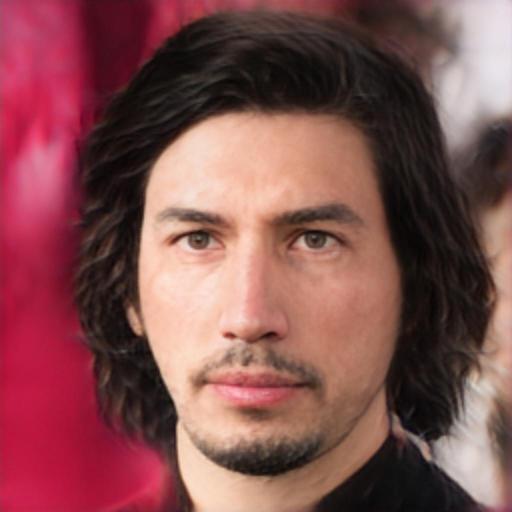} &
    \includegraphics[width=0.185\linewidth]{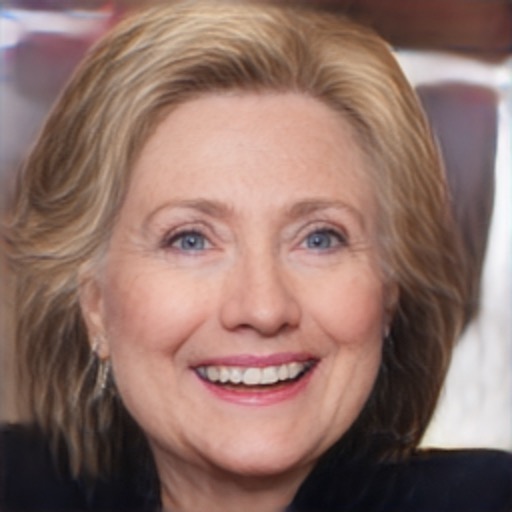} &
    \includegraphics[width=0.185\linewidth]{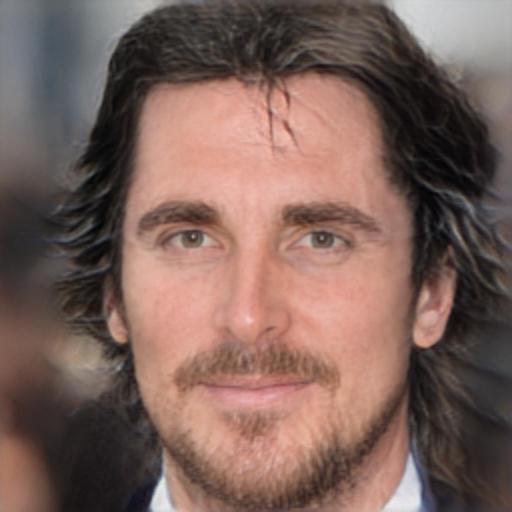} &
    \includegraphics[width=0.185\linewidth]{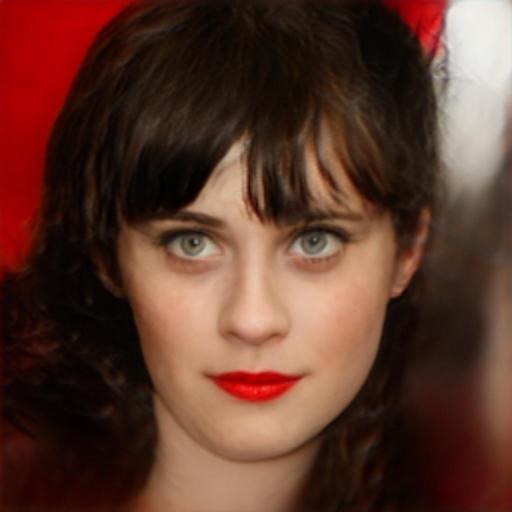} \\
    
    \end{tabular}
    }
    \vspace{-0.1cm}
    \caption{Reconstruction comparison of HyperStyle with and without the iterative refinement (IR) training and inference scheme. As shown, gradually predicting the desired weight offsets results in sharper images with less artifacts, particularly in finer details along the hair, for example. Best viewed zoomed-in.}
    \label{fig:ablation_iterations}
    \vspace{0.2cm}

\end{figure}

\begin{figure}[h!]
\setlength{\tabcolsep}{1pt}
    \centering
    { \small 
    \begin{tabular}{c c c c c}

    \includegraphics[width=0.185\linewidth]{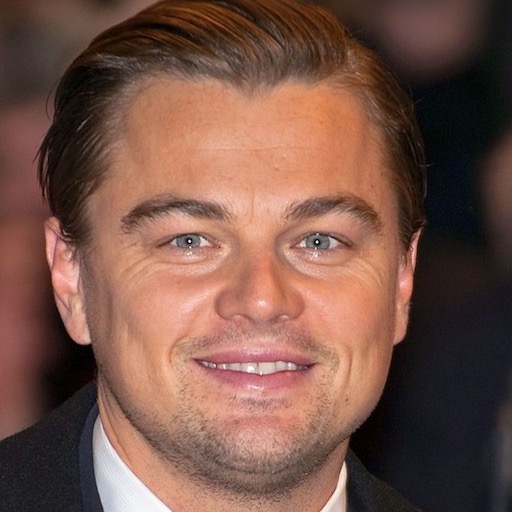} &
    \includegraphics[width=0.185\linewidth]{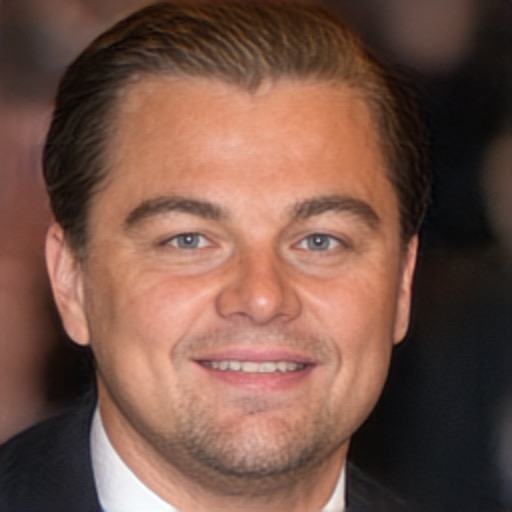} &
    \includegraphics[width=0.185\linewidth]{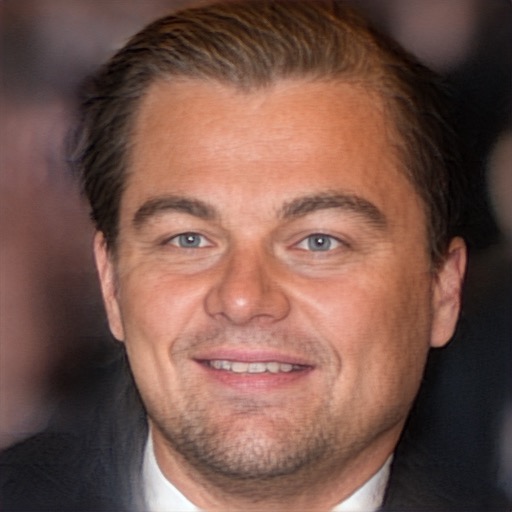} &
    \includegraphics[width=0.185\linewidth]{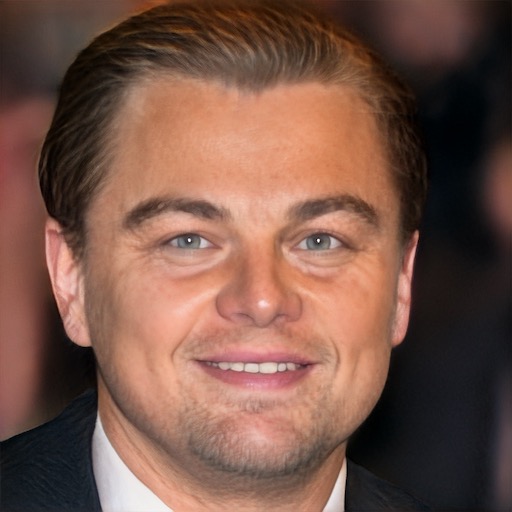} &
    \includegraphics[width=0.185\linewidth]{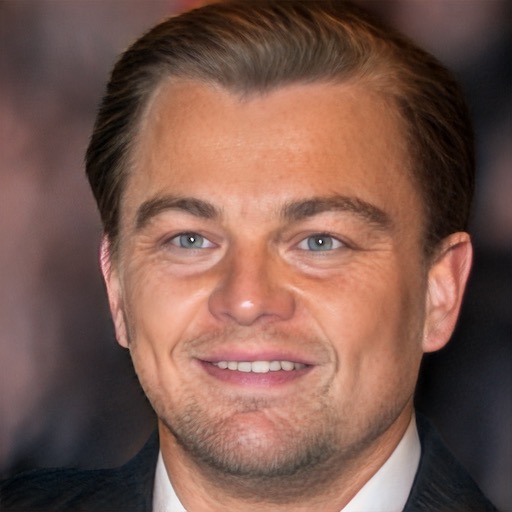} \\

    \includegraphics[width=0.185\linewidth]{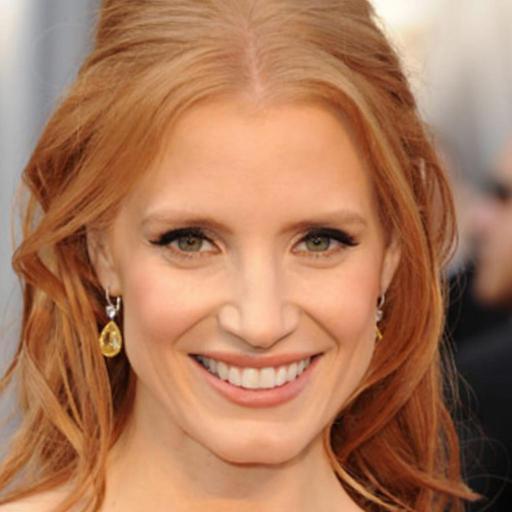} &
    \includegraphics[width=0.185\linewidth]{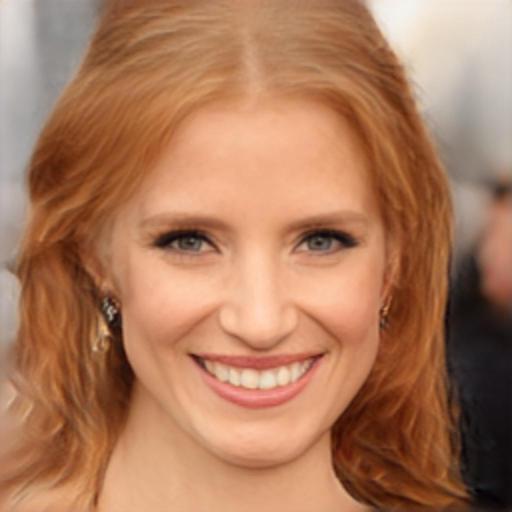} &
    \includegraphics[width=0.185\linewidth]{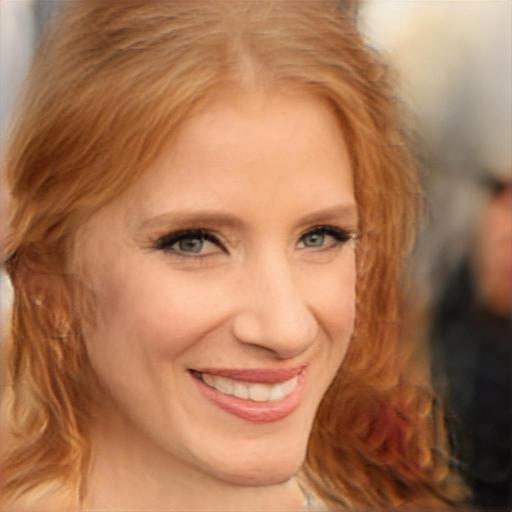} &
    \includegraphics[width=0.185\linewidth]{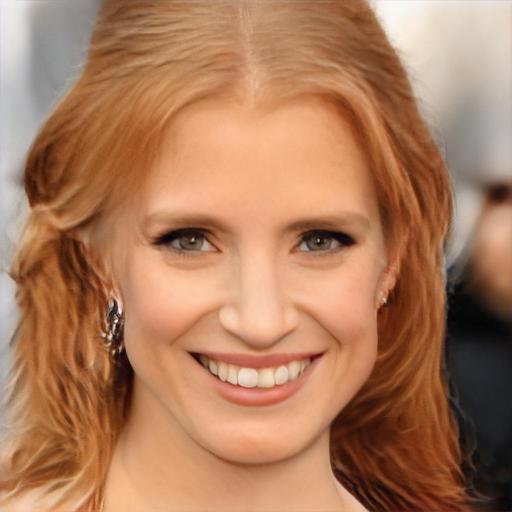} &
    \includegraphics[width=0.185\linewidth]{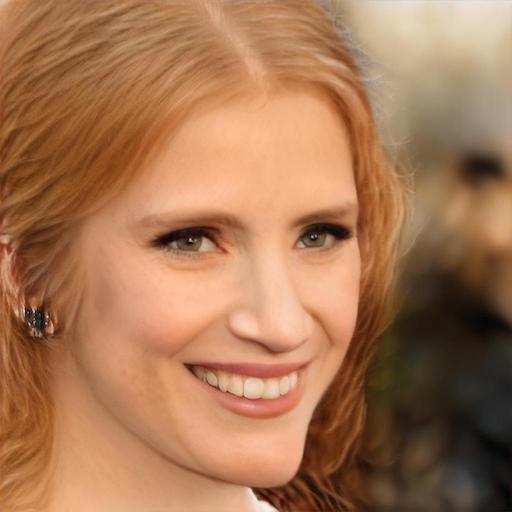} \\

    \includegraphics[width=0.185\linewidth]{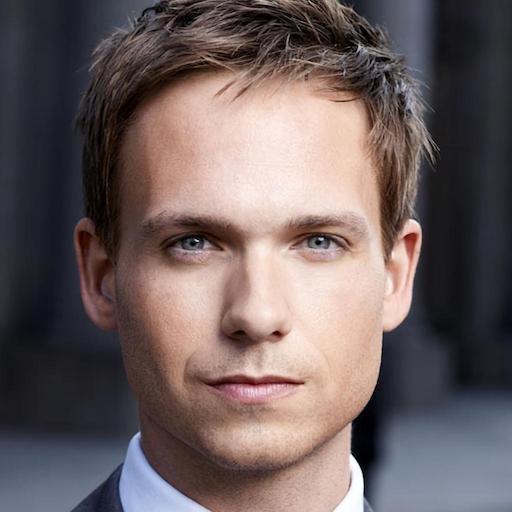} &
    \includegraphics[width=0.185\linewidth]{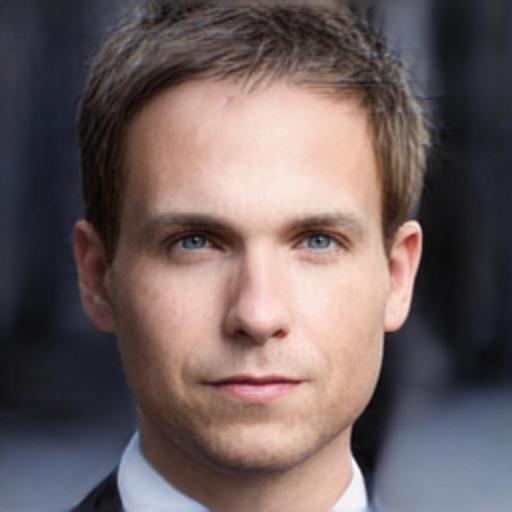} &
    \includegraphics[width=0.185\linewidth]{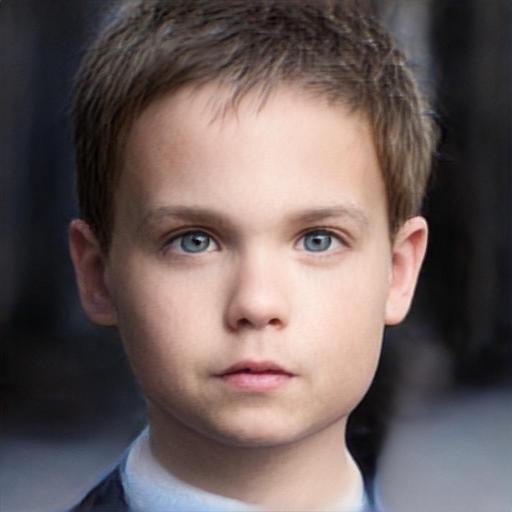} &
    \includegraphics[width=0.185\linewidth]{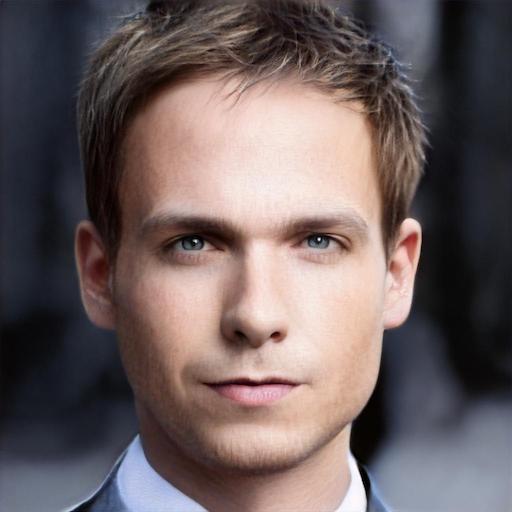} &
    \includegraphics[width=0.185\linewidth]{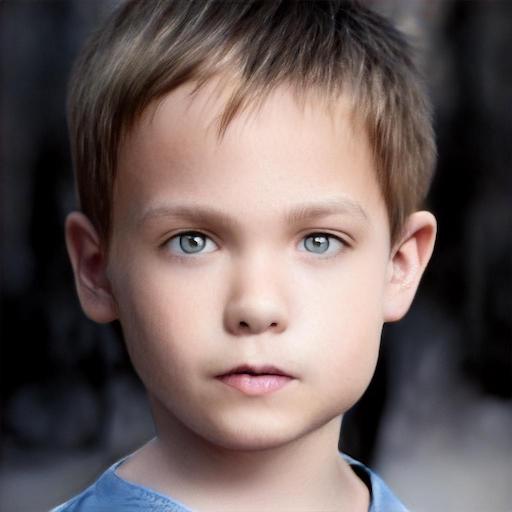} \\
    
    \includegraphics[width=0.185\linewidth]{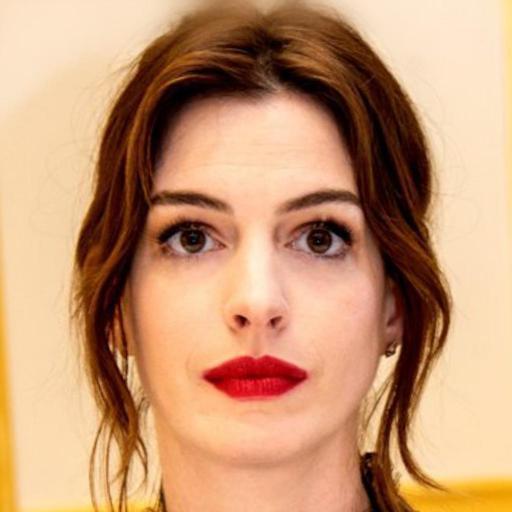} &
    \includegraphics[width=0.185\linewidth]{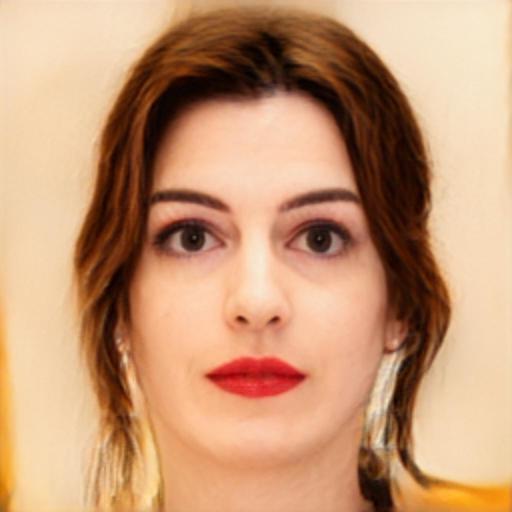} &
    \includegraphics[width=0.185\linewidth]{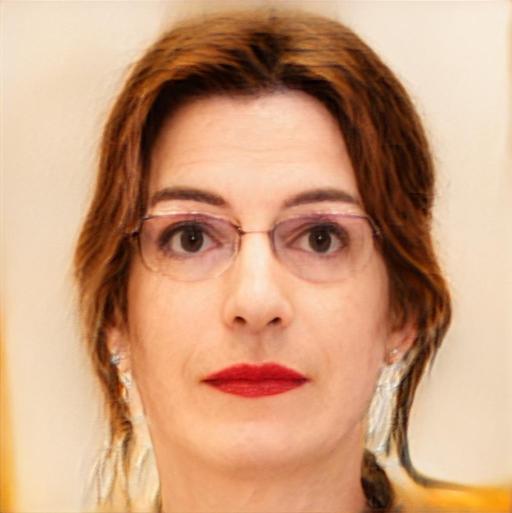} &
    \includegraphics[width=0.185\linewidth]{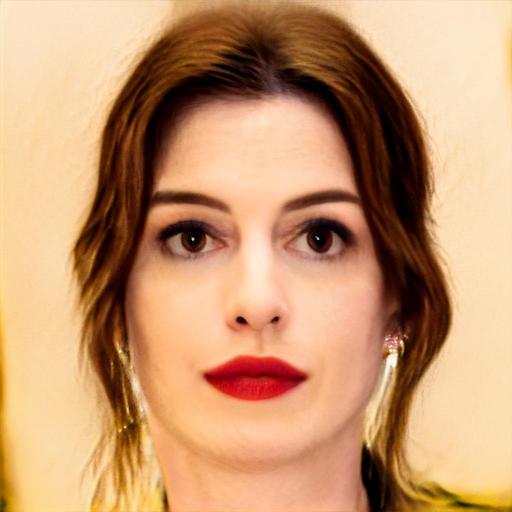} &
    \includegraphics[width=0.185\linewidth]{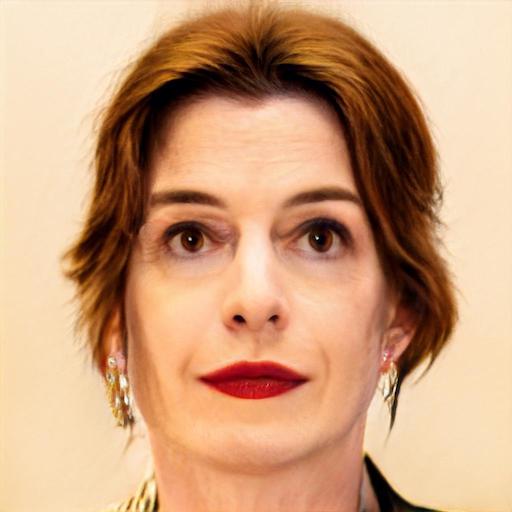} \\
    
    \multirow{1}{*}{Input} & 
    \begin{tabular}{@{}c@{}}All Layers \\ Inversion \end{tabular} & 
    \begin{tabular}{@{}c@{}}All Layers \\ Edit \end{tabular} &
    \begin{tabular}{@{}c@{}}HyperStyle \\ Inversion \end{tabular} &
    \begin{tabular}{@{}c@{}}HyperStyle \\ Edit \end{tabular} \\
    
    \end{tabular}
    }
    \vspace{-0.1cm}
    \caption{Comparison of HyperStyle trained with and without altering the toRGB layers of StyleGAN, denoted by ``All Layers" and ``HyperStyle", respectively. Altering toRGB layers produces more noticeable editing artifacts when making a global change.
    }
    \label{fig:ablation_torgb}
    \vspace{0.1cm}

\end{figure}

In Sec.~4.3 of the main paper, we presented a quantitative ablation study to validate the design choices of our hypernetworks. We now turn to provide visual comparisons to complement this. 
First, we illustrate the effectiveness of the iterative refinement scheme in \cref{fig:ablation_iterations}. Observe that iteratively predicting the weight offsets results in sharper reconstructions. This is best reflected in the preservation of fine details, most notably in the reconstruction of hairstyle and facial hair. 
In \cref{fig:ablation_torgb}, we show that altering the toRGB convolutional layers harms the editability of the resulting inversions. This is most noticeable in edits requiring global changes, such as pose and age. For instance, altering the pose in the second row results in blurred edits. Additionally, when modifying age in the bottom two rows, HyperStyle succeeds in realistically altering the clothing ($3$rd row) and hairstyle ($4$th row). 


\begin{table}
    \small
    \centering
    \vspace{0.1cm}
    \textbf{Cars} \\
    \vspace{0.025cm}
    \begin{tabular}{l@{\, }| c c c c c}
    \toprule
    Method & $\uparrow$ MS-SSIM & $\downarrow$ LPIPS & $\downarrow$ $L_2$ & $\downarrow$ Time (s) \\
    \midrule
    StyleGAN2~\cite{karras2020analyzing} &
    \multicolumn{1}{c}{$0.79$} &
    \multicolumn{1}{c}{$0.16$} &
    \multicolumn{1}{c}{$0.06$} &
    \multicolumn{1}{c}{$198.9$} \\ 
    PTI~\cite{roich2021pivotal} &
    \multicolumn{1}{c}{$0.93$} &
    \multicolumn{1}{c}{$0.11$} &
    \multicolumn{1}{c}{$0.01$} &
    \multicolumn{1}{c}{$33.71$} \\ 
    \midrule
    pSp~\cite{richardson2020encoding} &
    \multicolumn{1}{c}{$0.58$} &
    \multicolumn{1}{c}{$0.29$} &
    \multicolumn{1}{c}{$0.10$} &
    \multicolumn{1}{c}{$0.071$} \\
    e4e~\cite{tov2021designing} & 
    \multicolumn{1}{c}{$0.53$} &
    \multicolumn{1}{c}{$0.32$} &
    \multicolumn{1}{c}{$0.12$} &
    \multicolumn{1}{c}{$0.071$} \\
    $\text{ReStyle}_{pSp}$~\cite{alaluf2021restyle} & 
    \multicolumn{1}{c}{$0.66$} & 
    \multicolumn{1}{c}{$0.25$} & 
    \multicolumn{1}{c}{$0.07$} &
    \multicolumn{1}{c}{$0.359$} \\
    $\text{ReStyle}_{e4e}$~\cite{alaluf2021restyle} & 
    \multicolumn{1}{c}{$0.60$} & 
    \multicolumn{1}{c}{$0.29$} & 
    \multicolumn{1}{c}{$0.09$} &
    \multicolumn{1}{c}{$0.359$} \\
    \midrule
    HyperStyle & 
    \multicolumn{1}{c}{$0.67$} & 
    \multicolumn{1}{c}{$0.27$} & 
    \multicolumn{1}{c}{$0.07$} &
    \multicolumn{1}{c}{$0.491$} \\
    \bottomrule
    \end{tabular}
    \vspace{-0.2cm}
    \caption{Quantitative reconstruction results on the cars domain, computed over the Stanford Cars dataset~\cite{KrauseStarkDengFei-Fei_3DRR2013}.} 
    \label{tb:quantitative_inversion_cars}
\end{table}
\begin{table}
    \small
    \centering
    \vspace{0.1cm}
    \textbf{AFHQ Wild} \\
    \vspace{0.025cm}
    \begin{tabular}{l@{\, }| c c c c c}
    \toprule
    Method & $\uparrow$ MS-SSIM & $\downarrow$ LPIPS & $\downarrow$ $L_2$ & $\downarrow$ Time (s) \\
    \midrule
    StyleGAN2~\cite{karras2020analyzing} &
    \multicolumn{1}{c}{$0.82$} &
    \multicolumn{1}{c}{$0.13$} &
    \multicolumn{1}{c}{$0.03$} &
    \multicolumn{1}{c}{$203.0$} \\ 
    PTI~\cite{roich2021pivotal} &
    \multicolumn{1}{c}{$0.93$} &
    \multicolumn{1}{c}{$0.08$} &
    \multicolumn{1}{c}{$0.01$} &
    \multicolumn{1}{c}{$33.71$} \\ 
    \midrule
    pSp~\cite{richardson2020encoding} &
    \multicolumn{1}{c}{$0.51$} &
    \multicolumn{1}{c}{$0.35$} &
    \multicolumn{1}{c}{$0.13$} &
    \multicolumn{1}{c}{$0.075$} \\
    e4e~\cite{tov2021designing} & 
    \multicolumn{1}{c}{$0.47$} &
    \multicolumn{1}{c}{$0.36$} &
    \multicolumn{1}{c}{$0.14$} &
    \multicolumn{1}{c}{$0.075$} \\
    $\text{ReStyle}_{pSp}$~\cite{alaluf2021restyle} & 
    \multicolumn{1}{c}{$0.57$} &
    \multicolumn{1}{c}{$0.21$} & 
    \multicolumn{1}{c}{$0.05$} &
    \multicolumn{1}{c}{$0.303$} \\
    $\text{ReStyle}_{e4e}$~\cite{alaluf2021restyle} & 
    \multicolumn{1}{c}{$0.52$} &
    \multicolumn{1}{c}{$0.25$} & 
    \multicolumn{1}{c}{$0.07$} &
    \multicolumn{1}{c}{$0.303$} \\
    \midrule
    HyperStyle & 
    \multicolumn{1}{c}{$0.56$} &
    \multicolumn{1}{c}{$0.24$} & 
    \multicolumn{1}{c}{$0.06$} &
    \multicolumn{1}{c}{$0.551$} \\
    \bottomrule
    \end{tabular}
    \vspace{-0.2cm}
    \caption{Quantitative reconstruction results on the wild animals domain, computed over the AFHQ Wild~\cite{choi2020stargan} dataset.} 
    \label{tb:quantitative_inversion_afhq_wild}
\end{table}

\section{Additional Quantitative Results}
Following the quantitative reconstruction metrics provided in the main paper on the human facial domain, we provide quantitative results on the cars domain and wild animals domain in \cref{tb:quantitative_inversion_cars,tb:quantitative_inversion_afhq_wild}. 

\section{The HyperStyle Architecture}~\label{supp:hyperstyle_architecture}
Given the $6$-channel input, the HyperStyle architecture begins with a shared backbone which outputs a single $16\times16\times512$ feature map. 
This feature map is then passed to each \textit{Refinement Block}, which further down-samples the feature map using a set of $2$-strided $3\times3$ convolutions with LeakyReLU activations to obtain a $1\times1\times512$ representation. 
The standard Refinement Block then directly predicts the $1\times1\times C^{in}_\ell\times C^{out}_\ell$ using a single fully-connected layer. 

In addition to the standard Refinement Block, we introduce a \textit{Shared Refinement Block} that is shared between multiple hypernetwork layers.
These Shared Refinement Blocks make use of two fully-connected layers whose weights are shared between different output heads. The first fully-connected layer transforms the $1\times1\times512$ tensor to a $512\times512$ intermediate representation. This is followed by a per-channel fully-connected layer which maps each $1\times512$ channel to a $1\times1\times512$ tensor, resulting in the final $1\times1\times512\times512$ dimensional offsets. We apply these shared blocks to all generator layers with a convolutional dimension of $3\times3\times512\times512$. 

In both cases, we obtain a predicted offset of size $1\times1\times C^{in}_\ell\times C^{out}_\ell$ for a given generator layer $\ell$. This tensor is then broadcasted channel-wise to match the $k_\ell \times k_\ell$ kernel dimension of the generator's convolutional filters, resulting in a final offset of size $k_\ell \times k_\ell \times C^{in}_\ell\times C^{out}_\ell$. This final tensor can then be used to update the convolutional weights using Eq. (6) presented in the main paper.
We provide a breakdown of the two architectures in \cref{tb:supplementary_refinement_block} and \cref{tb:supplementary_shared_refinement_block}.

\section{The StyleGAN2 Architecture}
\begin{figure}
    \centering
    \setlength{\belowcaptionskip}{-5pt}
    \includegraphics[width=0.8\linewidth]{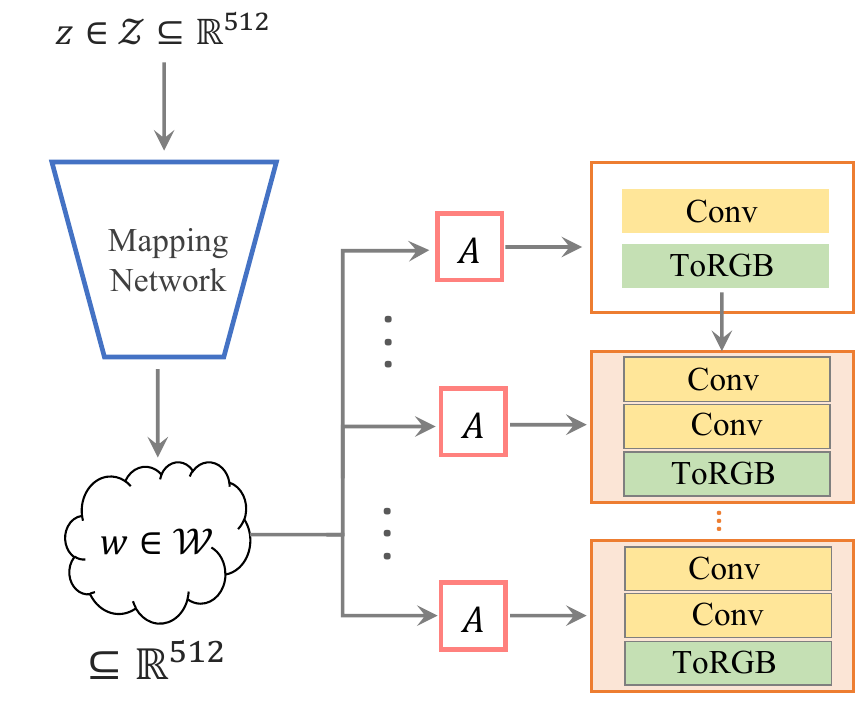}
    \caption{The StyleGAN2~\cite{karras2020analyzing} architecture.}
    \label{fig:stylegan2_arch}
\end{figure}

To determine which network parameters are most crucial to our inversion goal, it is important to understand the overall function of the components where these parameters reside. In the case of StyleGAN2~\cite{karras2020analyzing}, we consider four key components, as illustrated in \cref{fig:stylegan2_arch}. First, a mapping network converts the initial latent code $z \sim \mathcal{N}\left(0,1\right)^{512}$, into an equivalent code in a learned latent space $w \in \mathcal{W}$. 
These codes are then fed into a series of affine transformation blocks, one for each of the network's convolutional layers, which in turn predict a series of factors used to modulate the convolutional kernel weights. 
Lastly, the generator itself is built from two types of convolutional blocks: feature space convolutions, which learn increasingly complex representations of the data in some high-dimensional feature space, and toRGB blocks, which utilize convolutions to map these complex representations into residuals in a planar (or RGB) space. Of these four components, we restrict ourselves to modifying only the feature-space convolutions. We outline the different generator layers and their dimensions in ~\cref{tb:supplementary_stylegan2}.

\newcolumntype{P}[1]{>{\centering\arraybackslash}p{#1}}

\begin{table*}
    \setlength{\tabcolsep}{5pt}
    \centering
    \begin{tabular}{P{3cm} | P{3cm} | P{3cm} }
    \toprule
    Layer & Weight Dimension & Output Size \\
    
    \midrule
    
    Conv-LeakyReLU & $3\times3\times512\times256$ & $8\times8\times256$ \\
    Conv-LeakyReLU & $3\times3\times256\times256$ & $4\times4\times256$ \\
    Conv-LeakyReLU & $3\times3\times256\times512$ & $2\times2\times512$ \\
    AdaptivePool2D & - & $1\times1\times512$ \\
    Fully-Connected & $512 \times \left( C^{in}_{\ell} \cdot C^{out}_\ell \right)$ & $1\times1\times C^{in}_{\ell}\times C^{out}_\ell$ \\

    \bottomrule
    
    \end{tabular}

    \vspace{-0.15cm}
    \caption{The breakdown of the standard \textit{Refinement Block} described in Section 3 of the main paper. For a given generator layer $\ell$, Refinement Block receives as input the $16\times16\times512$ feature map extracted from the shared backbone and returns the predicted weight offsets of dimensions $1\times1\times C^{in}_{\ell}\times C^{out}_\ell$.  }
    \label{tb:supplementary_refinement_block}
\end{table*}

\newcolumntype{P}[1]{>{\centering\arraybackslash}p{#1}}

\begin{table*}
    \setlength{\tabcolsep}{5pt}
    \centering
    \begin{tabular}{P{5cm} | P{3cm} | P{3cm} }
    \toprule
    Layer & Weight Dimension & Output Size \\
    
    \midrule
    Conv-LeakyReLU & $3\times3\times512\times128$ & $16\times16\times128$ \\
    Conv-LeakyReLU & $3\times3\times128\times128$ & $8\times8\times128$ \\
    Conv-LeakyReLU & $3\times3\times128\times128$ & $4\times4\times128$ \\
    Conv-LeakyReLU & $3\times3\times128\times128$ & $2\times2\times128$ \\
    Conv-LeakyReLU & $3\times3\times128\times512$ & $1\times1\times512$ \\
    Fully-Connected & $512 \times 512$ & $1\times1\times512$ \\
    \hline 
    
    Shared Fully-Connected & $512 \times (512 \cdot 512)$ & $512\times512$ \\
    Shared Fully-Connected & $512 \times (512\cdot1\cdot1)$ & $1\times1\times512\times512$ \\

    \bottomrule
    
    \end{tabular}

    \vspace{-0.15cm}
    \caption{The breakdown of the \textit{Shared Refinement Block} described in Section 3 of the main paper. The Shared Refinement Block consists of a pair of fully-connected layers that are shared across multiple blocks (namely those tasked with predicting offsets for the generator's largest convolutional layers).
    }
    \label{tb:supplementary_shared_refinement_block}
\end{table*}

\section{Additional Qualitative Results}~\label{supp:comparisons}
Finally, we provide additional results and comparisons, as follows: 
\begin{enumerate}
    \item \cref{fig:supp_faces_inversion_comparison} provides additional reconstruction comparisons on the human facial domain. 
    \item \cref{fig:idinvert_comparison} contains a visual comparison between HyperStyle and IDInvert~\cite{zhu2020domain} on the human facial domain.
    \item \cref{fig:supp_cars_inversion_comparison} provides reconstruction comparisons on the cars domain. 
    \item \cref{fig:styleclip_faces_editing_comparison} shows additional editing comparisons on the human facial domain obtained with StyleCLIP~\cite{patashnik2021styleclip} and InterFaceGAN~\cite{shen2020interpreting}.
    \item \cref{fig:supp_editings_interfacegan} shows additional HyperStyle editing results on the human facial domain obtained with  InterFaceGAN~\cite{shen2020interpreting}.
    \item \cref{fig:supp_face_editings_styleclip} contains additional HyperStyle editing results on the human facial domain obtained with  StyleCLIP~\cite{patashnik2021styleclip}.
    \item \cref{fig:supp_car_editing_comparison} provides additional editing comparisons on the cars domain obtained with GANSpace~\cite{harkonen2020ganspace}.
    \item \cref{fig:supplementary_cars_edits} contains HyperStyle editing results on the cars domain obtained with GANSpace~\cite{harkonen2020ganspace}.
    \item \cref{fig:afhq_wild_edits} contains HyperStyle editing results obtained with StyleCLIP~\cite{patashnik2021styleclip} on the AFHQ Wild~\cite{choi2020stargan} test set.
    \item \cref{fig:supp_out_of_domain_e4e_psp} and \cref{fig:supp_out_of_domain_restyle} illustrate HyperStyle's reconstructions and edits on challenging out-of-domain images. 
    \item \cref{fig:supplementary_domain_adaption} illustrates additional domain adaptation results. 
\end{enumerate}

\section{Implementation Details}~\label{supp:implementation_details}
All hypernetworks employ a ResNet34~\cite{he2015deep} backbone pre-trained on ImageNet. The networks have a modified input layer to accommodate the $6$-channel inputs.  
We train our networks using the Ranger optimizer~\cite{Ranger} with a constant learning rate of $0.0001$ and a batch size of $8$. 

When applying the iterative refinement scheme from Alaluf~\etal~\cite{alaluf2021restyle}, our hypernetworks use $T=5$ iterative steps per batch during training. For each step $t$, we compute losses between the current reconstructions and inputs. That is, losses are computed $T$ times per batch. 

Following recent works~\cite{richardson2020encoding, tov2021designing,alaluf2021restyle}, we set $\lambda_{\text{LPIPS}}=0.8$. For the similarity loss human facial domain, we use a pre-trained ArcFace~\cite{deng2019arcface} network with $\lambda_{sim}=0.1$. For the remaining domains, we utilize a MoCo-based~\cite{chen2020improved} loss with $\lambda_{sim}=0.5$, as done in Tov~\etal~\cite{tov2021designing}. All experiments were conducted on a single NVIDIA Tesla P40 GPU.

\newcolumntype{P}[1]{>{\centering\arraybackslash}p{#1}}

\begin{table*}
    \setlength{\tabcolsep}{5pt}
    \centering
    \begin{tabular}{P{2cm} | P{2cm} | P{2cm} | P{5cm}}
    \toprule
    Group & Layer Index \quad $\ell$ & Layer Name & Filter Dimension $k\times k \times C^{in}_\ell \times C^{out}_\ell$ \\
    
    \midrule
    
    \multirow{5}{*}{Coarse} & 1 & Conv 1 & $3\times3\times512\times512$ \\
    & 2 & toRGB 1 & $1\times1\times512\times3$  \\
    & 3 & Conv 2 & $3\times3\times512\times512$  \\
    & 4 & Conv 3 & $3\times3\times512\times512$ \\
    & 5 & toRGB 2 & $1\times1\times512\times3$  \\
    
    \midrule
    
    \multirow{6}{*}{\underline{\textbf{Medium}}} & 6 & Conv 4 & $3\times3\times512\times512$ \\
    & 7 & Conv 5 & $3\times3\times512\times512$ \\
    & 8 & toRGB 3 & $1\times1\times512\times3$  \\
    & 9 & Conv 6 & $3\times3\times512\times512$  \\
    & 10 & Conv 7 & $3\times3\times512\times512$  \\
    & 11 & toRGB 4 & $1\times1\times512\times3$  \\
    
    \midrule
    
    \multirow{15}{*}{\underline{\textbf{Fine}}} & 12 & Conv 8 & $3\times3\times512\times512$ \\
    & 13 & Conv 9 & $3\times3\times512\times512$  \\
    & 14 & toRGB 5 & $1\times1\times512\times3$  \\
    & 15 & Conv 10 & $3\times3\times512\times256$ \\
    & 16 & Conv 11 & $3\times3\times256\times256$ \\
    & 17 & toRGB 6 & $1\times1\times256\times3$ \\
    & 18 & Conv 12 & $3\times3\times256\times128$ \\
    & 19 & Conv 13 & $3\times3\times128\times128$  \\
    & 20 & toRGB 7 & $1\times1\times128\times3$ \\
    & 21 & Conv 14 & $3\times3\times128\times64$ \\
    & 22 & Conv 15 & $3\times3\times64\times64$ \\
    & 23 & toRGB 8 & $1\times1\times64\times3$ \\
    & 24 & Conv 16 & $3\times3\times64\times32$ \\
    & 25 & Conv 17 & $3\times3\times32\times32$ \\
    & 26 & toRGB 9 & $1\times1\times32\times3$ \\
    
    \bottomrule
    
    \end{tabular}

    \vspace{-0.1cm}
    \caption{The breakdown of the StyleGAN2~\cite{karras2020analyzing} layer weights and their filter dimensions, split into the coarse, medium, and fine sets. Our final hypernetwork configuration alters the non-toRGB, feature-space convolutions from the medium and fine layers.}
    \label{tb:supplementary_stylegan2}
\end{table*}

\newcommand{\ccbync}{\href{https://creativecommons.org/licenses/by-nc/4.0/legalcode}{CC BY-NC 4.0}}

\newcommand{\cczero}{\href{https://creativecommons.org/publicdomain/zero/1.0/}{CC0 1.0}}

\newcommand{\ccbyncsa}{\href{https://creativecommons.org/licenses/by-nc-sa/4.0/}{CC BY-NC-SA 4.0}}

\newcommand{\nvsrc}{\href{https://nvlabs.github.io/stylegan2/license.html}{Nvidia Source Code License-NC}}

\newcommand{\bsd}{\href{https://opensource.org/licenses/BSD-3-Clause}{BSD 3-Clause}}

\newcommand{\mitlic}{\href{https://opensource.org/licenses/MIT}{MIT License}}

\newcommand{\apache}{\href{http://www.apache.org/licenses/LICENSE-2.0}{Apache 2.0 License}}

\begin{table*}
    \small
    \centering
    \makebox[\linewidth]{

    }
    \caption{Datasets and models used in our work and their respective licenses.}
    \label{tbl:licenses}
\end{table*}

\section{Licenses}
We provide the licenses of all datasets and models used in our work in \cref{tbl:licenses}.

\clearpage

\begin{figure*}

    \vspace{2cm}

    \centering
    \setlength{\belowcaptionskip}{-6pt}
    \setlength{\tabcolsep}{1.5pt}
    {\small
    \begin{tabular}{c | c c | c c c c c}

        \includegraphics[width=0.12\textwidth]{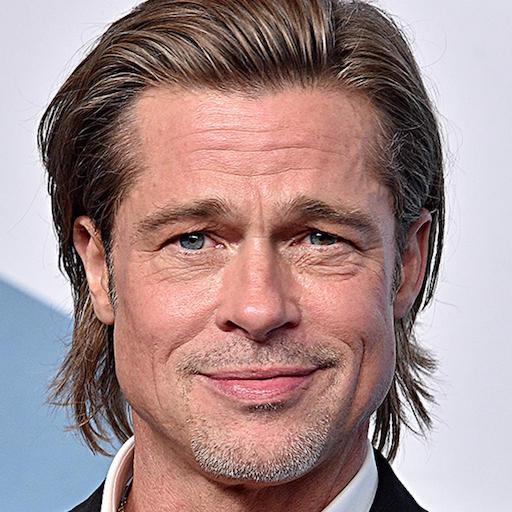} &
        \includegraphics[width=0.12\textwidth]{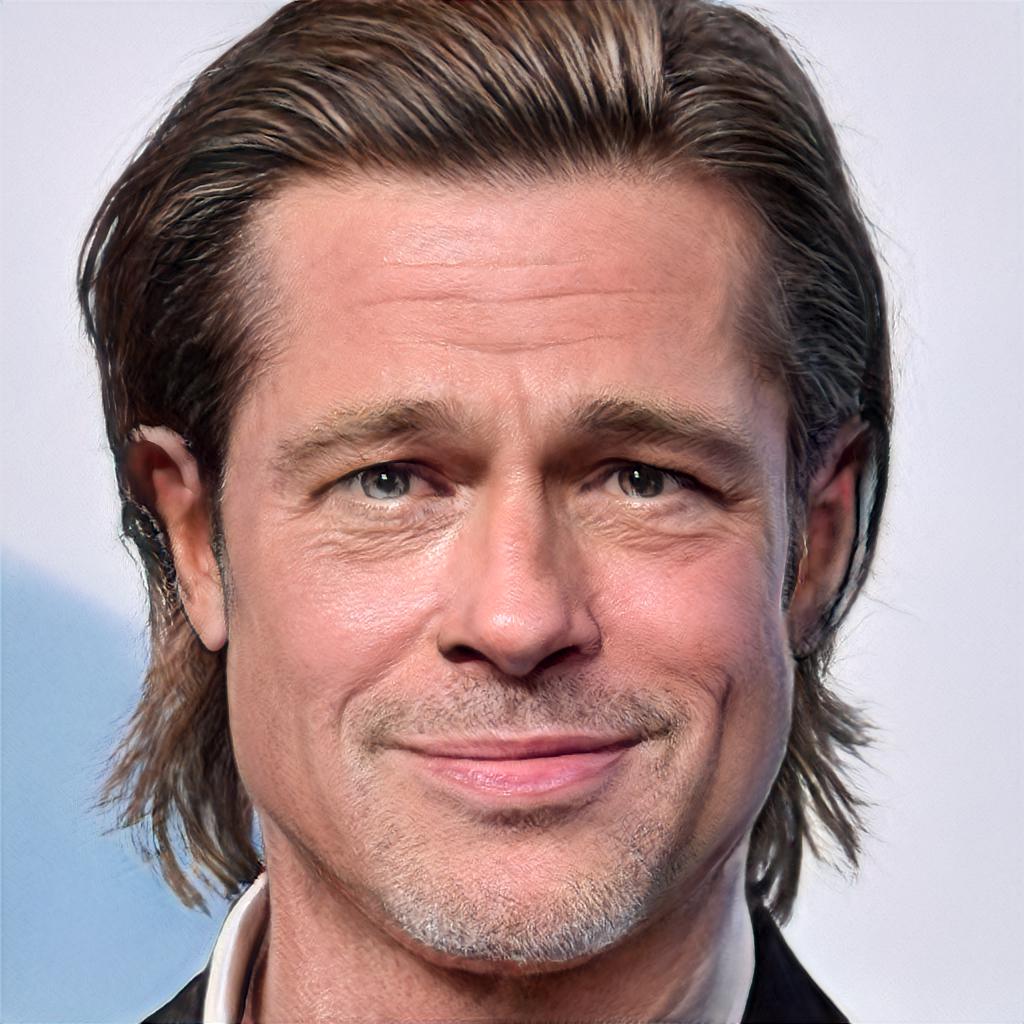} &
        \includegraphics[width=0.12\textwidth]{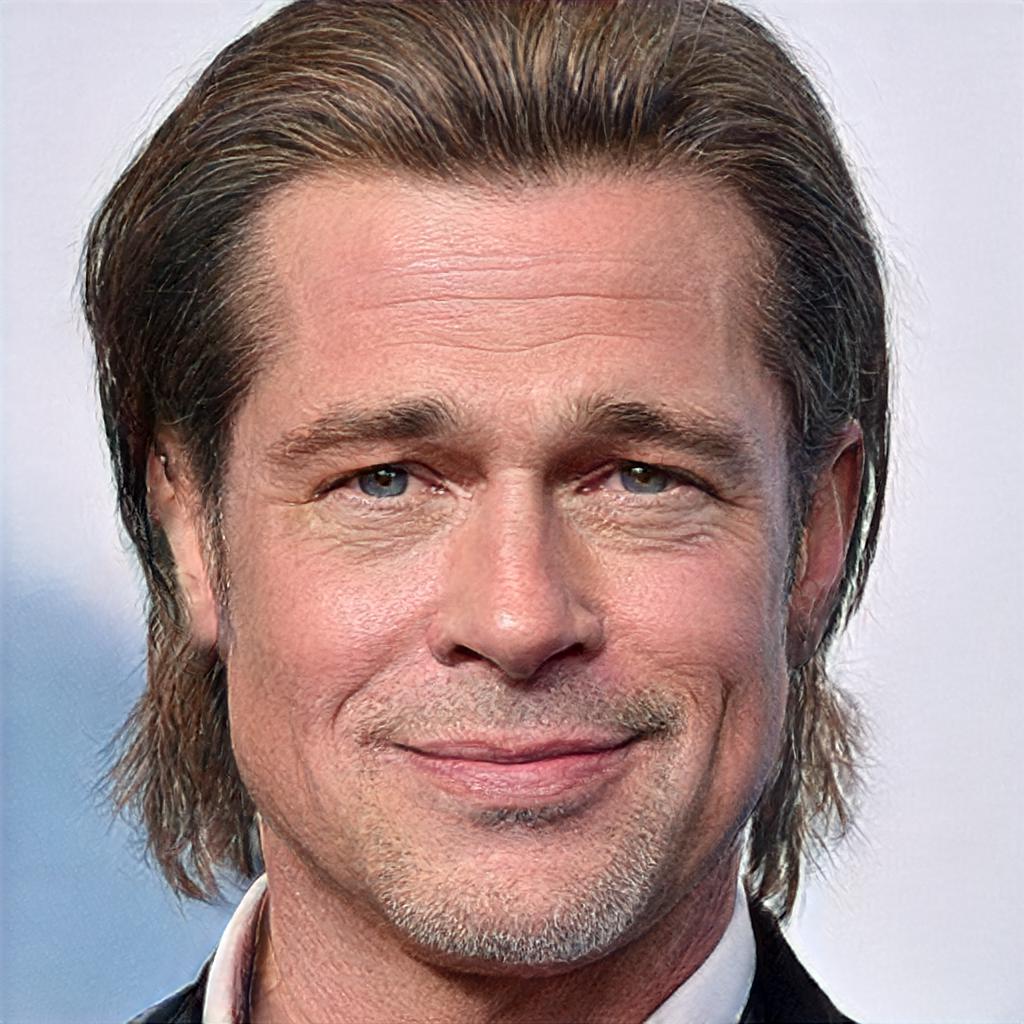} &
        \includegraphics[width=0.12\textwidth]{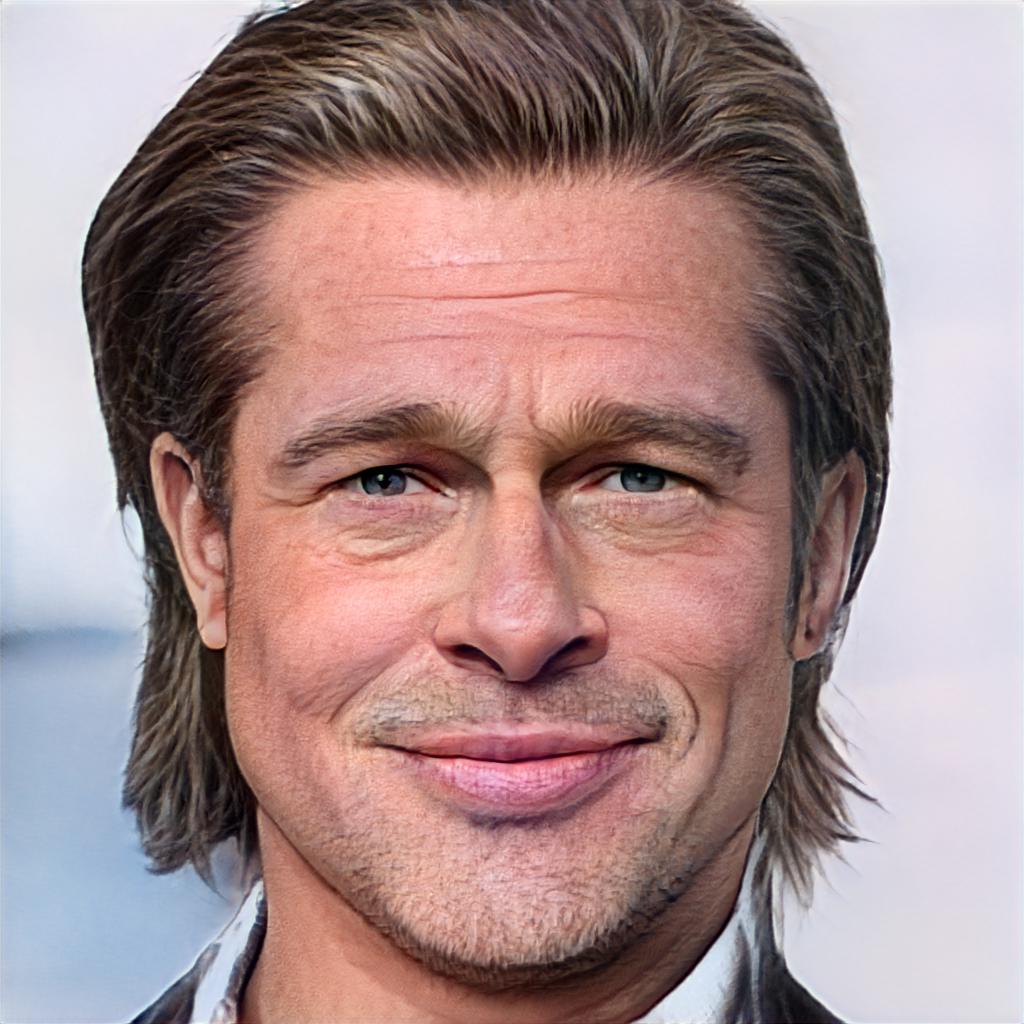} &
        \includegraphics[width=0.12\textwidth]{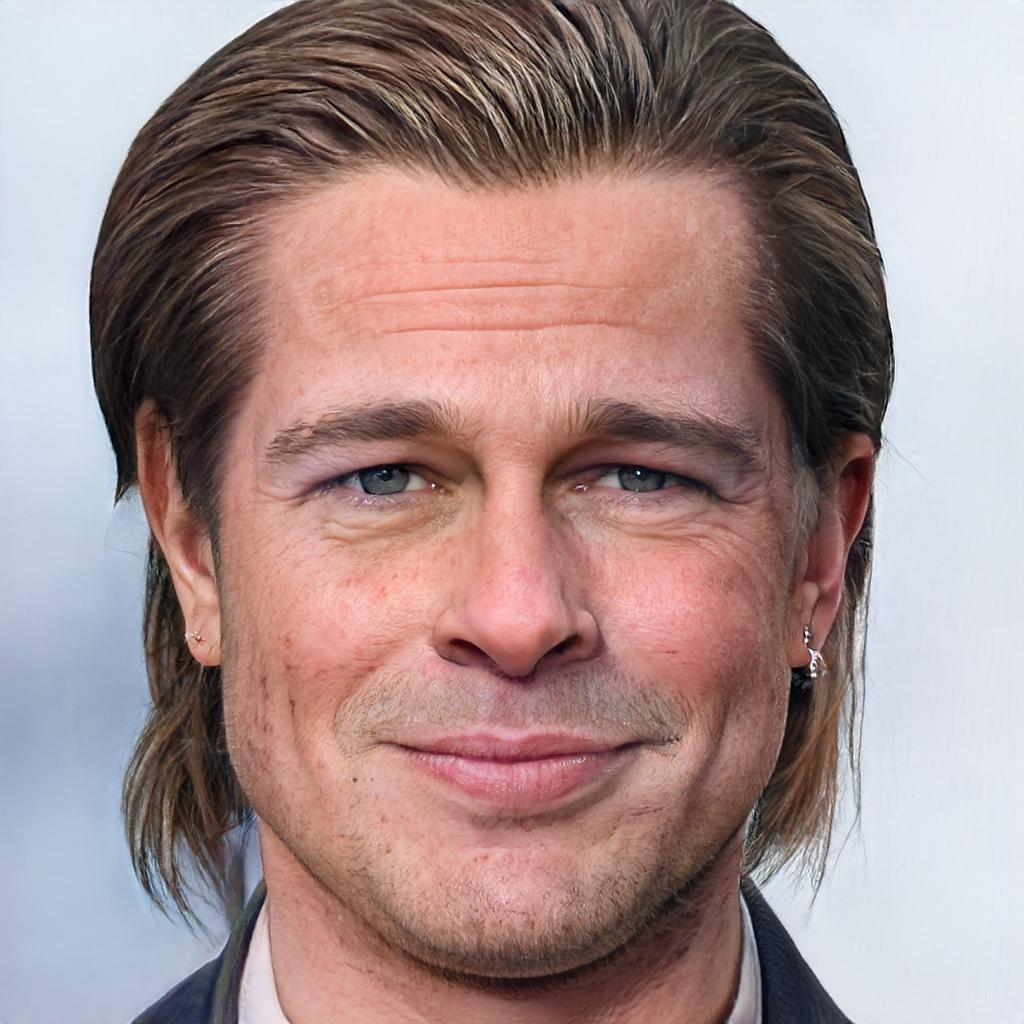} &
        \includegraphics[width=0.12\textwidth]{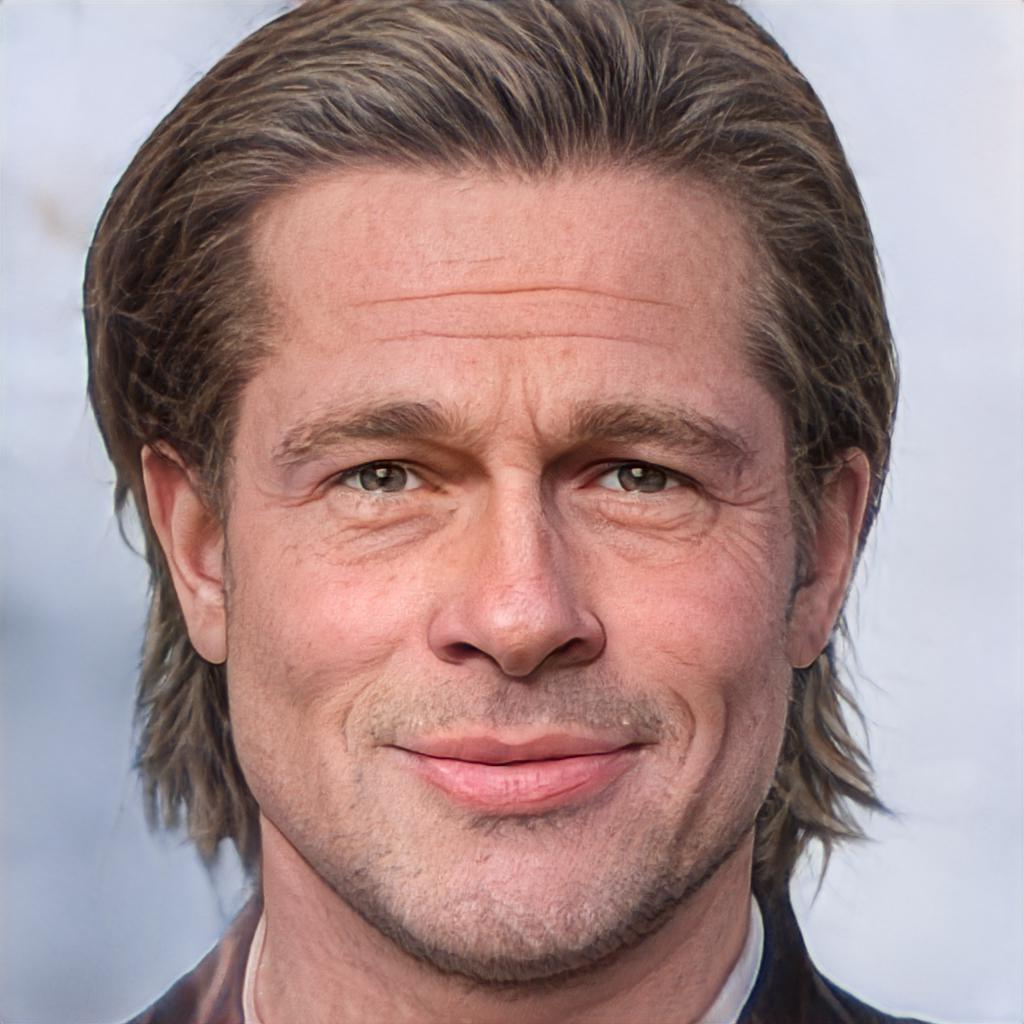} &
        \includegraphics[width=0.12\textwidth]{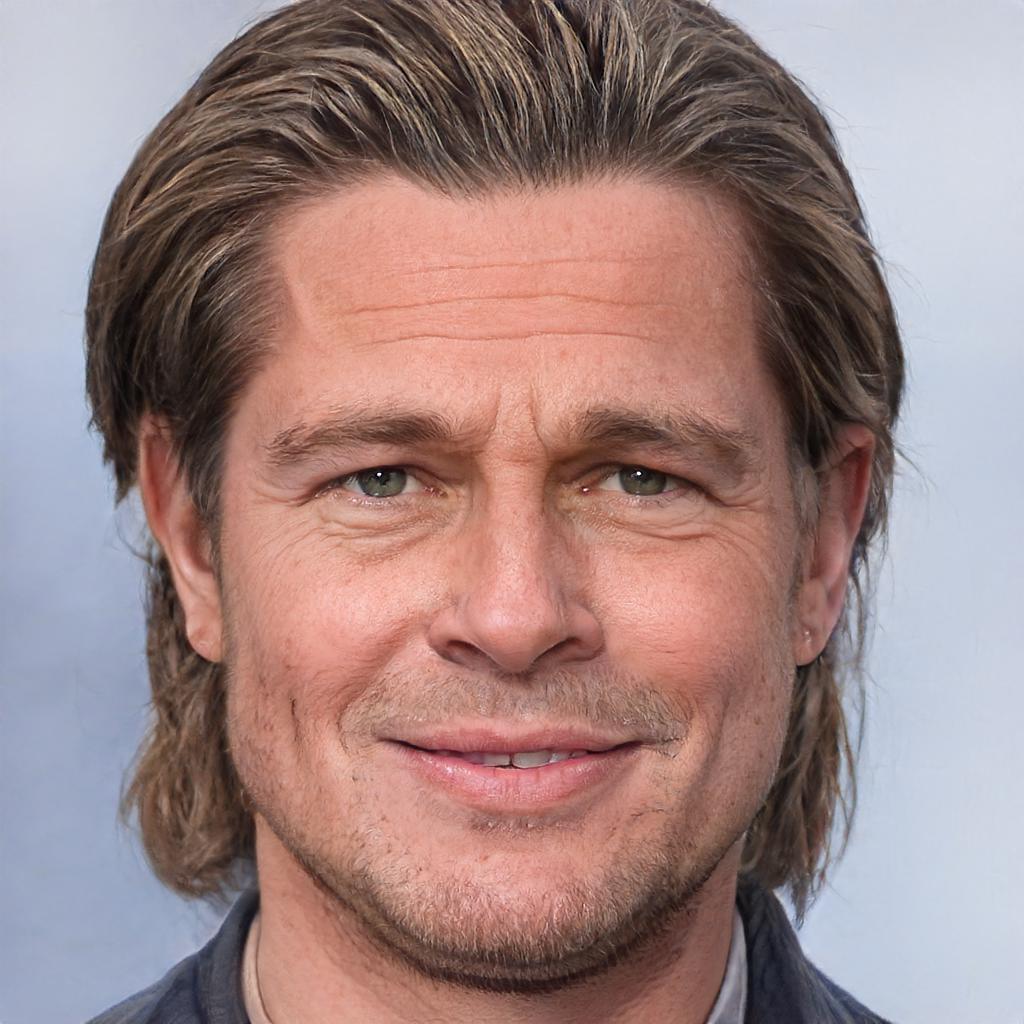} &
        \includegraphics[width=0.12\textwidth]{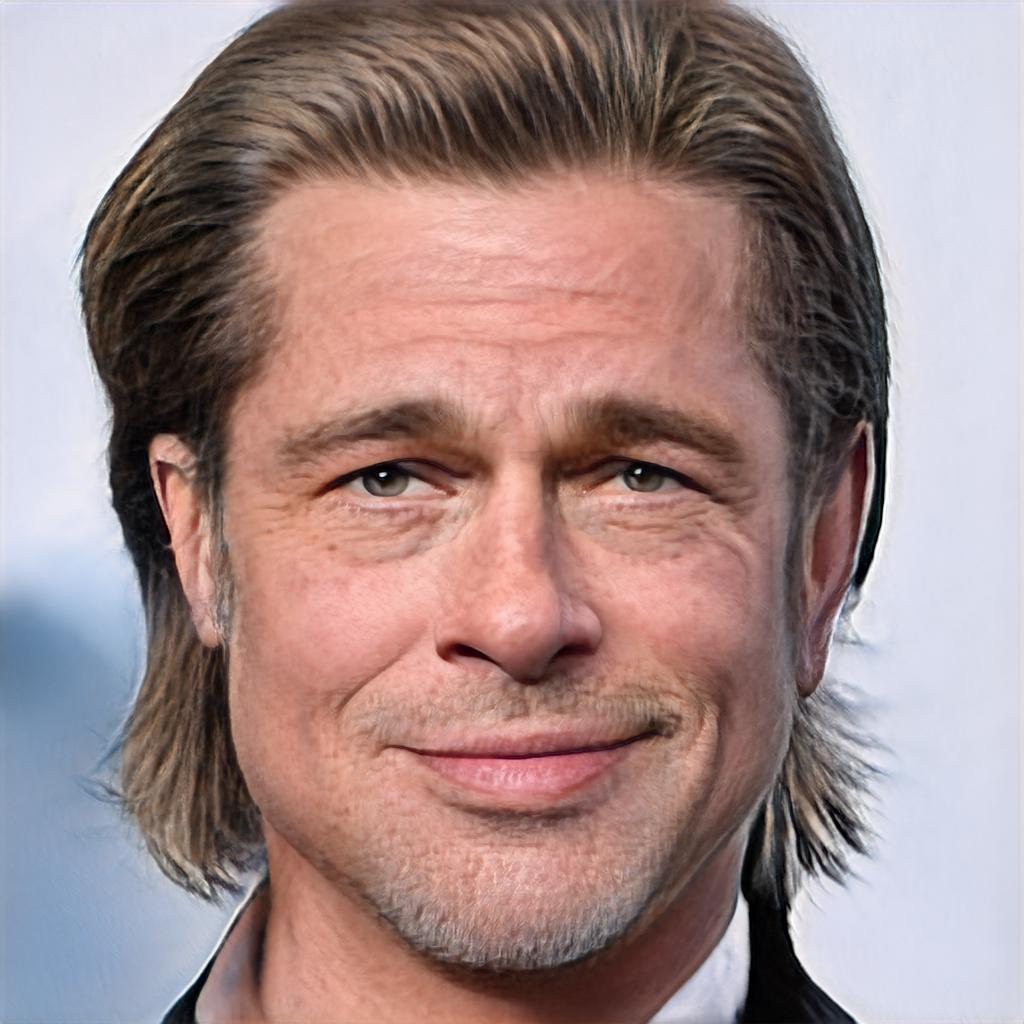} \\
        
        \includegraphics[width=0.12\textwidth]{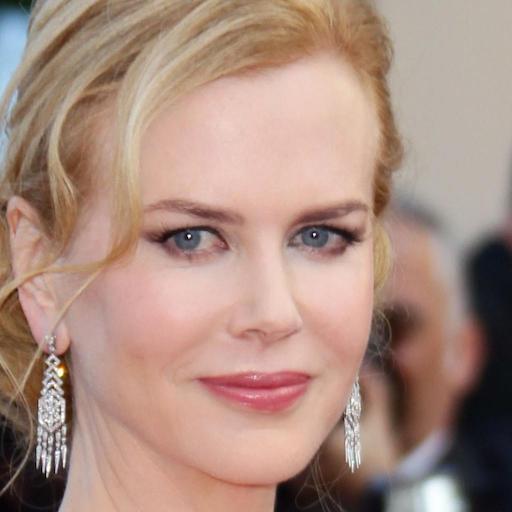} &
        \includegraphics[width=0.12\textwidth]{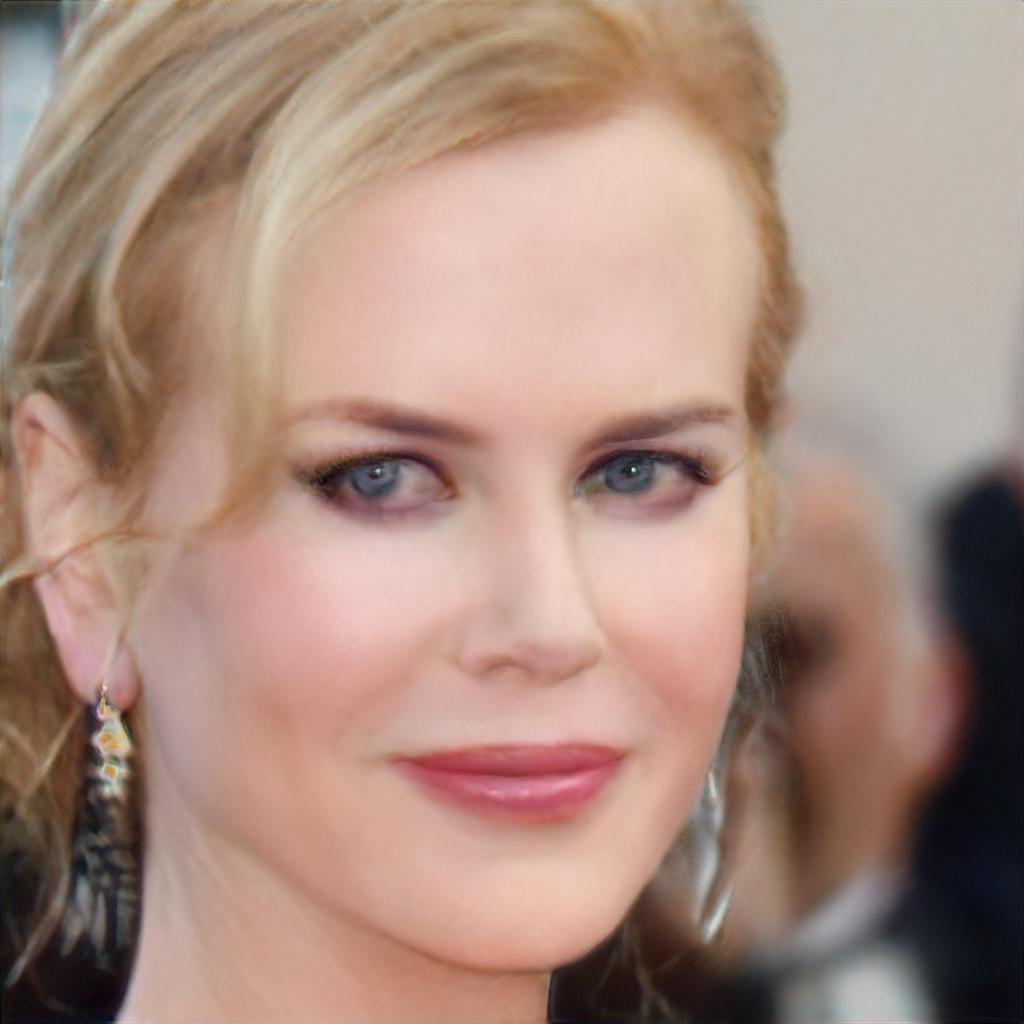} &
        \includegraphics[width=0.12\textwidth]{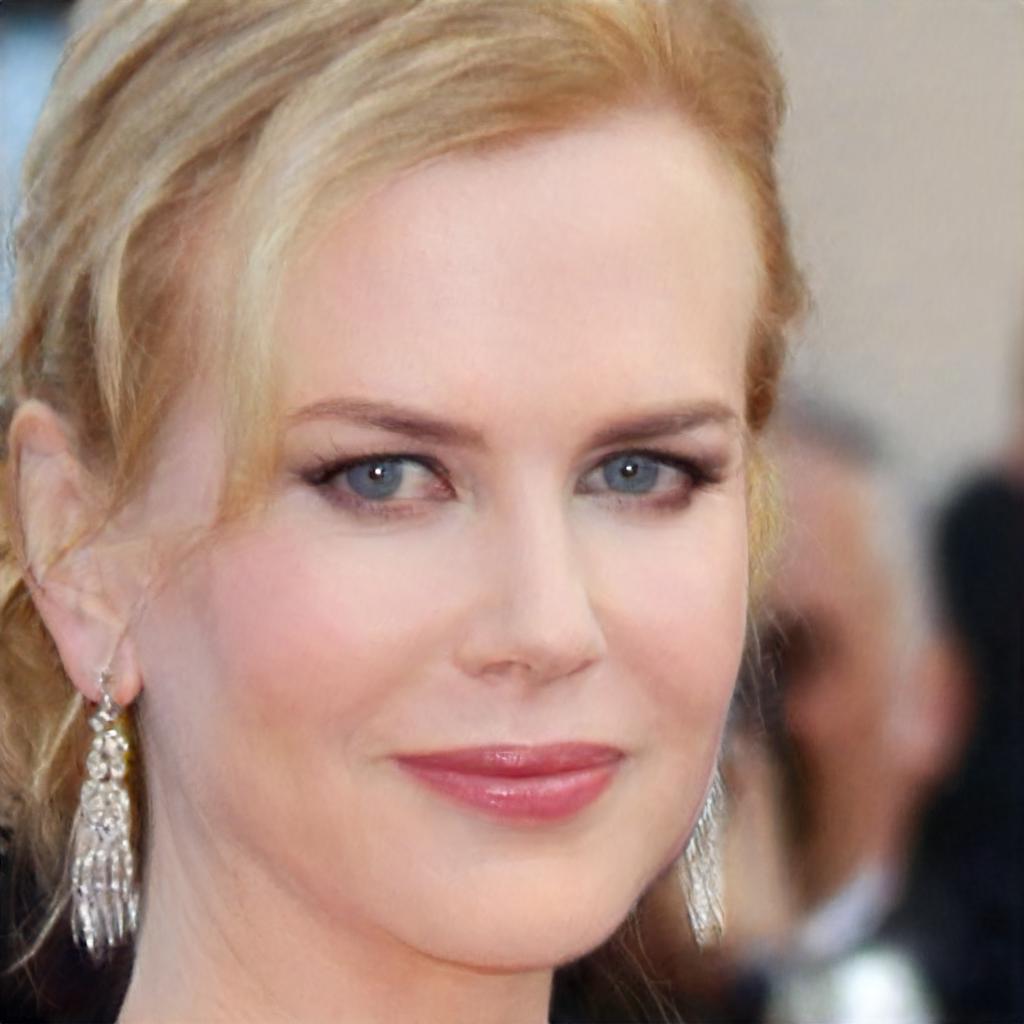} &
        \includegraphics[width=0.12\textwidth]{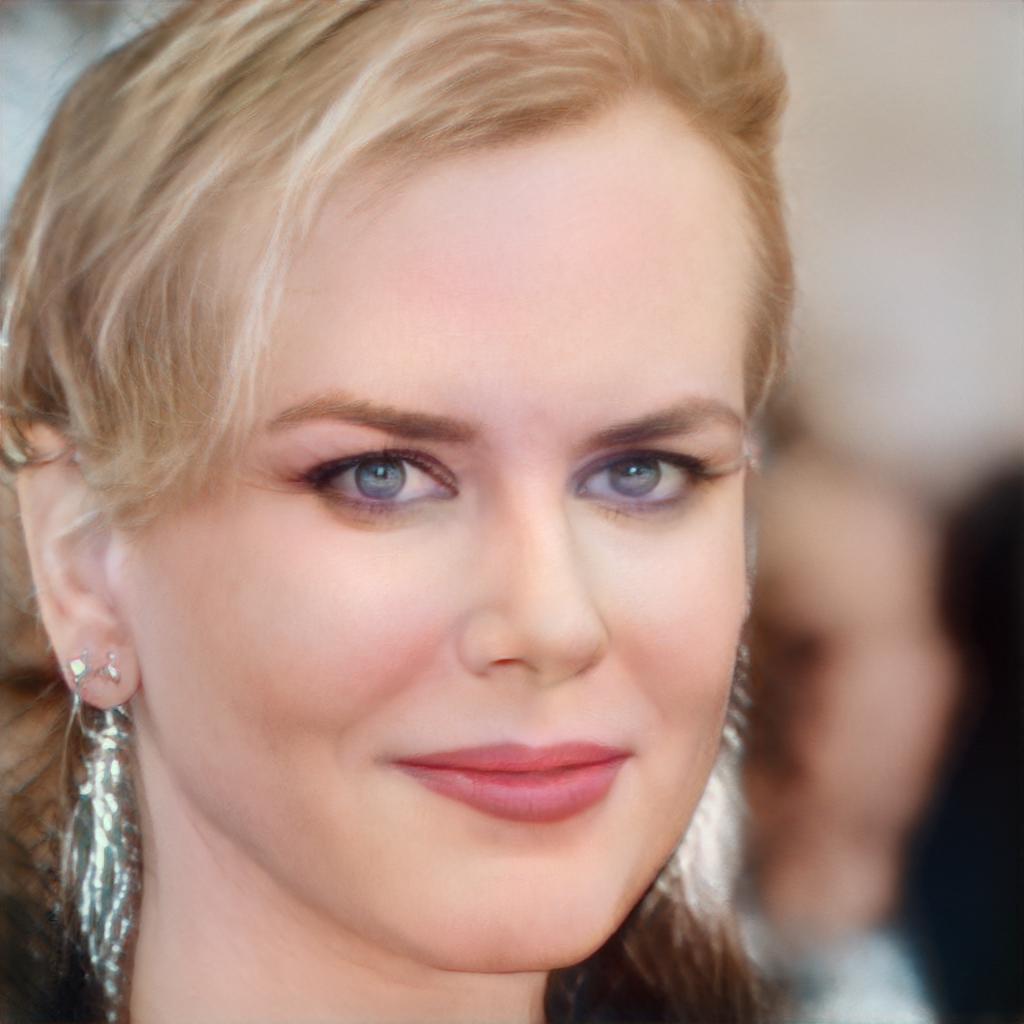} &
        \includegraphics[width=0.12\textwidth]{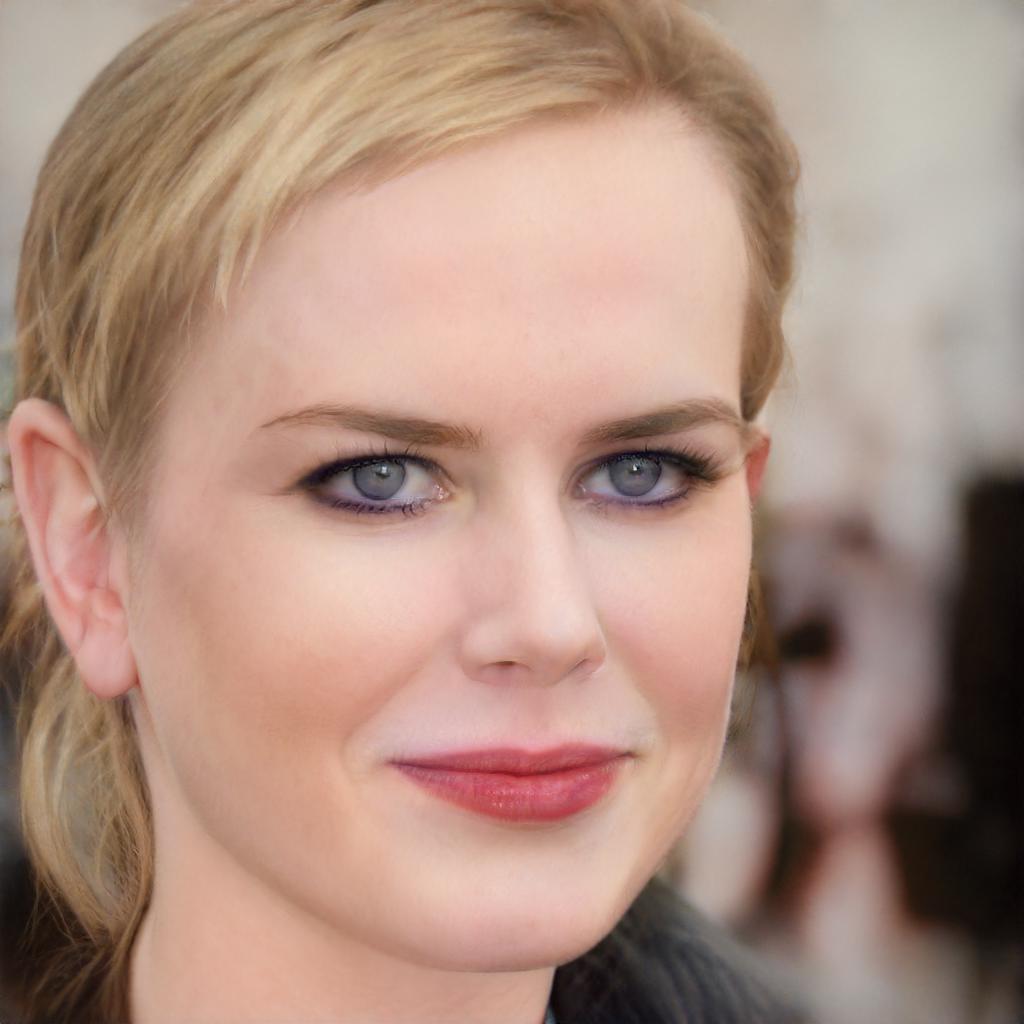} &
        \includegraphics[width=0.12\textwidth]{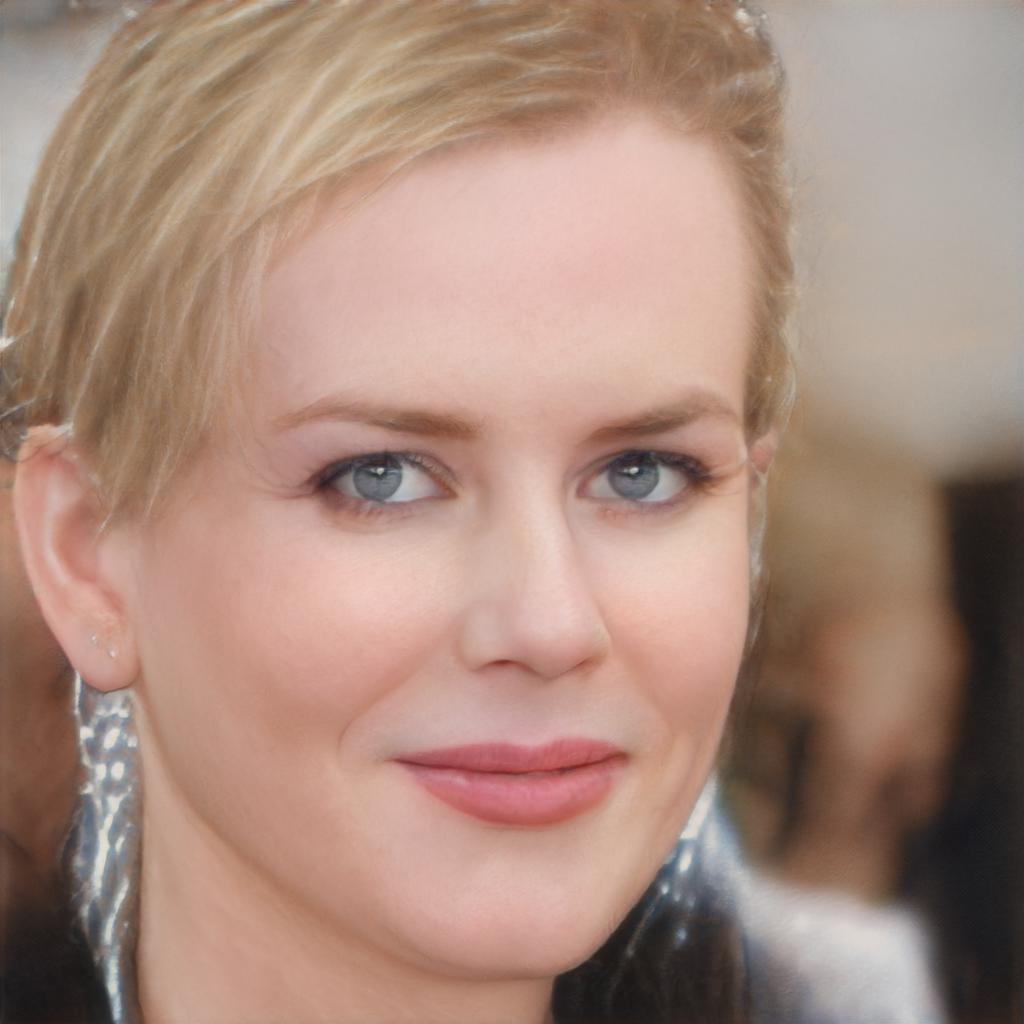} &
        \includegraphics[width=0.12\textwidth]{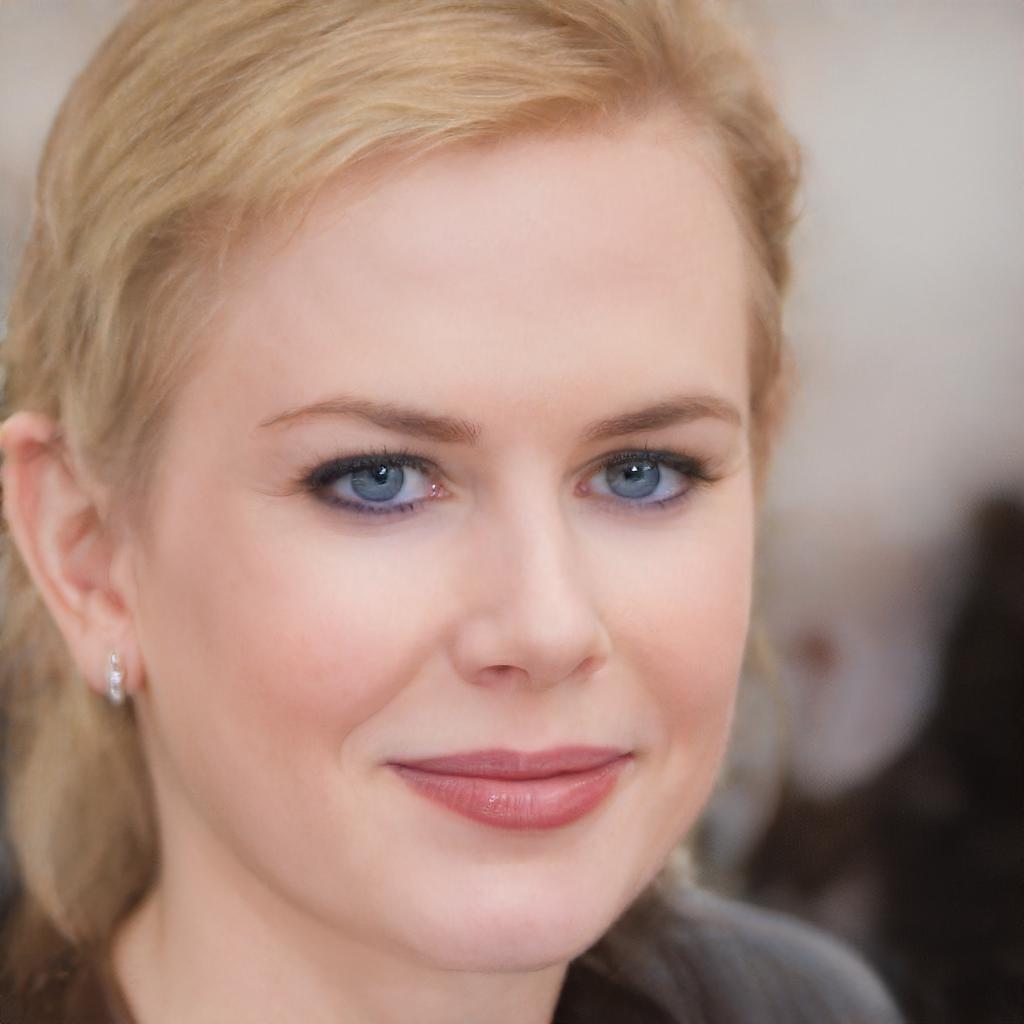} &
        \includegraphics[width=0.12\textwidth]{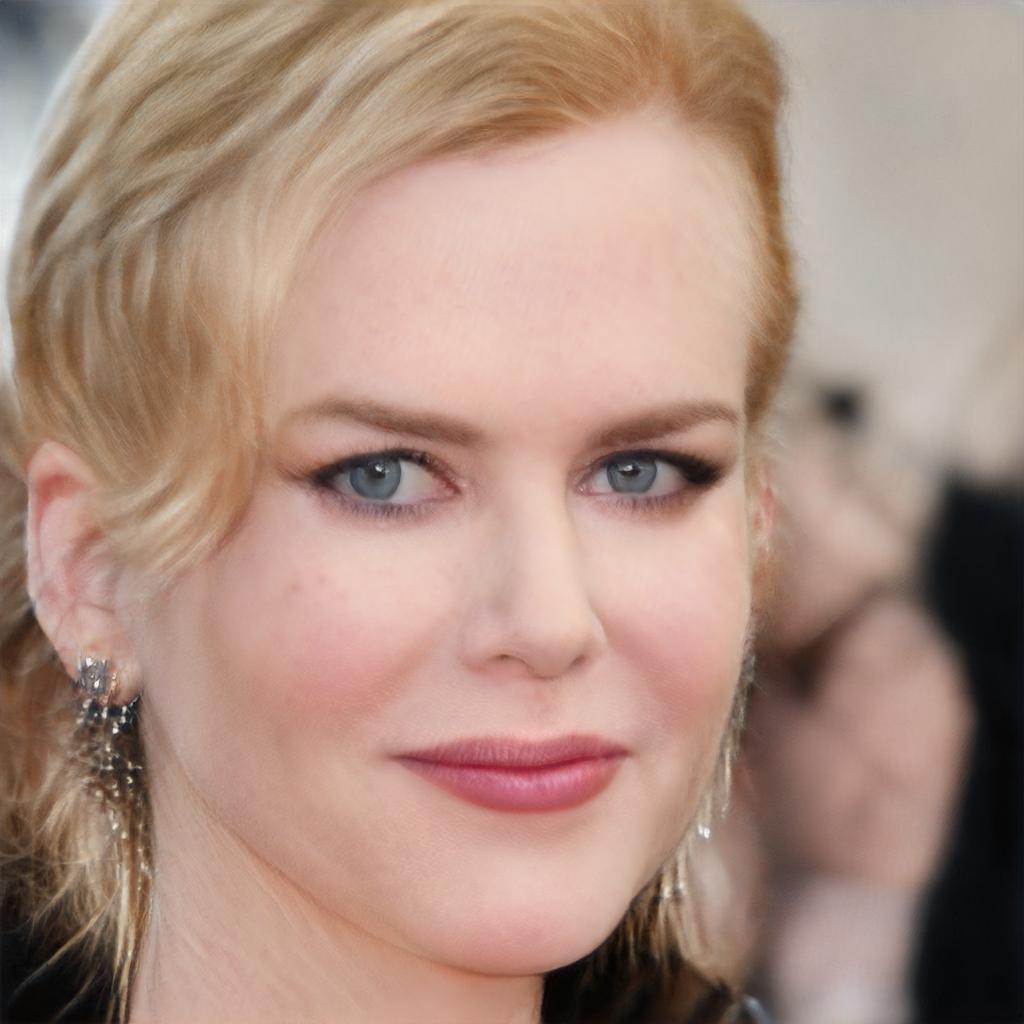} \\
\end{tabular}
    
    }
    \vspace{-0.2cm}
    \caption{Additional reconstruction quality comparisons on the human facial domain between the various inversion techniques. Best viewed zoomed-in.}
    \vspace{-0.25cm}
    \label{fig:supp_faces_inversion_comparison}
\end{figure*}
\begin{figure*}
\setlength{\tabcolsep}{1pt}
    \centering
    { \small 
    \begin{tabular}{c c c c c c c c c}

    \raisebox{0.2in}{\rotatebox{90}{Input}} &
    \includegraphics[width=0.125\textwidth]{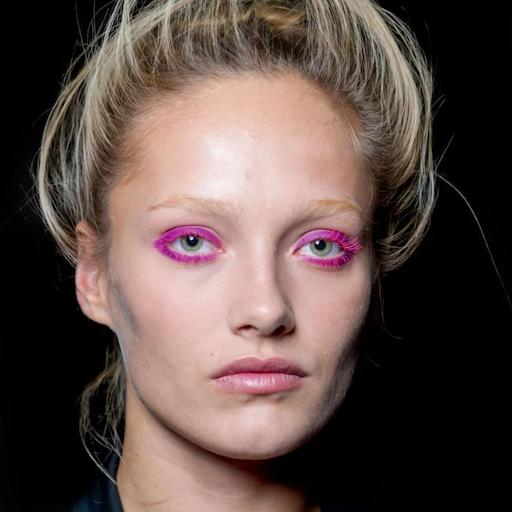} &
    \includegraphics[width=0.125\textwidth]{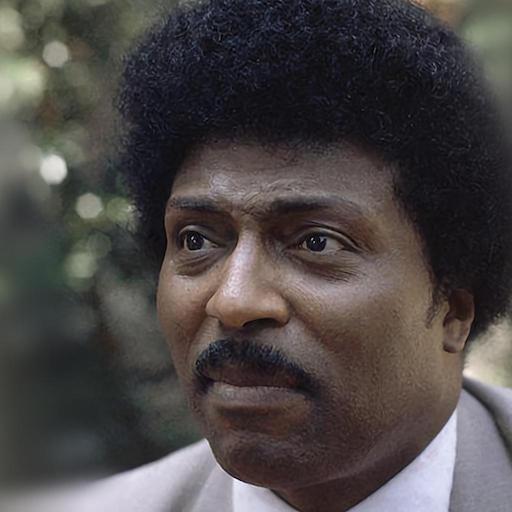} &
    \includegraphics[width=0.125\textwidth]{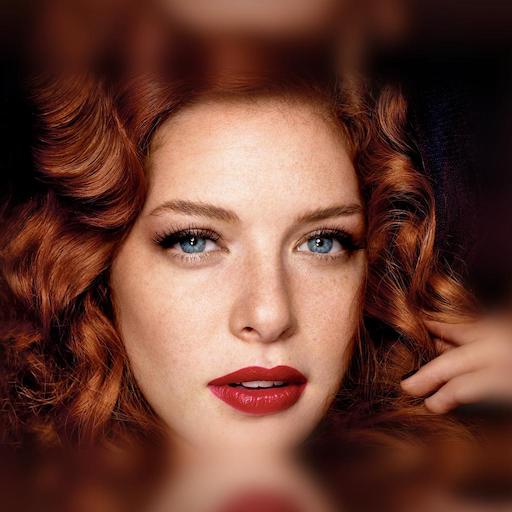} &
    \includegraphics[width=0.125\textwidth]{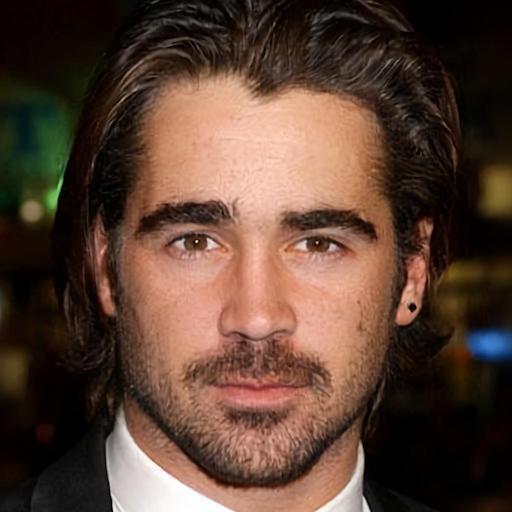} &
    \includegraphics[width=0.125\textwidth]{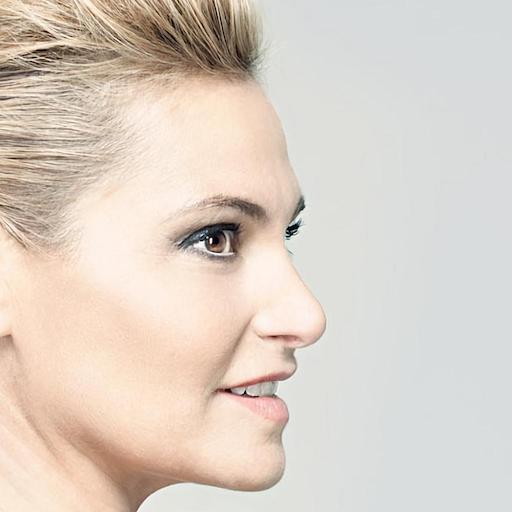} &
    \includegraphics[width=0.125\textwidth]{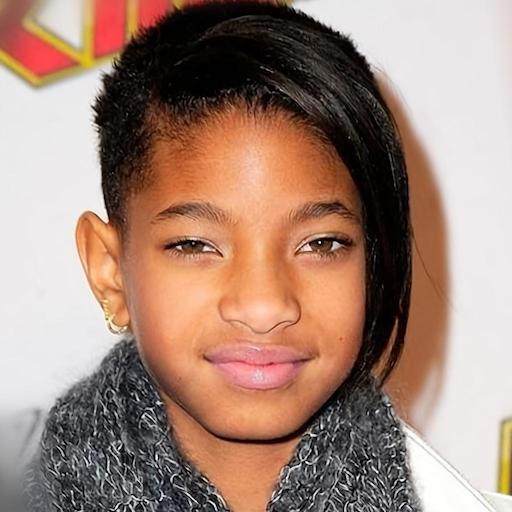} &
    \includegraphics[width=0.125\textwidth]{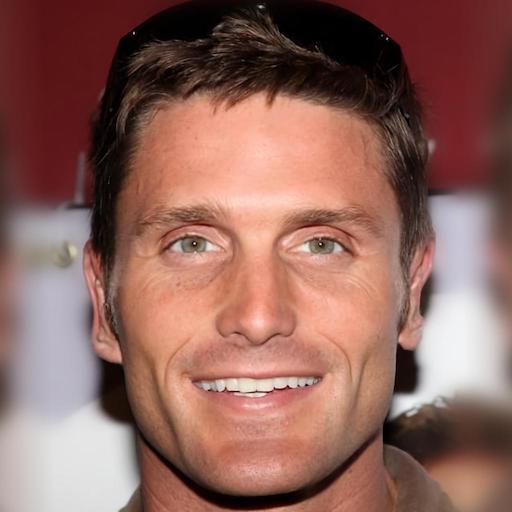} \\

    \raisebox{0.15in}{\rotatebox{90}{IDInvert}} &
    \includegraphics[width=0.125\textwidth]{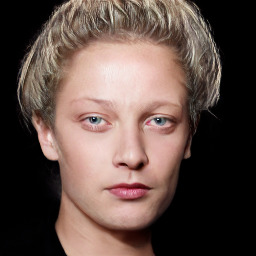} &
    \includegraphics[width=0.125\textwidth]{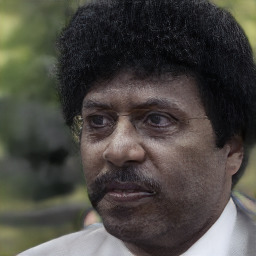} &
    \includegraphics[width=0.125\textwidth]{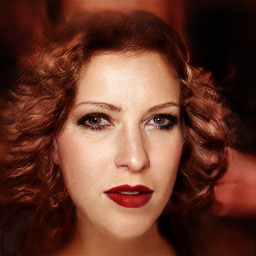} &
    \includegraphics[width=0.125\textwidth]{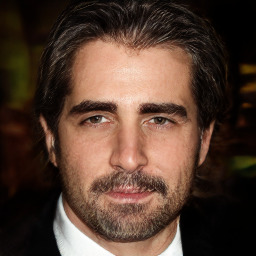} &
    \includegraphics[width=0.125\textwidth]{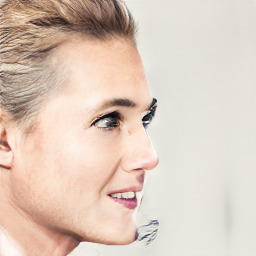} &
    \includegraphics[width=0.125\textwidth]{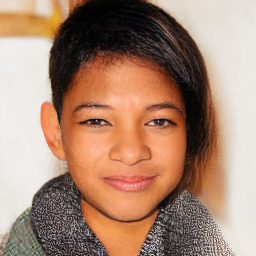} &
    \includegraphics[width=0.125\textwidth]{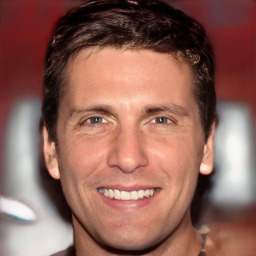} \\
\end{tabular}
    }
    \vspace{-0.325cm}
    \caption{Reconstruction comparison of HyperStyle and IDInvert~\cite{zhu2020domain} on the CelebA-HQ~\cite{karras2017progressive,liu2015deep} test set.} 
    \label{fig:idinvert_comparison}

\end{figure*}
\begin{figure*}
    
    \vspace{3cm}
    \centering
    \setlength{\belowcaptionskip}{-6pt}
    \setlength{\tabcolsep}{1.5pt}
    {\small
    \begin{tabular}{c | c c | c c c c c}

        \includegraphics[width=0.12\textwidth]{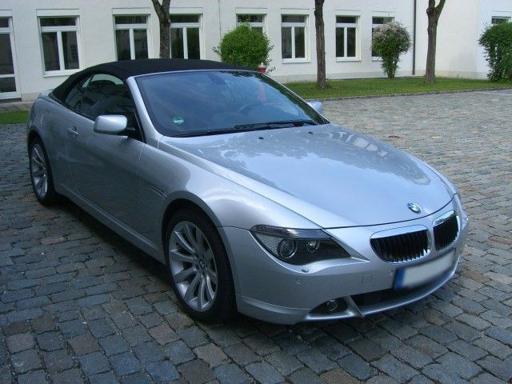} &
        \includegraphics[width=0.12\textwidth]{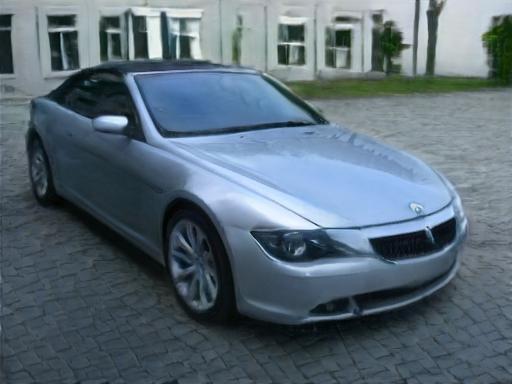} &
        \includegraphics[width=0.12\textwidth]{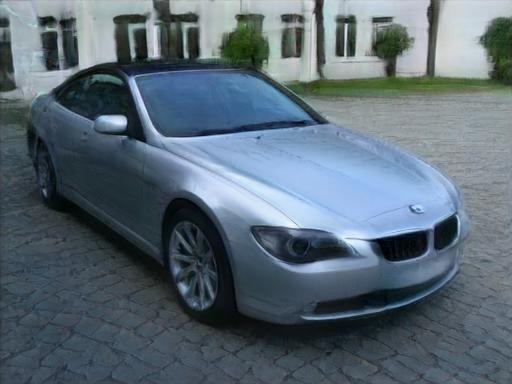} &
        \includegraphics[width=0.12\textwidth]{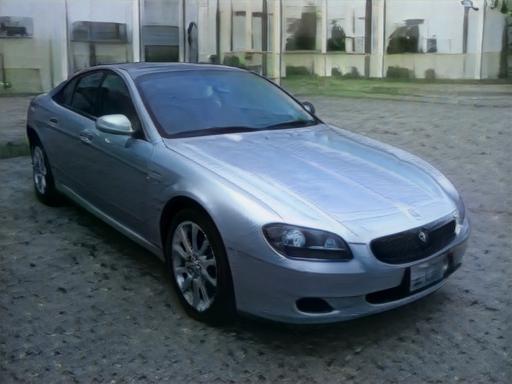} &
        \includegraphics[width=0.12\textwidth]{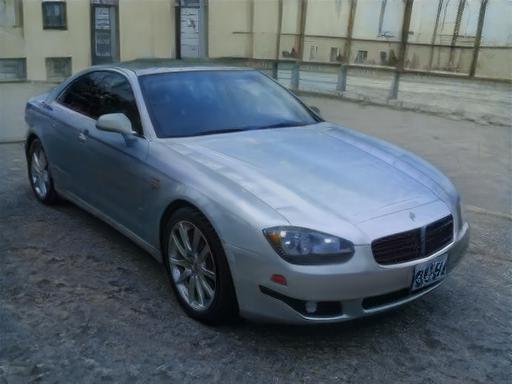} &
        \includegraphics[width=0.12\textwidth]{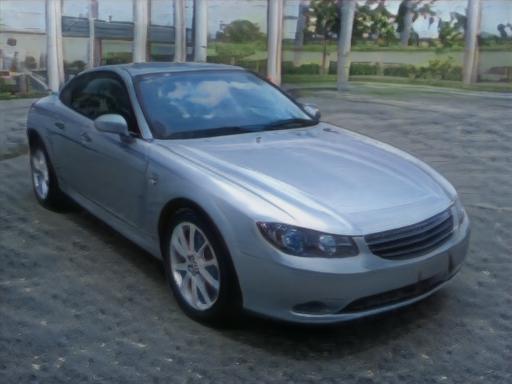} &
        \includegraphics[width=0.12\textwidth]{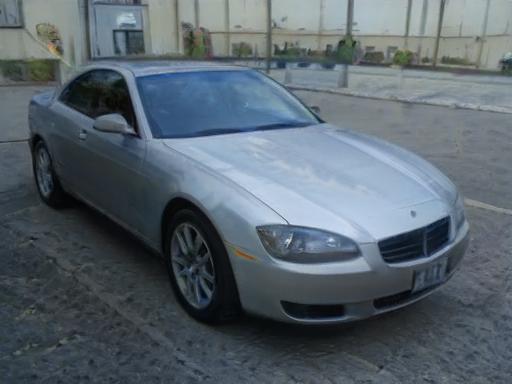} &
        \includegraphics[width=0.12\textwidth]{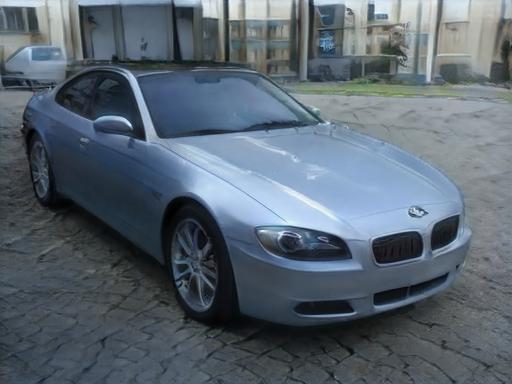} \\
        
        \includegraphics[width=0.12\textwidth]{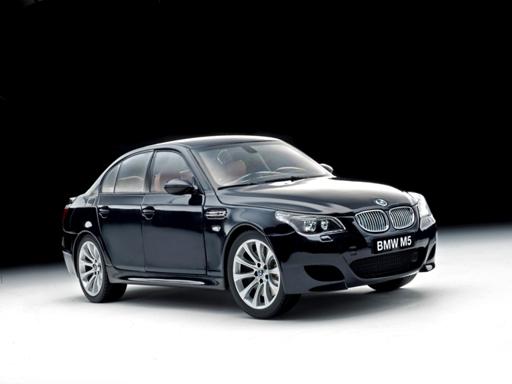} &
        \includegraphics[width=0.12\textwidth]{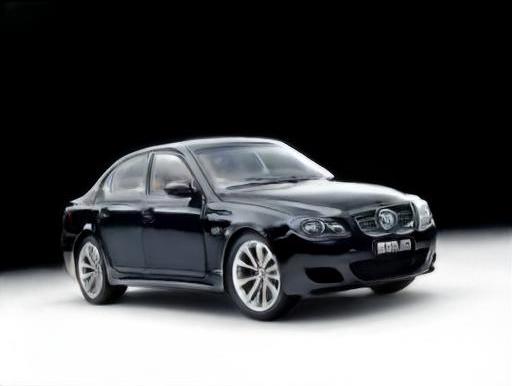} &
        \includegraphics[width=0.12\textwidth]{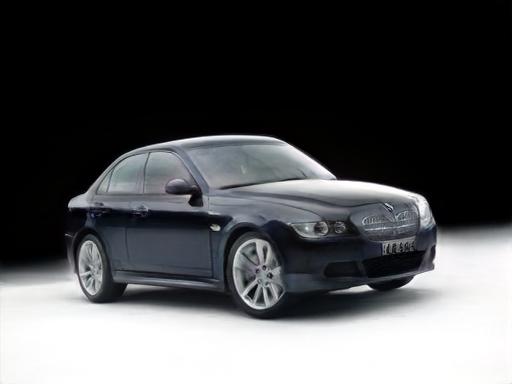} &
        \includegraphics[width=0.12\textwidth]{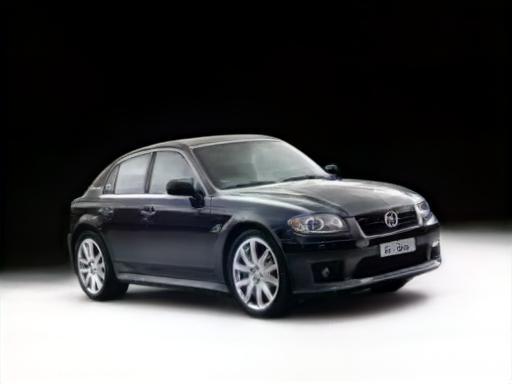} &
        \includegraphics[width=0.12\textwidth]{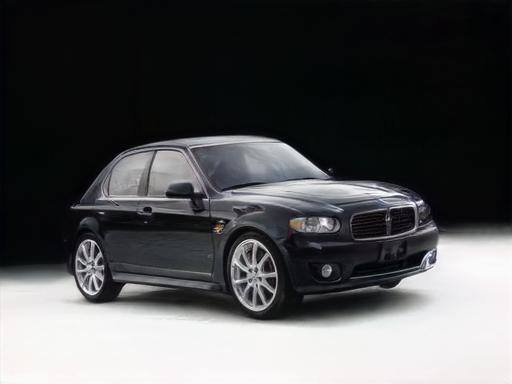} &
        \includegraphics[width=0.12\textwidth]{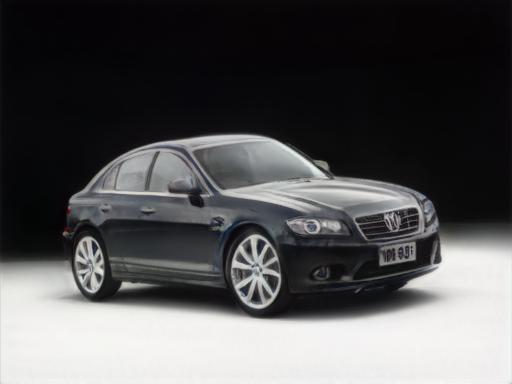} &
        \includegraphics[width=0.12\textwidth]{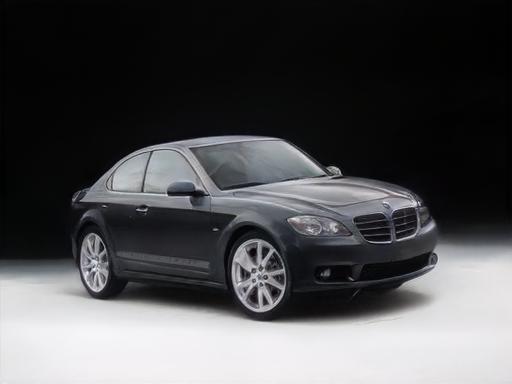} &
        \includegraphics[width=0.12\textwidth]{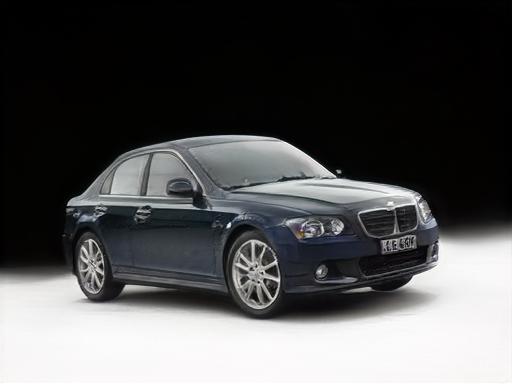} \\
\end{tabular}
    
    }
    \vspace{-0.2cm}
    \caption{Additional reconstruction quality comparisons on the cars domain between the various inversion techniques.}
    \vspace{-0.25cm}
    \label{fig:supp_cars_inversion_comparison}
\end{figure*}
\begin{figure*}
\setlength{\tabcolsep}{1pt}
    \centering
    { \small
    
    \begin{tabular}{c c c c c c c c c}

    \raisebox{0.15in}{\rotatebox{90}{$+$Beard}} &
    \includegraphics[width=0.1125\textwidth]{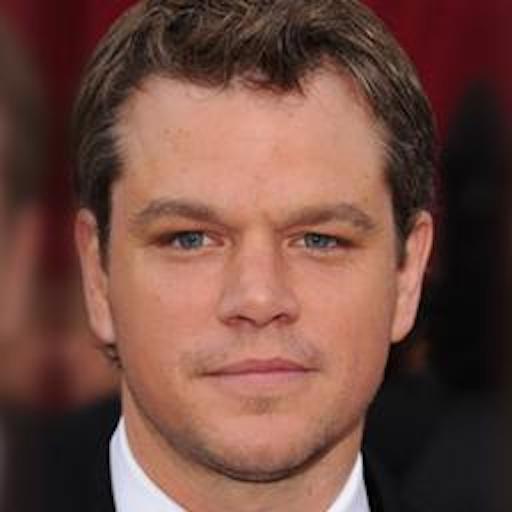} &
    \includegraphics[width=0.1125\textwidth]{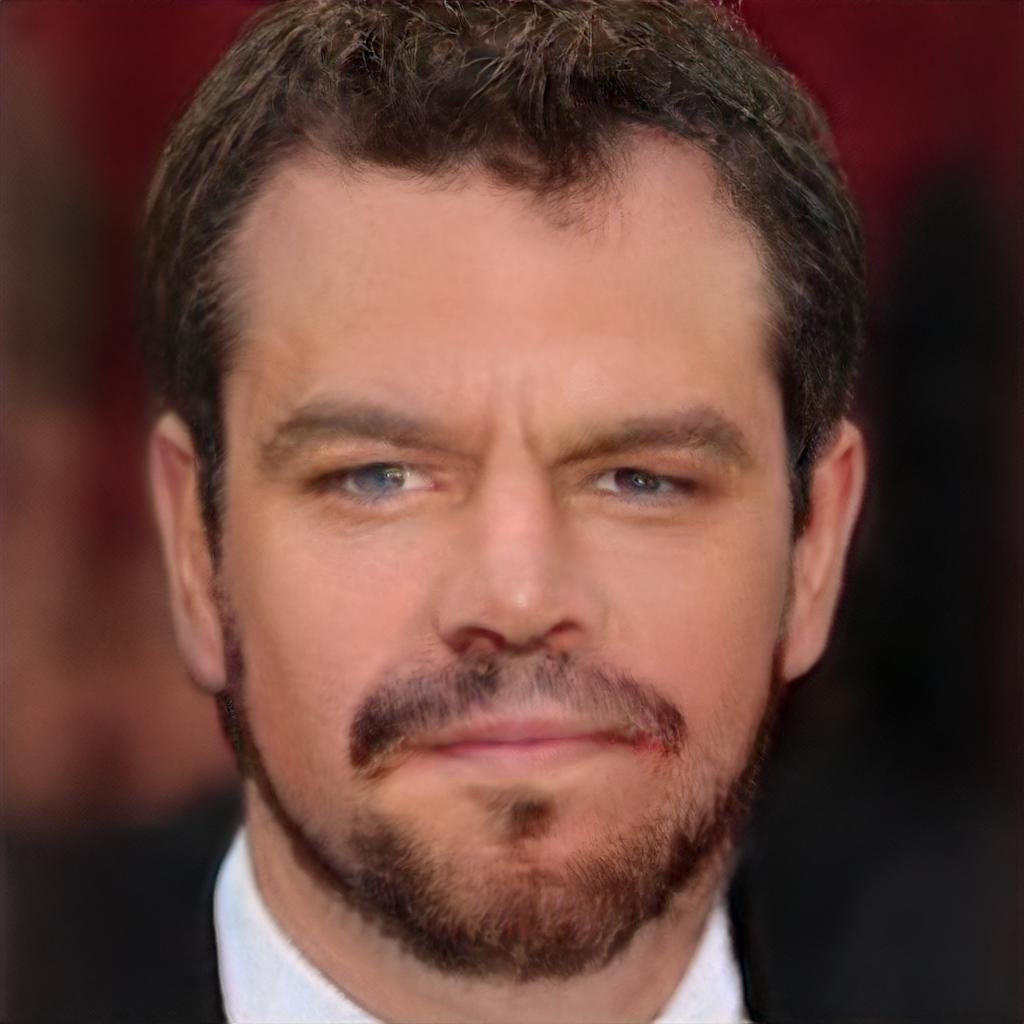} &
    \includegraphics[width=0.1125\textwidth]{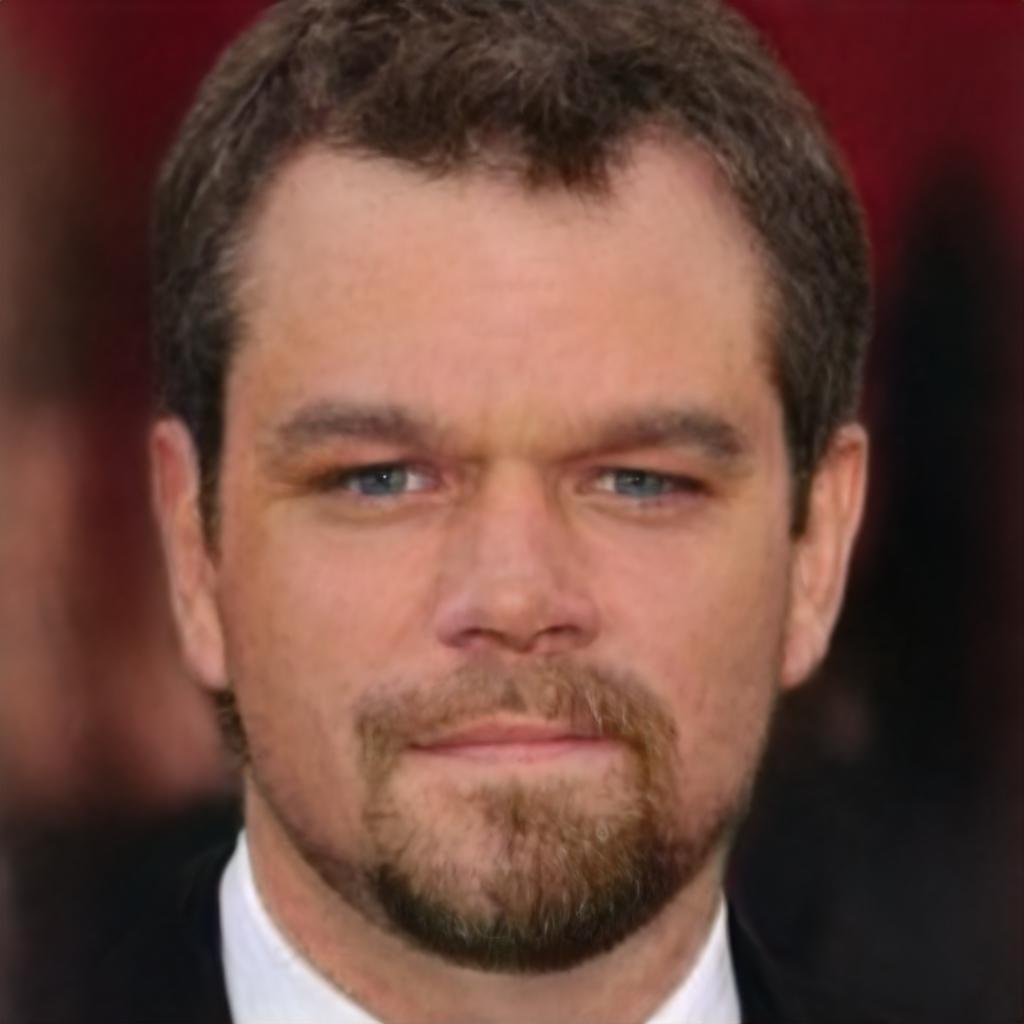} &
    \includegraphics[width=0.1125\textwidth]{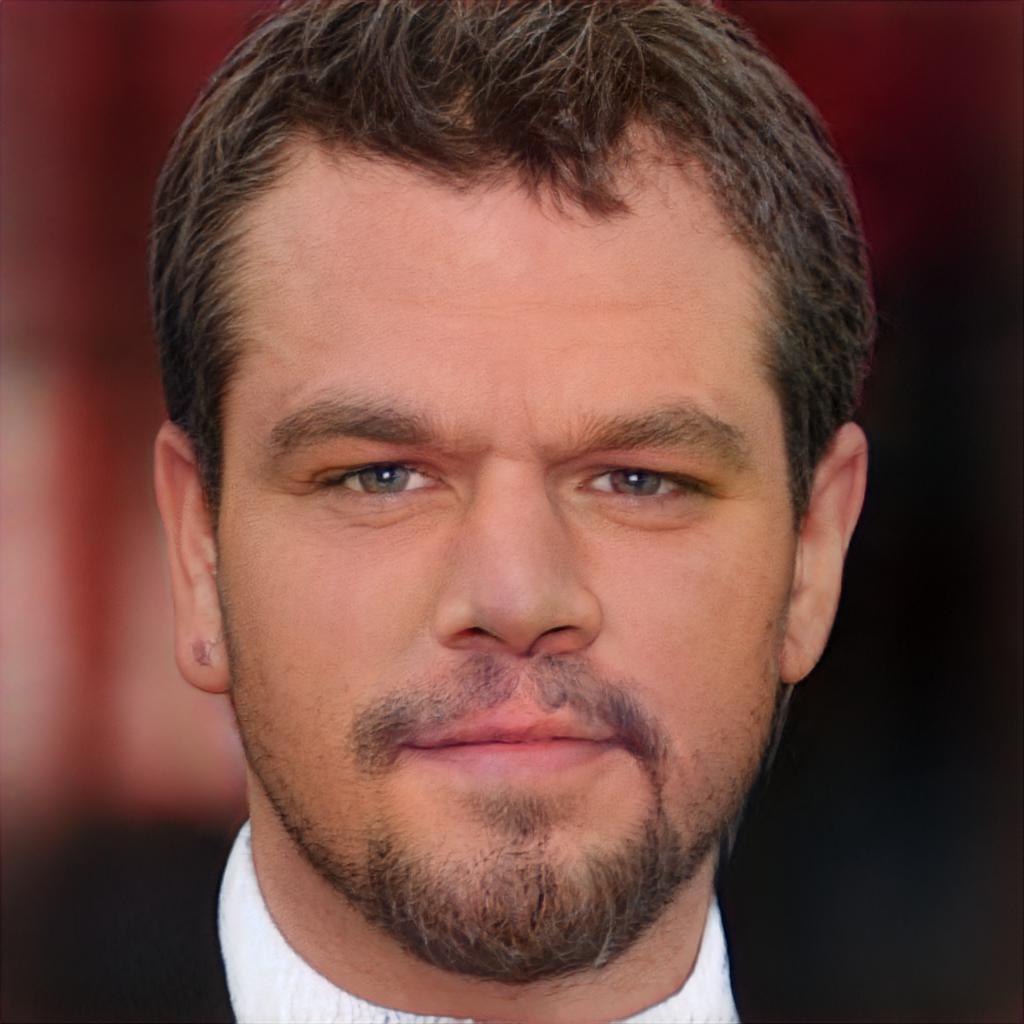} &
    \includegraphics[width=0.1125\textwidth]{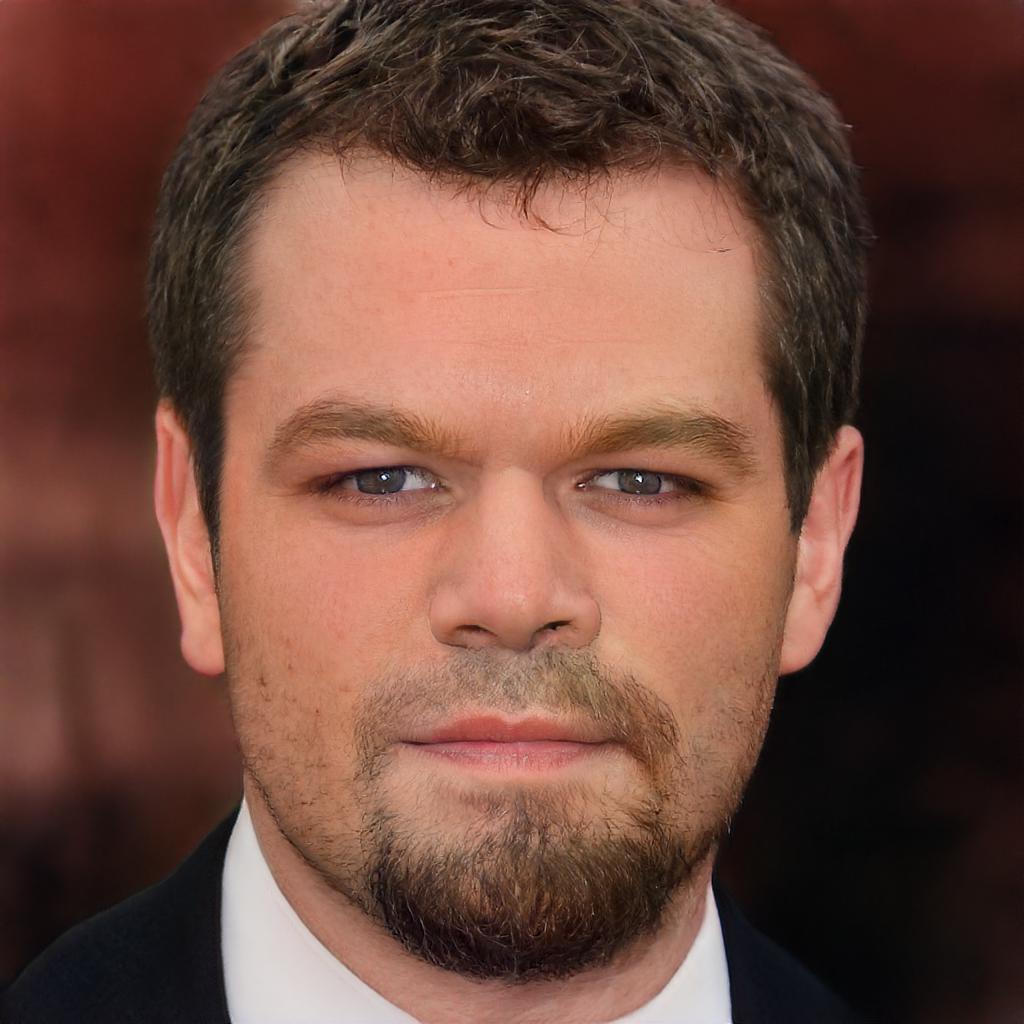} &
    \includegraphics[width=0.1125\textwidth]{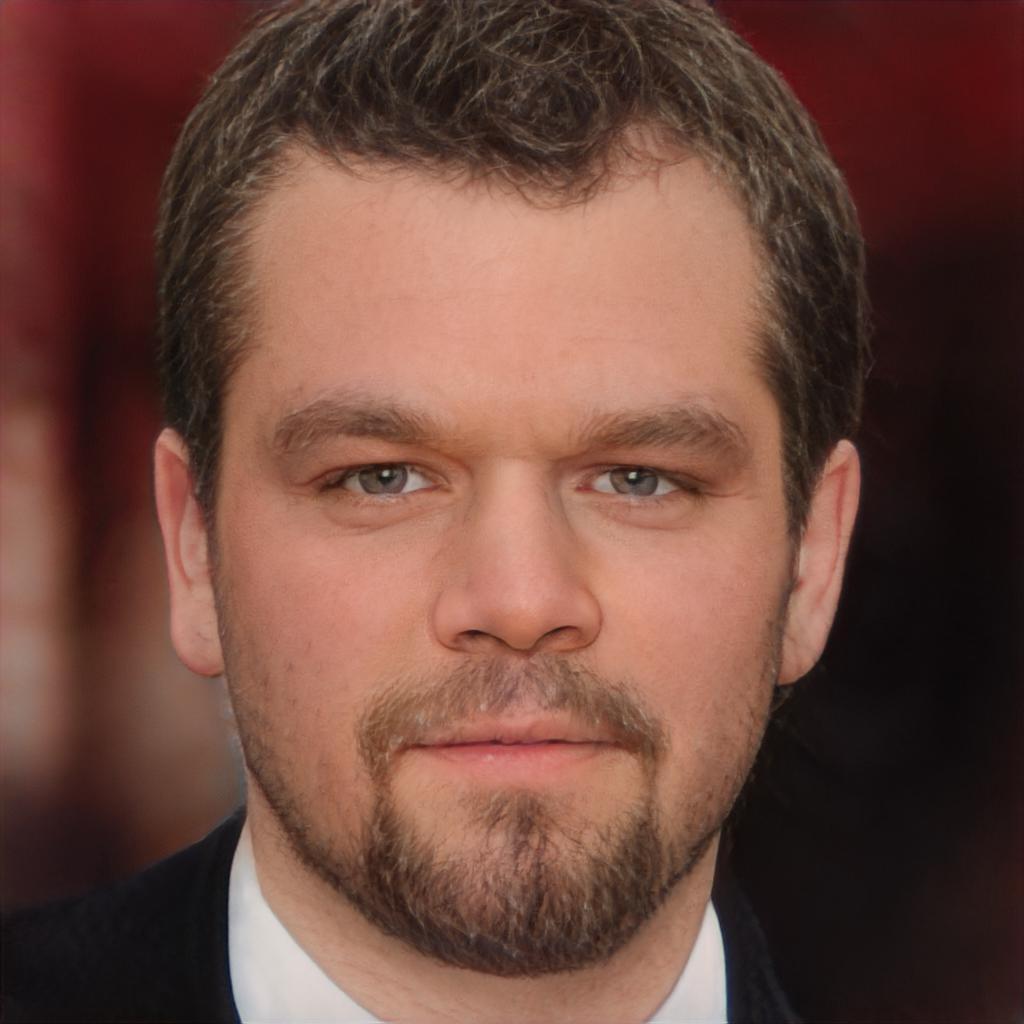} &
    \includegraphics[width=0.1125\textwidth]{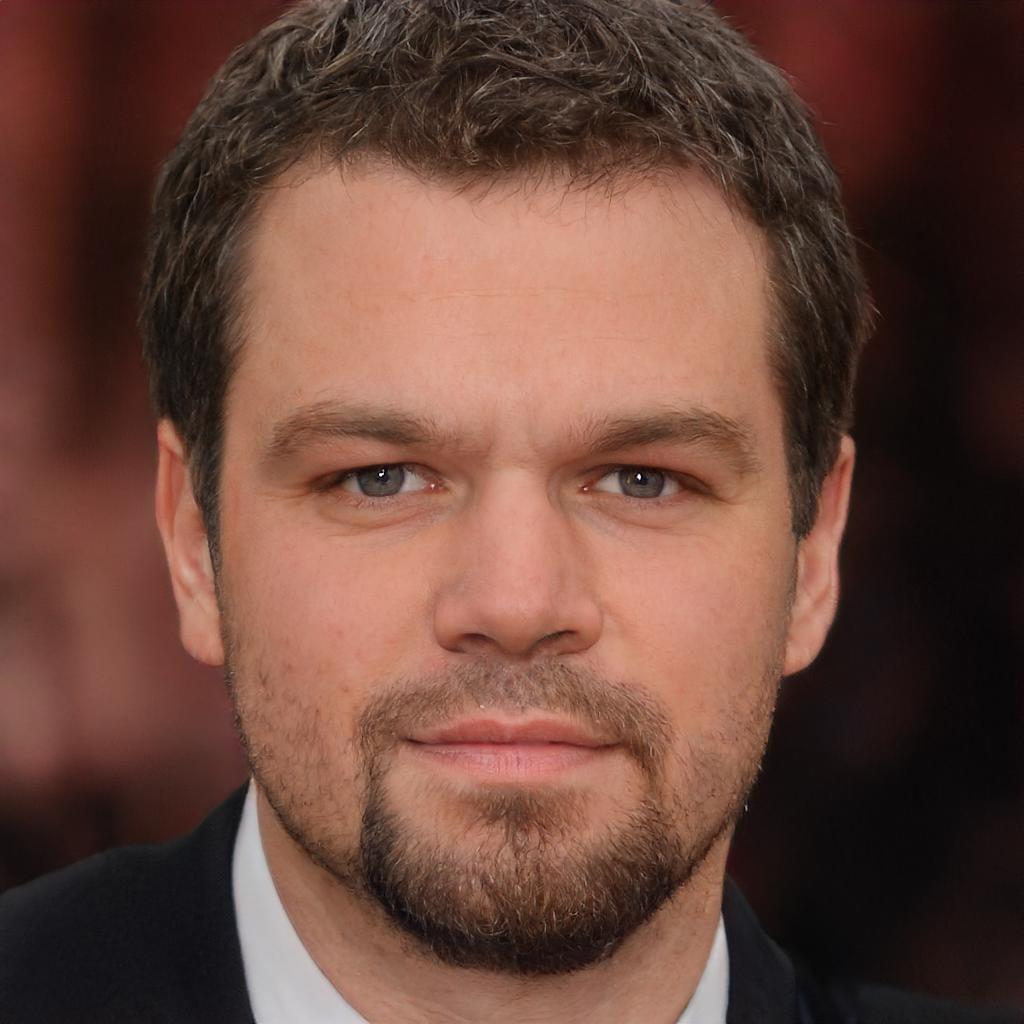} &
    \includegraphics[width=0.1125\textwidth]{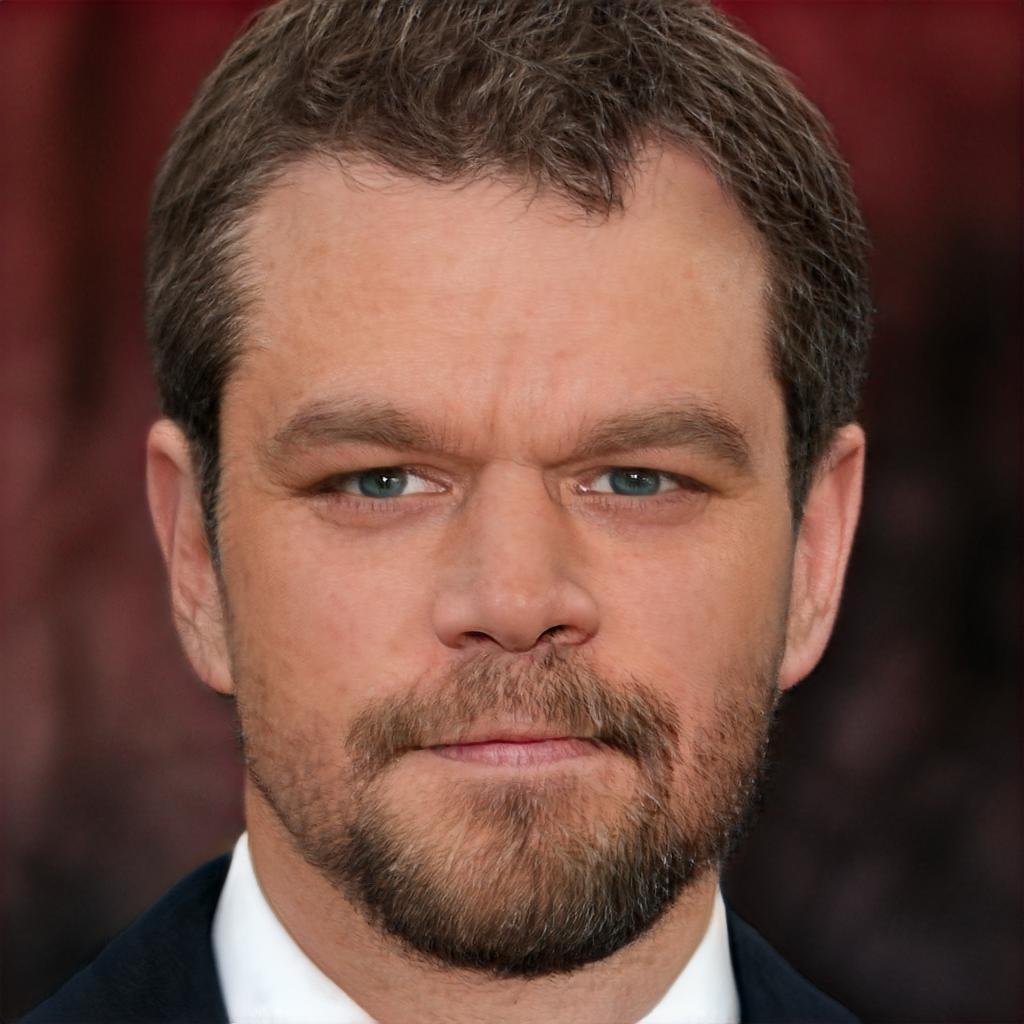} \\

    \raisebox{0.05in}{\rotatebox{90}{$+$Brown Hair}} &
    \includegraphics[width=0.1125\textwidth]{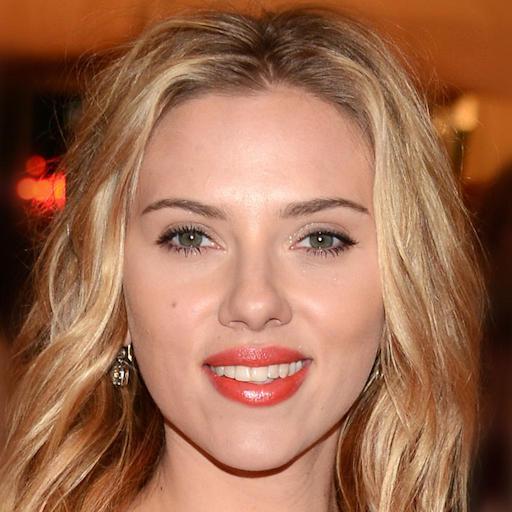} &
    \includegraphics[width=0.1125\textwidth]{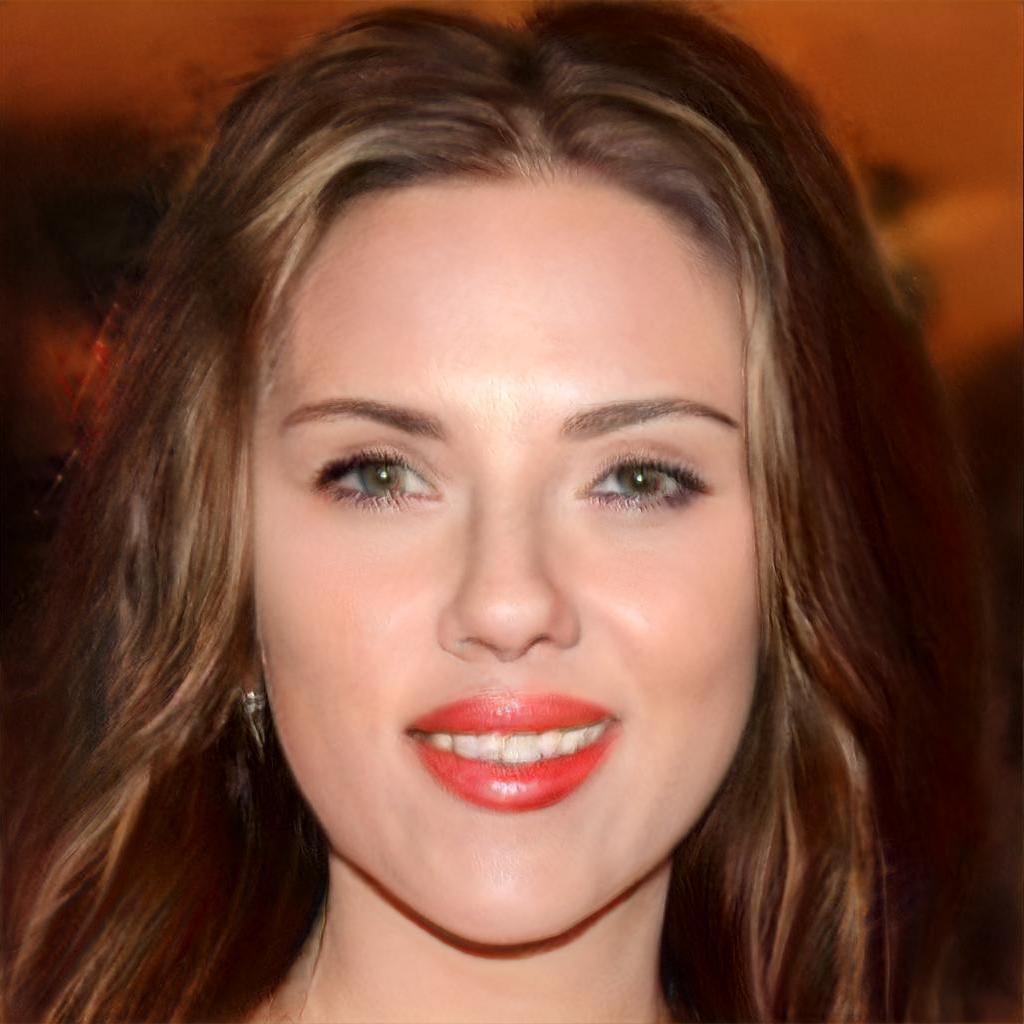} &
    \includegraphics[width=0.1125\textwidth]{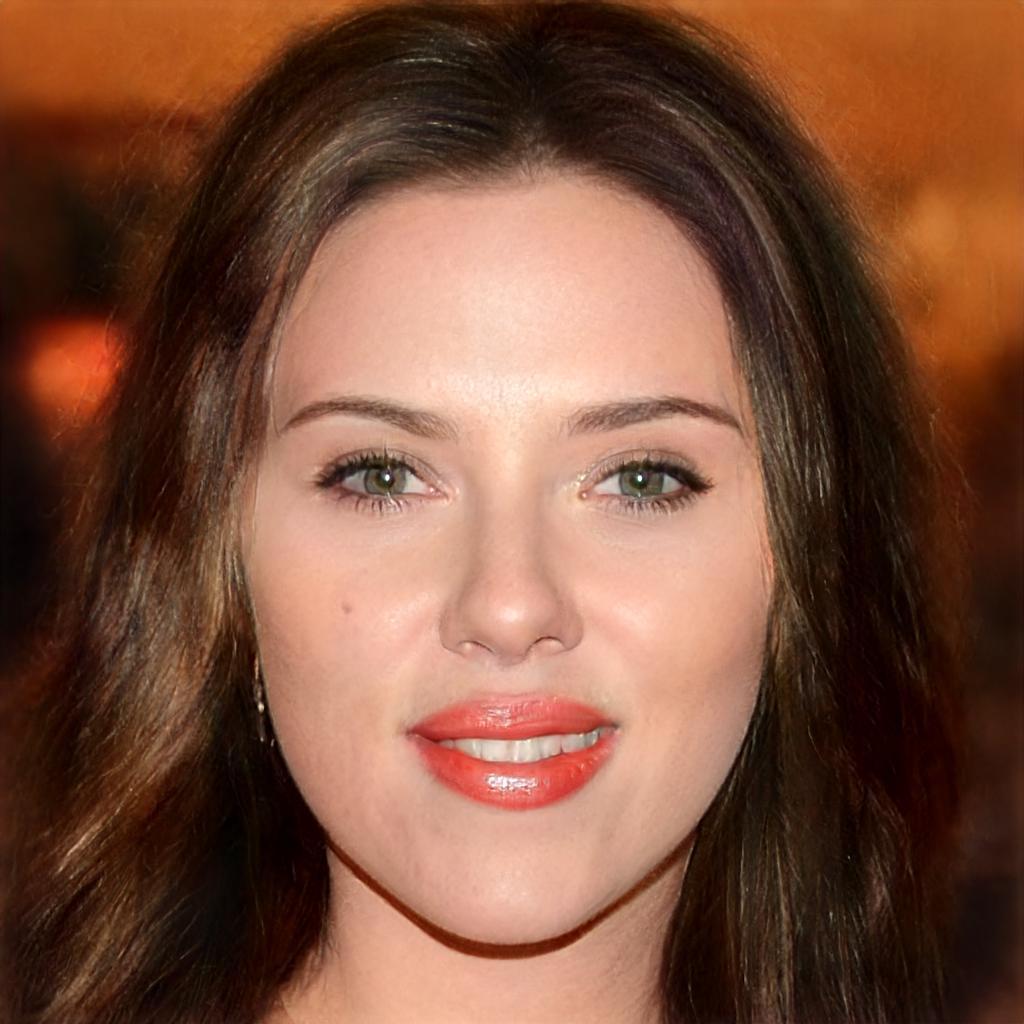} &
    \includegraphics[width=0.1125\textwidth]{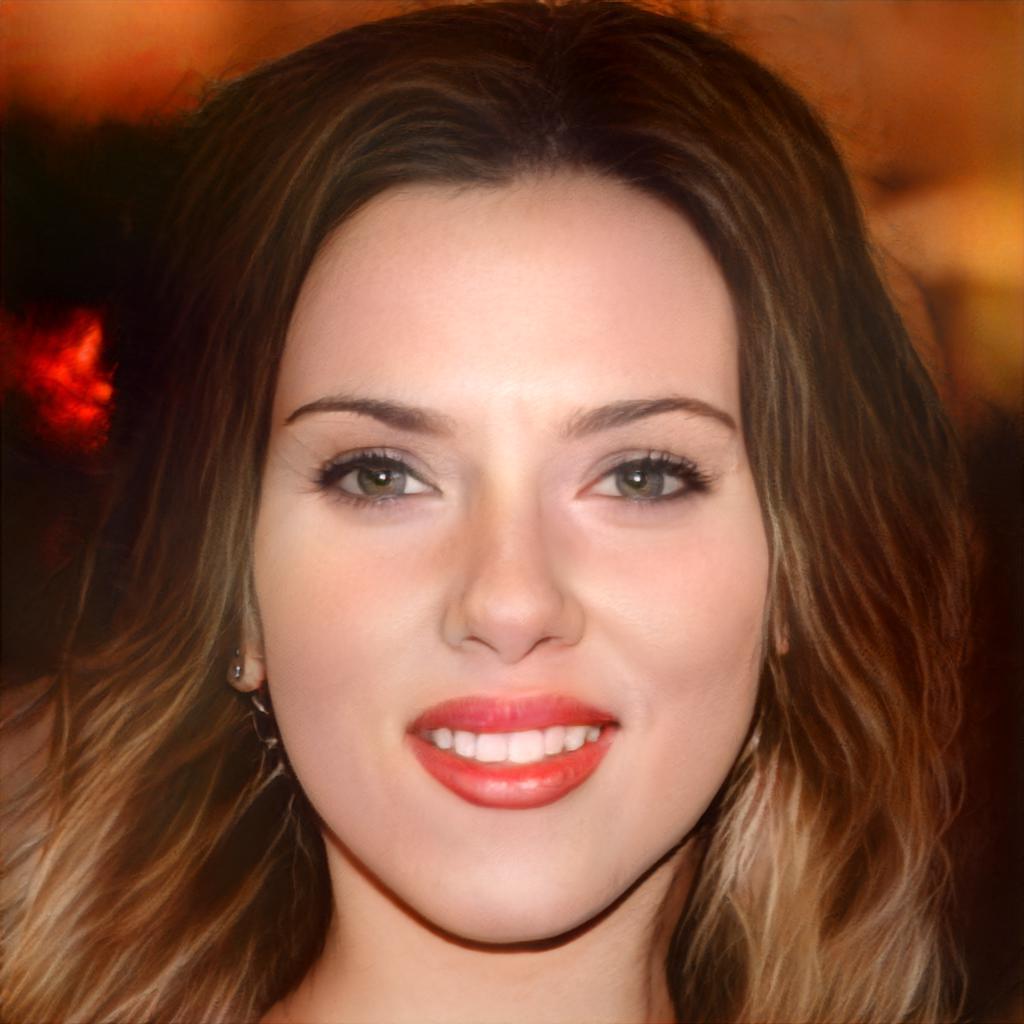} &
    \includegraphics[width=0.1125\textwidth]{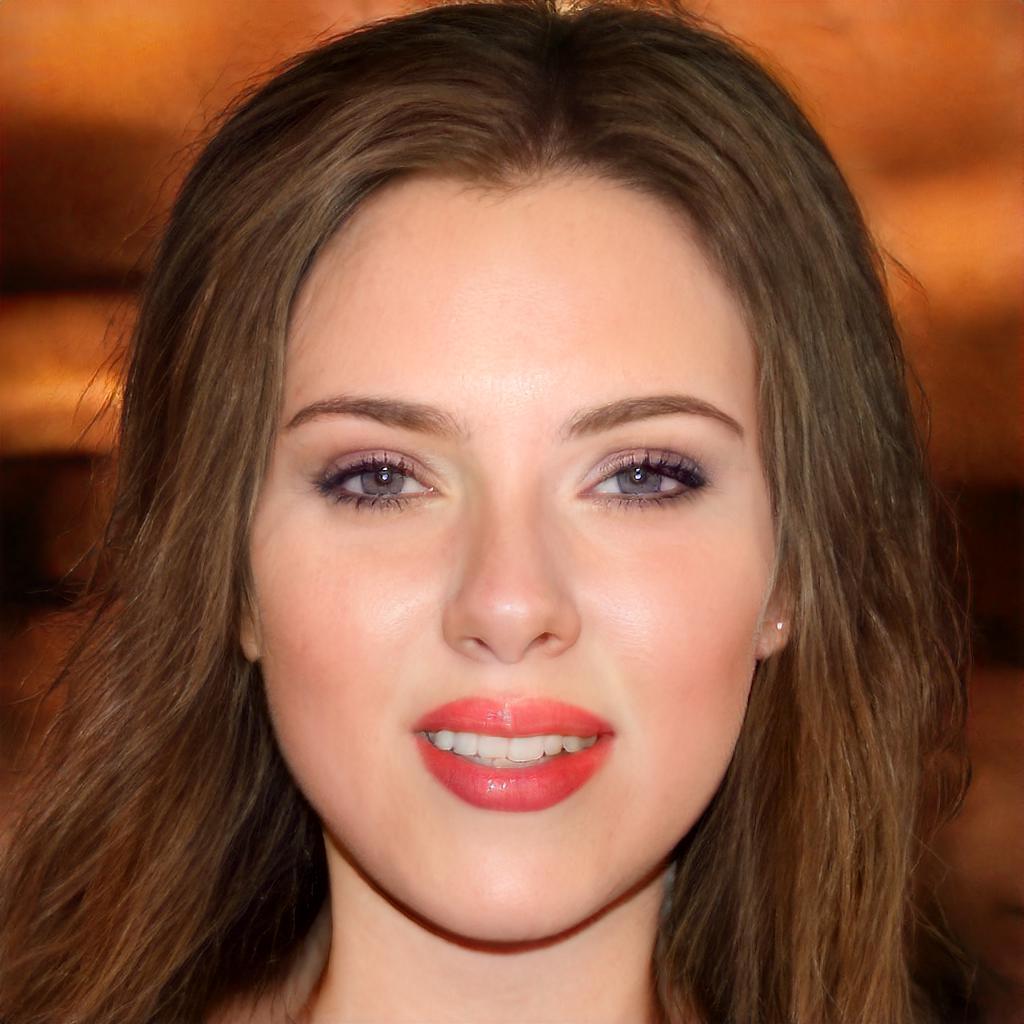} &
    \includegraphics[width=0.1125\textwidth]{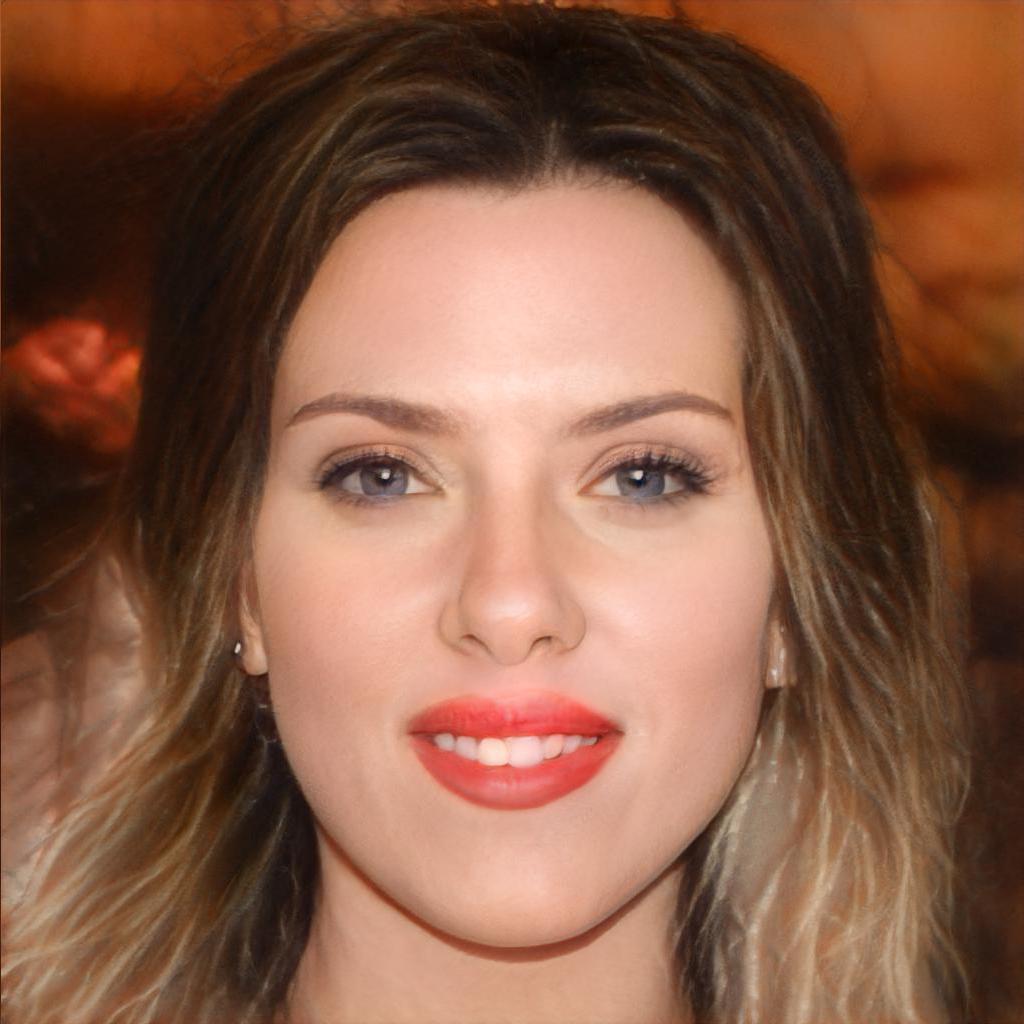} &
    \includegraphics[width=0.1125\textwidth]{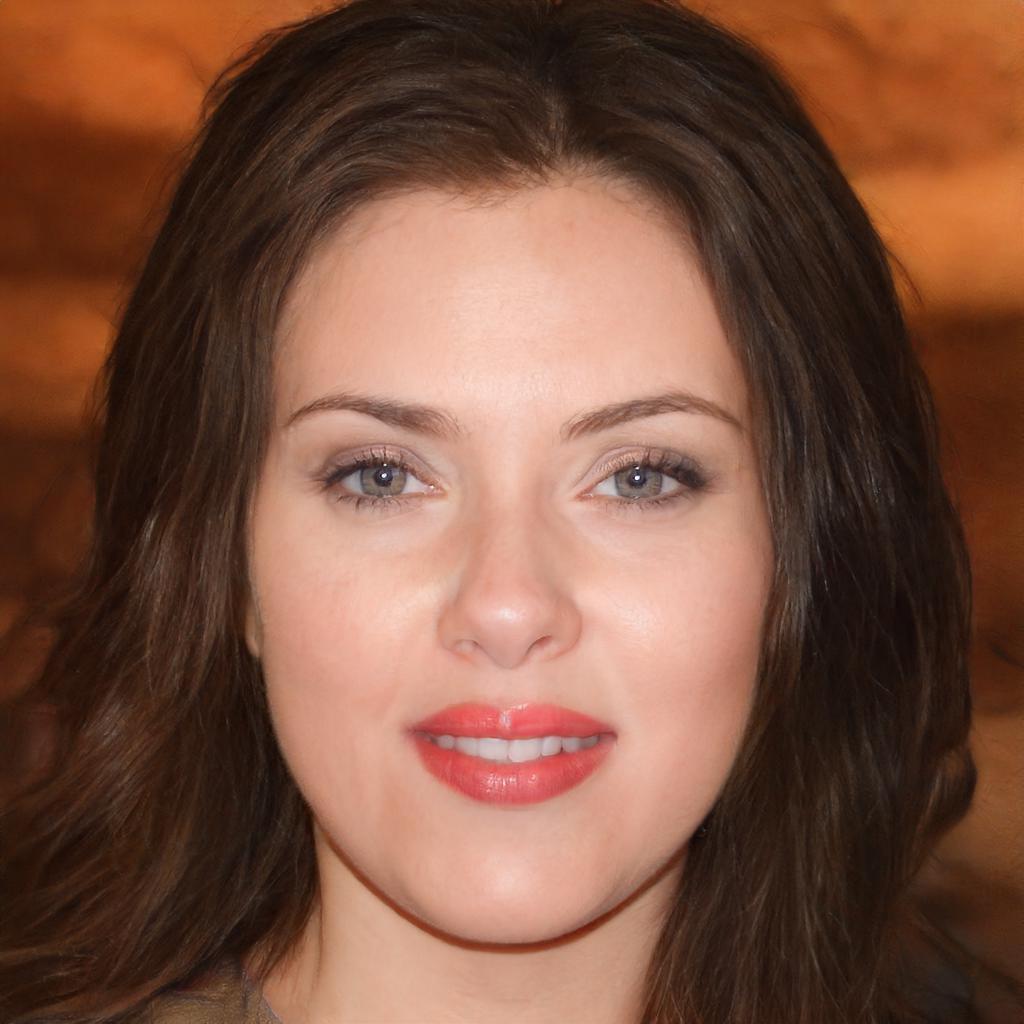} &
    \includegraphics[width=0.1125\textwidth]{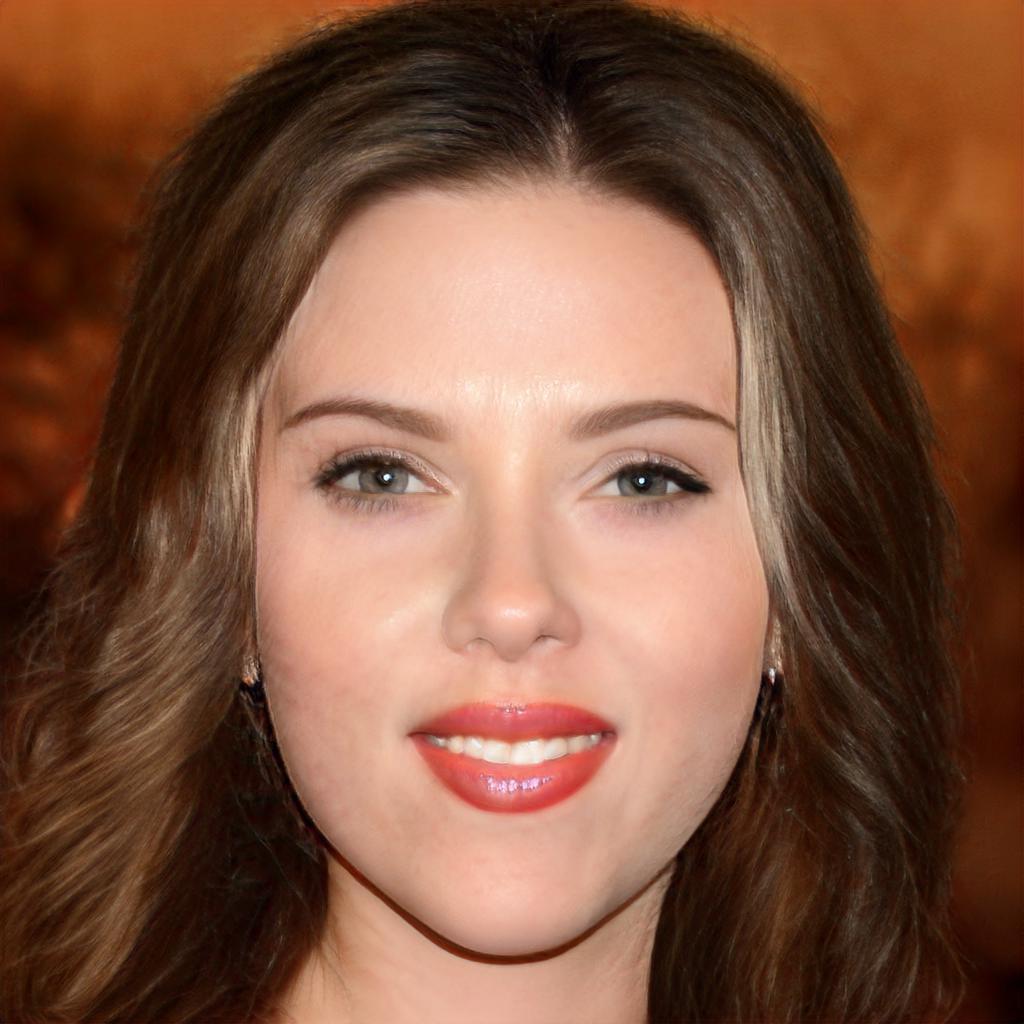} \\
\end{tabular}
    }
    \caption{Additional editing comparisons between inversion techniques obtained using StyleClip~\cite{patashnik2021styleclip} (first five rows) and InterFaceGAN~\cite{shen2020interpreting} (bottom four rows).} 
    \label{fig:styleclip_faces_editing_comparison}
\end{figure*}
\begin{figure*}

    \vspace{1cm}

    \setlength{\tabcolsep}{1pt}
    \centering
    { \small
    \begin{tabular}{c c c c c c c c}
    \includegraphics[width=0.12\linewidth]{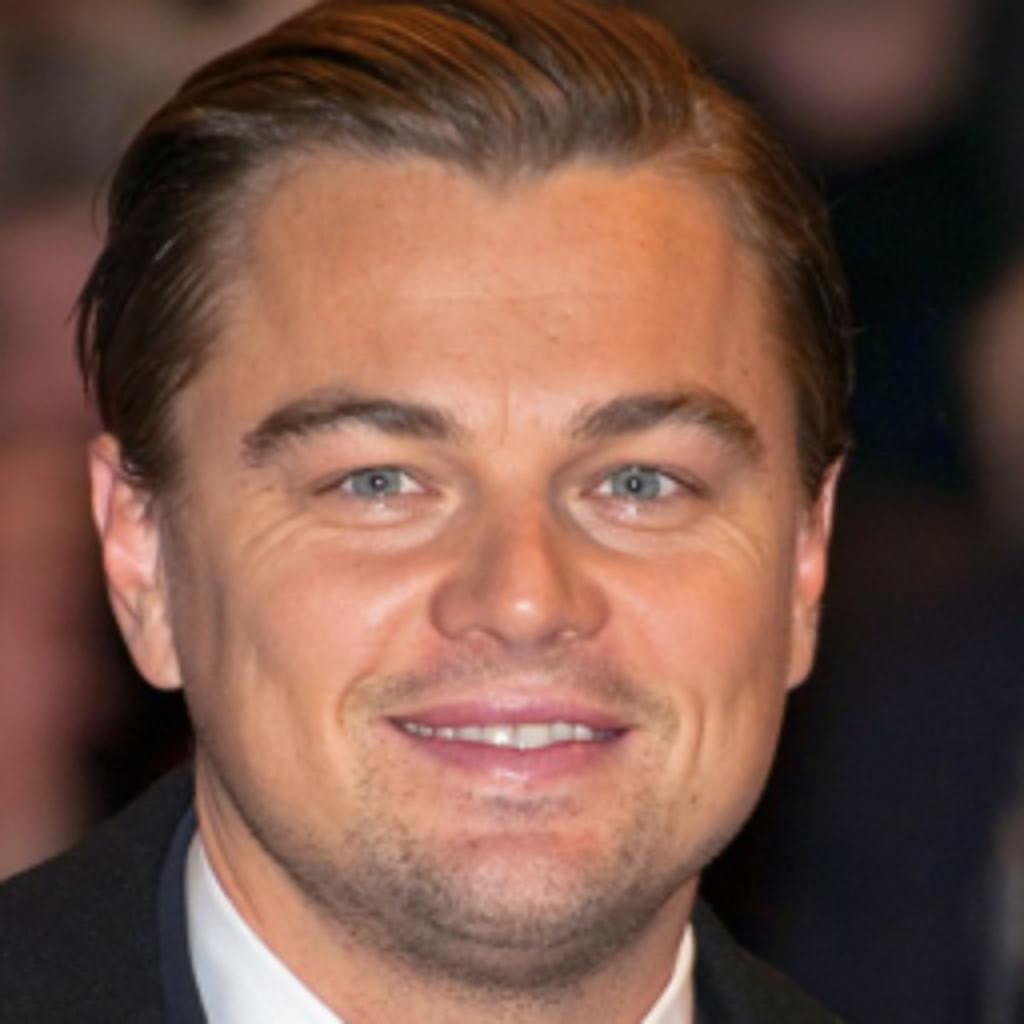} &
    \includegraphics[width=0.12\linewidth]{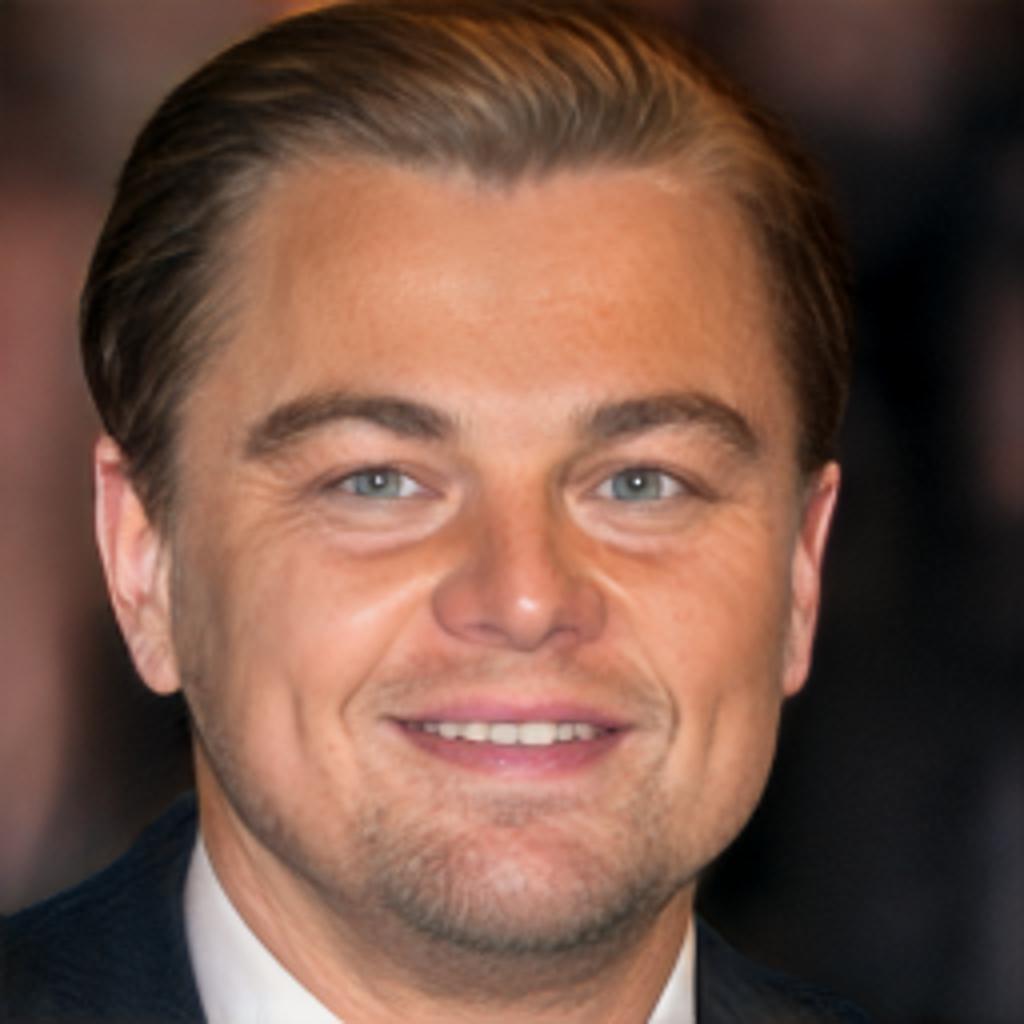} &
    \includegraphics[width=0.12\linewidth]{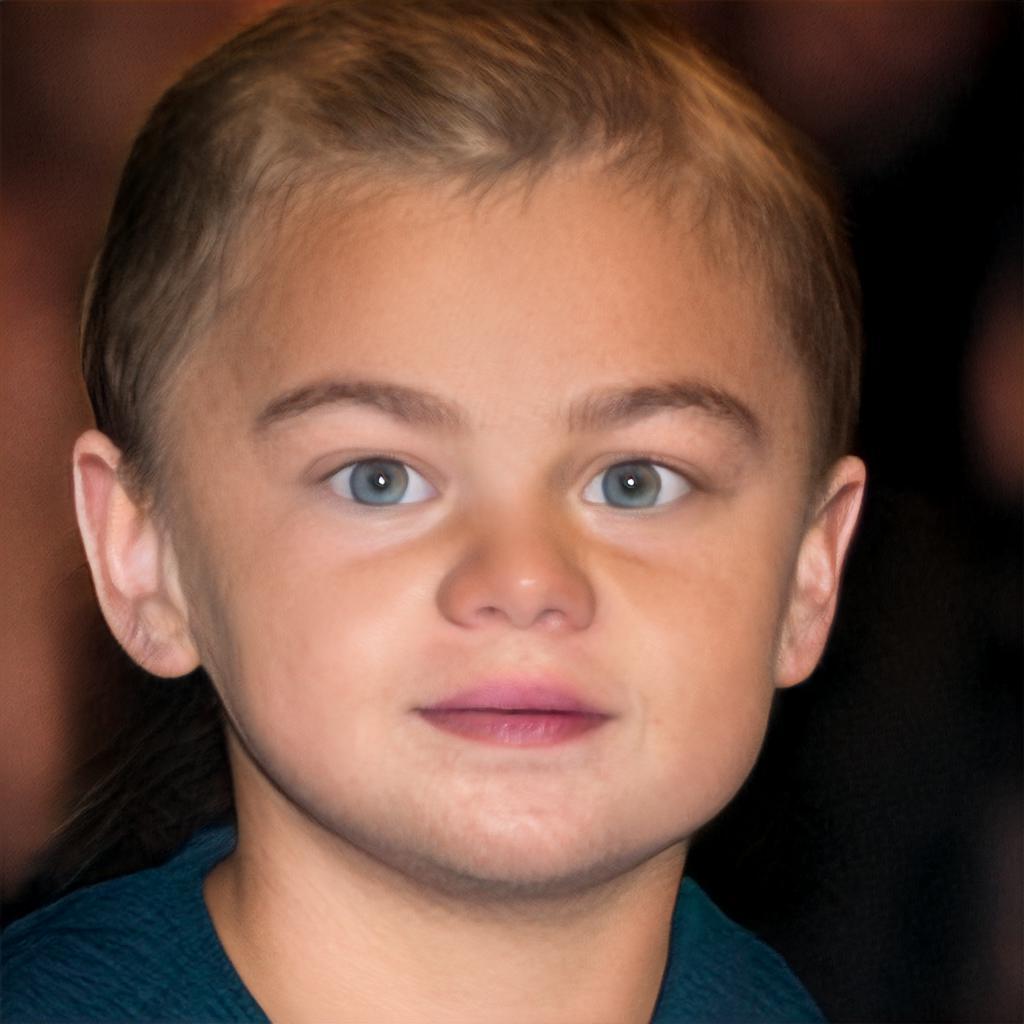} &
    \includegraphics[width=0.12\linewidth]{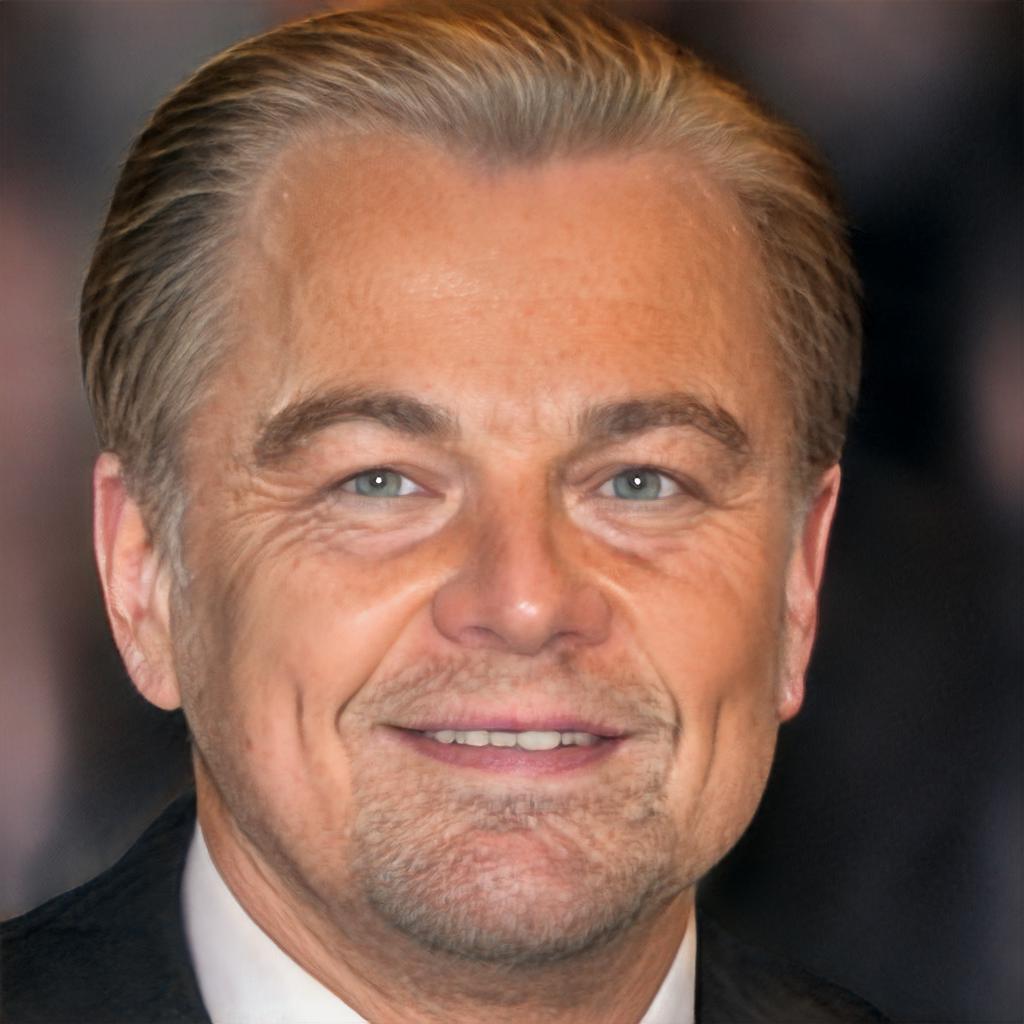} &
    \includegraphics[width=0.12\linewidth]{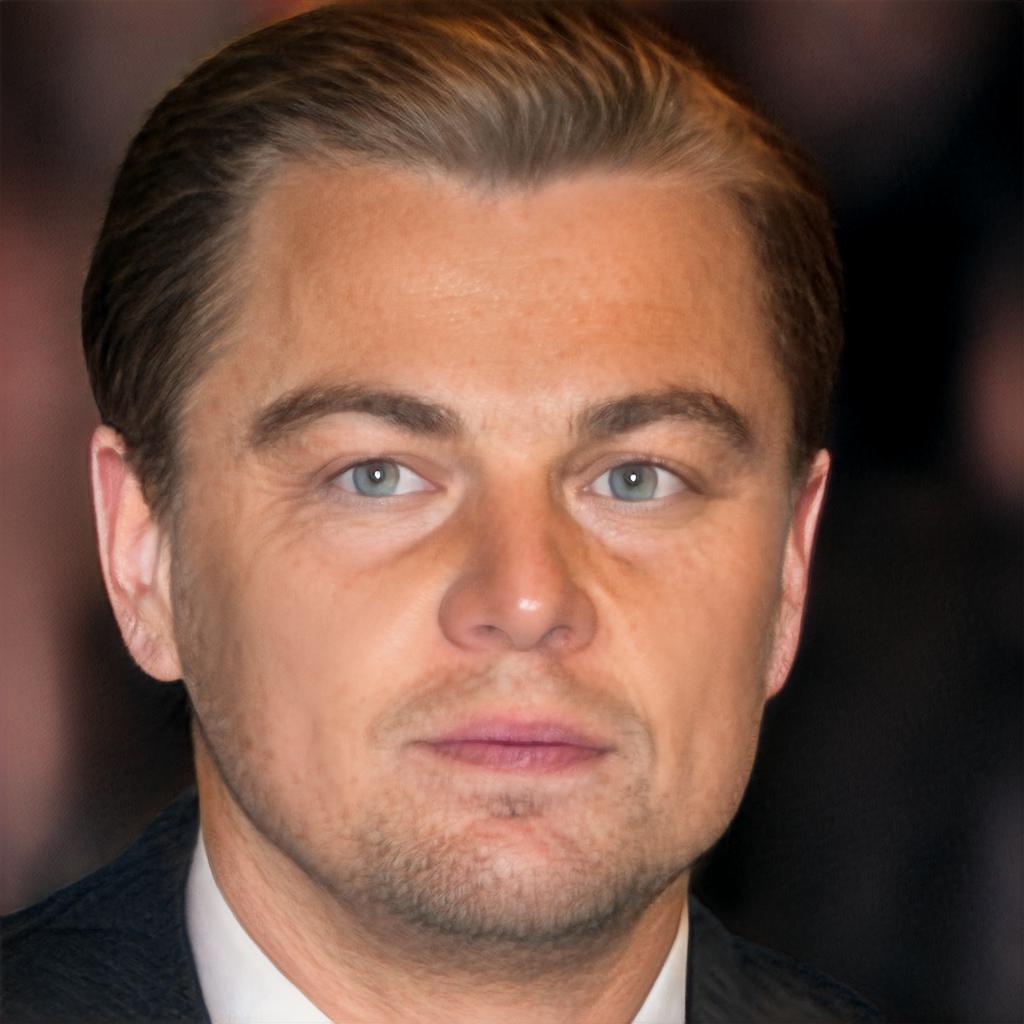} &
    \includegraphics[width=0.12\linewidth]{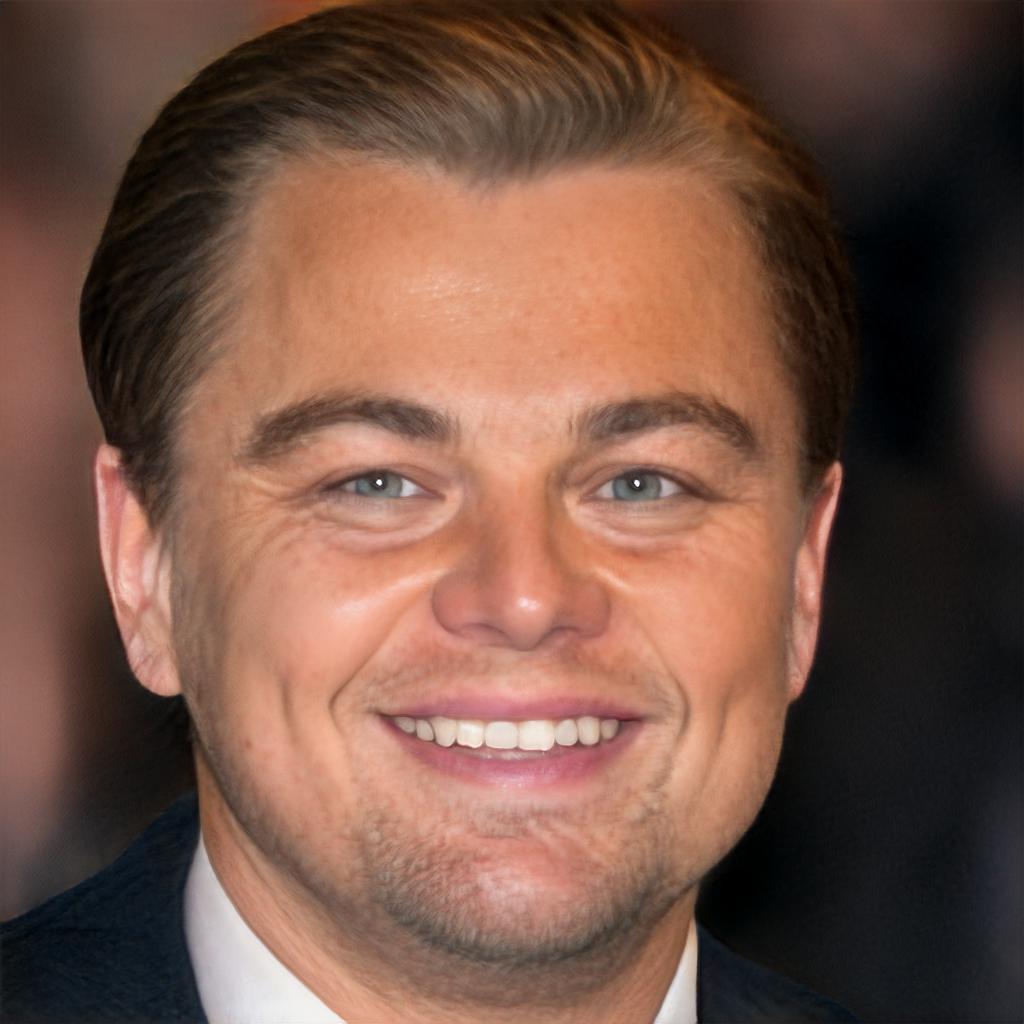} &
    \includegraphics[width=0.12\linewidth]{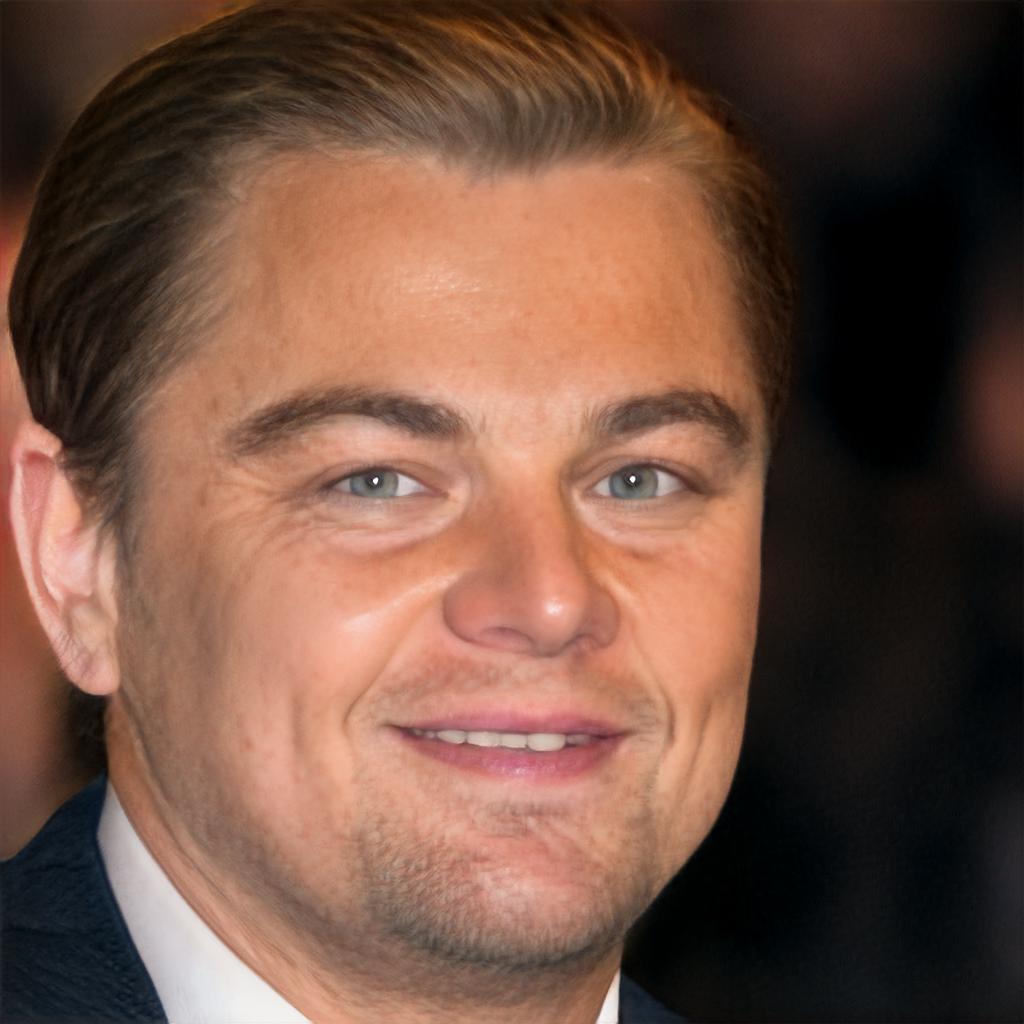} &
    \includegraphics[width=0.12\linewidth]{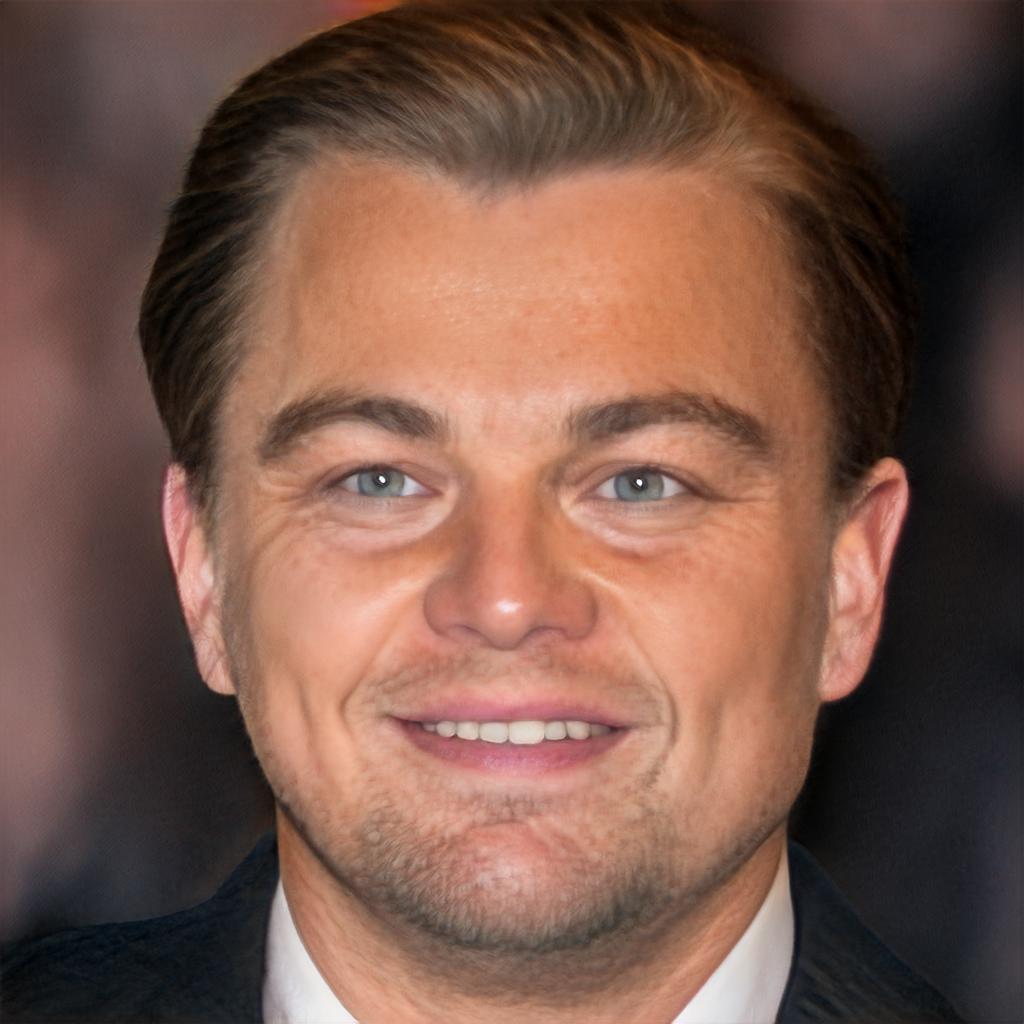} \\
    
    \includegraphics[width=0.12\linewidth]{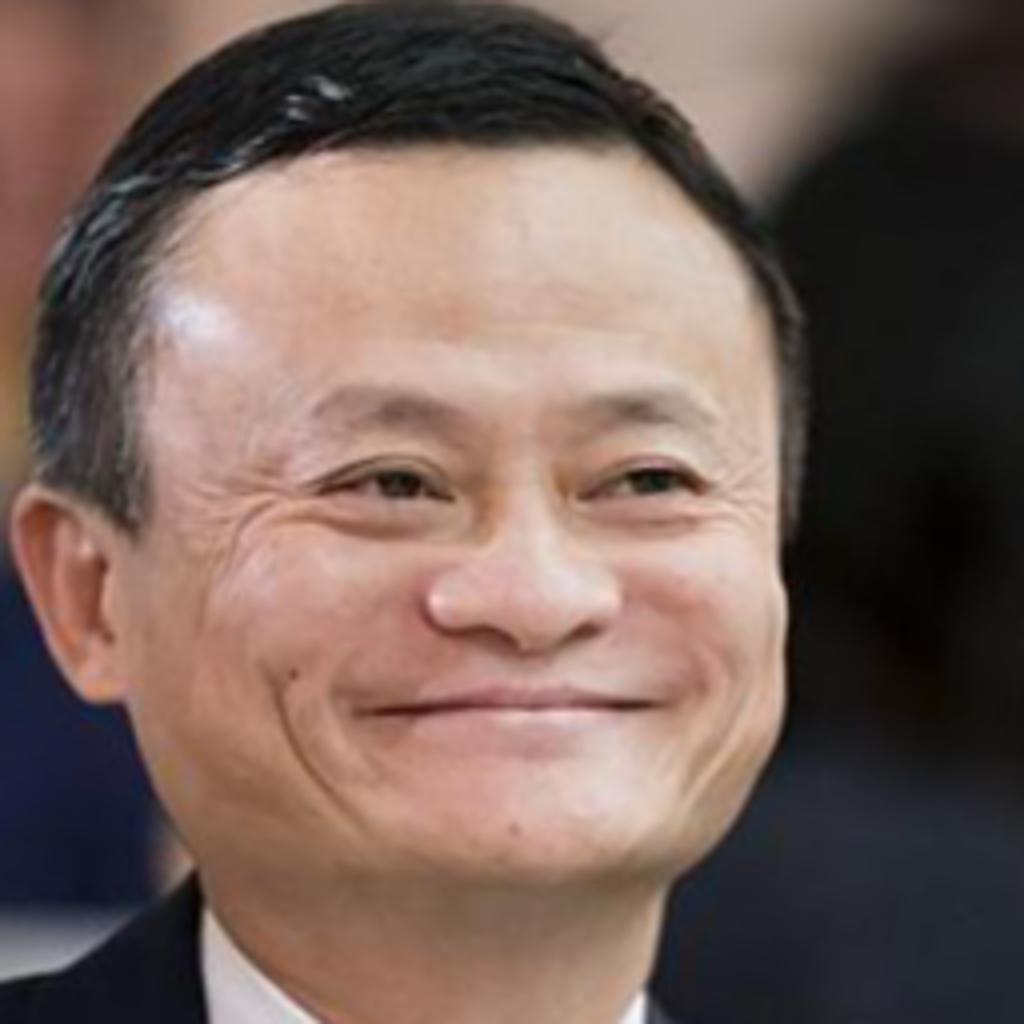} &
    \includegraphics[width=0.12\linewidth]{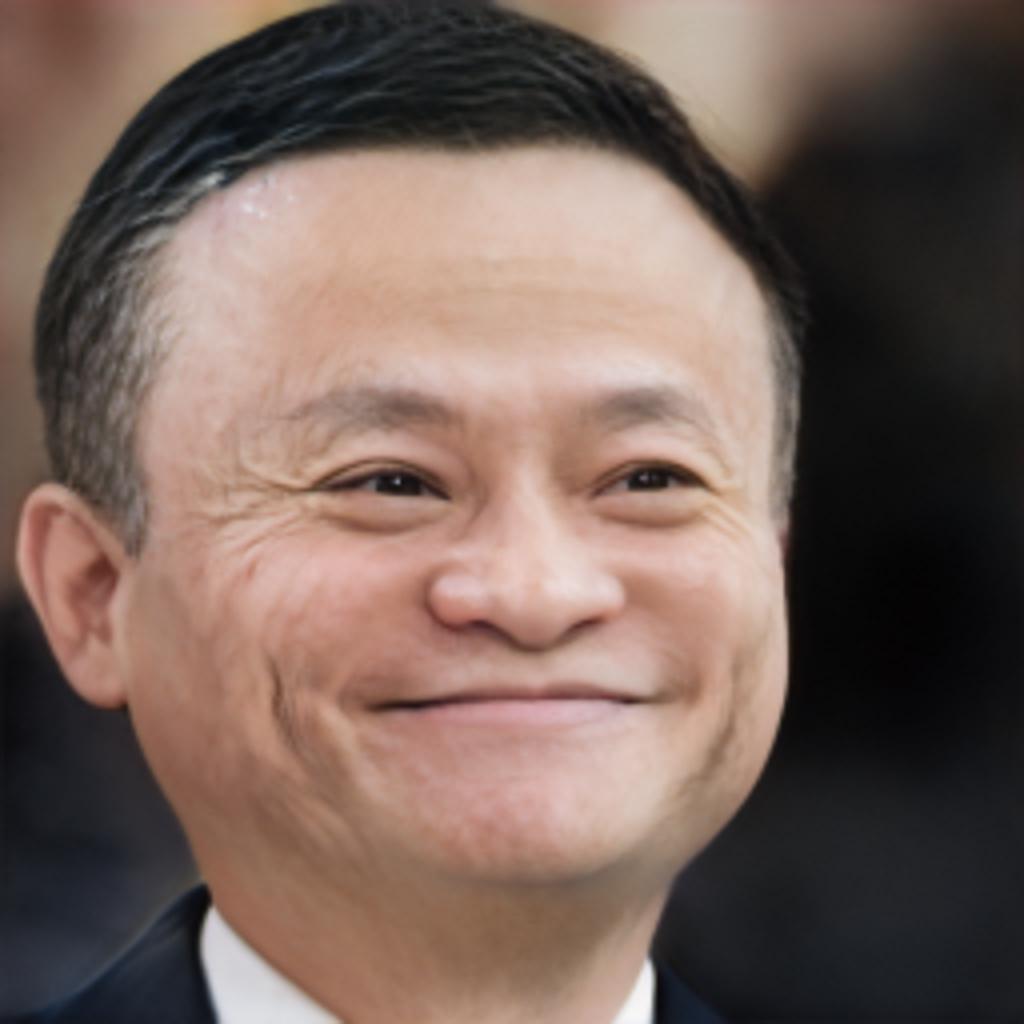} &
    \includegraphics[width=0.12\linewidth]{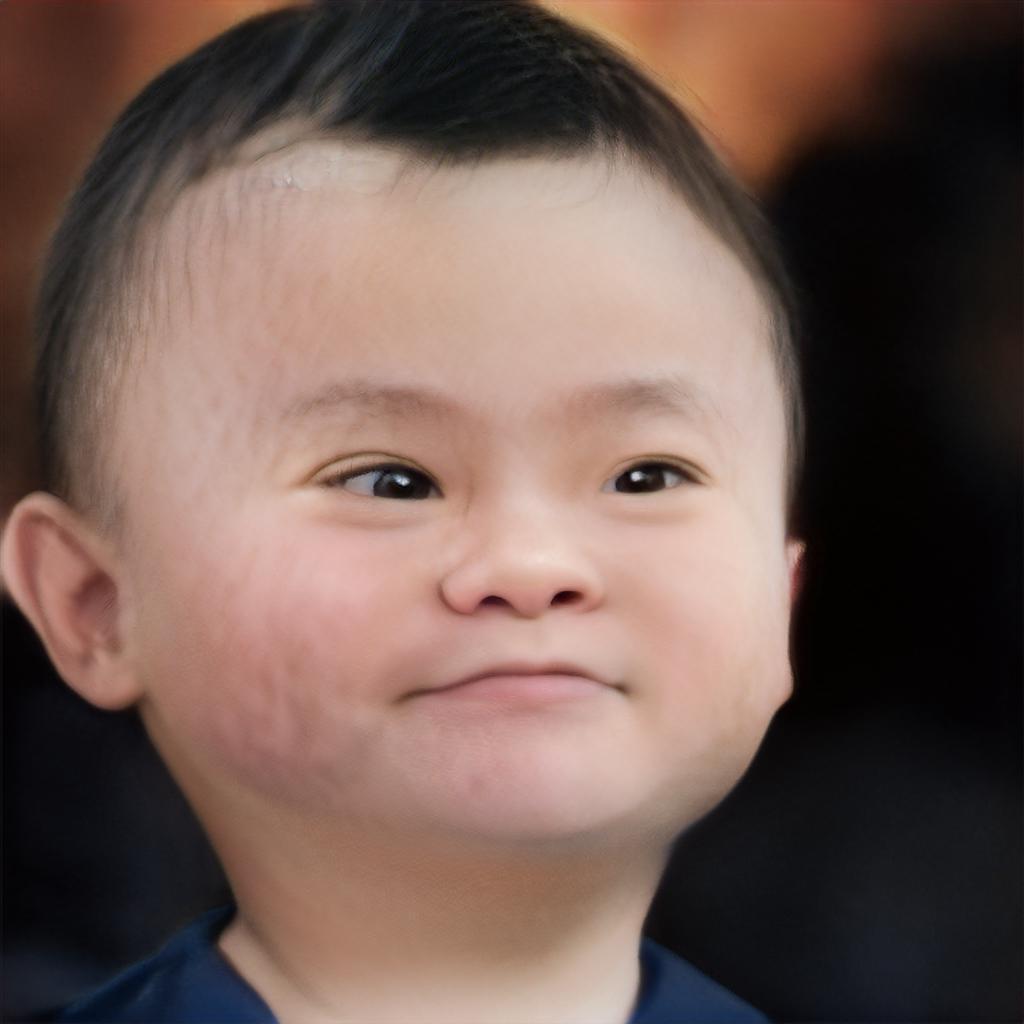} &
    \includegraphics[width=0.12\linewidth]{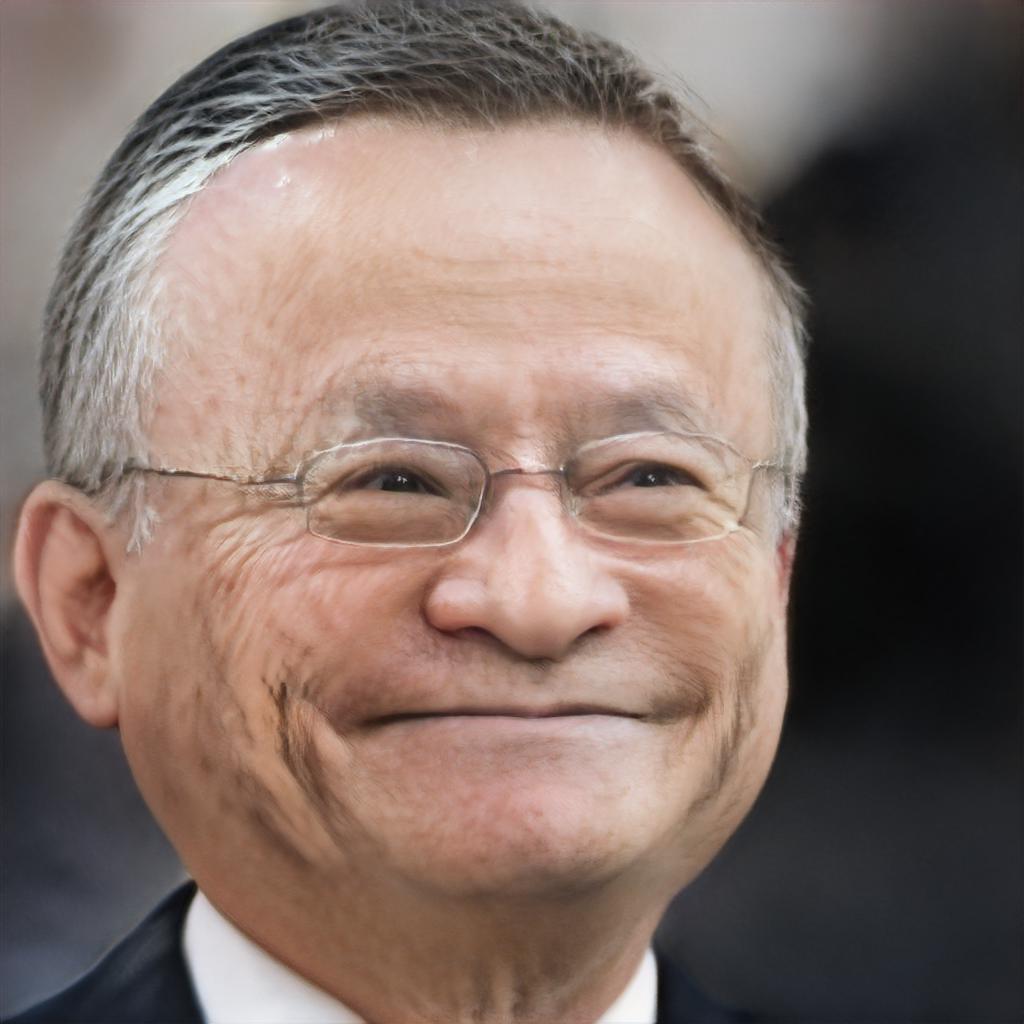} &
    \includegraphics[width=0.12\linewidth]{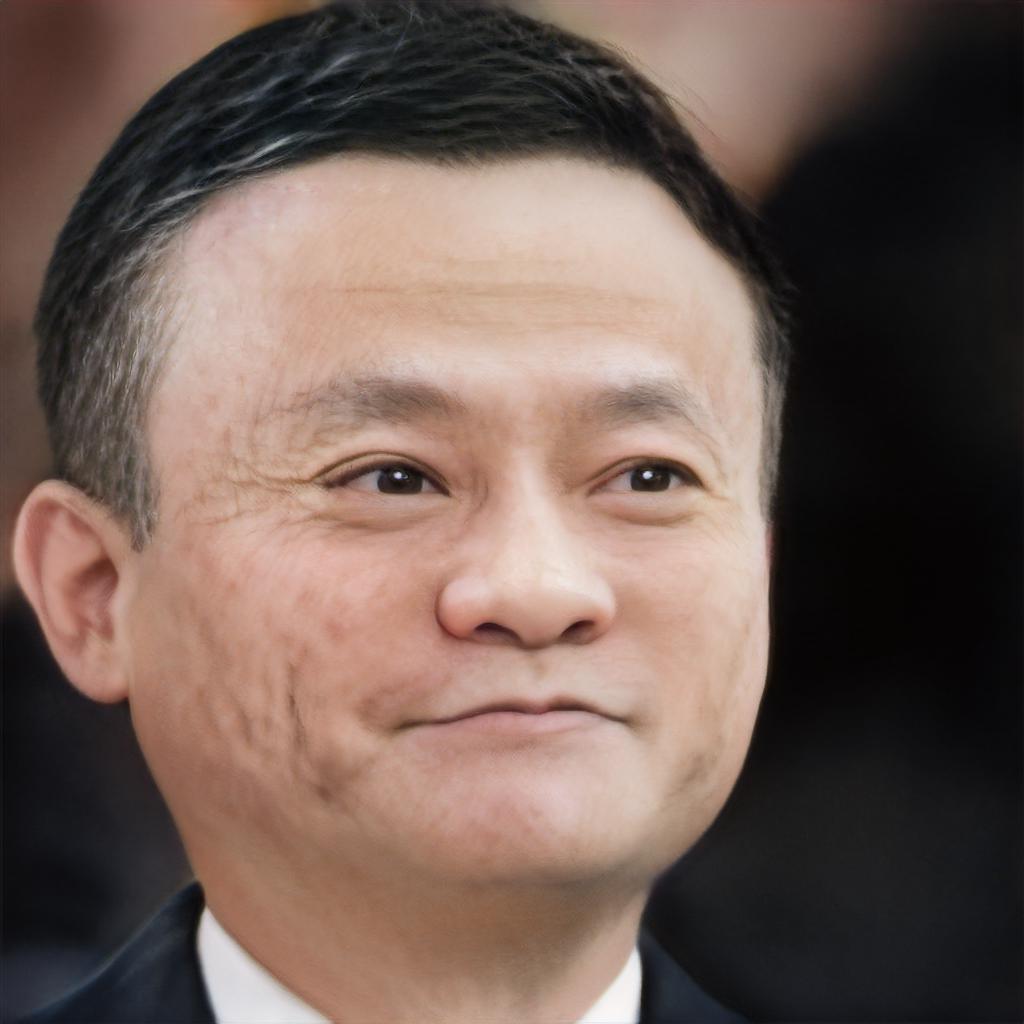} &
    \includegraphics[width=0.12\linewidth]{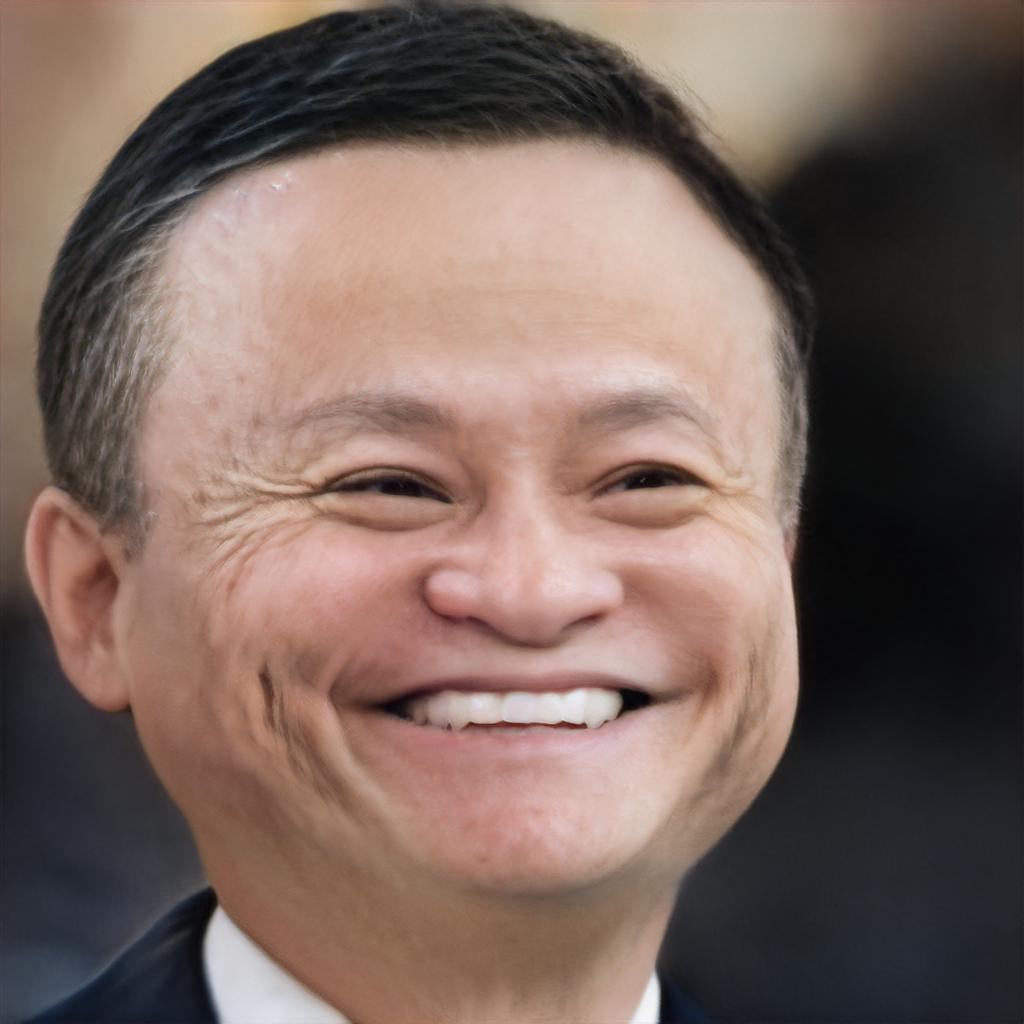} &
    \includegraphics[width=0.12\linewidth]{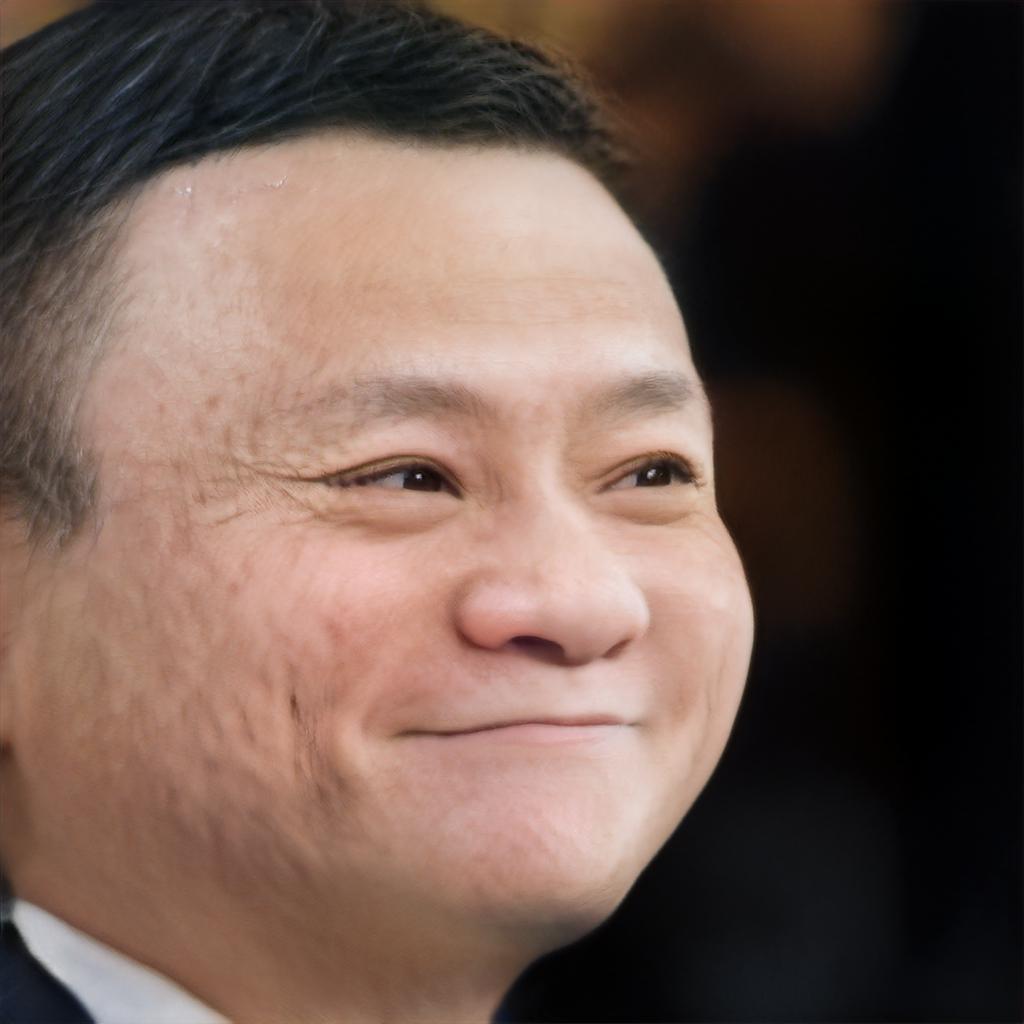} &
    \includegraphics[width=0.12\linewidth]{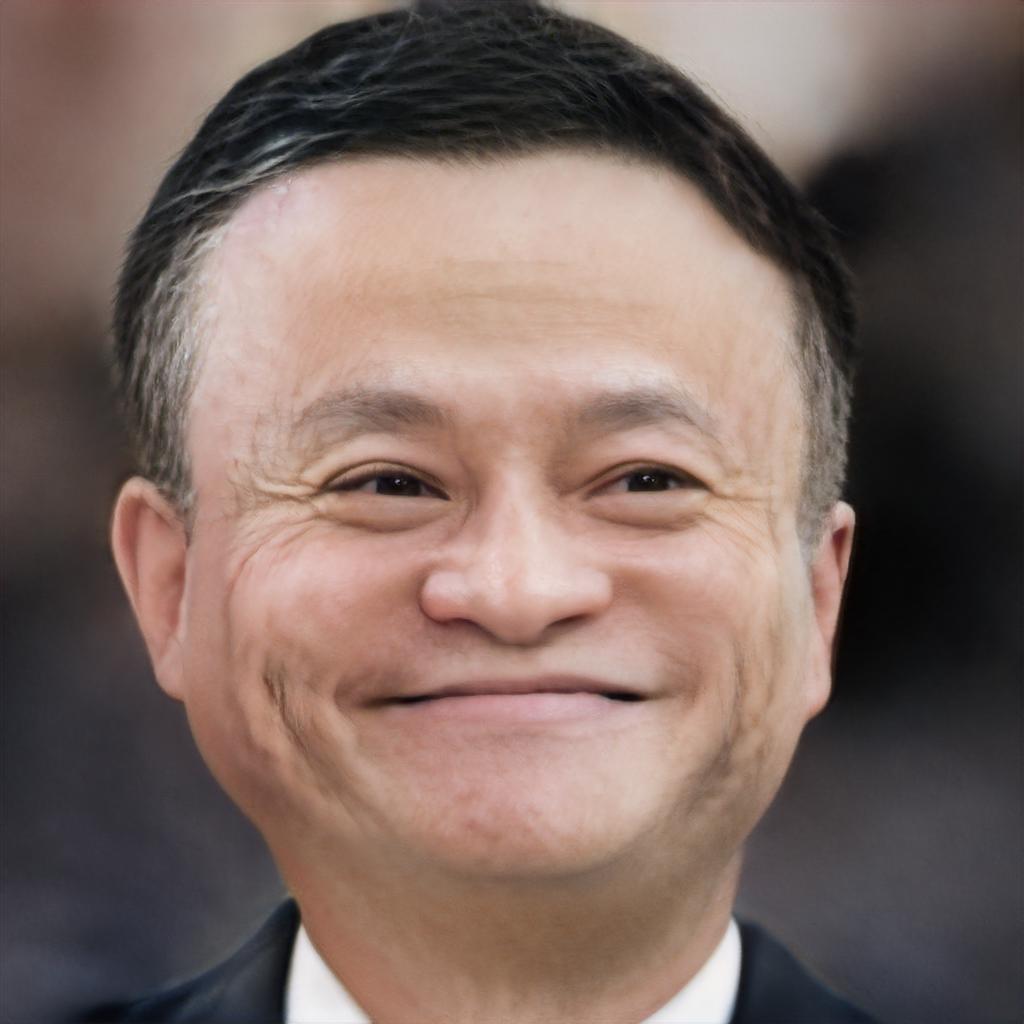} \\
\end{tabular}
    }
    \caption{Additional reconstruction and editing results obtained by HyperStyle over the facial domain using InterFaceGAN~\cite{shen2020interpreting}.}
    \label{fig:supp_editings_interfacegan}
\end{figure*}
\begin{figure*}

    \setlength{\tabcolsep}{1pt}
    \centering
    { \small
    \begin{tabular}{c c | c c c c c}

    \includegraphics[width=0.12\linewidth]{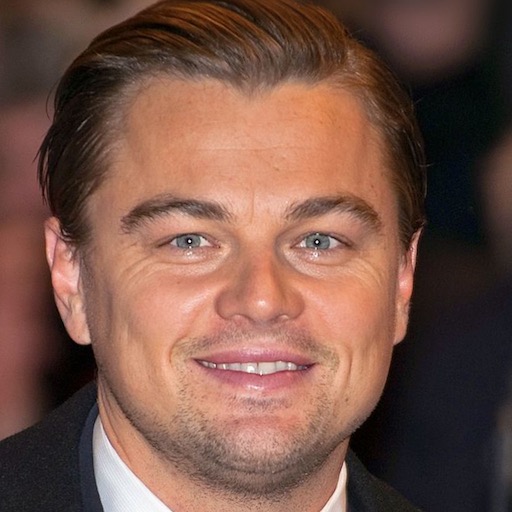} &
    \includegraphics[width=0.12\linewidth]{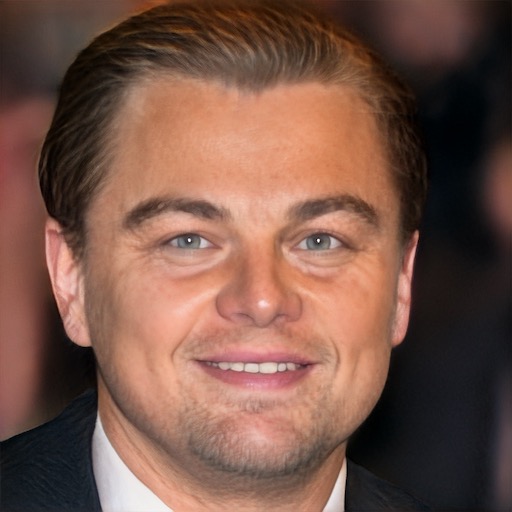} &
    \includegraphics[width=0.12\linewidth]{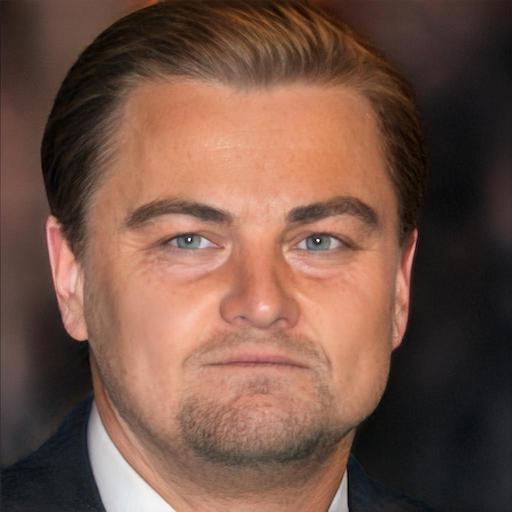} &
    \includegraphics[width=0.12\linewidth]{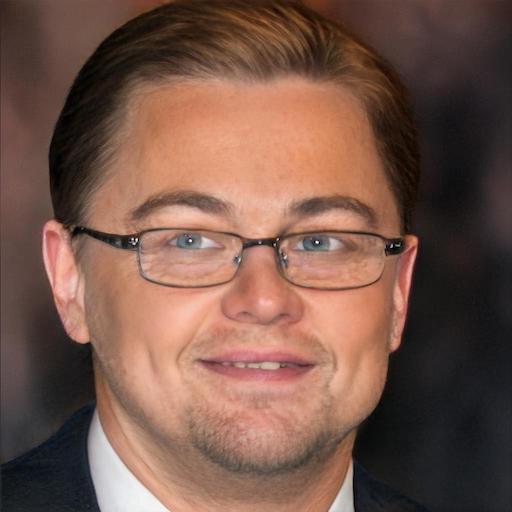} &
    \includegraphics[width=0.12\linewidth]{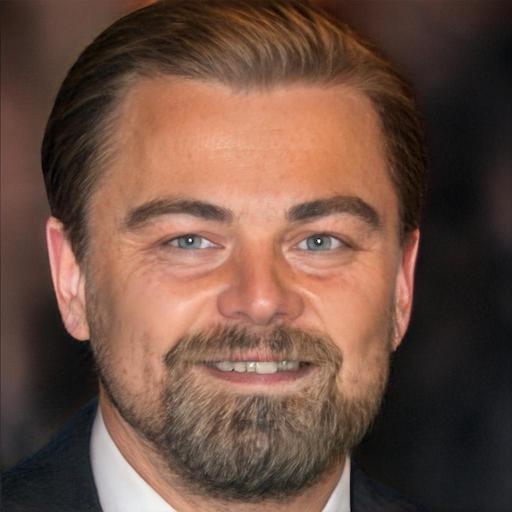} &
    \includegraphics[width=0.12\linewidth]{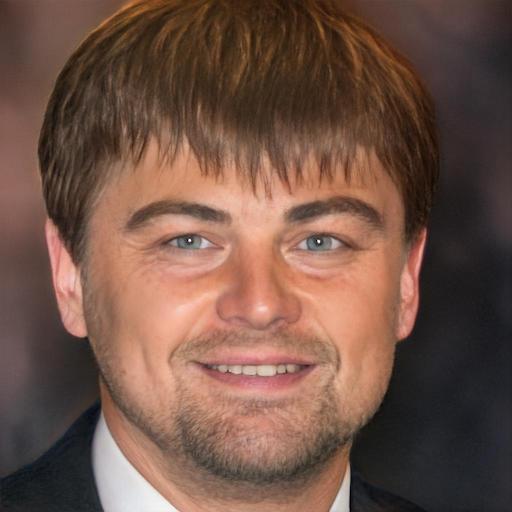} &
    \includegraphics[width=0.12\linewidth]{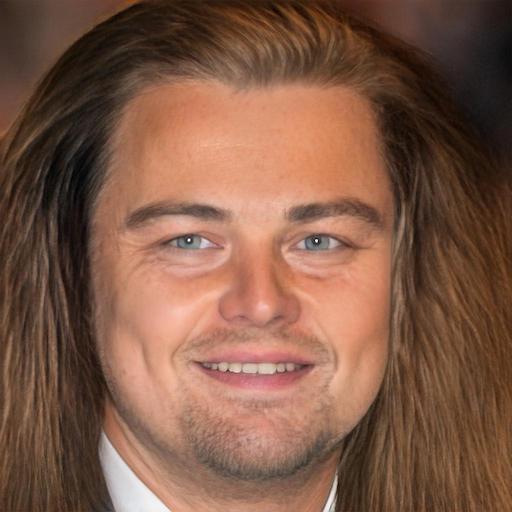} \\
    
    Input & Inversion & Sad & Glasses & Beard & Bowl Cut & Long Hair \\

    \includegraphics[width=0.12\linewidth]{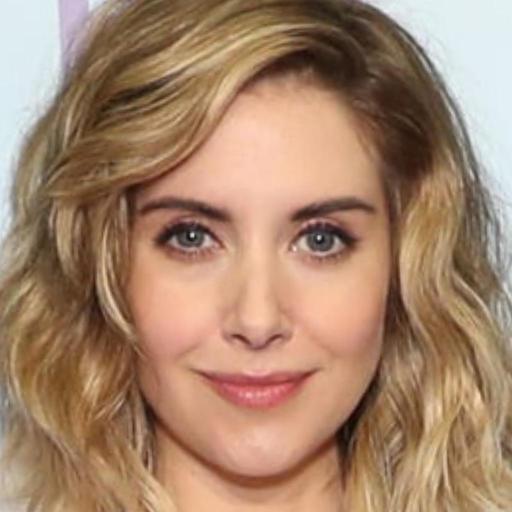} &
    \includegraphics[width=0.12\linewidth]{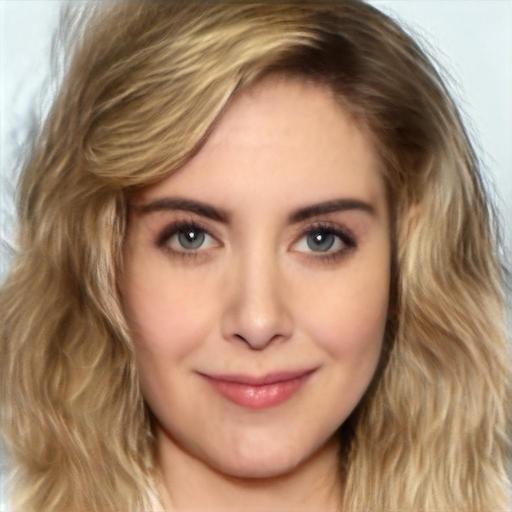} &
    \includegraphics[width=0.12\linewidth]{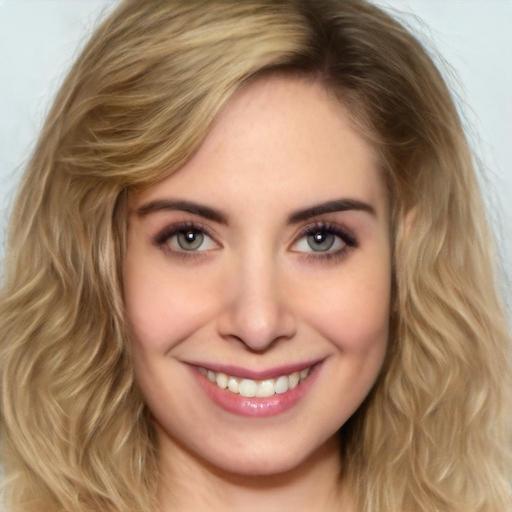} &
    \includegraphics[width=0.12\linewidth]{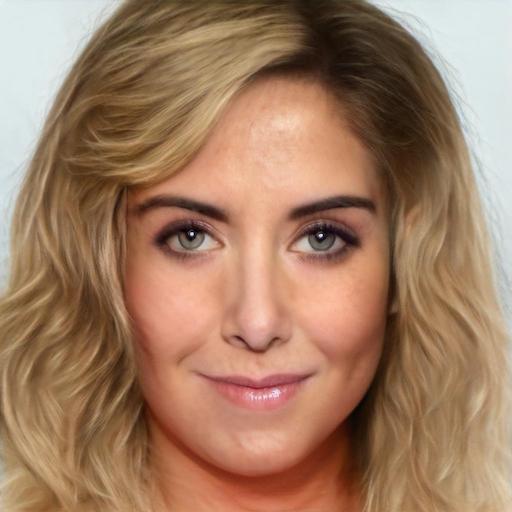} &
    \includegraphics[width=0.12\linewidth]{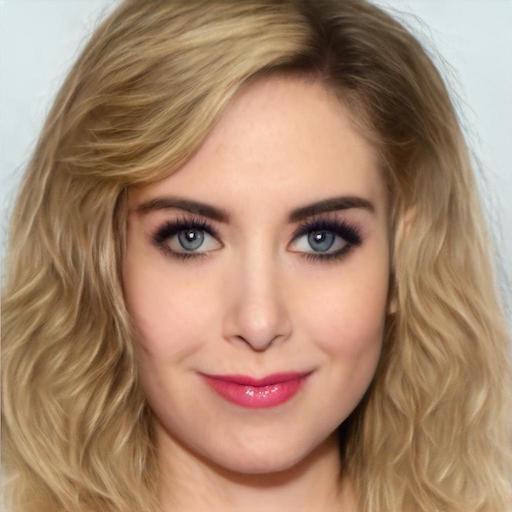} &
    \includegraphics[width=0.12\linewidth]{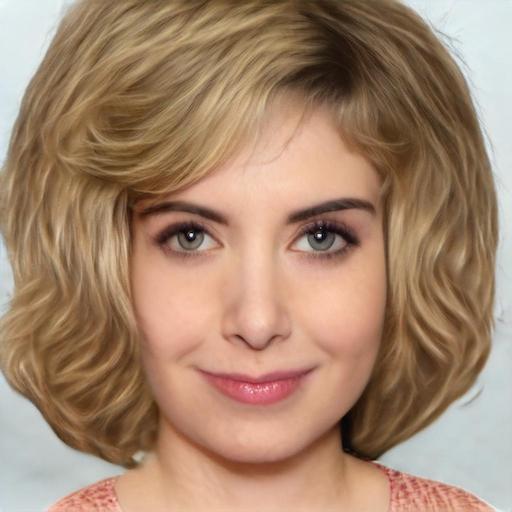} &
    \includegraphics[width=0.12\linewidth]{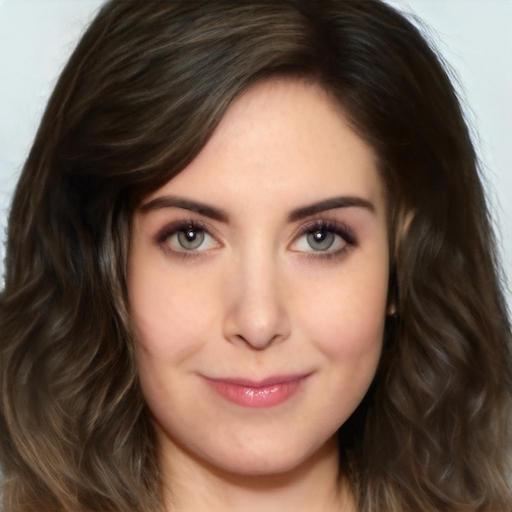} \\
    
    Input & Inversion & Smile & Tanned & Makeup & Short Hair & Brown Hair \\
\end{tabular}
    }
    \caption{Additional reconstruction and editing results obtained by HyperStyle over the facial domain using StyleCLIP's~\cite{shen2020interpreting} global direction approach. We illustrate a wide range of edits including changes to expression, facial hair, makeup, and hairstyle.}
    \label{fig:supp_face_editings_styleclip}
\end{figure*}
\begin{figure*}
    \centering
    \setlength{\belowcaptionskip}{-6pt}
    \setlength{\tabcolsep}{1.5pt}
    {\small
    \begin{tabular}{c c | c c | c c c c c}

        \raisebox{0.1in}{\rotatebox{90}{$+$Grass}} &
        \includegraphics[width=0.1125\textwidth]{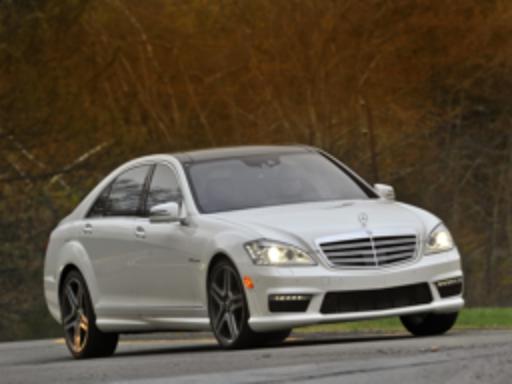} &
        \includegraphics[width=0.1125\textwidth]{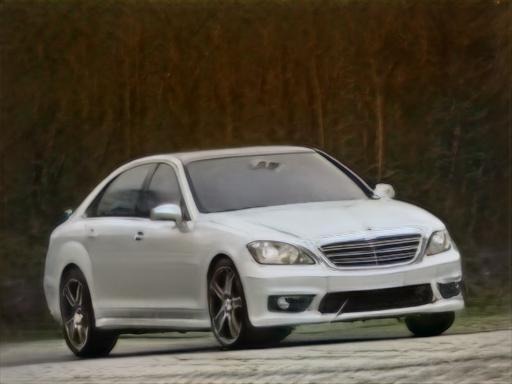} &
        \includegraphics[width=0.1125\textwidth]{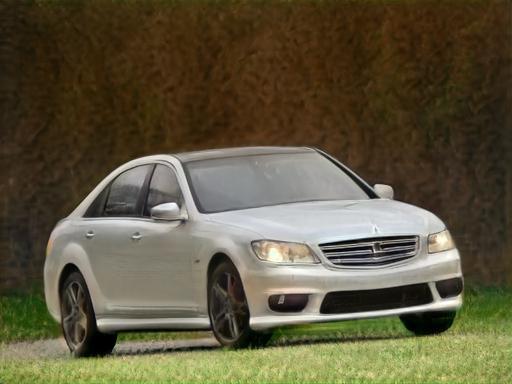} &
        \includegraphics[width=0.1125\textwidth]{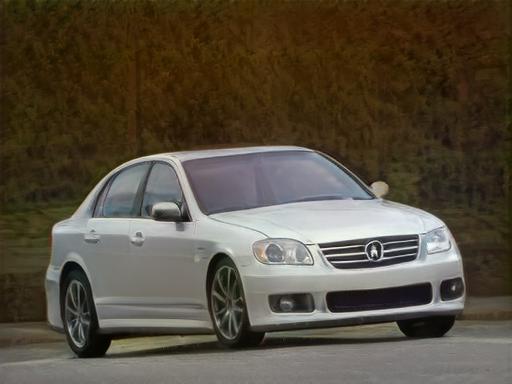} &
        \includegraphics[width=0.1125\textwidth]{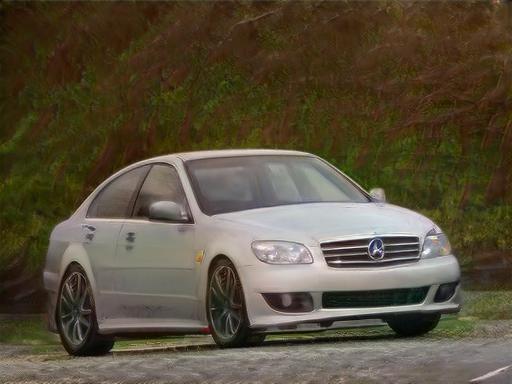} &
        \includegraphics[width=0.1125\textwidth]{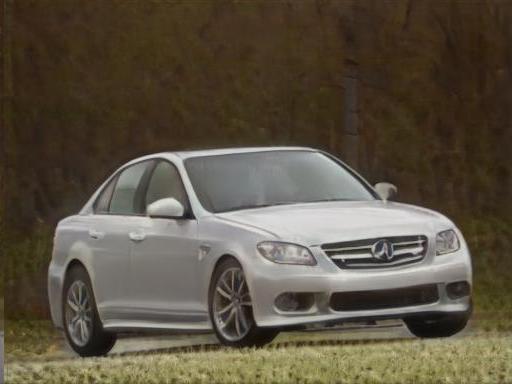} &
        \includegraphics[width=0.1125\textwidth]{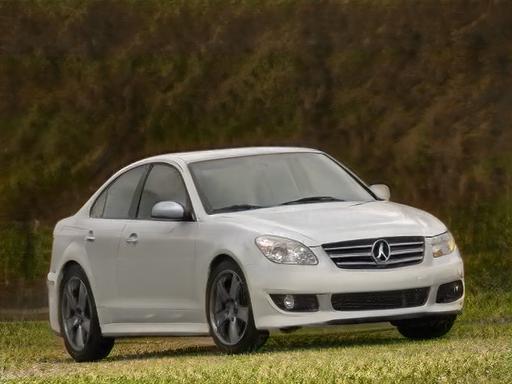} &
        \includegraphics[width=0.1125\textwidth]{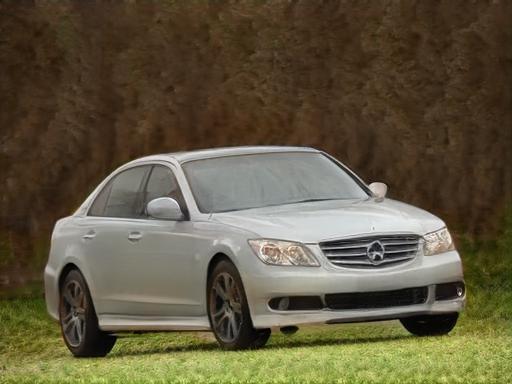} \\
        
        \raisebox{0.1in}{\rotatebox{90}{$+$Grass}} &
        \includegraphics[width=0.1125\textwidth]{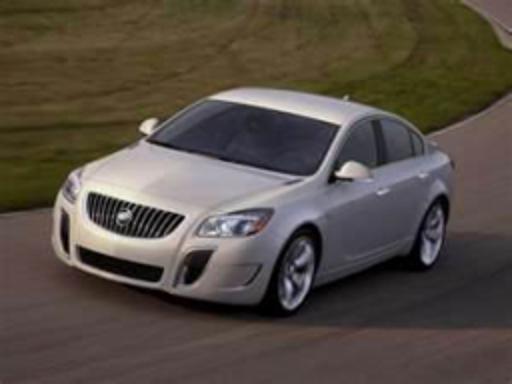} &
        \includegraphics[width=0.1125\textwidth]{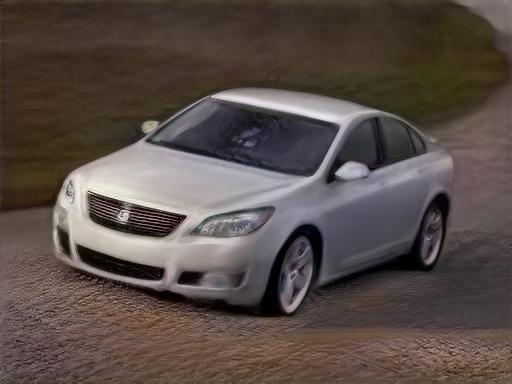} &
        \includegraphics[width=0.1125\textwidth]{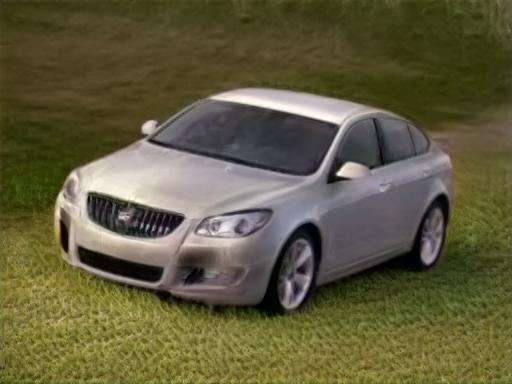} &
        \includegraphics[width=0.1125\textwidth]{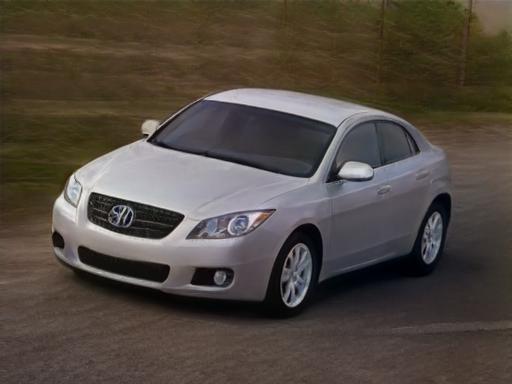} &
        \includegraphics[width=0.1125\textwidth]{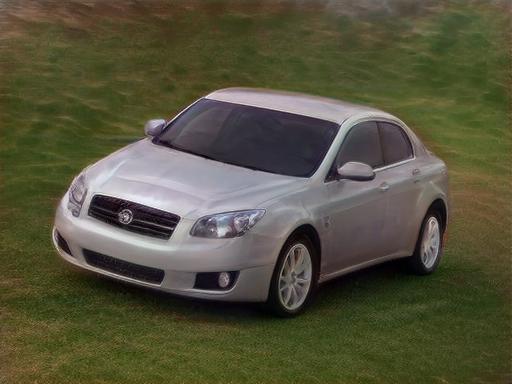} &
        \includegraphics[width=0.1125\textwidth]{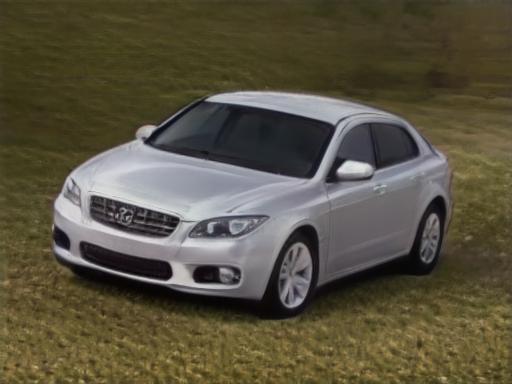} &
        \includegraphics[width=0.1125\textwidth]{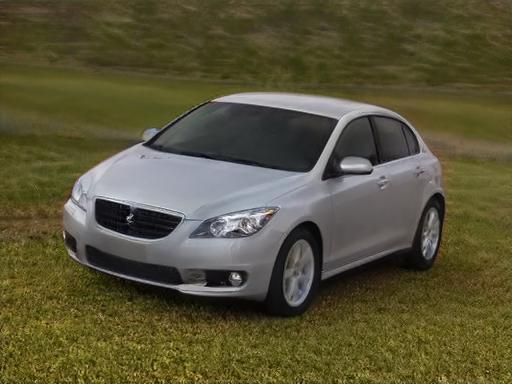} &
        \includegraphics[width=0.1125\textwidth]{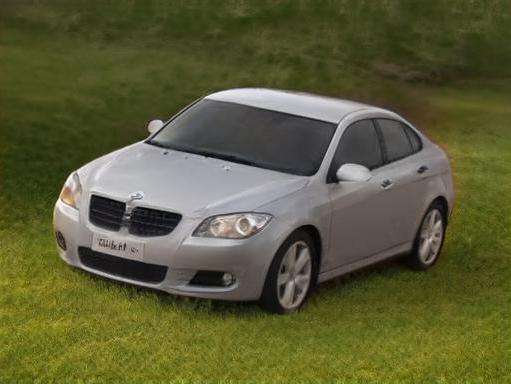} \\
\end{tabular}
    }
    \caption{Additional editing comparisons over the cars domain obtained using GANSpace~\cite{harkonen2020ganspace}.}
    \label{fig:supp_car_editing_comparison}
\end{figure*}
\begin{figure*}
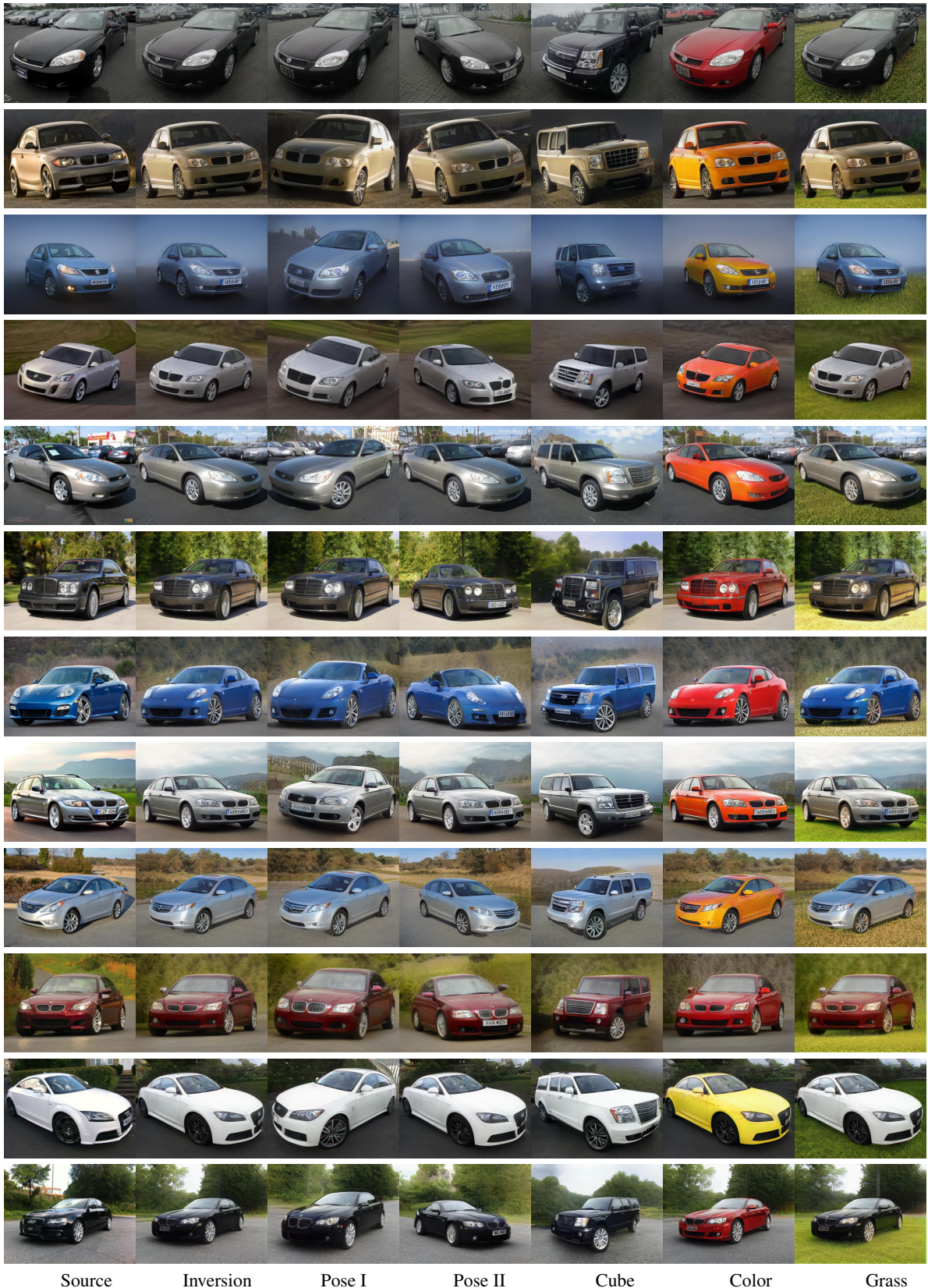

\setlength{\tabcolsep}{1pt}
    \centering
    { \small
    %
    }
    \caption{Additional editing results obtained with HyperStyle over the cars domain obtained using GANSpace~\cite{harkonen2020ganspace}.}
    \label{fig:supplementary_cars_edits}
\end{figure*}
\begin{figure*}
    \centering
    \setlength{\belowcaptionskip}{-6pt}
    \setlength{\tabcolsep}{1pt}
    {\small
    \begin{tabular}{c c c c c c}

        \includegraphics[width=0.14\textwidth]{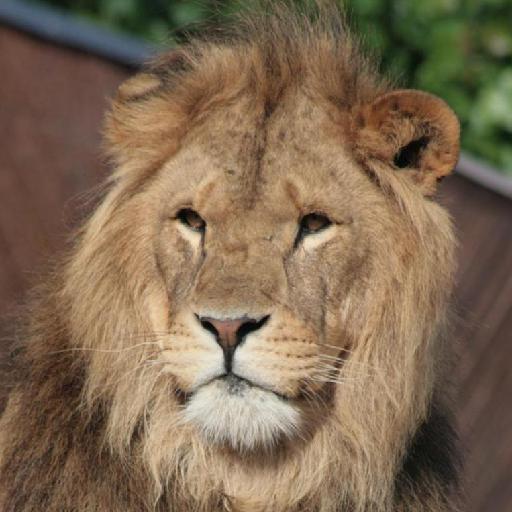} &
        \includegraphics[width=0.14\textwidth]{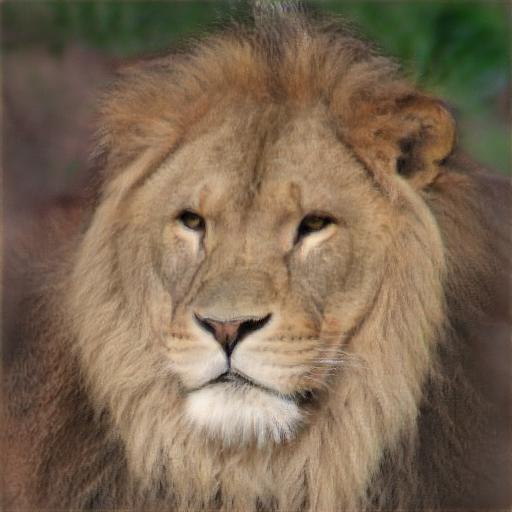} &
        \includegraphics[width=0.14\textwidth]{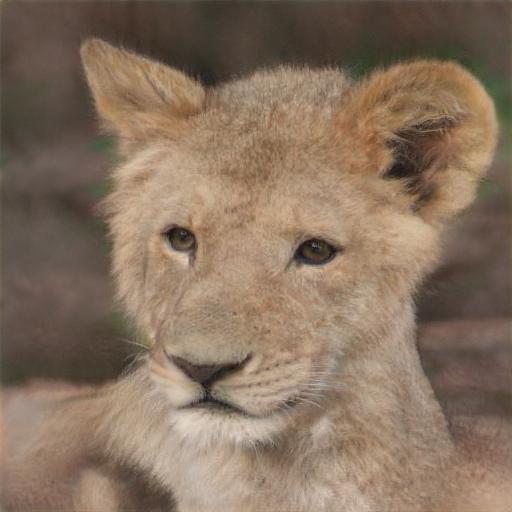} &
        \hspace{0.5cm}
        \includegraphics[width=0.14\textwidth]{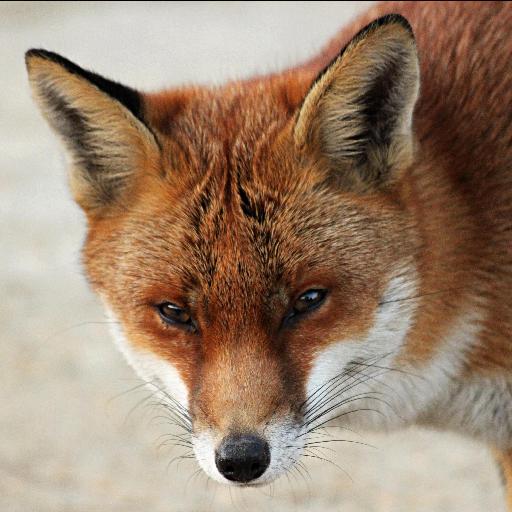} &
        \includegraphics[width=0.14\textwidth]{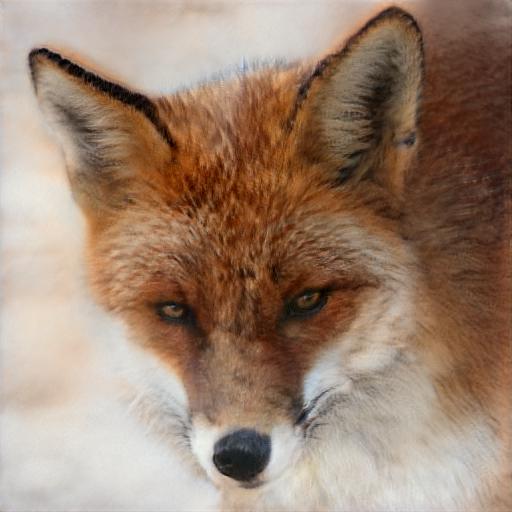} &
        \includegraphics[width=0.14\textwidth]{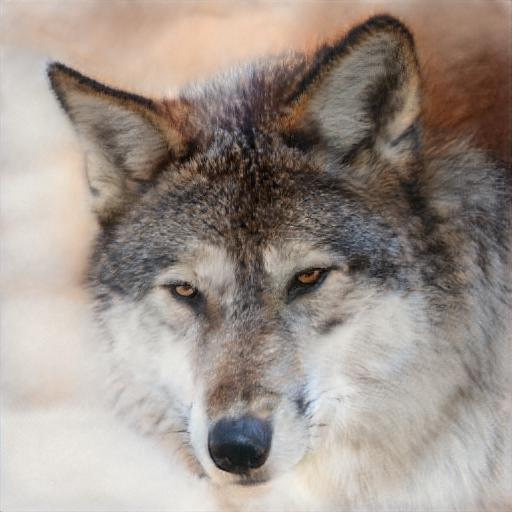} \\
        
        \includegraphics[width=0.14\textwidth]{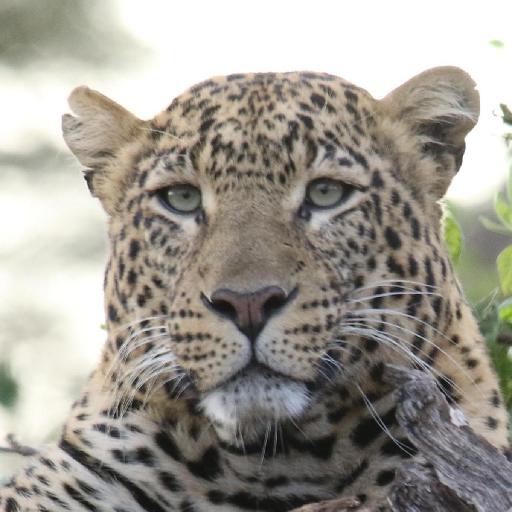} &
        \includegraphics[width=0.14\textwidth]{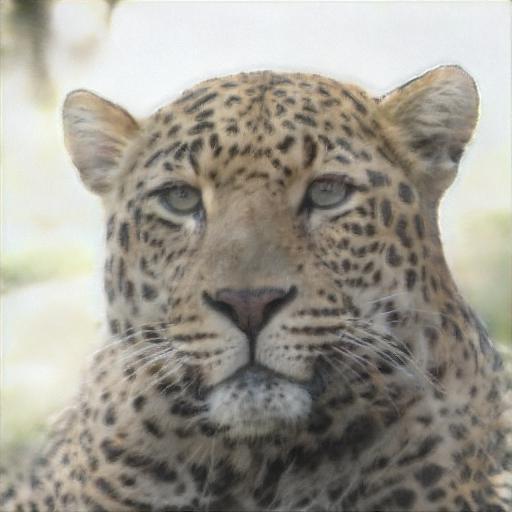} &
        \includegraphics[width=0.14\textwidth]{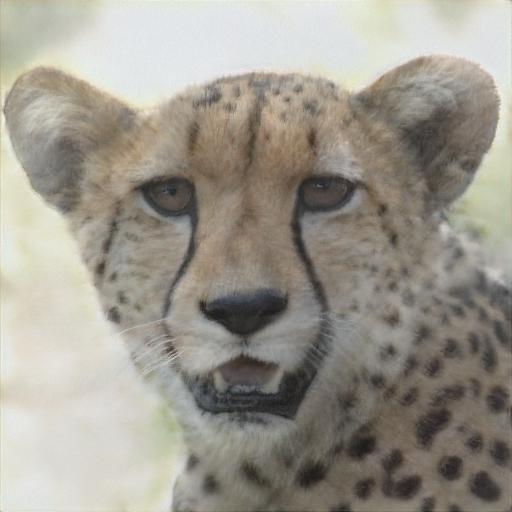} &
        \hspace{0.5cm}
        \includegraphics[width=0.14\textwidth]{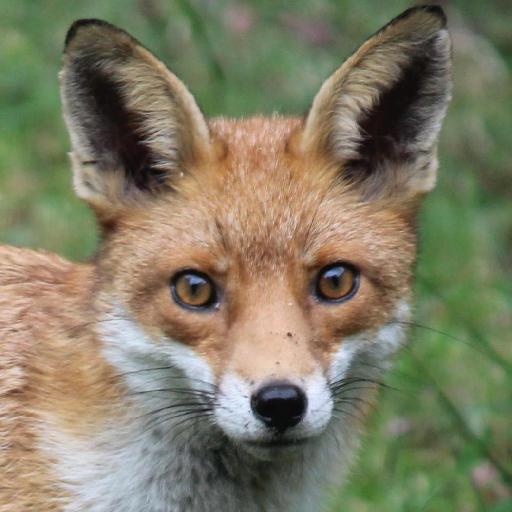} &
        \includegraphics[width=0.14\textwidth]{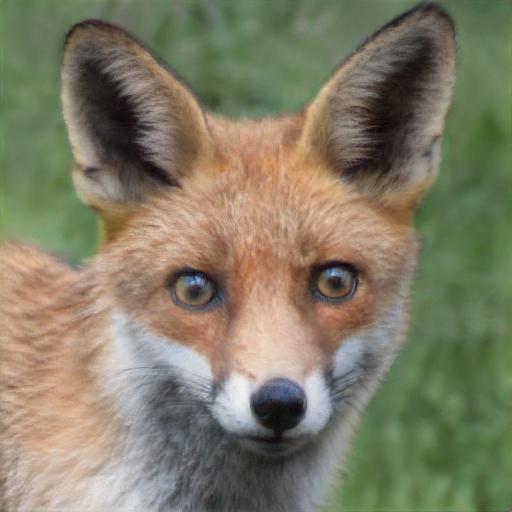} &
        \includegraphics[width=0.14\textwidth]{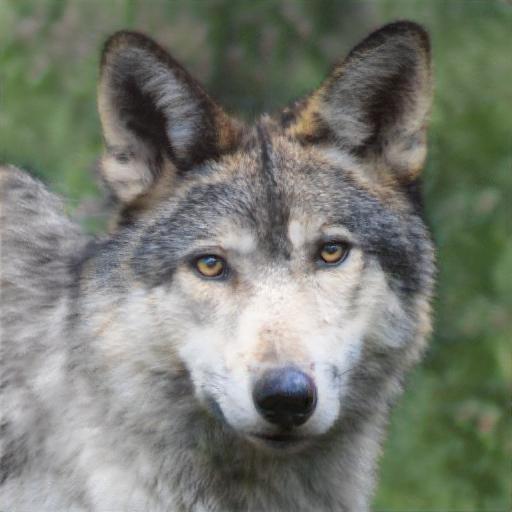} \\
\end{tabular}
    }
    \caption{Reconstruction and editing results obtained with HyperStyle on the AFHQ Wild~\cite{choi2020stargan} test set. Edits were obtained using StyleCLIP's~\cite{patashnik2021styleclip} global directions approach.}
    \label{fig:afhq_wild_edits}
\end{figure*}
\begin{figure*}

    \vspace{3cm}

    \setlength{\tabcolsep}{1pt}
    \centering%
    {\small%
    \begin{tabular}{c | c c | c c c}

    \vspace{-0.08cm}

    \includegraphics[width=0.14\linewidth]{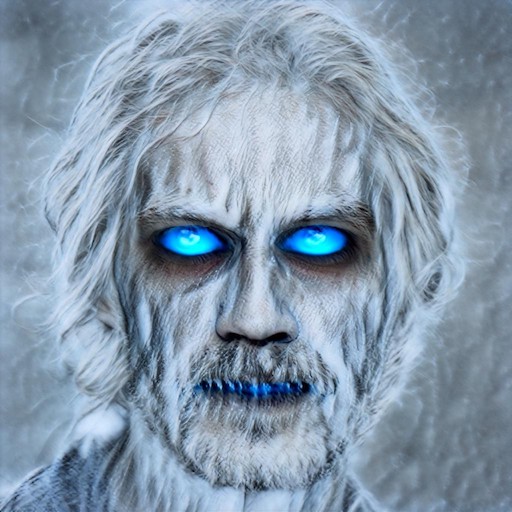} &
    \includegraphics[width=0.14\linewidth]{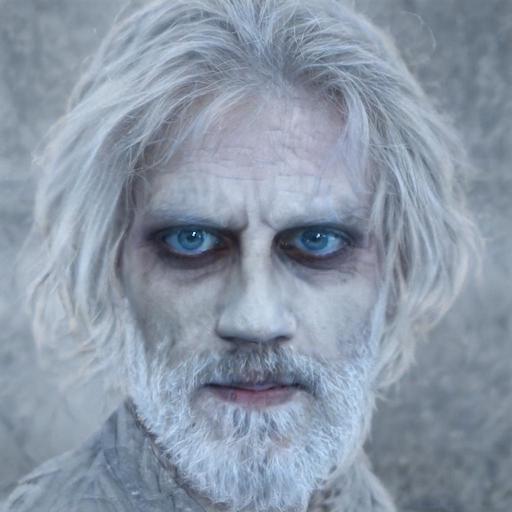} &
    \includegraphics[width=0.14\linewidth]{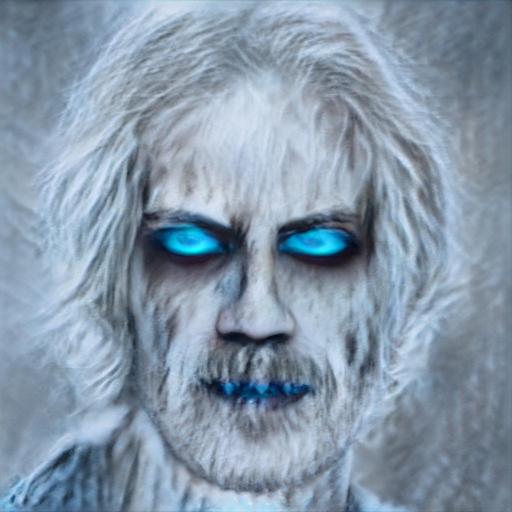} &
    \includegraphics[width=0.14\linewidth]{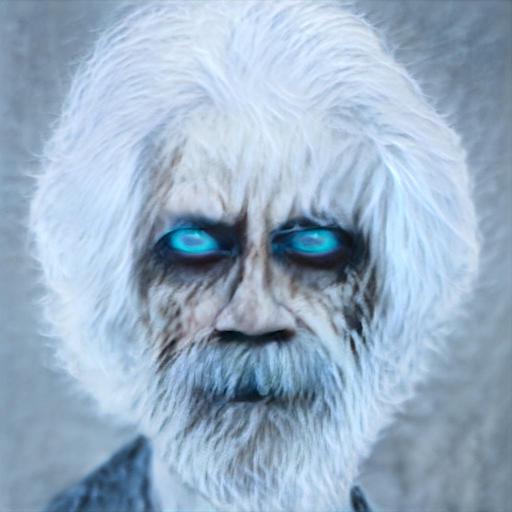} &
    \includegraphics[width=0.14\linewidth]{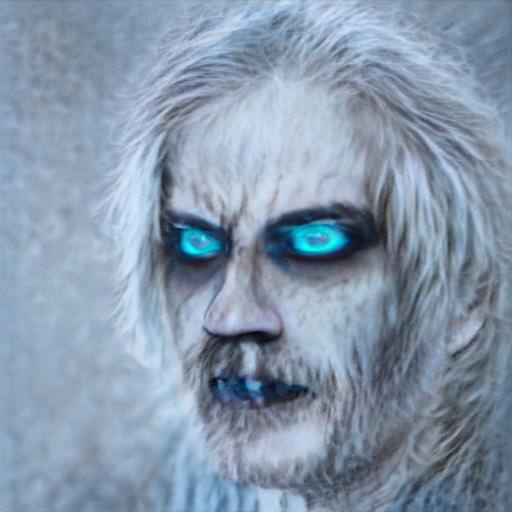} &
    \includegraphics[width=0.14\linewidth]{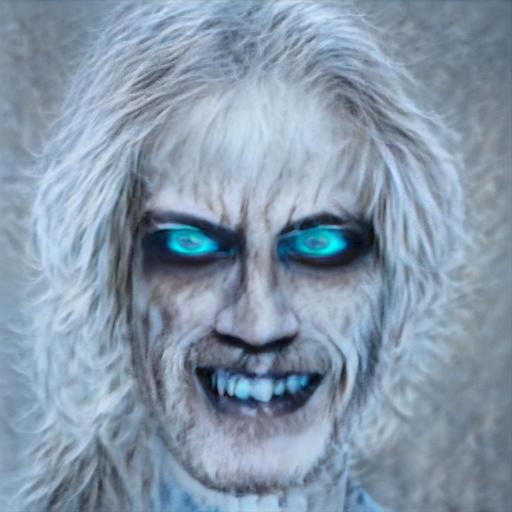} \\
    \vspace{-0.08cm}
    \includegraphics[width=0.14\linewidth]{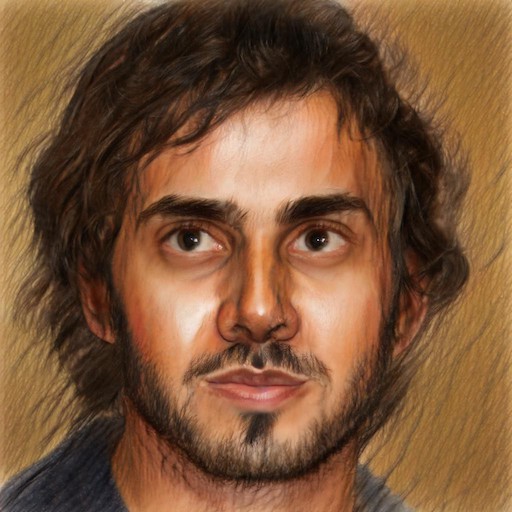} &
    \includegraphics[width=0.14\linewidth]{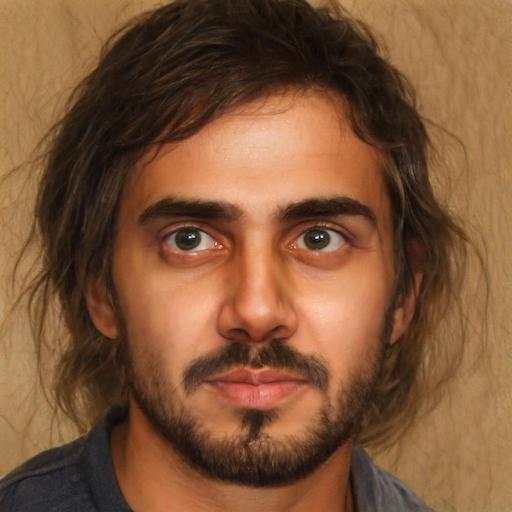} &
    \includegraphics[width=0.14\linewidth]{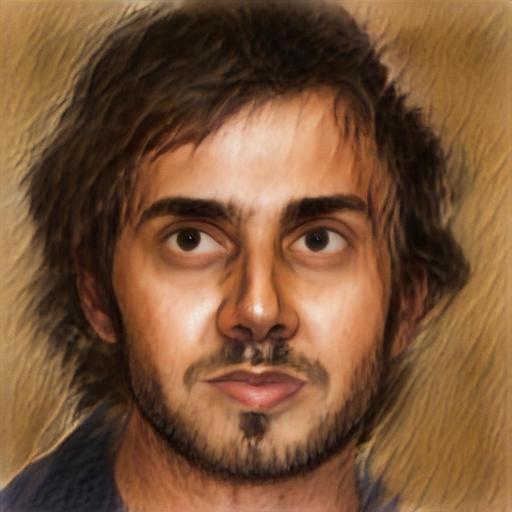} &
    \includegraphics[width=0.14\linewidth]{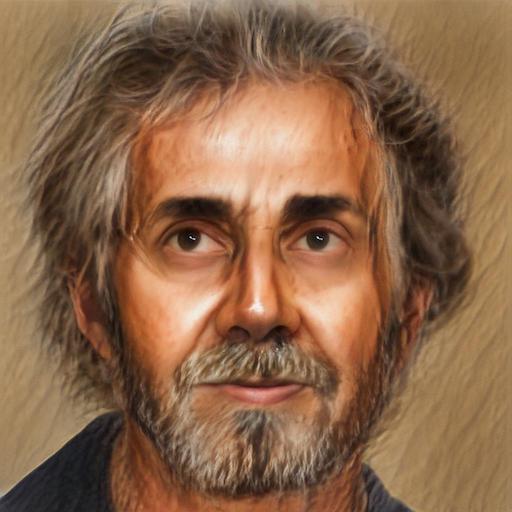} &
    \includegraphics[width=0.14\linewidth]{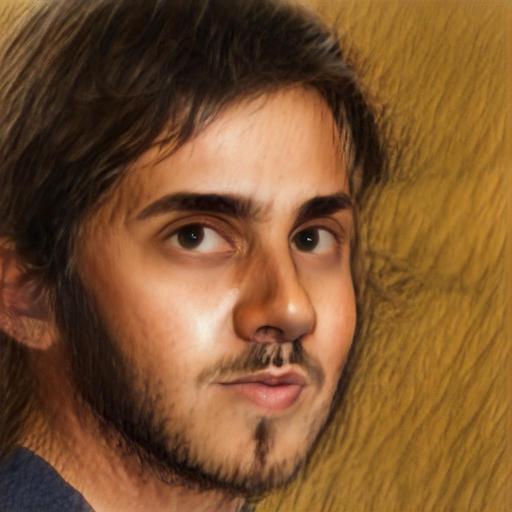} &
    \includegraphics[width=0.14\linewidth]{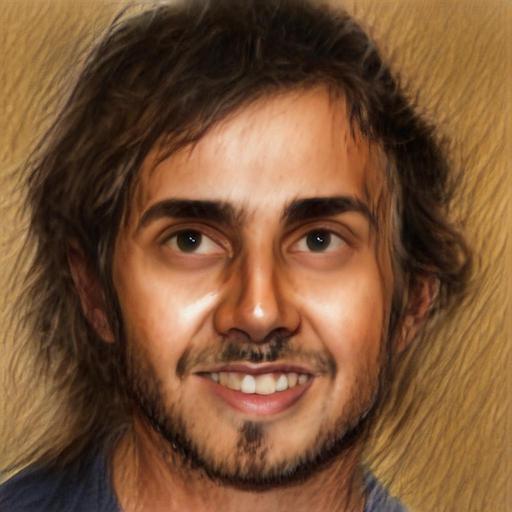} \\
    
    \includegraphics[width=0.14\linewidth]{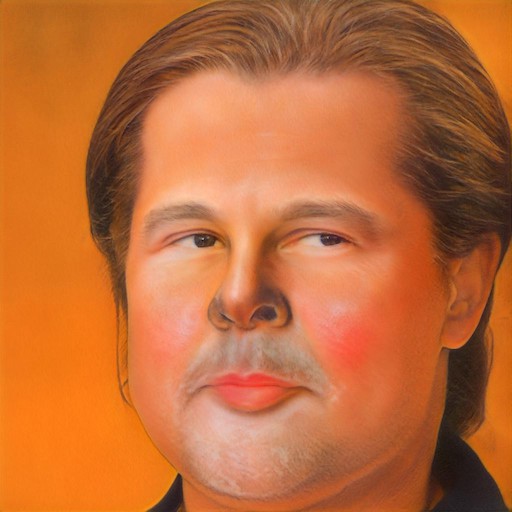} &
    \includegraphics[width=0.14\linewidth]{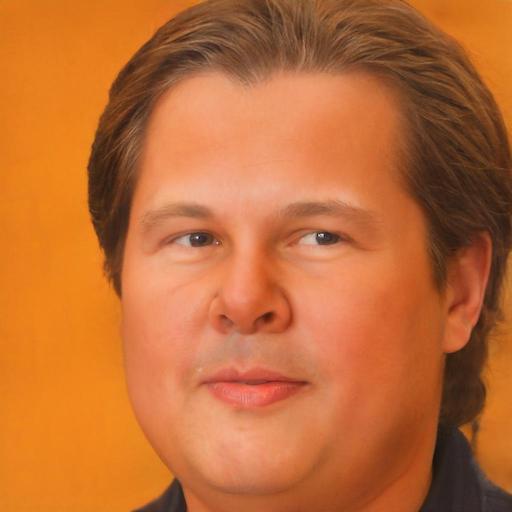} &
    \includegraphics[width=0.14\linewidth]{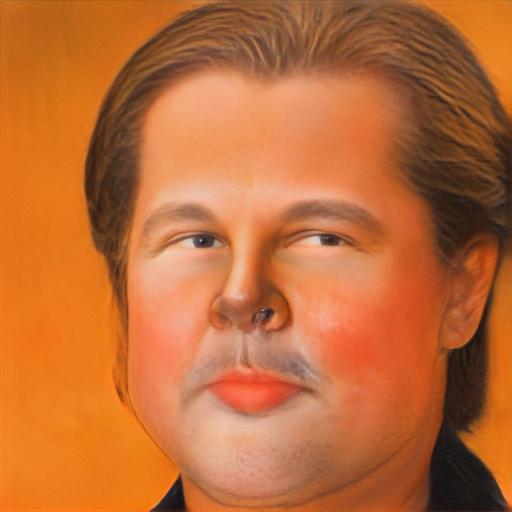} &
    \includegraphics[width=0.14\linewidth]{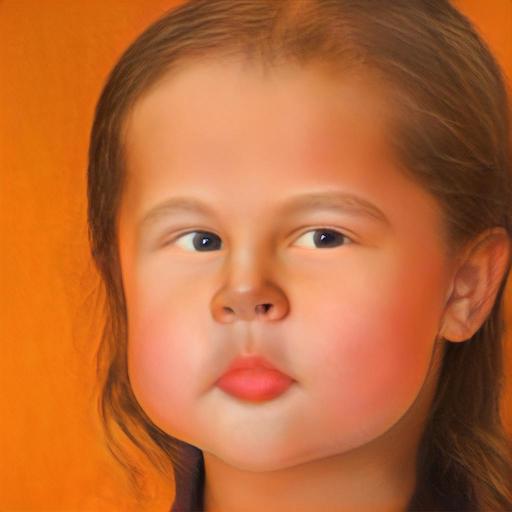} &
    \includegraphics[width=0.14\linewidth]{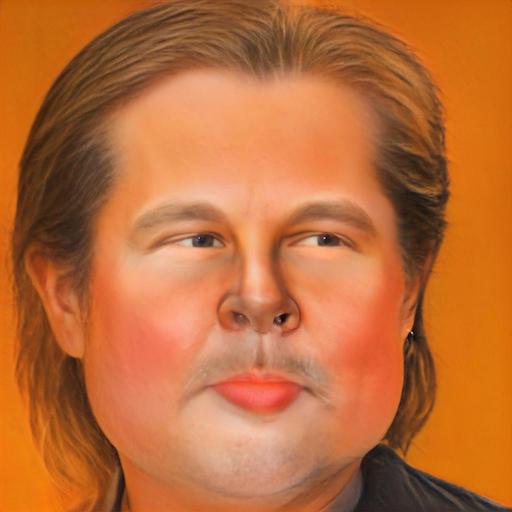} &
    \includegraphics[width=0.14\linewidth]{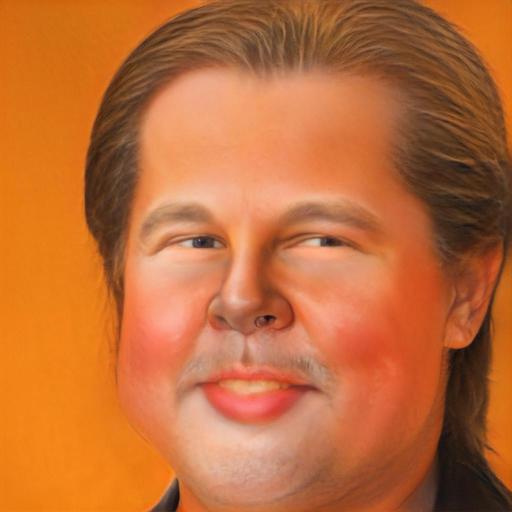} \\

    Input & e4e & HyperStyle & \multicolumn{3}{c}{ ------------ HyperStyle Edits ------------ }
    
    \end{tabular}
    }
    
    \vspace{0.5cm}
    
    {\small
    \begin{tabular}{c | c c | c c c}

    \vspace{-0.08cm}
    
    \includegraphics[width=0.14\linewidth]{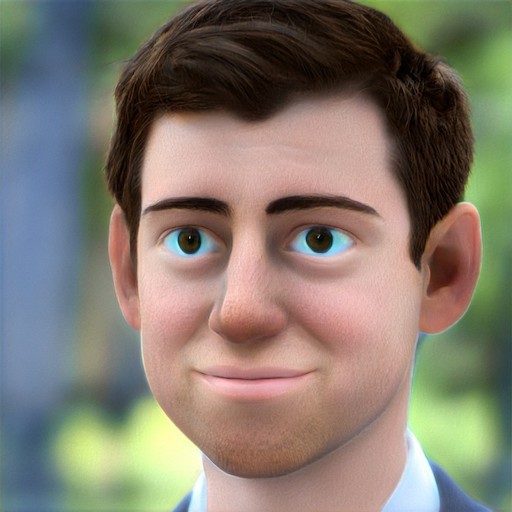} &
    \includegraphics[width=0.14\linewidth]{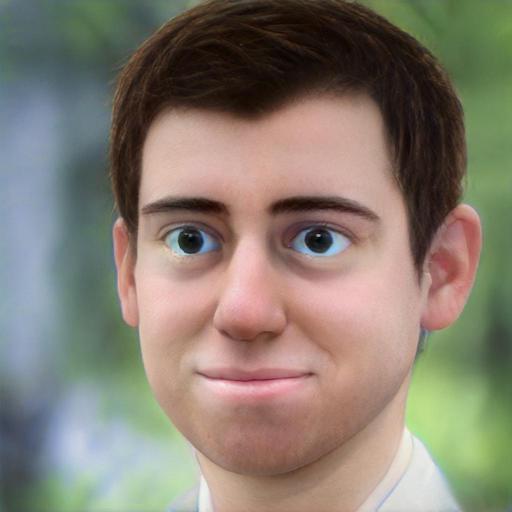} &
    \includegraphics[width=0.14\linewidth]{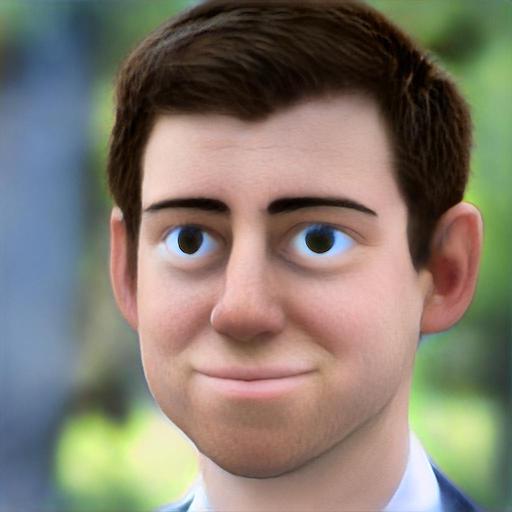} &
    \includegraphics[width=0.14\linewidth]{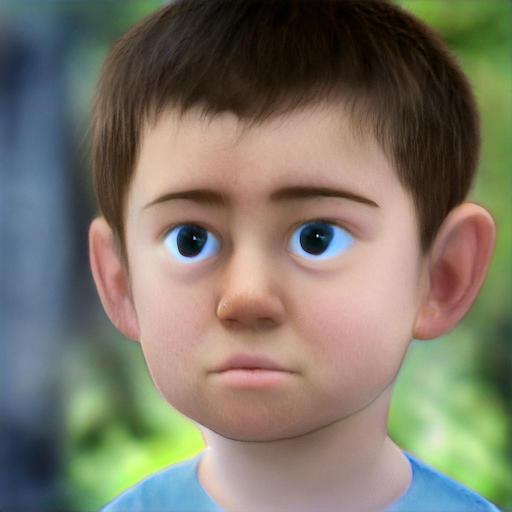} &
    \includegraphics[width=0.14\linewidth]{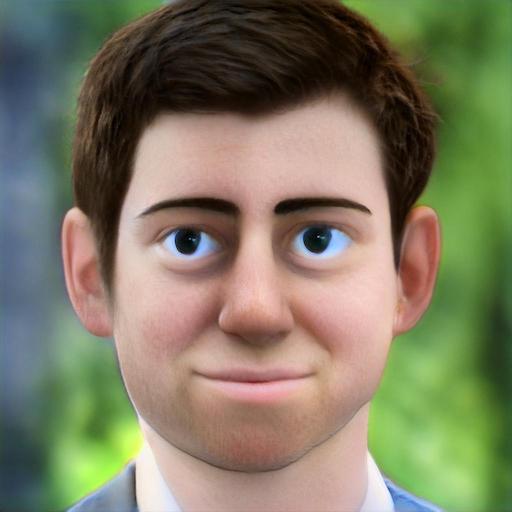} &
    \includegraphics[width=0.14\linewidth]{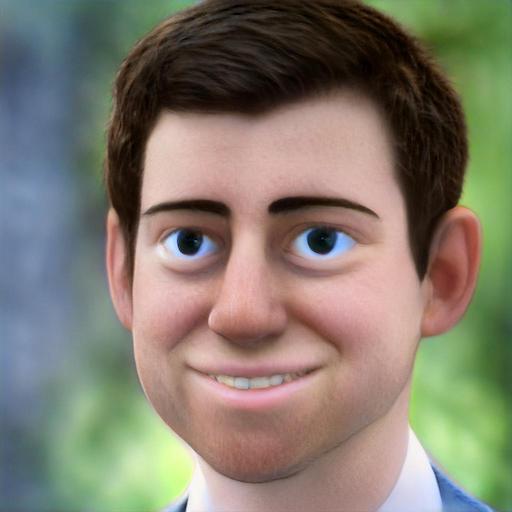} \\
    \vspace{-0.08cm}
    \includegraphics[width=0.14\linewidth]{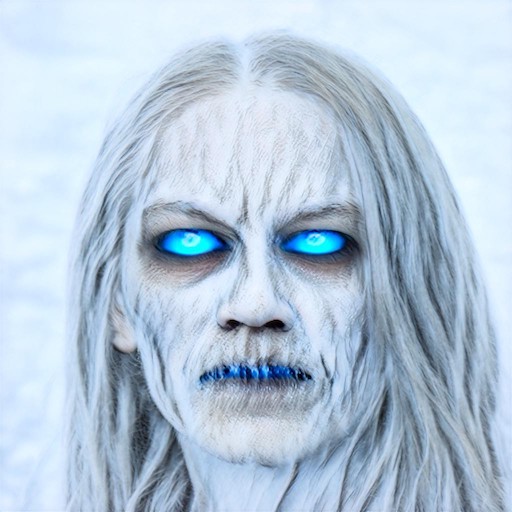} &
    \includegraphics[width=0.14\linewidth]{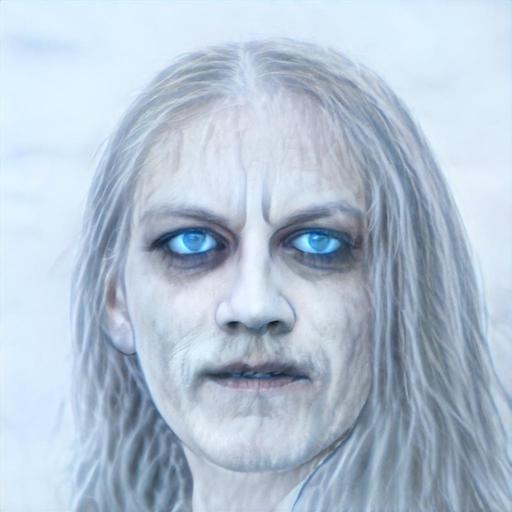} &
    \includegraphics[width=0.14\linewidth]{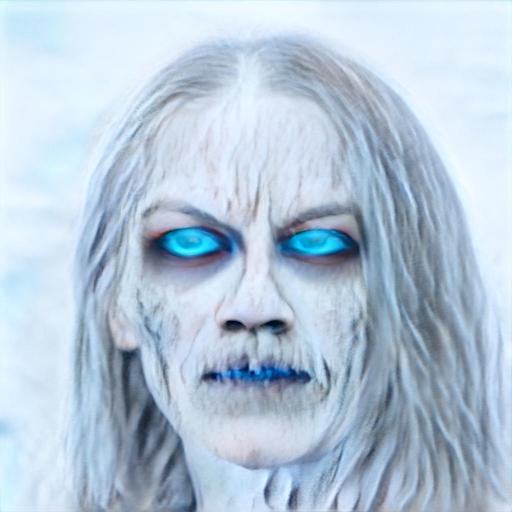} &
    \includegraphics[width=0.14\linewidth]{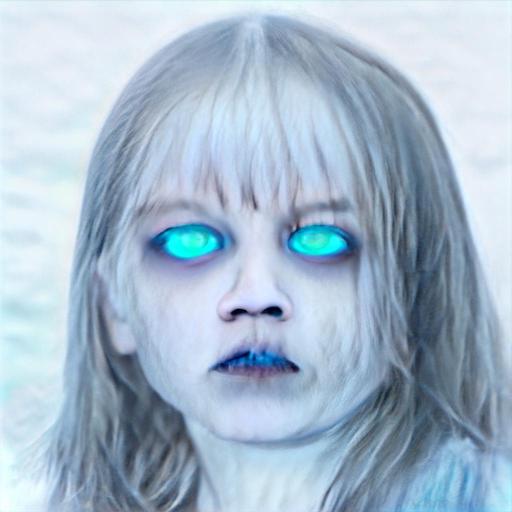} &
    \includegraphics[width=0.14\linewidth]{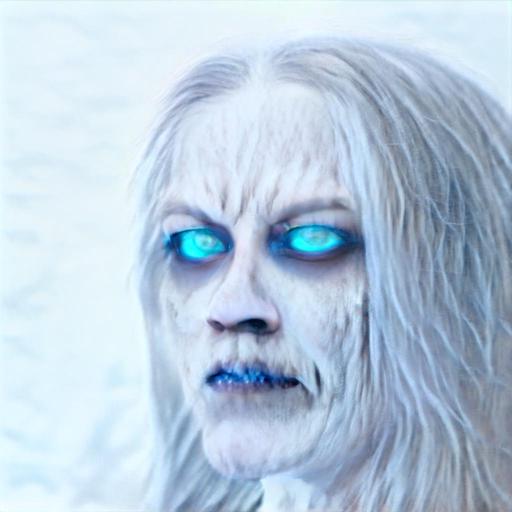} &
    \includegraphics[width=0.14\linewidth]{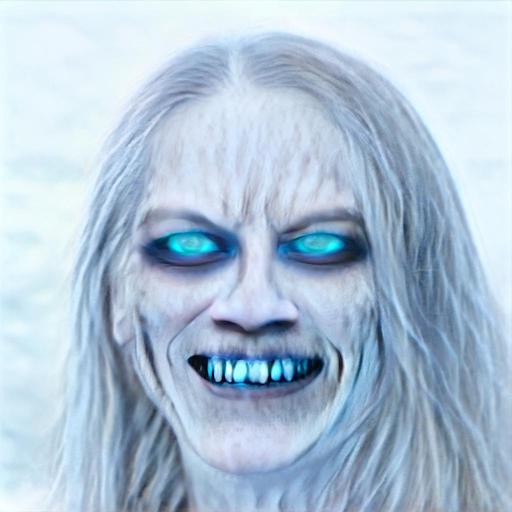} \\
    
    \includegraphics[width=0.14\linewidth]{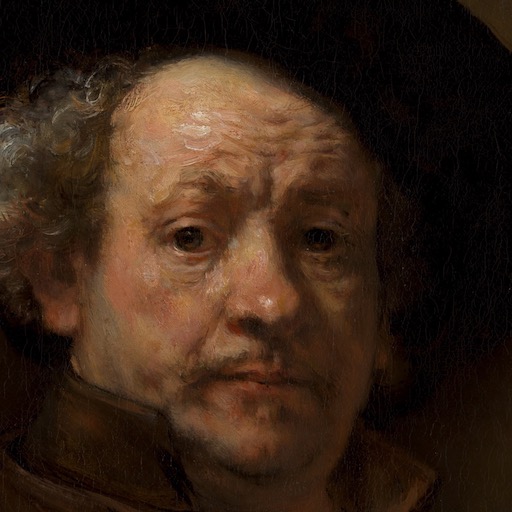} &
    \includegraphics[width=0.14\linewidth]{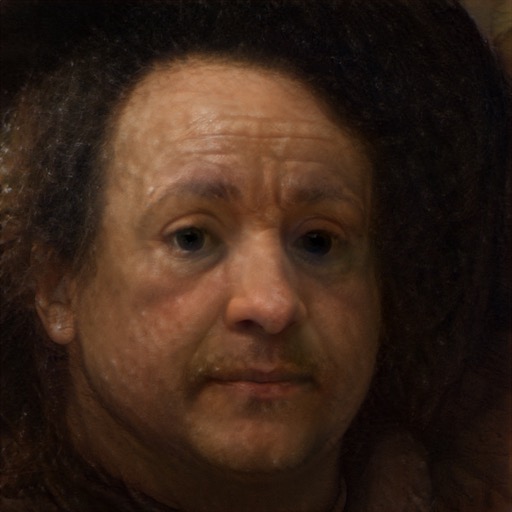} &
    \includegraphics[width=0.14\linewidth]{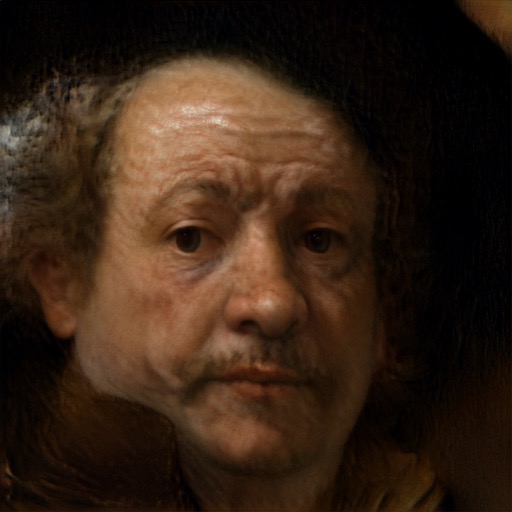} &
    \includegraphics[width=0.14\linewidth]{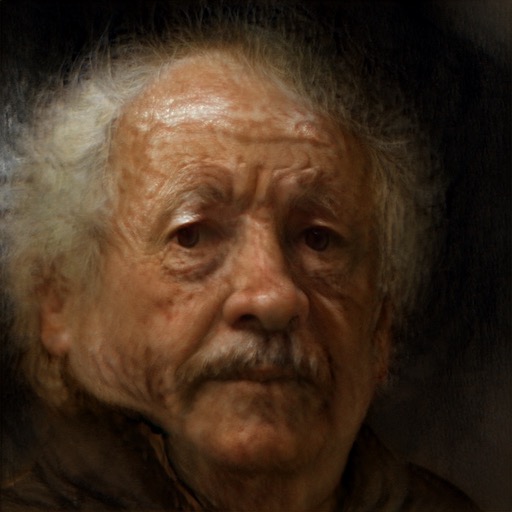} &
    \includegraphics[width=0.14\linewidth]{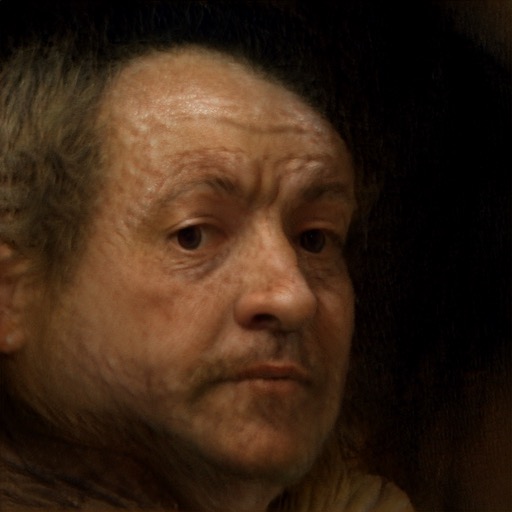} &
    \includegraphics[width=0.14\linewidth]{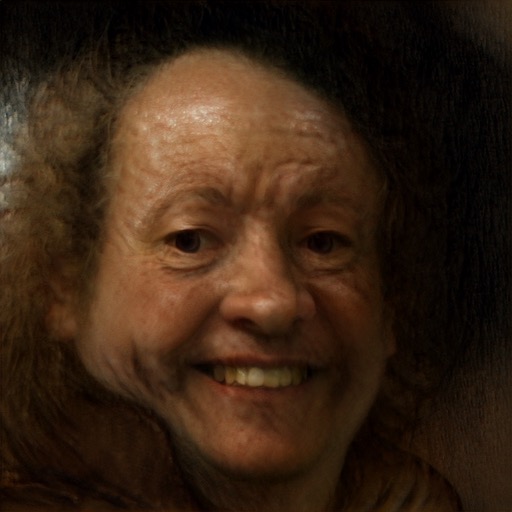} \\

    Input & pSp & HyperStyle & \multicolumn{3}{c}{ ------------ HyperStyle Edits ------------ }

    \end{tabular}
    }
    \caption{
    Reconstruction and editing results obtained on challenging input styles not observed during training. 
    All reconstructions and editing results are obtained with a hypernetwork and StyleGAN generator trained on the FFHQ~\cite{karras2019style} dataset. 
    Note, some of the input images were generated from a StyleGAN model fine-tuned with StyleGAN-NADA~\cite{gal2021stylegannada}. Even in such cases, prior encoders struggle in accurately reconstructing the input.  
    } \vspace{-0.5cm}
    \label{fig:supp_out_of_domain_e4e_psp}

\end{figure*}
\begin{figure*}
    
    \vspace{3cm}

    \setlength{\tabcolsep}{1pt}
    \centering
    {\small

    \begin{tabular}{c | c c | c c c}

    \vspace{-0.08cm}
    
    \includegraphics[width=0.14\linewidth]{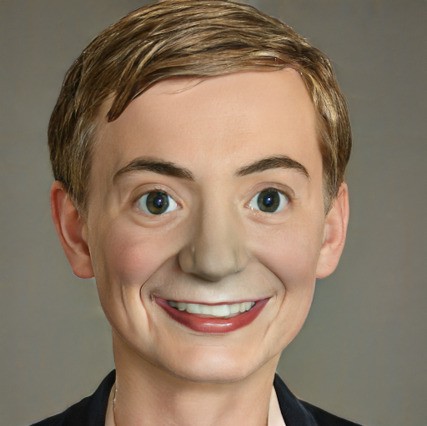} &
    \includegraphics[width=0.14\linewidth]{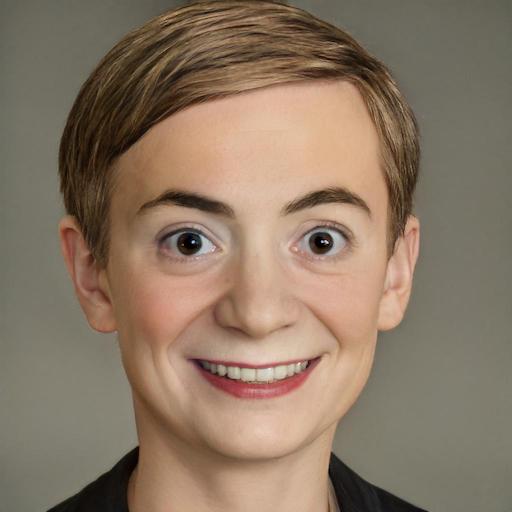} &
    \includegraphics[width=0.14\linewidth]{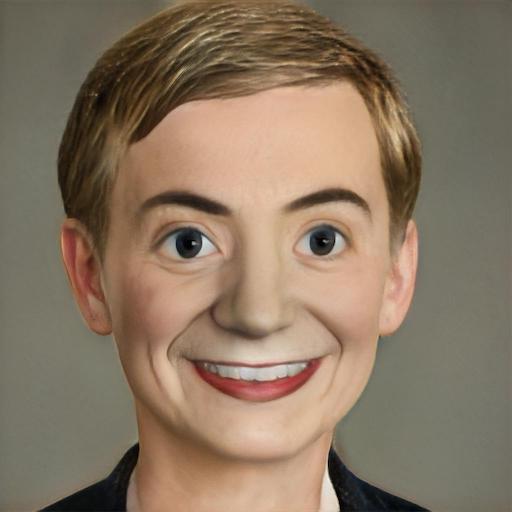} &
    \includegraphics[width=0.14\linewidth]{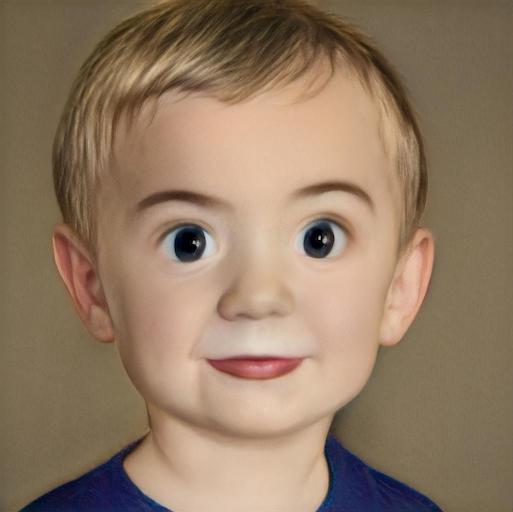} &
    \includegraphics[width=0.14\linewidth]{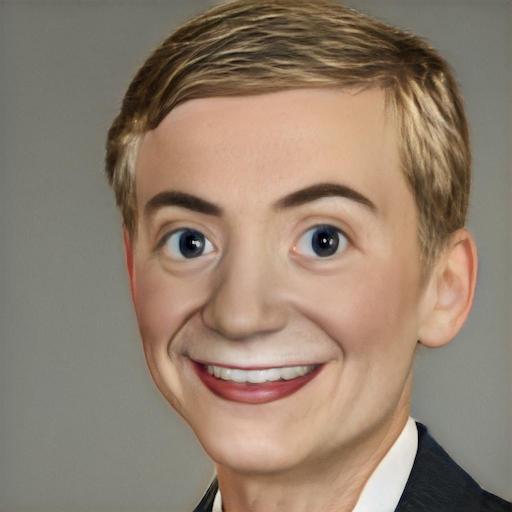} &
    \includegraphics[width=0.14\linewidth]{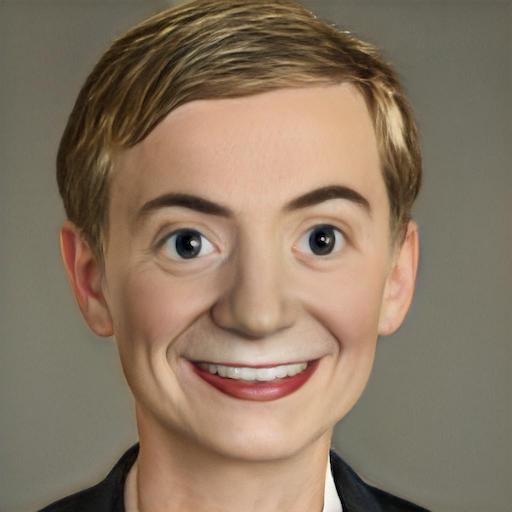} \\
    \vspace{-0.08cm}
    \includegraphics[width=0.14\linewidth]{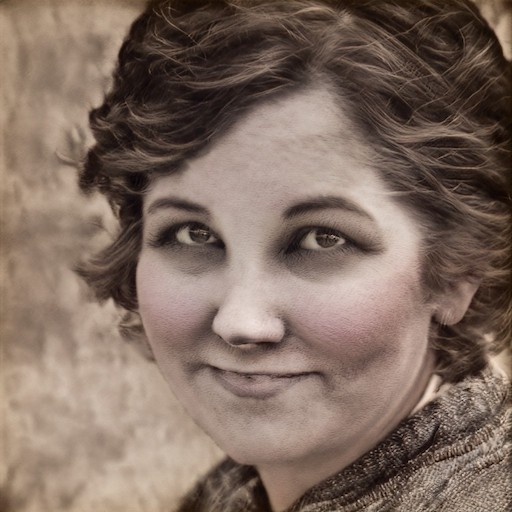} &
    \includegraphics[width=0.14\linewidth]{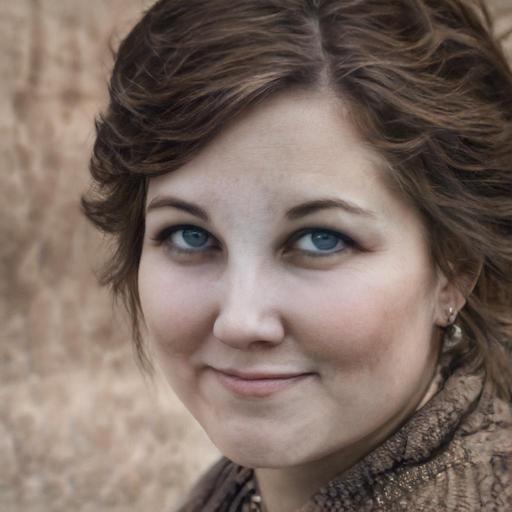} &
    \includegraphics[width=0.14\linewidth]{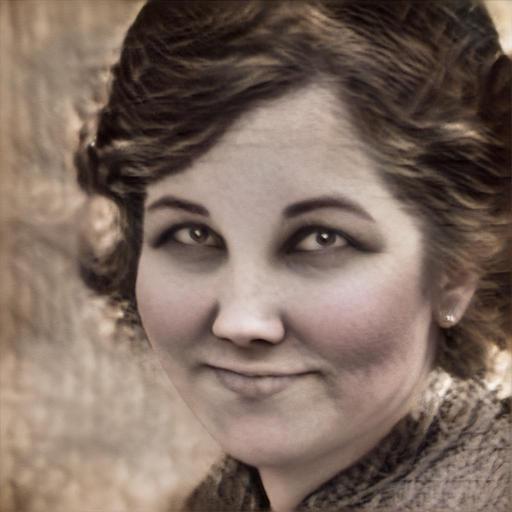} &
    \includegraphics[width=0.14\linewidth]{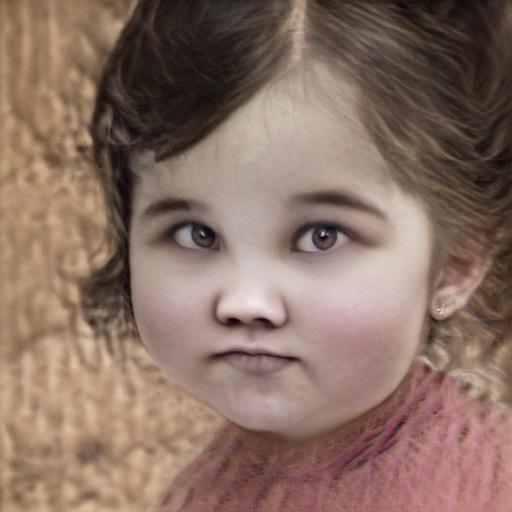} &
    \includegraphics[width=0.14\linewidth]{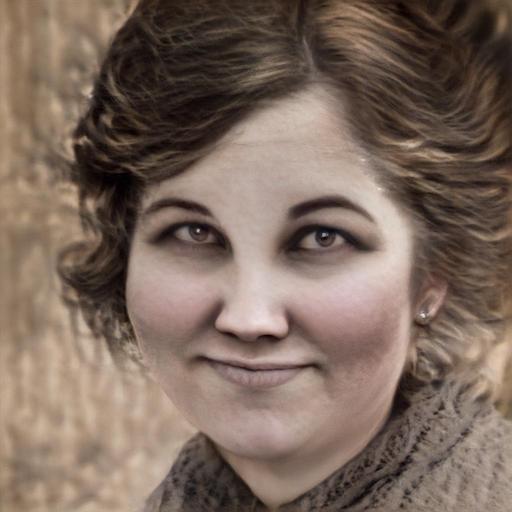} &
    \includegraphics[width=0.14\linewidth]{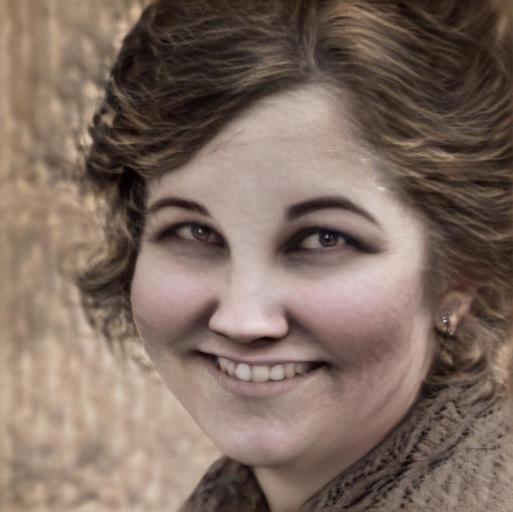} \\
    
    \includegraphics[width=0.14\linewidth]{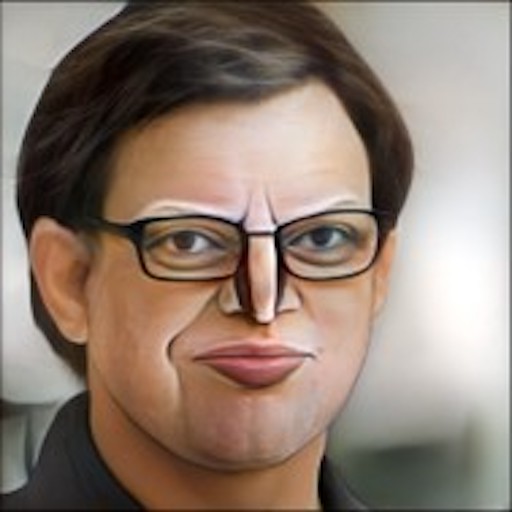} &
    \includegraphics[width=0.14\linewidth]{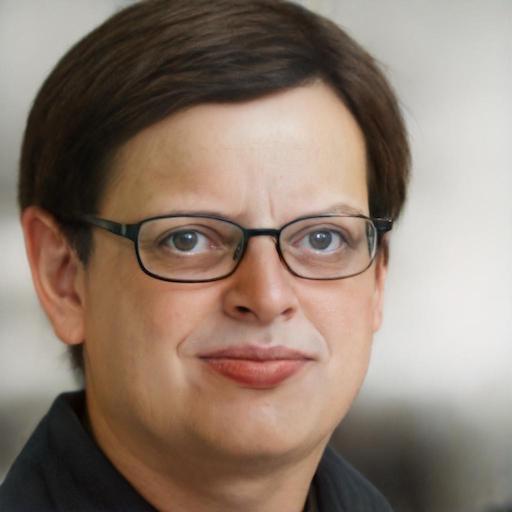} &
    \includegraphics[width=0.14\linewidth]{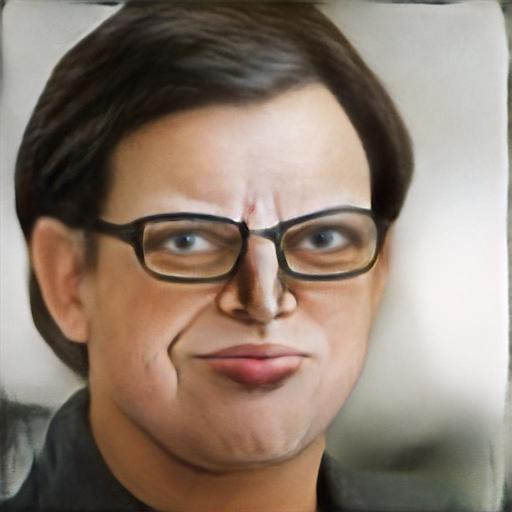} &
    \includegraphics[width=0.14\linewidth]{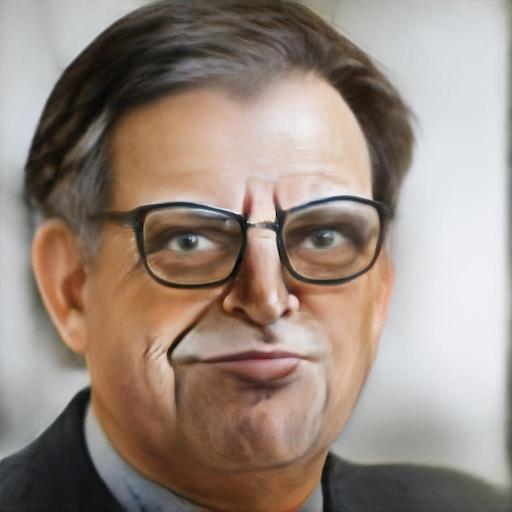} &
    \includegraphics[width=0.14\linewidth]{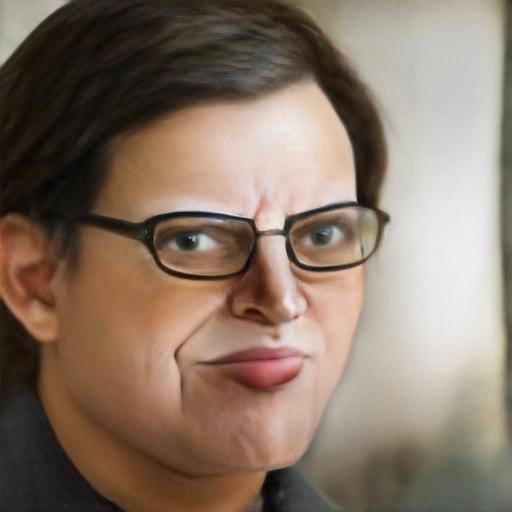} &
    \includegraphics[width=0.14\linewidth]{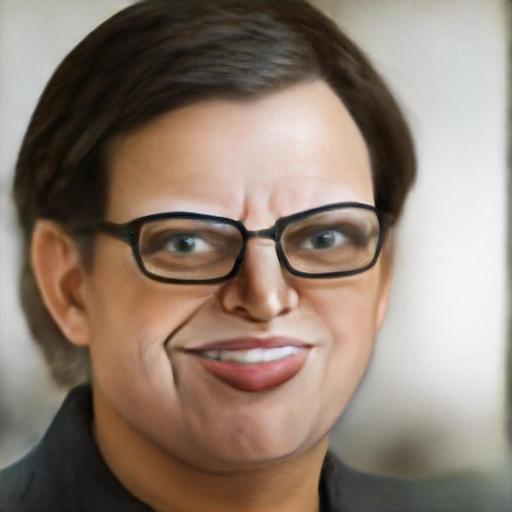} \\

    Input & $\text{ReStyle}_{e4e}$ & HyperStyle & \multicolumn{3}{c}{ ------------ HyperStyle Edits ------------ }
    
    \end{tabular}
    }
    
    {\small
    \vspace{0.5cm}

    \begin{tabular}{c | c c | c c c}

    \vspace{-0.08cm}
    
    \includegraphics[width=0.14\linewidth]{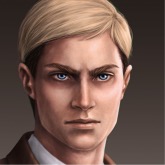} &
    \includegraphics[width=0.14\linewidth]{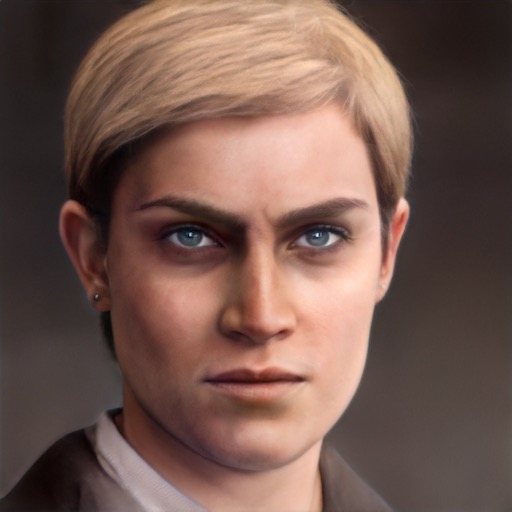} &
    \includegraphics[width=0.14\linewidth]{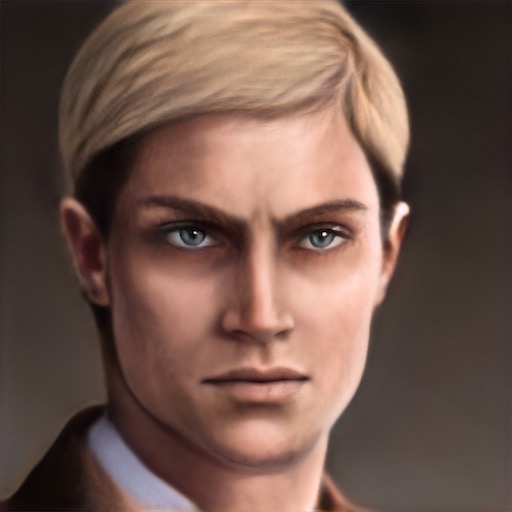} &
    \includegraphics[width=0.14\linewidth]{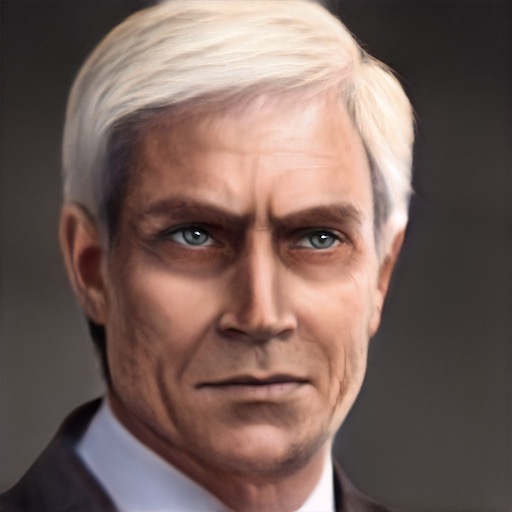} &
    \includegraphics[width=0.14\linewidth]{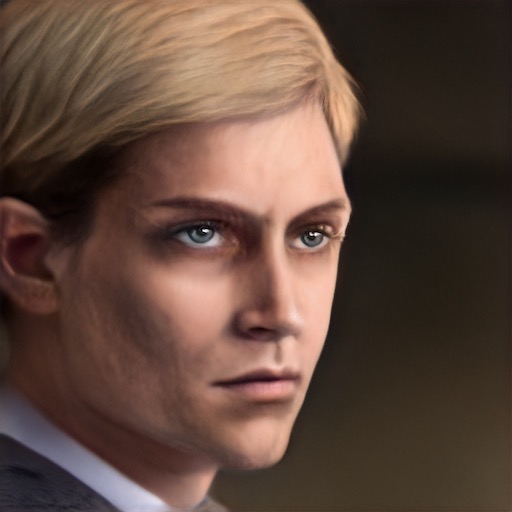} &
    \includegraphics[width=0.14\linewidth]{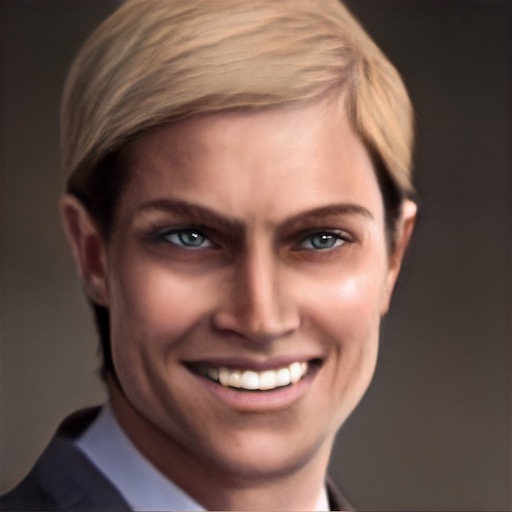} \\
    \vspace{-0.08cm}
    \includegraphics[width=0.14\linewidth]{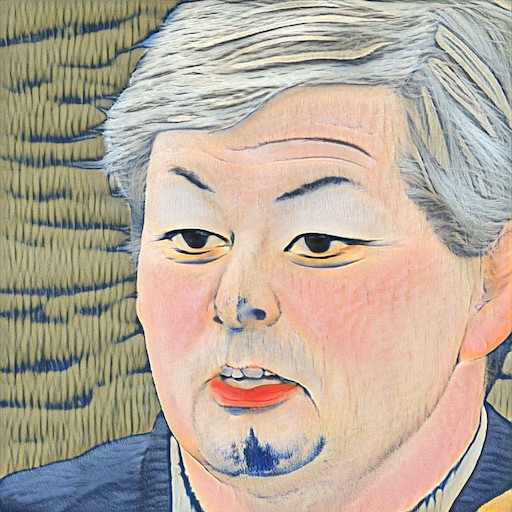} &
    \includegraphics[width=0.14\linewidth]{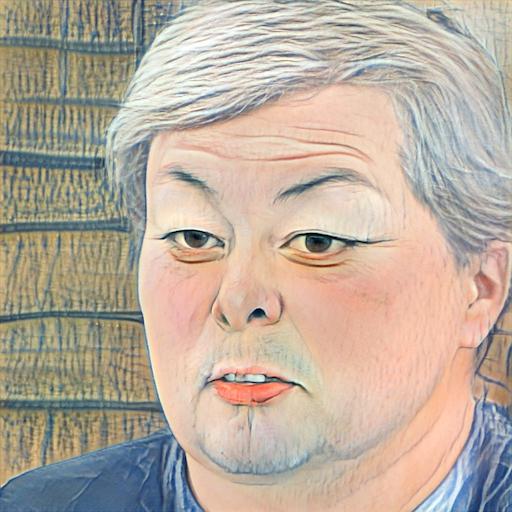} &
    \includegraphics[width=0.14\linewidth]{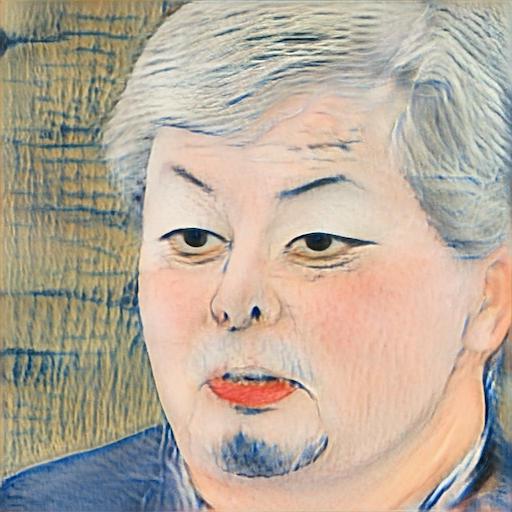} &
    \includegraphics[width=0.14\linewidth]{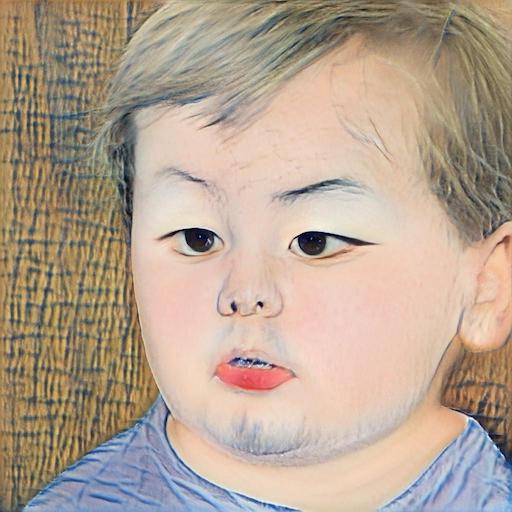} &
    \includegraphics[width=0.14\linewidth]{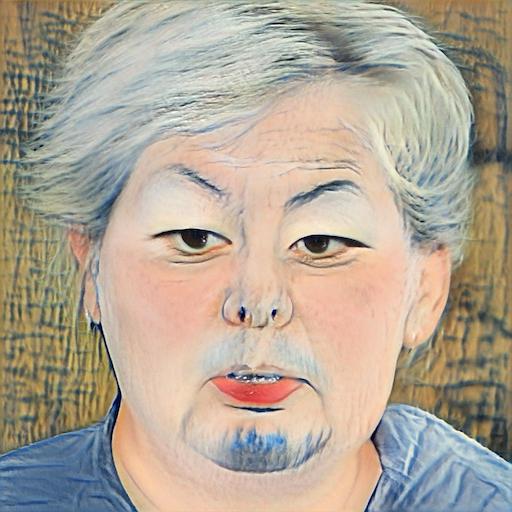} &
    \includegraphics[width=0.14\linewidth]{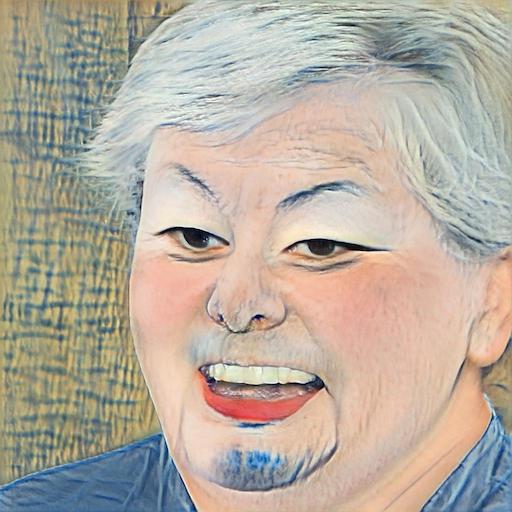} \\
    
    \includegraphics[width=0.14\linewidth]{images/supp/out_of_domain/1_ww_input.jpg} &
    \includegraphics[width=0.14\linewidth]{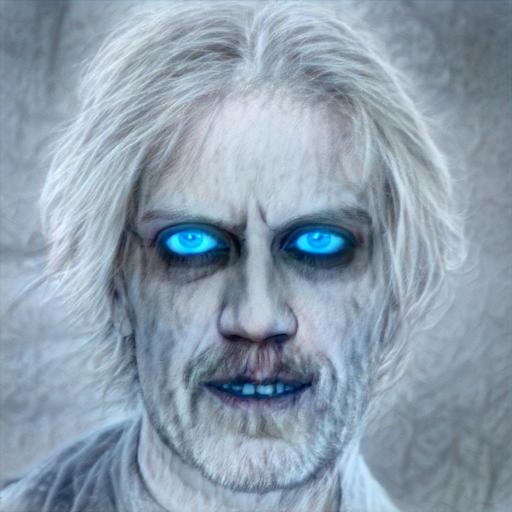} &
    \includegraphics[width=0.14\linewidth]{images/supp/out_of_domain/1_ww_hyperstyle.jpg} &
    \includegraphics[width=0.14\linewidth]{images/supp/out_of_domain/1_ww_age.jpg} &
    \includegraphics[width=0.14\linewidth]{images/supp/out_of_domain/1_ww_pose.jpg} &
    \includegraphics[width=0.14\linewidth]{images/supp/out_of_domain/1_ww_smile.jpg} \\

    Input & $\text{ReStyle}_{pSp}$ & HyperStyle & \multicolumn{3}{c}{ ------------ HyperStyle Edits ------------ }
    
    \end{tabular}
    }
    \caption{
    Reconstruction and editing results obtained on challenging input styles unobserved during training, using the same settings as in \cref{fig:supp_out_of_domain_e4e_psp}. 
    } \vspace{-0.5cm}
    \label{fig:supp_out_of_domain_restyle}

\end{figure*}
\begin{figure*}
\setlength{\tabcolsep}{1pt}
    \centering
    { \small 
    \begin{tabular}{c c c c c c}

    \raisebox{0.20in}{\rotatebox{90}{Toonify}} &
    \includegraphics[width=0.125\linewidth]{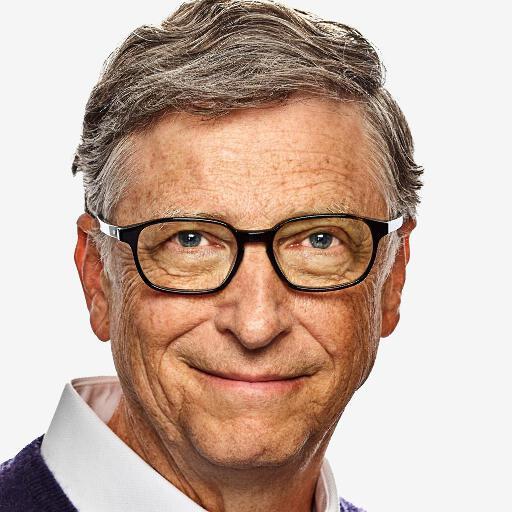} &
    \includegraphics[width=0.125\linewidth]{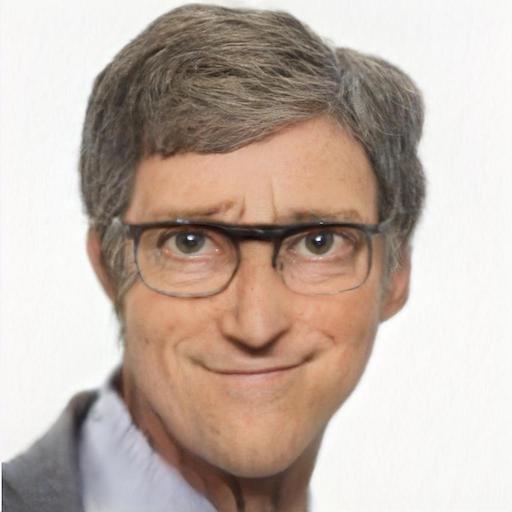} &
    \includegraphics[width=0.125\linewidth]{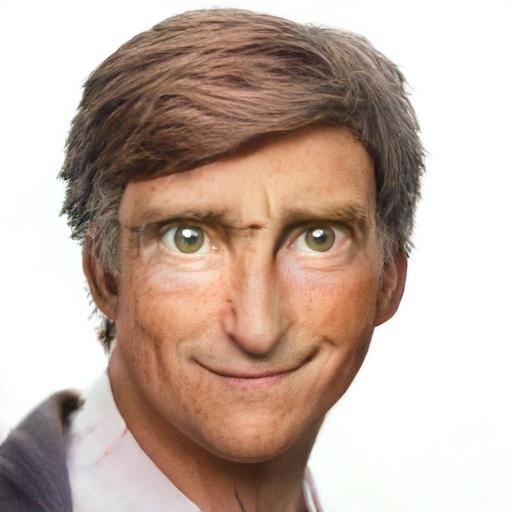} &
    \includegraphics[width=0.125\linewidth]{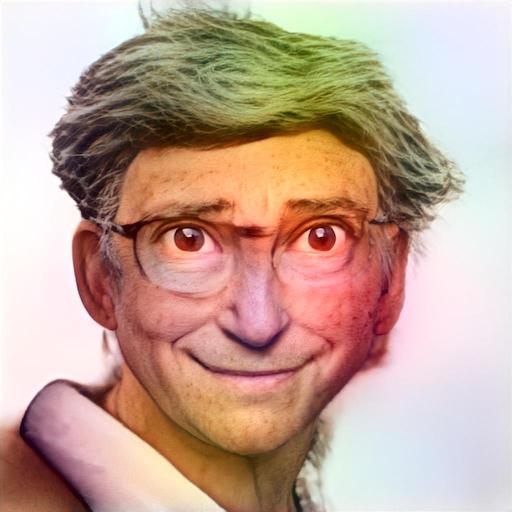} &
    \includegraphics[width=0.125\linewidth]{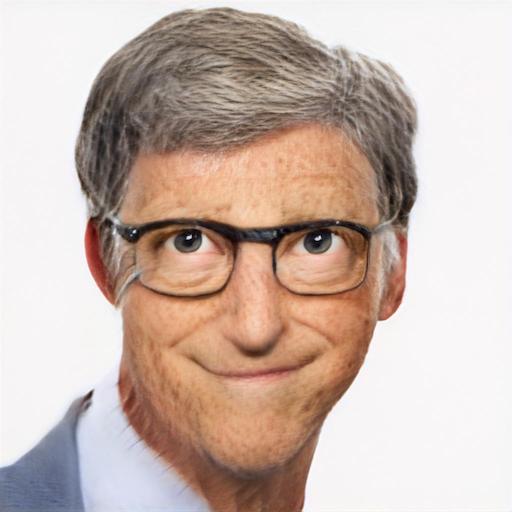} \\
    
    \raisebox{0.20in}{\rotatebox{90}{Toonify}} &
    \includegraphics[width=0.125\linewidth]{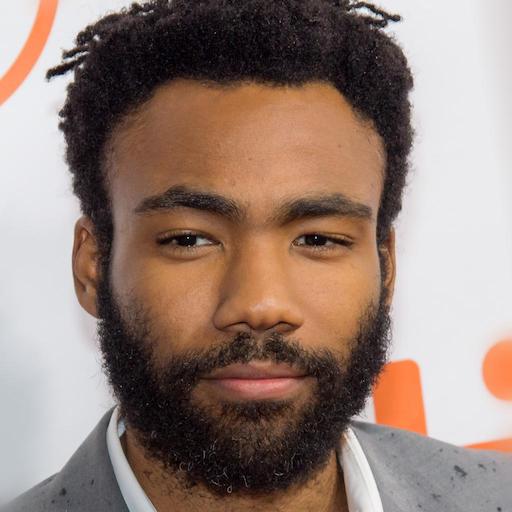} &
    \includegraphics[width=0.125\linewidth]{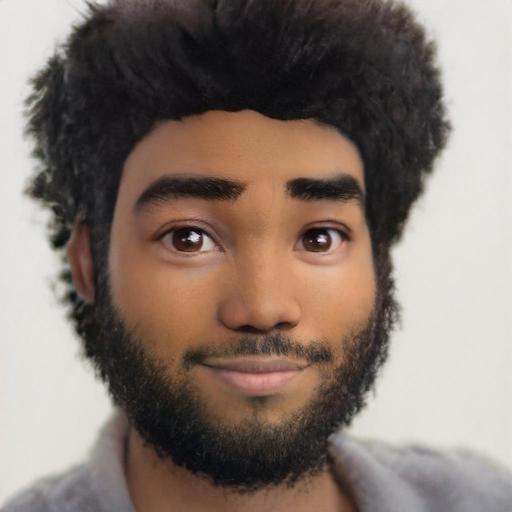} &
    \includegraphics[width=0.125\linewidth]{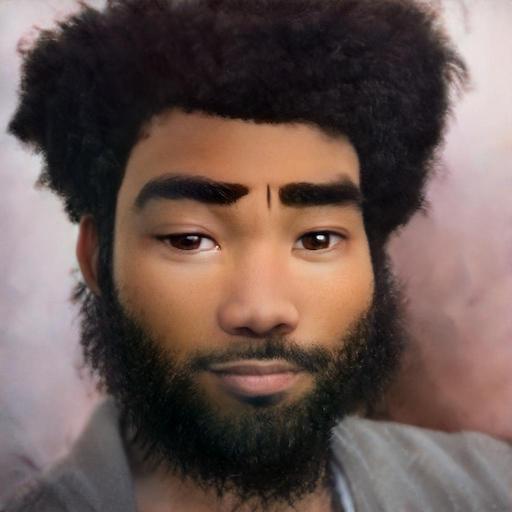} &
    \includegraphics[width=0.125\linewidth]{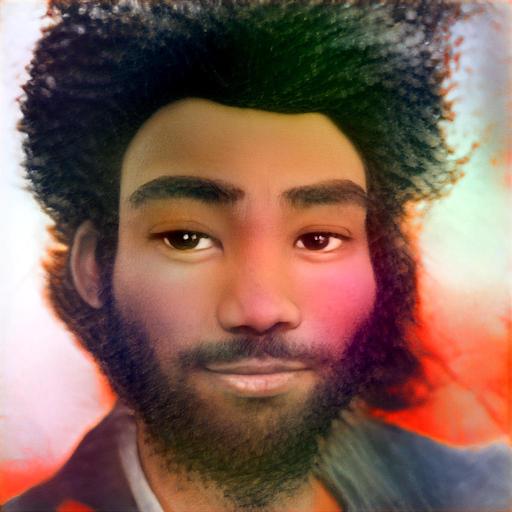} &
    \includegraphics[width=0.125\linewidth]{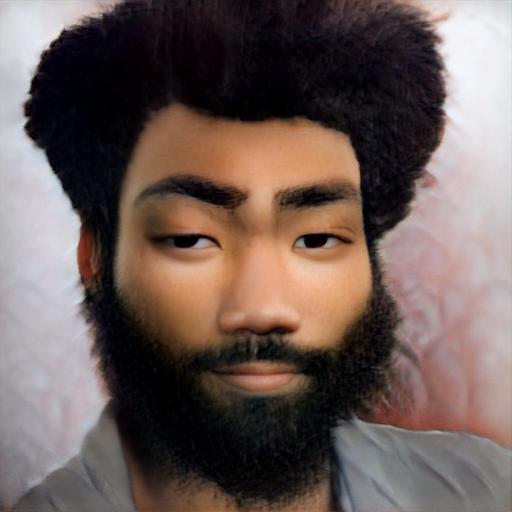} \\
    
    \raisebox{0.20in}{\rotatebox{90}{Toonify}} &
    \includegraphics[width=0.125\linewidth]{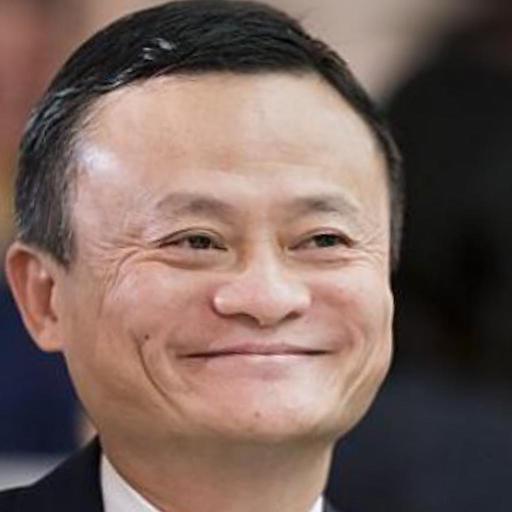} &
    \includegraphics[width=0.125\linewidth]{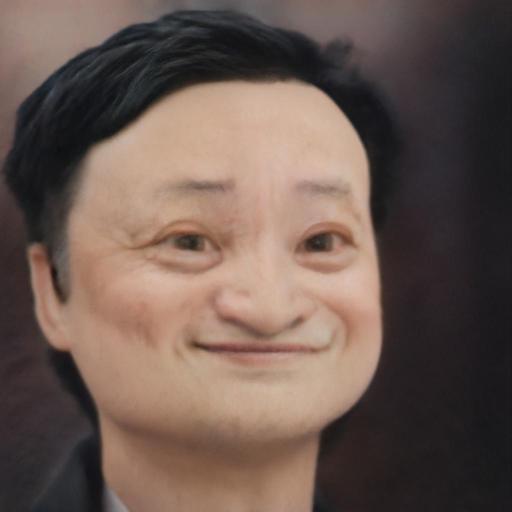} &
    \includegraphics[width=0.125\linewidth]{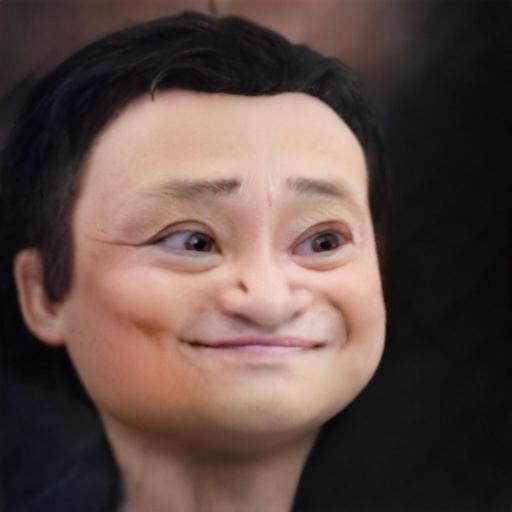} &
    \includegraphics[width=0.125\linewidth]{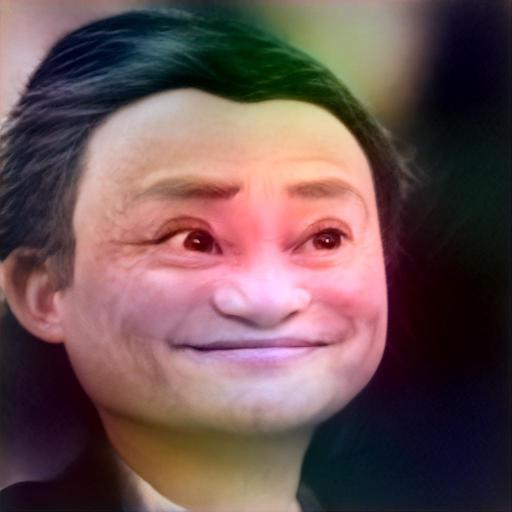} &
    \includegraphics[width=0.125\linewidth]{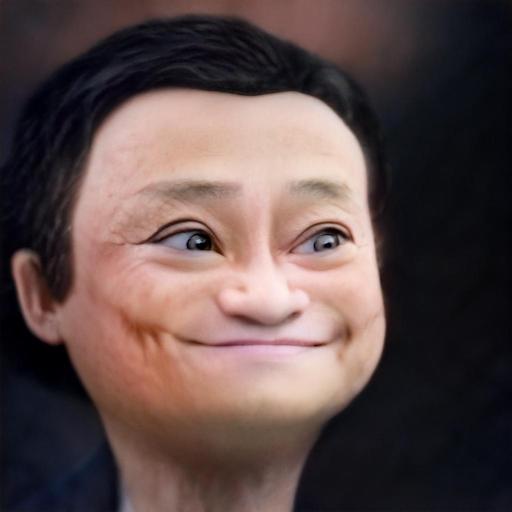} \\

    \raisebox{0.225in}{\rotatebox{90}{Sketch}} &
    \includegraphics[width=0.125\linewidth]{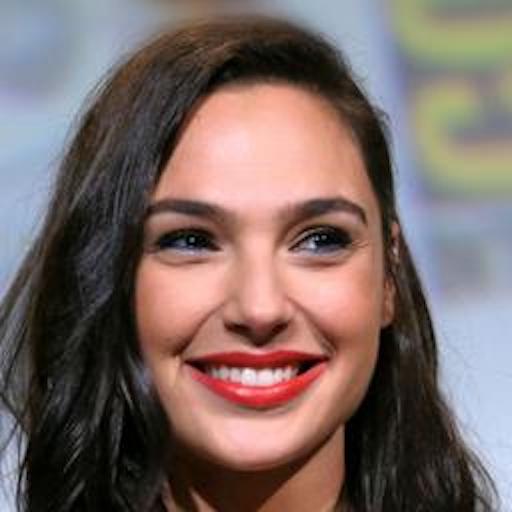} &
    \includegraphics[width=0.125\linewidth]{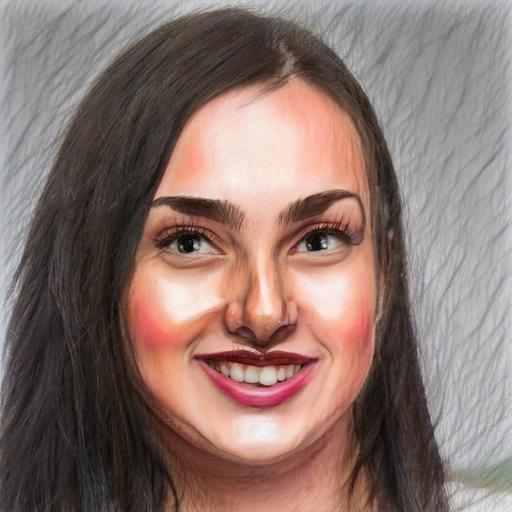} &
    \includegraphics[width=0.125\linewidth]{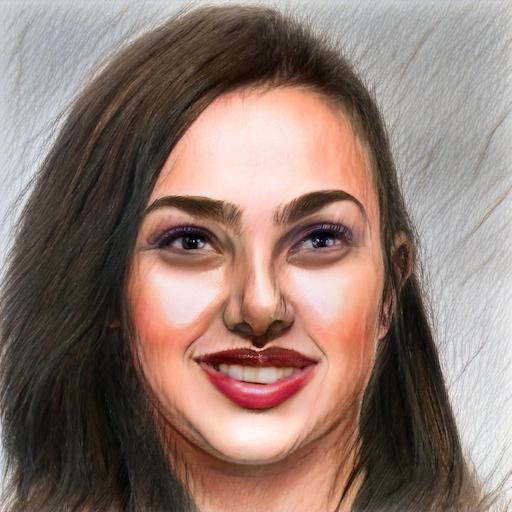} &
    \includegraphics[width=0.125\linewidth]{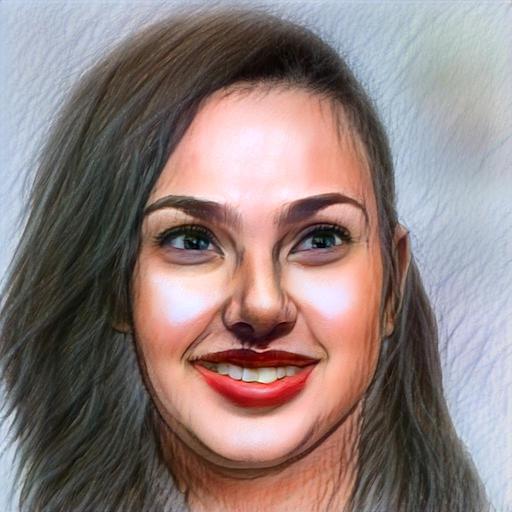} &
    \includegraphics[width=0.125\linewidth]{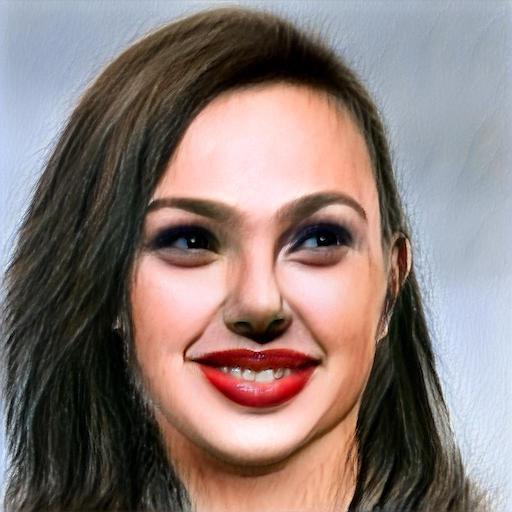} \\
    
    \raisebox{0.225in}{\rotatebox{90}{Sketch}} &
    \includegraphics[width=0.125\linewidth]{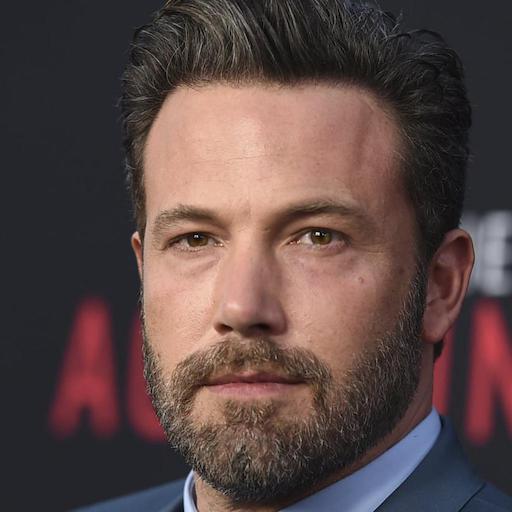} &
    \includegraphics[width=0.125\linewidth]{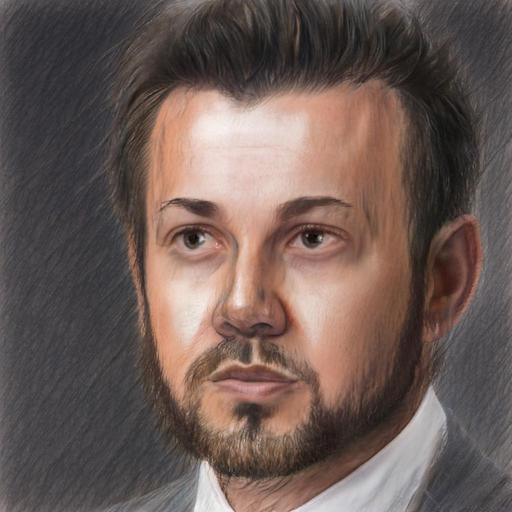} &
    \includegraphics[width=0.125\linewidth]{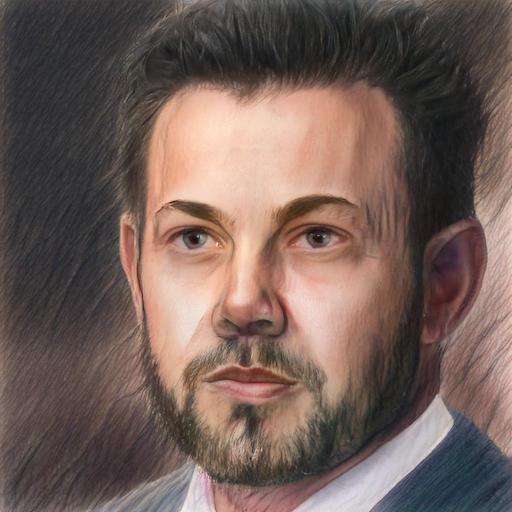} &
    \includegraphics[width=0.125\linewidth]{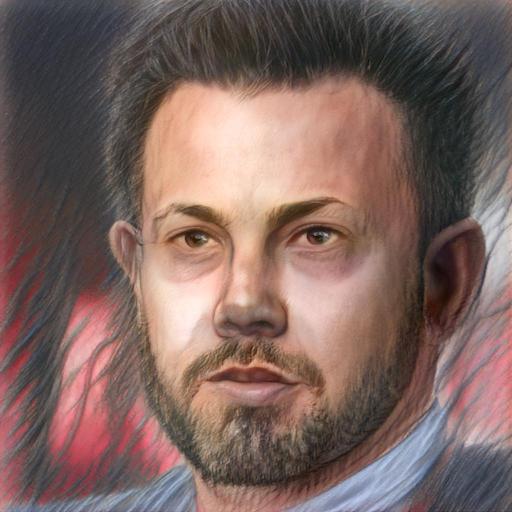} &
    \includegraphics[width=0.125\linewidth]{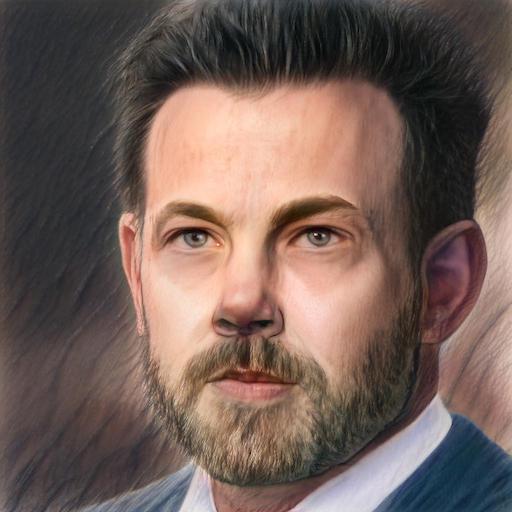} \\
    
    \raisebox{0.225in}{\rotatebox{90}{Sketch}} &
    \includegraphics[width=0.125\linewidth]{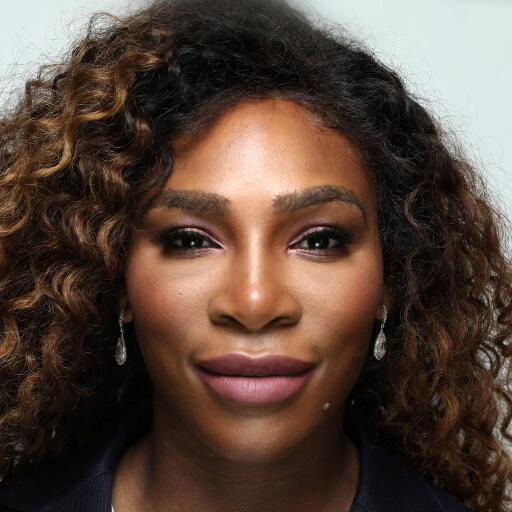} &
    \includegraphics[width=0.125\linewidth]{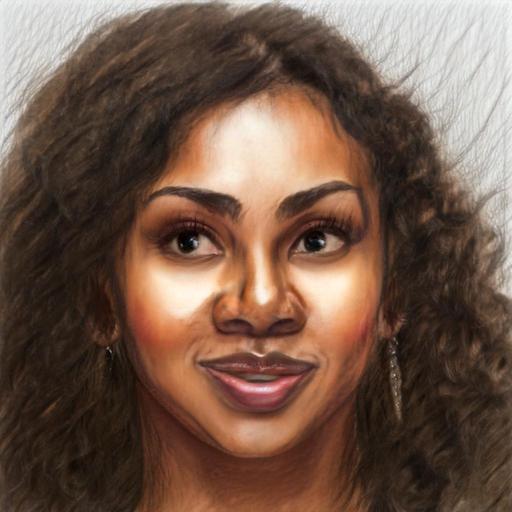} &
    \includegraphics[width=0.125\linewidth]{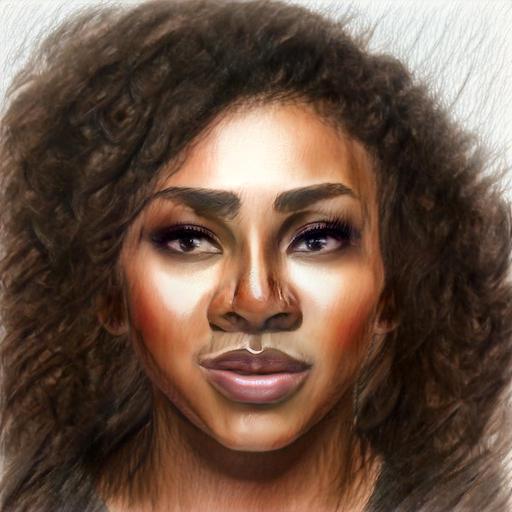} &
    \includegraphics[width=0.125\linewidth]{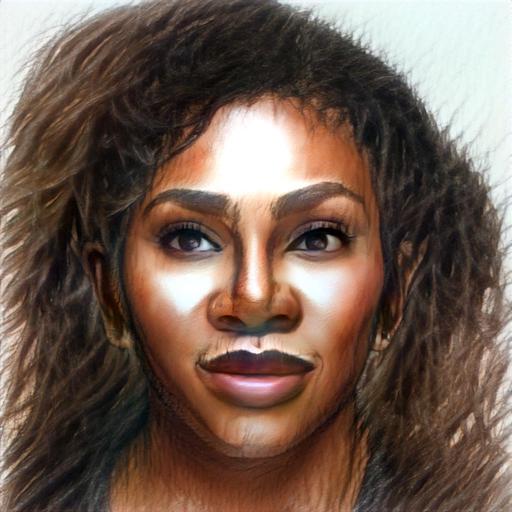} &
    \includegraphics[width=0.125\linewidth]{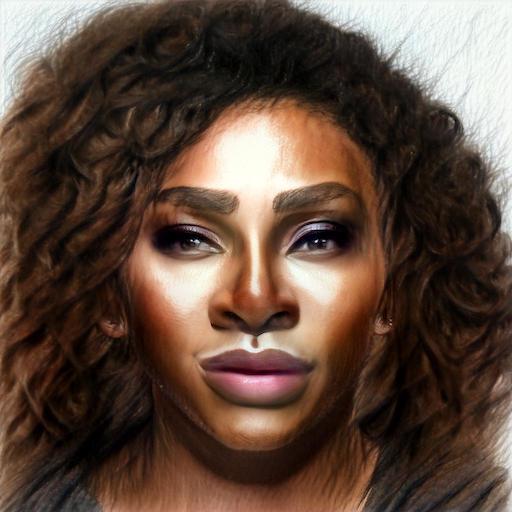} \\
    
    \raisebox{0.275in}{\rotatebox{90}{Pixar}} &
    \includegraphics[width=0.125\linewidth]{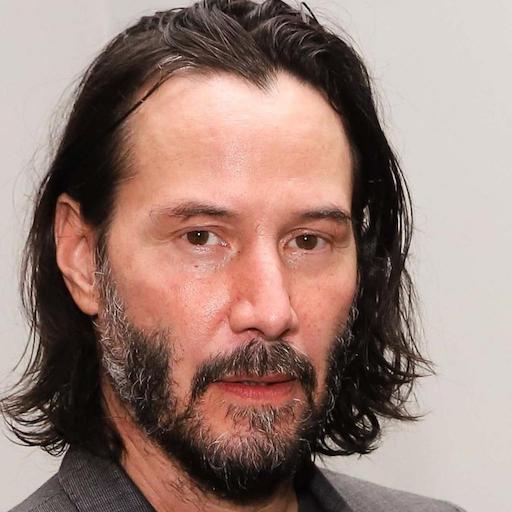} &
    \includegraphics[width=0.125\linewidth]{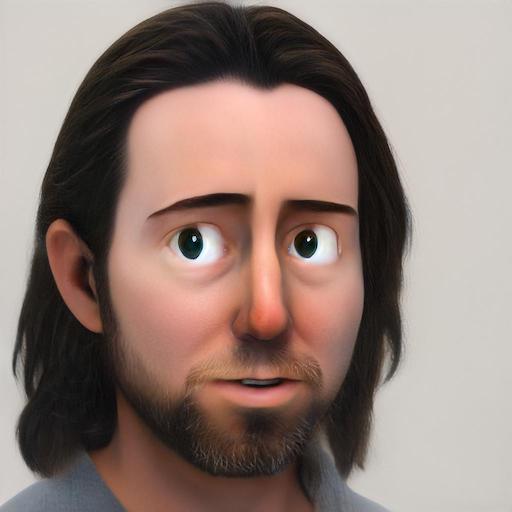} &
    \includegraphics[width=0.125\linewidth]{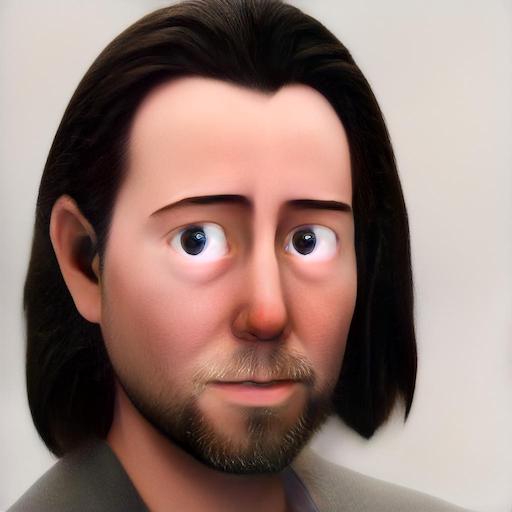} &
    \includegraphics[width=0.125\linewidth]{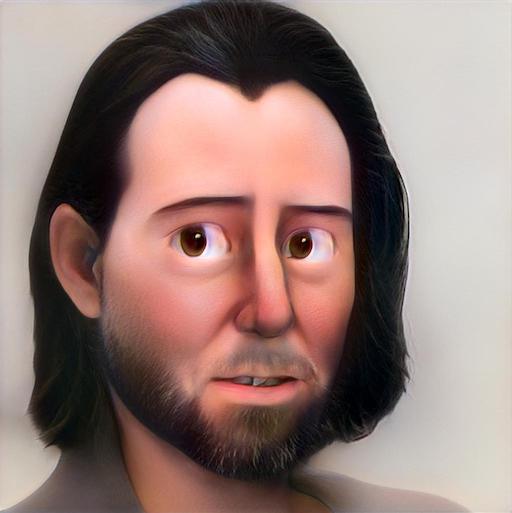} &
    \includegraphics[width=0.125\linewidth]{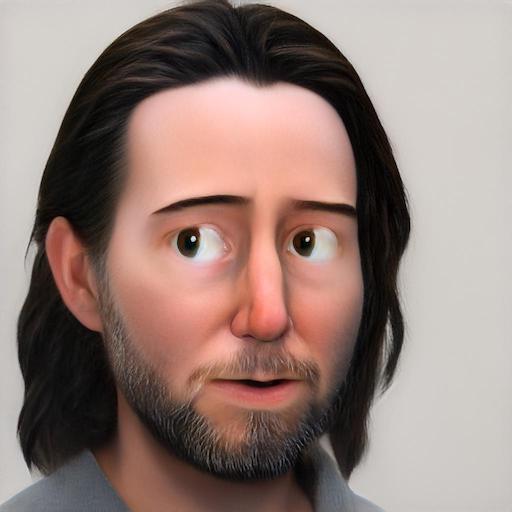} \\
    
    \raisebox{0.275in}{\rotatebox{90}{Pixar}} &
    \includegraphics[width=0.125\linewidth]{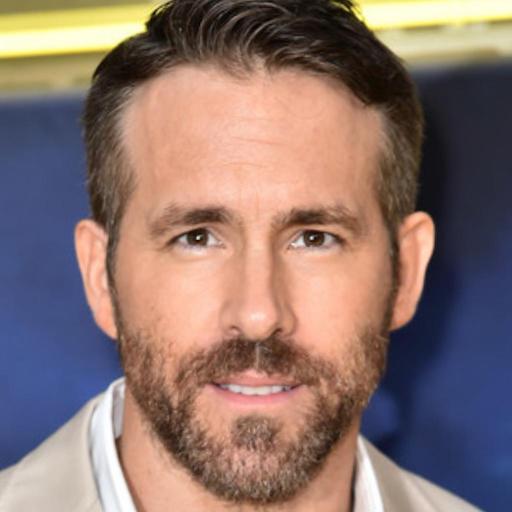} &
    \includegraphics[width=0.125\linewidth]{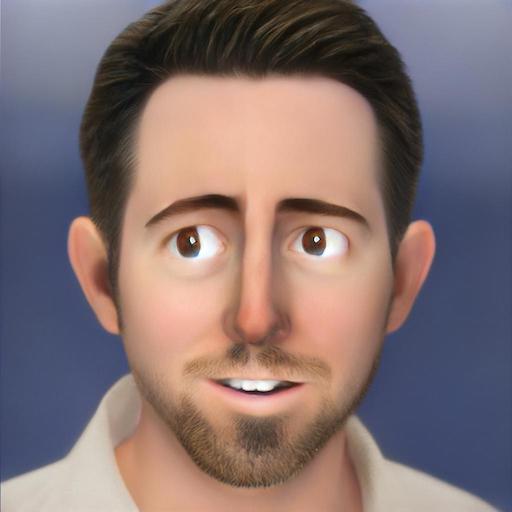} &
    \includegraphics[width=0.125\linewidth]{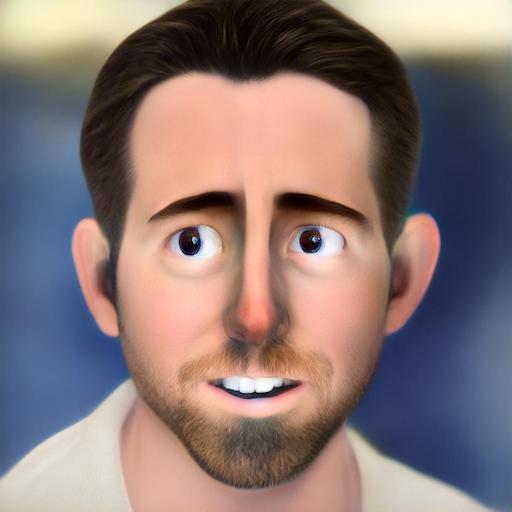} &
    \includegraphics[width=0.125\linewidth]{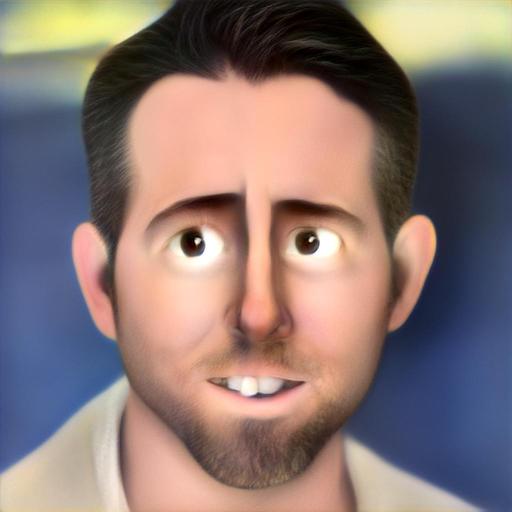} &
    \includegraphics[width=0.125\linewidth]{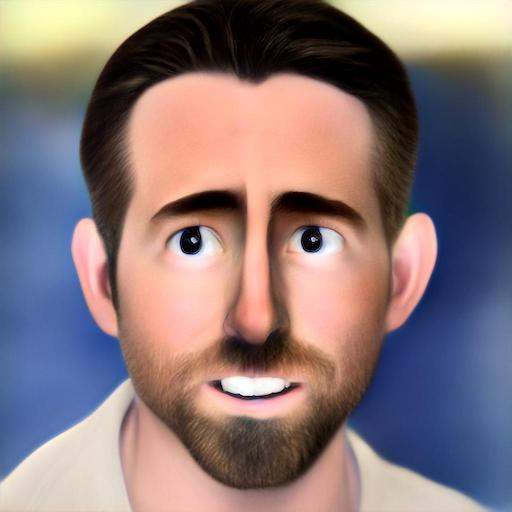} \\
    
    \raisebox{0.275in}{\rotatebox{90}{Pixar}} &
    \includegraphics[width=0.125\linewidth]{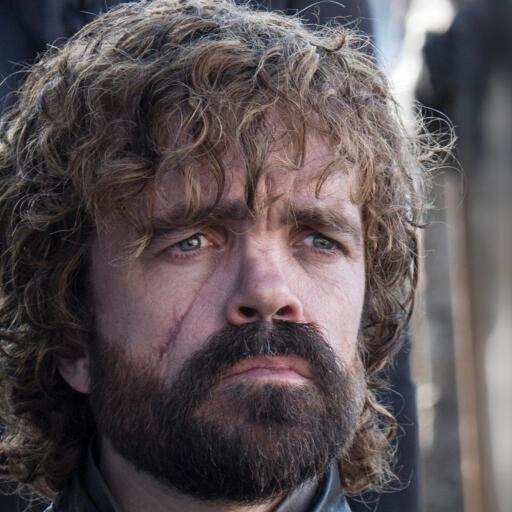} &
    \includegraphics[width=0.125\linewidth]{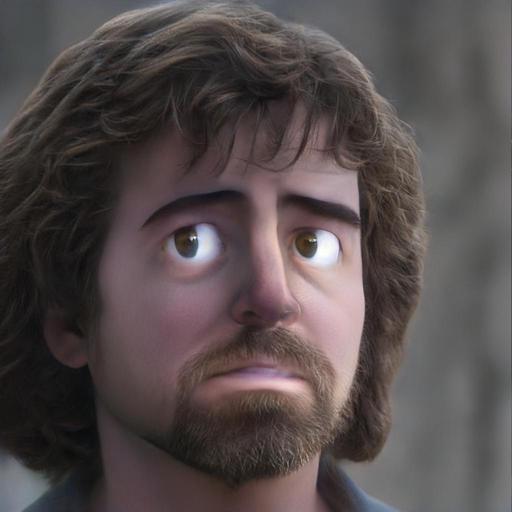} &
    \includegraphics[width=0.125\linewidth]{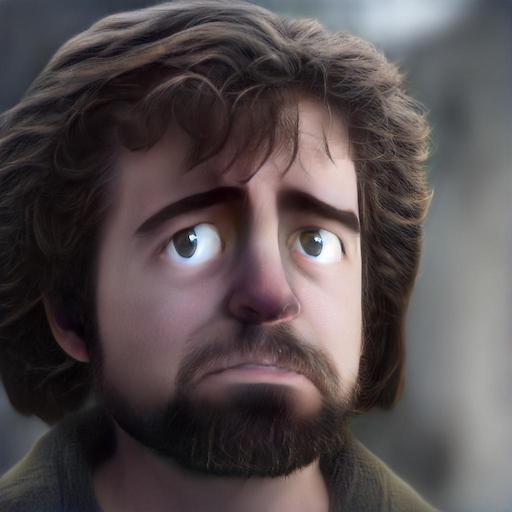} &
    \includegraphics[width=0.125\linewidth]{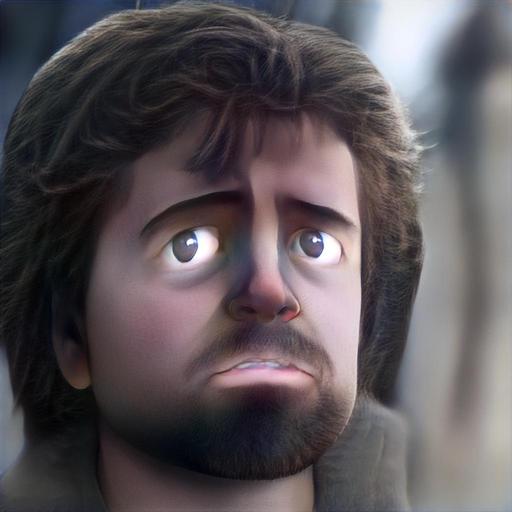} &
    \includegraphics[width=0.125\linewidth]{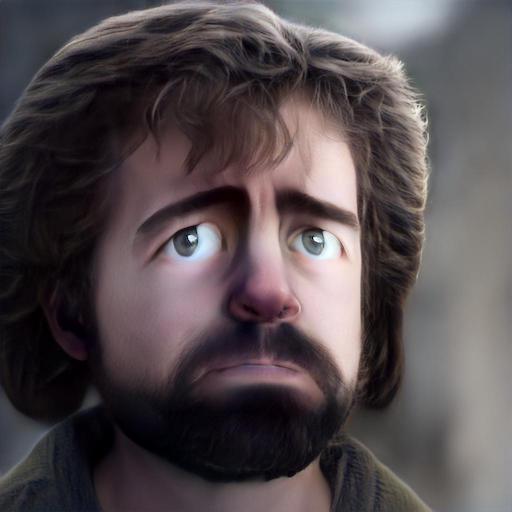} \\
    
    & Input & e4e & $\text{ReStyle}_{e4e}$ & $\text{ReStyle}_{pSp}$ & HyperStyle
    
    \end{tabular}
    }
    \caption{Additional domain adaptation comparisons for various fine-tuned models such as Toonify~\cite{pinkney2020resolution} and those obtained using StyleGAN-NADA~\cite{gal2021stylegannada}.}
    \label{fig:supplementary_domain_adaption}
\end{figure*}

\end{document}